\documentstyle[epsf]{julie}

\newcommand{\myendfig}{\vskip -0.14in \end{figure} }
\newcommand{\myendtab}{\vskip -0.01in \end{table} }

\newtheorem{ex}{Example}[chapter]
\newtheorem{theor}{Theorem}[chapter]
\newtheorem{mtd}{Method}[chapter]
\newtheorem{mydef}{Definition}[chapter]
\newtheorem{mylem}[theor]{Lemma}

\newcommand{\qed}{\mbox{$\ \Box$}}
\newcommand{\lqed}{\qed\vspace{.1in}}

\newenvironment{proof-of}[1]{\noindent {\bf Proof of #1:}}{\hfill \lqed}

\begin{document}
 \Name{Pedersen}{Ted}{ } 
 \Title{Learning Probabilistic Models}{of Word Sense Disambiguation}{}
 \DegreeA{B.A., Drake University}
 \DegreeB{M.S., University of Arkansas}
 \DegreeSought{Doctor of Philosophy}
 \Major{Computer Science}
 \University{Southern Methodist University}
 \School{School of Engineering and Applied Science}
 \DegreeDate{May 16, 1998}
 \ThesisDate{May 16, 1998}
 \ThesisType{Dissertation}

 \Advisor{Dan Moldovan}
 \CommitteeMemberA{Dr. Rebecca Bruce}
 \CommitteeMemberB{Dr. Weidong Chen}
 \CommitteeMemberC{Dr. Frank Coyle}
 \CommitteeMemberD{Dr. Margaret Dunham}
 \CommitteeMemberE{Dr. Mandyam Srinath}


 \ApprovalTitlePages

 \begin{Acknowledgment}
 I am indebted to Dr. Rebecca Bruce for sharing freely of her time, 
knowledge,  and insights throughout this research. Certainly none of this 
would have been possible without her. 

Dr. Weidong Chen, Dr. Frank Coyle, Dr. Maggie Dunham, Dr. Dan Moldovan,
and Dr. Mandyam Srinath have all made important contributions to this
dissertation. They   are also among the main reasons why my time at SMU
has been both happy  and productive.  

I am also grateful to Dr. Janyce Wiebe, Lei Duan, Mehmet Kayaalp, Ken 
McKeever, and Tom O'Hara for many valuable comments and suggestions that 
influenced the direction of this research. 

This work was supported by the Office of Naval Research under grant number 
N00014-95-1-0776.

 \end{Acknowledgment}

 \begin{Abstract}
 Selecting the most appropriate sense for an ambiguous word is a common
problem in natural language  processing. This dissertation pursues
corpus--based approaches that learn  probabilistic models of word sense
disambiguation from large amounts of  text. These models consist of a 
parametric form and parameter estimates. The parametric form characterizes 
the interactions among the  contextual  features and the sense  of the 
ambiguous word. Parameter estimates  describe the probability of observing 
different combinations of feature  values. These models disambiguate by 
determining the most probable sense  of an ambiguous word given the 
context in which it occurs. 

This dissertation presents several enhancements to existing supervised
methods of learning probabilistic models of disambiguation from 
sense--tagged text. A new search strategy, forward sequential, guides the 
selection process through the space of possible models. Each model 
considered for selection is judged by a new class of evaluation metric, 
the  information criteria. The combination of forward sequential search
and Akaike's Information Criteria is shown to consistently select highly
accurate models of disambiguation.  The same search strategy and
evaluation criterion also serve as the basis of the Naive Mix, a new
supervised learning algorithm that is shown to be competitive with leading
machine learning methodologies. In these comparisons the Naive Bayesian 
classifier also fares well which seems surprising since it is based on a 
model where the parametric form is simply assumed. However, an
explanation for this success is presented in terms of learning rates and 
bias--variance decompositions of classification error.    

Unfortunately, sense--tagged text only exists in small quantities and is 
expensive to create. This substantially limits the portability of
supervised learning approaches to word sense disambiguation. This 
bottleneck is addressed by developing unsupervised methods that learn
probabilistic models from raw untagged text. However, such text does not
contain enough information to automatically select a parametric form. 
Instead, one must simply be assumed. Given a form, the senses of ambiguous
words are treated  as missing data and their values are imputed via the
Expectation Maximization algorithm and Gibbs Sampling. Here the parametric 
form of the Naive Bayesian classifier is employed. However, this
methodology is appropriate for any parametric form in the class of
decomposable models. Several local--context, frequency--based  feature 
sets are also developed and shown to be appropriate for unsupervised 
learning of word senses from raw untagged text. 

 \end{Abstract}

 \PreliminaryPages



\begin{thesis}

\chapter{INTRODUCTION}

This dissertation is about computational methods that resolve the
meanings of ambiguous words in natural language text. Here, disambiguation 
is defined as the selection of the intended sense of an ambiguous word
from a known and finite set of possible meanings.  This choice is based 
upon a probabilistic model that tells which member of the set of possible
meanings is the most likely given the context in which the ambiguous word 
occurs. 

Resolving ambiguity is a routine process for a human; it requires
little conscious effort since a broad understanding of both language
and the real--world are utilized to make decisions about the intended 
sense of a word. For a human, the context in which an ambiguous
word occurs includes a wealth of knowledge beyond that which is contained 
in the text. Modeling this vast amount of information in a
representation a computer program can access and make inferences from is  
an, as yet, unachieved goal of Artificial Intelligence. Given the lack of 
such resources, this dissertation does not attempt to duplicate the 
process a human uses to resolve ambiguity. 

Instead, {\it corpus--based} methods are employed which make
disambiguation decisions based on probabilistic models learned from
large quantities of naturally occurring text. In these approaches, context 
is defined in a very limited way and consists of information that can 
easily be extracted from the sentence in which an ambiguous word occurs; 
no deep understanding of the linguistic structure or real--world 
underpinnings of a text is required. This results in methods that
take advantage of the abundance of text available online
and do not require the availability of rich sources of real--world 
knowledge. 

\newpage
\section{Word Sense Disambiguation}

Most words have multiple possible senses, each of which is appropriate
in certain contexts. Such ambiguity can result in the misunderstanding
of a sentence. For example, the newspaper headline {\it Drunk
Gets 9 Years in Violin Case} causes momentary confusion due to word
sense ambiguity. Does this imply that someone has been sentenced to
spend 9 years in a box used to store a musical instrument? Or has
someone has been sentenced  to prison for 9 years for a crime
involving a violin? Clearly the latter interpretation is
intended. The key to making this determination is  resolving the
intended sense of {\it case}. This is not terribly difficult for a human
since it is widely known that people are not imprisoned in violin
cases. However, a  computer program that attempts to resolve
this same
ambiguity will have a more challenging task since it is not likely  
to have this particular piece of knowledge available. 

The difficulty of resolving word sense ambiguity with a computer
program was first noted by Yehoshua Bar--Hillel, an early researcher in
machine translation. In \cite{BarHillel60} he presented the following
example: 
\begin{quote}
Little John was looking for his toy box. Finally, he found it. The box
was in the pen. John was very happy. 
\end{quote}
Bar-Hillel assumed that {\it pen} can have two senses: a writing
instrument or an enclosure where small children can play.
He concluded that:
\begin{quote}
\ldots no existing or imaginable program will enable an electronic
computer to determine that the word {\it pen} in the given sentence
within the given context has the second of the above meanings. 
\end{quote}

Disambiguating {\it pen} using a knowledge--based approach
requires rather esoteric pieces of information; ``toy boxes are 
smaller than play pens'' and ``toy boxes are larger than writing
pens,'' plus some mechanism for making inferences given these facts. 
To have this available for all potential ambiguities is
indeed an impossibility. In that regard Bar--Hillel is
correct. However, while such approaches require an impractical amount
of real--world knowledge, corpus--based methods that learn from large
amounts  of naturally occurring text offer a viable alternative.  

Computational approaches that automatically perform  word sense
disambiguation have potentially wide application. Resolving ambiguity
is an important issue in machine translation, document categorization,
information retrieval, and language understanding.   

Consider an example from machine translation. The noun {\it
bill} can refer to a piece of legislation that is not yet law or to a
statement requesting payment for services rendered. However, in
Spanish these two  senses of {\it bill} have two distinct
translations; {\it proyecto de ley} and {\it cuenta}. To translate
{\it The Senate bill is being voted on tomorrow} from English to
Spanish, the intended sense of {\it bill} must be resolved.  Even a
simple word by word translation to Spanish is not possible without
resolving this ambiguity. 

Document classification can also hinge upon the interpretation of an
ambiguous word. Suppose that there are two documents where the word
{\it bill} occurs a large number of times. If a classification
decision is made based on this fact and the sense of {\it bill} is not
known, it is possible that {\it Peterson's Field Guide to North American 
Birds} and the {\it Federal Register} will be considered the same type
of document as both contain frequent usages of {\it bill}. 

\section{Learning from Text}

This dissertation focuses on corpus--based approaches to learning 
probabilistic models that resolve the meaning of ambiguous words. These 
models indicate which sense of an ambiguous  word is most probable given 
the context in which it occurs. In this framework disambiguation consists 
of classifying an ambiguous word into one of several predetermined 
senses.  

These probabilistic models are learned via supervised and unsupervised 
approaches. If manually disambiguated examples are available to serve
as {\it training data} then supervised learning is most effective.  These  
examples take the form of  {\it sense--tagged text} which is created by 
collecting a large number of sentences that contain a particular ambiguous 
word.  Each instance of the ambiguous word is manually annotated to  
indicate the most appropriate sense for that  usage.  Supervised learning  
builds a generalized model from this set of examples and uses this 
model to disambiguate instances of the ambiguous word found in {\it test
data} that is separate from the training data. 

If there are no training examples available then learning is
unsupervised and is based upon {\it raw} or {\it untagged} text. An
unsupervised algorithm divides all the usages of an ambiguous word into a
specified number of groups based upon the context in which each instance
of the word occurs. There is no separation of the data into a training and
test sample. 

Before either kind of learning can take place, a feature set
must be developed. This defines the context of the ambiguous
word and consists of those properties of both the ambiguous 
word and the sentence in which it occurs that are relevant to making a sense
distinction. These properties are generally referred to as {\it
contextual features} or simply {\it features}. Human intuition and
linguistic insight are certainly desirable at this stage. The
development of a feature set is a subjective process; given the
complexity of human language there are a huge number of possible
contextual features and it is not possible to empirically examine even
a fraction of them. This dissertation uses an existing feature set for
supervised learning and develops several new feature sets appropriate
for unsupervised learning.

Regardless of whether a probabilistic model is learned via supervised
or unsupervised techniques, the nature of the resulting model is
the same. These models consist of a {\it parametric form} and {\it
parameter estimates}.  The parametric form shows which contextual
features affect the values of other contextual features as well as
which contextual features affect the sense of the ambiguous word. The
parameter estimates tell how likely certain combinations of values for
the contextual features  are to occur with a particular sense of an
ambiguous word. 

Thus, there are two steps to learning a probabilistic model of
disambiguation. First, the parametric form must either be specified by
the user or learned from sense--tagged text. Second, parameter
estimates are made based upon evidence in the text.   The following
sections summarize how each of these steps is performed during
supervised and unsupervised learning. More details of the learning
processes are contained in Chapters 3 and 4. Empirical evaluation 
of these methods is presented in Chapters 6 and 7. 

\subsection{Supervised Learning}

The supervised approaches in this dissertation generally follow the
model selection method introduced by Bruce and Wiebe (e.g,
\cite{Bruce95},  \cite{BruceW94A}, and \cite{BruceW94B}). Their method
learns both the parametric form and parameter estimates of a special
class  of probabilistic models, {\it decomposable  log--linear
models}. This dissertation extends their approach by identifying
alternative criteria for evaluating the suitability of a model for
disambiguation and also identifies an alternative strategy for
searching the space of possible models.  

The approach of Bruce and Wiebe and the extensions described in this 
dissertation all have the objective of learning a single probabilistic 
model that adequately characterizes a training sample for a given
ambiguous word. However, this dissertation shows
that different models selected by different methodologies often result
in similar levels of disambiguation performance. This suggests that
model selection is somewhat uncertain and that a single ``best'' model
may not exist for a particular word. A new variation on  the
sequential model selection methodology, the {\it Naive Mix}, is
introduced and addresses this type of uncertainly. The Naive Mix
is an averaged probabilistic model that is based on an entire sequence
of models found during a selection process rather than just a single
model. Empirical comparison shows that the Naive Mix improves
on the disambiguation performance of a single selected model and 
is competitive with leading machine learning algorithms.   

The {\it Naive Bayesian classifier} is a supervised learning method
where the  parametric form is assumed and only parameter estimates are 
learned from sense--tagged text. Despite some history of success in word 
sense disambiguation and other applications, the behavior of Naive Bayes 
has  been  poorly understood. This dissertation includes an analysis  
that offers an explanation for its ability to perform at relatively high 
levels of accuracy. 

\subsection{Unsupervised Learning}

A general limitation of supervised learning approaches to word sense
disambiguation is that sense--tagged text is not available for most
domains. While sense--tagged text is not as complicated to create as
more elaborate representations of real--world knowledge, it is still a
time--consuming activity and limits the portability of methods that
require it. In order to overcome this difficulty, this
dissertation develops  {\it knowledge--lean} approaches that learn
probabilistic models from raw untagged text. 

Raw text only consists of the words and punctuation that normally
appear in a document; there are no manually attached sense
distinctions to ambiguous words nor is any other kind of information
augmented to the raw text. Even without sense--tagged text it is still
possible to learn a probabilistic model using an unsupervised approach.
In this case the parametric form must be specified by the user and
then parameter estimates can be made from the text.  Based on its success 
in supervised learning, this dissertation uses the parametric form of the 
Naive Bayesian classifier when performing unsupervised learning of 
probabilistic models. However, estimating parameters is more complicated 
in unsupervised learning than in the supervised case. 

The parametric form of any probabilistic model of disambiguation 
must include a feature 
representing the sense of the ambiguous word; however, raw text contains 
no values for this feature. The sense is  treated as a {\it latent} or 
{\it missing} feature.   Two different approaches to estimating parameters
given missing data  are evaluated;  the EM algorithm and Gibbs Sampling. 
The probabilistic models that result are also compared to two well--known 
agglomerative clustering algorithms, Ward's minimum--variance method and
McQuitty's  similarity analysis. The application of these methodologies to
word sense  sense disambiguation  is an important development since it
eliminates the requirement for sense--tagged text made by supervised
learning algorithms.   

\section{Basic Assumptions}

There are several assumptions that underly both the supervised and
unsupervised approaches to word sense disambiguation presented in this
dissertation:

\begin{enumerate}
\item A separate probabilistic model is learned for each ambiguous
word.  
\item  Any part--of--speech ambiguity is resolved prior to sense
disambiguation.\footnote{For example, {\it share} can be used as a noun,
{\it I have a share of stock}, or as a verb, {\it It would be nice to
share your stock}.} 
\item Contextual features are only defined within the boundaries
of the sentence in which an ambiguous word occurs. In other words,
only information that occurs in the same sentence is used to resolve
the meaning of an ambiguous word.   
\item The possible senses of a word are defined by a dictionary and
are known prior to disambiguation. In this dissertation Longman's
Dictionary of Contemporary English \cite{Procter78} and WordNet
\cite{Miller93} are the sources of word meanings.   
\end{enumerate}

The relaxation or elimination of any of these assumptions presents
opportunities for future work that will be discussed further in Chapter
9.  

\section{Chapter Summaries}

Chapter 2 develops background material regarding probabilistic models
and their use as classifiers. Particular emphasis is placed on the
class of decomposable models since they are used throughout this
dissertation. 

Chapter 3 discusses supervised learning approaches to word sense
disambiguation. The statistical  model selection method of Bruce 
and Wiebe is outlined here and alternatives to their
model evaluation criteria and search strategy are presented. 
The Naive Mix is introduced.  This is a new supervised learning
algorithm that extends model selection from a process that selects a
single probabilistic model to one that finds an averaged model based
on a sequence of probabilistic models. Each succeeding model in the
sequence characterizes the training data increasingly well. The Naive
Bayesian classifier is also presented.  

Chapter 4 addresses unsupervised learning of word senses from raw,
untagged text. This chapter shows how the EM algorithm and Gibbs
Sampling can be employed to estimate the parameters of a model given
the parametric form and the systematic absence of data; in this case
the sense of an ambiguous word is treated as missing data. 
Two agglomerative clustering algorithms, Ward's minimum--variance
method and McQuitty's similarity analysis, are also presented and used
as points of comparison.  

Chapter 5 describes the words that are disambiguated as part of the
empirical evaluation of the methods described in Chapters 3 and 4. The 
possible senses for each word are defined and an empirical study of the 
distributional characteristics of each word is presented.  Four feature 
sets are also discussed. The feature set for supervised learning is 
due to Bruce and Wiebe. There are three new feature sets introduced for  
unsupervised learning. 

Chapter 6 presents an empirical evaluation of the supervised learning
algorithms described in Chapter 3. There are four principal
experiments. The first compares the  overall accuracy of a range of
sequential model selection methods. The second compares the accuracy
of the Naive Mix to several leading machine learning algorithms. The
third determines the learning rate of the most accurate methods from
the first two experiments. The fourth decomposes the classification
errors of the most accurate methods into more fundamental components. 

Chapter 7 makes several comparisons among the unsupervised learning methods
presented in Chapter 5.
The first is between the accuracy of probabilistic models where the 
parametric form is assumed and parameter estimates are made via the EM
algorithm and Gibbs Sampling. The second employs two agglomerative 
clustering algorithms, Ward's minimum--variance method and McQuitty's
similarity analysis, and determines which is the more accurate. 
Finally, the two most accurate approaches, Gibbs Sampling and McQuitty's
similarity analysis,  are compared. 

Chapter 8 reviews related work in word sense disambiguation.
Methodologies are grouped together based upon the type of
knowledge source or data they require to perform disambiguation. 
There are discussions of  work  based on semantic  networks, 
machine readable  dictionaries, parallel translations of text,
sense--tagged text, and raw untagged text. 

Chapter 9 summarizes the contributions of this dissertation and
provides a discussion of future research directions. 

\chapter{PROBABILISTIC MODELS}

This chapter introduces the basics of probabilistic models and shows how   
such models can be used as classifiers to perform word sense 
disambiguation. Particular  attention is paid to a special class of 
probabilistic model known as {\it decomposable log--linear models}
\cite{DarrochLS80} since they are  well suited for use with the
supervised and unsupervised learning methodologies described in
Chapters 3 and 4. 

\section{Inferential Statistics} \label{sec:probabilistic}

The purpose of inferential statistics is to learn something about a
population of interest.  The characteristics of a population are
described by {\it parameters}.  Since it is generally not possible to
exhaustively study a population, estimated values for parameters are
learned from randomly selected samples of data from the population.

Each parameter is associated with a distinct {\it event} that can
occur in the population. An event is the state of a process at
a particular moment in time. A common example is coin
tossing.  This is a {\it binomial} process since there are only two
possible events; the coin toss comes up heads or tails. A process with
more than two possible events is {\it multinomial}. Tossing a die is
an example since there are 6 possible events.    

The events in this dissertation are sentences in which an ambiguous
word occurs. Each sentence is  represented by a combination  of
discrete values for a set of {\it random variables}. Each random
variable represents a property or {\it feature} of the sentence.
The dependencies among these features are characterized by the 
{\it parametric form} of a probabilistic model. 

A {\it feature vector} is a particular instantiation of the random
variables. Each feature vector represents an observation or an
instance of an event, i.e., a sentence with an
ambiguous word.  The exhaustive collection of all possible events given
a set of feature  variables defines the {\it event space}. 

The {\it joint probability distribution} of a set of feature variables
indicates how likely each event in the event space is to occur. 
The probability of observing a  particular event is described by a
parameter. In addition to the parametric form, a probabilistic model
also includes estimated values for all of these parameters.  

Suppose that in a random sample of events from a population there
are $N$ observations of $q$ distinct events, i.e., feature vectors,
where each observation is described by $n$ discrete feature variables
($F_1, F_2,$ $\ldots,$ $F_{n-1},F_n$). Let $f_i$ and $\theta_i$ be the
frequency and probability of the $i^{th}$ feature vector,
respectively. Then the data sample $D = (f_1,f_2, \ldots, f_q)$ has a
multinomial distribution with parameters $(N;\Theta)$, where
$\Theta = (\theta_1, \theta_2, \ldots, \theta_q)$
defines the joint probability distribution of the feature
variables ($F_1, F_2, \ldots, F_{n-1},F_n$). 

The parameters of a probabilistic model can be estimated using a
number of approaches; maximum likelihood and Bayesian estimation are
described in the following sections. The model selection methodologies
described in Chapter 3  and the EM algorithm from Chapter 4 employ
maximum likelihood estimates. Gibbs Sampling, also described in
Chapter 4, makes use of Bayesian estimates. 

\subsection{Maximum Likelihood Estimation}

Values for the parameters of a probabilistic model can be estimated
using  maximum likelihood estimates such that $\hat\theta_i =
\frac{f_i}{N}$.  In this framework, a parameter can only be estimated
if the associated event is observed in a sample of data. 

A maximum likelihood estimate maximizes the probability of obtaining
the data sample that was observed, $D$, by maximizing the likelihood
function, $p(D|\Theta)$. The likelihood function for a multinomial
distribution is defined as follows:\footnote{Other distributions will
have different formulations of the likelihood function.}  
\begin{equation}
p(D|\Theta) = \frac{N!}{\prod_{i=1}^{q} f_{i}!} \prod_{i=1}^{q}
\hat\theta_i^{f_i}
\end{equation}

Implicit in the multinomial distribution is the assumption that all the
features of an event are {\it dependent}. When this is the case the
value of any single feature variable is directly affected by the
values of all the other feature variables.  A probabilistic model
where all features are dependent is considered {\it saturated}.  

The danger of relying on a saturated probabilistic model is
that reliable parameter estimates may be difficult to obtain. When
using maximum likelihood estimates, any event that is not observed in
the data sample will have a zero--valued parameter estimate associated
with it. This is undesirable since the model  regards the associated
event as an impossibility. It is more likely that the event is simply
unusual and that the sample is not large enough to gather adequate
information regarding rare events when using a saturated model.   

However, if the event space is very small it may be reasonable to
assume that all feature variables are dependent on one another and
that every possible event can be observed in a data
sample. For example, if  an event space is defined by two binary
feature variables,  $(F_1, F_2)$,  then the saturated model has four
parameters, each representing the probability of observing one of the
four possible events. Table \ref{tab:4parms} shows a scenario where
a sample consists of $N=150$ events. The frequency counts of these
events are shown in column $freq(F_1,F_2)$,  and the resulting
maximum likelihood estimates are calculated and displayed in column MLE. 

\begin{table}
\begin{center}
\caption{Maximum Likelihood Estimates} \label{tab:4parms}
\begin{tabular}{|ccc|c|} \hline
$F_1$ & $F_2$& $freq(F_1,F_2)$ &  MLE \\ \hline
0   & 0     & 21 & $\hat\theta_{1} = \frac{21}{150} = .14$ \\
0   & 1     & 38 & $\hat\theta_{2} = \frac{38}{150} = .25$ \\
1   & 0     & 60 & $\hat\theta_{3} = \frac{60}{150} = .40$ \\
1   & 1     & 31 & $\hat\theta_{4} = \frac{31}{150} = .21$ \\ \hline 
\end{tabular}
\end{center}
\myendtab

It is more often the case in real world problems that the number of
possible events is somewhat larger than four. The number of parameters
needed to represent these events in a probabilistic model is
determined by the number of dependencies among the feature
variables. If  the model is saturated then all of the features are
dependent on one  another and the number of parameters in the
probabilistic model is equal  to the number of possible events in the
event space.   

Suppose that
an event space is defined by a set of 20 binary feature variables
($F_1, F_2, \cdots, F_{20}$). The joint probability distribution of
this feature set consists of $2^{20}$ parameters. Unless the number of 
observations  in the data
sample is greater than $2^{20}$, it is inevitable that there will be a
great many parameter estimates with zero values. If $q < 2^{20}$,
where $q$ represents the number of distinct events in a sample, then
$2^{20} - q$ events will have zero estimates. This situation is
exacerbated if the distribution of events in the data sample is {\it
skewed}, i.e., $q \ll N$. Unfortunately, it is often the case in
natural language that the distribution of events is quite skewed
(e.g. \cite{PedersenKB96}, \cite{Zipf35}). 

An alternative to using a saturated model is  to find a probabilistic 
model with fewer dependencies among the feature
variables that still maintains a good {\it fit} to the data sample. 
Such a model is more 
parsimonious and yet retains a reasonably close 
characterization of the data. Given such a  model,   the joint
probability distribution can be expressed in terms of a smaller number
of parameters.   

Dependencies among feature variables can be eliminated if a pair of
variables are identified as {\it conditionally independent}.
Feature variables $F_1$ and $F_2$ are conditionally independent given
S if:  
\begin{equation}
p(F_1=f_1|F_2=f_2,S=s) \ \ \ = \ \ \ p(F_1=f_1|S=s)  
\label{eq:ci1}
\end{equation}
or:
\begin{equation}
p(F_2=f_2|F_1=f_1,S=s) \ \ \ = \ \ \ p(F_2=f_2|S=s)    
\label{eq:ci2}
\end{equation}

In Equation \ref{eq:ci1}, the probability of observing feature variable 
$F_1$ with value $f_1$ is not affected by the value of feature variable 
$F_2$ if it is already known that feature variable $S$ has value $s$. A
similar interpretation applies to Equation \ref{eq:ci2}.\footnote{In the 
remainder of this dissertation, a simplified notation will be employed 
where feature variable names are not specified when they can be
inferred from the feature values. For example, in $p(f_1|f_2,s) =
p(f_1|s)$ it is understood that the lower case letters refer to
particular values for a feature variable of the same name.}     

An automatic method for selecting probabilistic models with fewer
dependencies among the feature variables is described by Bruce and
Wiebe (e.g., \cite{Bruce95}, \cite{BruceW94A}, \cite{BruceW94B}).
This method selects models from the class of decomposable log--linear
models  and will be described in greater detail in Chapter 3.   

\subsection{Bayesian Estimation}

Bayesian estimation of parameters is an alternative to maximum
likelihood estimation. Such an estimate is the product of the
likelihood function, $p(D|\Theta)$, and the prior probability,  
$p(\Theta)$. This product defines the {\it posterior probability
function}, $p({ \Theta}|D)$, defined by Bayes Rule as:   
\begin{equation}
p(\Theta|D) = \frac{p(D|\Theta)p(\Theta)}{p(D)}
\label{eq:bayesrule}
\end{equation}

The posterior function represents the probability of
estimating the parameters, $\Theta$, given the observed sample,
$D$. The likelihood function, $p(D|{ \Theta})$, represents the
probability of observing the sample, $D$, given that it comes from the
population characterized by the parameters, $\Theta$. The
prior probability function, $p(\Theta)$, represents the unconditional
probability that the parameters have values $\Theta$. This is a
subjective probability that is estimated prior to sampling. 
Finally, $p(D)$ is the probability of observing a 
sample, $D$, regardless of the actual value of the parameters,
$\Theta$.    

When making a Bayesian estimate some care must be taken in specifying
the distribution of the prior probability $p(\Theta)$. The
nature of the likelihood function must be taken into account,
otherwise the product of the likelihood function and the
prior function may lead to invalid results. Prior probabilities whose
distributions lend themselves to fundamentally sound computation of
the posterior probability from the likelihood function are known as 
{\it conjugate priors}. A prior probability is a conjugate prior if it is
related to the events represented by the likelihood function in such a
way that both the posterior and prior  probabilities are members of
the same family of distributions.   

For example, suppose a binomial process such as coin tossing  is
being modeled, where the observations in a sample are classified into
two mutually exclusive categories; heads or tails. The beta
distribution is known to be conjugate to observations in a binomial
process. If the prior probability of observing a heads or tails is
assigned via a beta distribution, then the posterior probability will
also  be a member of the beta family.   

The multinomial distribution is the $n$--event generalization of the
2--event binomial distribution. The  Dirichlet distribution is the
$n$--event generalization of the 2--event beta distribution. Since the
beta distribution is the conjugate prior of the binomial distribution,
it follows that the Dirichlet distribution is the conjugate prior of
the multinomial distribution. When the likelihood function is
multinomial and the prior function is specified using the Dirichlet
distribution, the resulting posterior probability function is
expressed in terms of the Dirichlet distribution. 

\section{Decomposable Models} \label{sec:decomposable}

Decomposable models \cite{DarrochLS80} are a subset of the class of
Graphical Models \cite{Whittaker90} which is in turn a subset of the
class of log-linear models \cite{BishopFH75}. Decomposable models can
also be categorized as the class of models that are both 
Bayesian Networks \cite{Pearl88} and Graphical Models. 
They were first applied to 
natural language processing and word sense disambiguation by Bruce and 
Wiebe (e.g., \cite{Bruce95},  \cite{BruceW94A}, \cite{BruceW94B}).  

In any Graphical Model, feature variables are either dependent or
conditionally independent of one another.  The parametric form of
these models have a graphical representation such that each feature
variable in the model is represented by  a node in the graph, and there is
an undirected edge between each pair of nodes corresponding to
dependent feature variables. Any two nodes that are not directly
connected by an edge are conditionally independent given the values of
the nodes on the path that connects them. 

The graphical representation of a decomposable  model
corresponds to an undirected chordal graph whose set of maximal
cliques defines the joint probability distribution of the model.   
A graph is chordal if every cycle of length four or more has a
shortcut, i.e., a chord. A maximal clique is the largest set of 
nodes that are completely connected, i.e., dependent.   

In general, parameter estimates are based on {\it sufficient statistics}. 
These provide all the information from the data sample that is needed
to estimate the value of a parameter. The sufficient statistics of the
parameters of a decomposable model are the {\it marginal frequencies} of
the events represented by the feature variables that form maximal cliques 
in the graphical representation. Each maximal clique is made up of a 
subset of the feature variables that are all dependent. Together these 
features  define a {\it marginal event space}. The probability of 
observing any specific instantiation of these features, i.e., a {\it
marginal event}, is defined by the {\it marginal probability 
distribution}. 

The joint probability distribution of a decomposable model is expressed as 
the product of the  marginal distributions of the  variables in the 
maximal cliques of the graphical representation, scaled by the marginal 
probability distributions of feature variables common to two or more
of these maximal sets.   Because their joint distributions have such
closed--form expressions, the  parameters of a decomposable model  can
be estimated directly from the data sample without the need   for an
iterative fitting procedure as is required, for example, to  estimate
the parameters of maximum entropy models (e.g.,  \cite{BergerDD96}). 

\newpage
\subsection{Examples}

To clarify these concepts, both the graphical representation and
parameter estimates associated with several examples of decomposable
models are presented in terms of a
simple word sense disambiguation example. The task is to
disambiguate various instances of {\it bill} by selecting one of two 
possible senses; a piece of pending legislation or a statement requesting
payment for services rendered.  

Each sentence containing {\it bill} is represented using five binary
feature variables. The classification variable $S$ represents the
sense of {\it bill}. Four contextual feature variables indicate
whether or not a given word has occurred in the sentence with the
ambiguous use of {\it bill}. The presence or absence of {\it
Congress,} {\it veto}, {\it restaurant} and {\it tip}, are represented
by binary variables $C, V, R$ and $T$, respectively. These variables
have a value of $yes$ if the word occurs in the sentence and $no$ if
it does not.  

\begin{table}
\begin{center}
\caption{Sense--tagged text for {\it bill}} \label{tab:inventedwsd}
\begin{tabular}{|l||ccccc|} \hline
\multicolumn{1}{|c||}{Sense--tagged sentences} &
\multicolumn{5}{c|}{Feature vectors} \\ \cline{2-6}
 & C & V & R & T & S \\ \hline
I paid the bill/pay at the restaurant. & no & no & yes & no & pay \\ \hline
Congress overrode the veto of that bill/law. & yes & yes & no & no & law\\ \hline
Congress passed a new bill/law today. & yes & no & no & no & law \\ \hline
The restaurant bill/pay does not include the tip. & no & no & yes & yes &pay\\ \hline
The bill/law was killed in committee.& no & no & no & no & law\\ \hline
\end{tabular} 
\end{center}
\myendtab

A sample of $N$ sentences that contain {\it bill} is collected. The
instances of {\it bill} are manually annotated with sense values by a
human judge. These sense--tagged sentences are converted by a {\it feature 
extractor} into the feature vectors shown in Table 
\ref{tab:inventedwsd}.           

Given five binary feature variables,  there are $32$ possible events
in the event space. If the parametric form is the saturated model then
there are also $32$ parameters to estimate.  For this example the
saturated model is notated $(CVRTS)$ and its graphical representation
is shown in Figure \ref{fig:satur}. This  model is decomposable as
there is a path of length one between any two feature variables in the
graphical representation.   

\begin{figure}
\centerline{\epsfbox{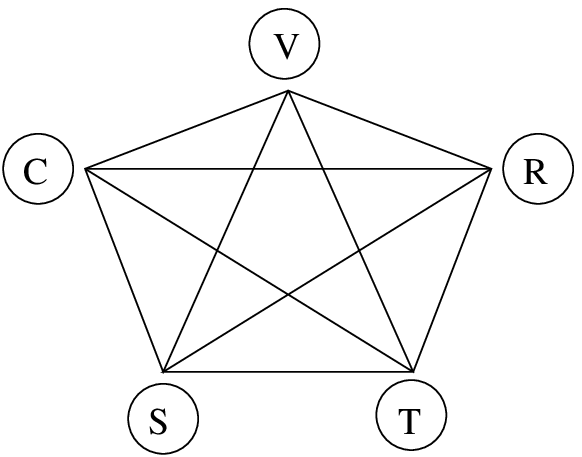}}
\caption{Saturated Model $(CVRTS)$}
\label{fig:satur}
\myendfig

In order to estimate values for all the
parameters of the saturated model, every possible event must be
observed in the sample data.  Let the parameter estimate
$\hat\theta_{f_1,f_2,\ldots,f_{n-1},s}^{F_1,F_2,\ldots,F_{n-1},S}$
represent the probability that a feature vector
($f_1,f_2,\ldots,f_{n-1},s$) is observed in the data sample where each
sentence is represented by the random variables $(F_1,F_2,\ldots,
F_{n-1},S)$. The parameter estimates of the saturated model are
calculated as follows:
\begin{equation}
\hat\theta_{c,v,r,t,s}^{C,V,R,T,S} \ \ = \ \ \hat p(c,v,r,t,s) 
\ \ = \ \ \frac{freq(c,v,r,t,s)}{N}
\end{equation}

However, an alternative to the saturated model is to use the model
selection process described in Chapter 3 to find a
more parsimonious probabilistic model that contains only the most
important dependencies among the feature variables. This model can then 
be used as a classifier to disambiguate subsequent occurrences of the
ambiguous word.  

Suppose that the model selection process finds that  the model
$(CSV)(RST)$, shown in Figure \ref{fig:decomp}, is an adequate
characterization of the data sample. There are a number of properties
of the model revealed in the graphical representation. First, it is a
decomposable model since all cycles of length four or more have a
chord. Second, conditional independence relationships can be read off
the graph.  For example, the values of features $R$ and $V$  are
conditionally independent given the value of $S$;  $p(r| v, s)  =
p(r|s)$, or $p(v|r,s) = p(v|s)$.   Third, $(CSV)$ and $(RST)$ are the
maximal cliques. The variables in each clique are all dependent and
each clique defines a marginal distribution.  Each marginal
distribution defines a marginal event space with eight possible
events. Thus the total number of parameters  needed to define the
joint probability distribution reduces from 32 to 16 when using this
model rather than the saturated model.   

\begin{figure}
\centerline{\epsfbox{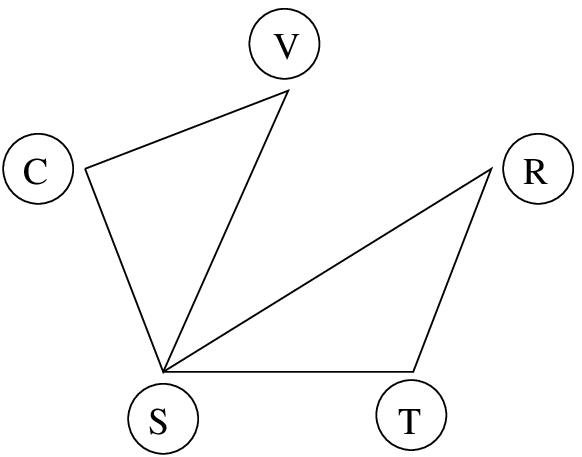}}
\caption{Decomposable Model $(CSV)(RST)$}
\label{fig:decomp}
\myendfig

The maximum likelihood estimates for the parameters of the joint
probability distribution are expressed in terms of the parameters of
the decomposable model.   The sufficient statistics of a decomposable
model are the marginal frequencies of the variables represented in the
maximal cliques of the graphical representation. Given the parametric
form $(CSV)(RST)$, the sufficient statistics are the marginal
frequencies $freq(c,s,v)$ and $freq(r,s,t)$. The parameters  of the
decomposable model are $\hat\theta_{c,s,v}^{C,S,V}$ and
$\hat\theta_{r,s,t}^{R,S,T}$.  These represent the probability that
the marginal events $(c,s,v)$ and  $(r,s,t)$ will be observed in a
data  sample. These estimates are made by normalizing the marginal
frequencies by the sample size $N$:     
\begin{equation}
\hat\theta_{c,s,v}^{CSV} \ \  = \ \ \hat p(c,s,v) = \ \
\frac{freq(c,s,v)}{N} 
\end{equation}
and
\begin{equation}
\hat\theta_{r,s,t}^{RST} \ \  = \ \ \hat p(r,s,t) = \ \
\frac{freq(r,s,t)}{N}
\end{equation}

Each  parameter of the joint probability distribution can be expressed in
terms of these decomposable model parameters. The joint probability of
observing the event $(c,v,r,t,s)$ is expressed as the product
of the marginal probabilities of observing marginal events $(c,s,v)$ and
$(r,s,t)$:  

\begin{equation}
\hat\theta_{c,v,r,t,s}^{CVRTS} = \frac{\hat\theta_{c,s,v}^{CSV} \times
\hat\theta_{r,s,t}^{RST}}
{\hat\theta_s^S}
\label{eq:mle1}
\end{equation}

While the denominator $\hat\theta_s^S$ represents an estimate of a
marginal distribution, it is not technically a parameter since it is
completely determined by the numerator. The denominator does not add
any new information to the model, it simply factors out any marginal
distributions that occur in more than one of the marginal
distributions found in the numerator. 

In contrast to the saturated model, the model of independence assumes
that there are no dependencies among any of the feature variables. For
this example the model of independence is notated $(C)(V)(R)(T)(S)$
and the graphical representation is shown in Figure \ref{fig:indep}.
This model has five maximal cliques, each containing one node and no
dependencies. This defines five marginal distributions, each of which has
two possible values. The number of parameters needed to define the
joint probability distribution is reduced to 10.  These parameters are
estimated as follows: 
\begin{equation}
\hat\theta_{c,v,r,t,s}^{C,V,R,T,S} = \hat\theta_c^C \times
\hat\theta_v^V \times \hat\theta_r^R \times \hat\theta_t^T \times \hat\theta_s^S  
\end{equation}

\begin{figure}
\centerline{\epsfbox{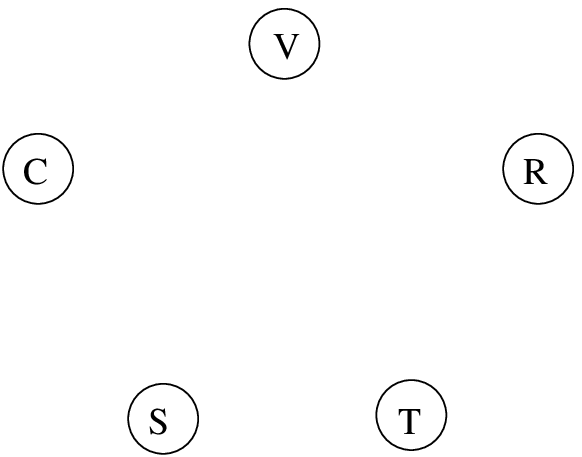}}
\caption{Model of Independence $(C)(V)(R)(T)(S)$}
\label{fig:indep}
\myendfig

This model indicates that the probability of observing a particular
value for a feature variable is not influenced by the values of any of
the other feature variables. No features affect the values of any
other features. The model of independence is trivially decomposable as
there are no cycles in the graphical representation of the
model. Despite its simplicity, the model of independence is used throughout
the experimental evaluation described in Chapter 6.  It serves as the
basis of the {\it majority classifier}, a probabilistic model that
assigns the most frequent sense of an ambiguous word in a sample
of data to every instance of the ambiguous word it subsequently
encounters.  

The Naive Bayesian classifier \cite{DudaH73} also plays a role later
in this dissertation. This is a decomposable model that has a
significant history in natural language processing and a range of
other applications. This  model assumes that all of the contextual
features are conditionally independent given the value  of the
classification variable. 

\begin{figure}
\centerline{\epsfbox{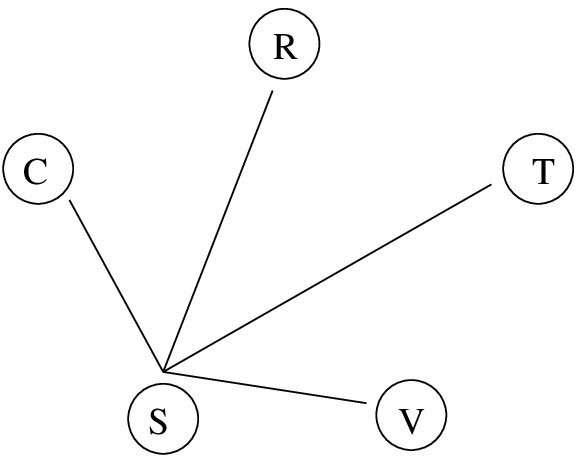}}
\caption{Naive Bayes Model $(C S)(R S)(T S)(V S)$}
\label{fig:naivebayes}
\myendfig

For the example in this chapter,  the
parametric form of Naive Bayes is notated $(C S)(R S)(T S)(V S)$ and
has a graphical representation as shown in Figure \ref{fig:naivebayes}.   
In this model there are four maximal cliques, each with two nodes and one
dependency. The variables are binary so each of the four marginal
distributions represents four possible events. The parameter estimates 
for Naive Bayes are computed as follows: 
\begin{equation}
\hat\theta_{c,v,r,t,s}^{C,V,R,T,S} = \frac{\hat\theta_{c,s}^{C,S}
\times \hat\theta_{v,s}^{V,S} \times \hat\theta_{r,s}^{R,S} \times
\hat\theta_{t,s}^{T,S}}
{\hat\theta_{s}^{S} \times \hat\theta_{s}^{S} \times \hat\theta_{s}^{S}}  
\end{equation}
By applying the following identities, 
\begin{equation} 
\hat\theta_{x,s}^{X,S} \ \ = \ \ \hat p(x,s) \ \  = \ \  \hat p(x|s) 
\times \hat p(s) \ \
\ \ and \ \ \ \ \ \hat\theta_{s}^{S} \ \ = \ \ \hat p(s) 
\end{equation}
the Naive Bayesian classifier can also be expressed in its more
traditional representation:
\begin{equation}
\hat p(c,v,r,t,s) = \hat p(s) \times \hat p(c|s) \times \hat p(v|s) \times
\hat p(r|s) \times \hat p(t|s)    
\end{equation}

\subsection{Decomposable Models as Classifiers}
\label{sec:classification}

A probabilistic model consists of a parametric form that describes the
dependencies among the features and parameter estimates that tell how
likely each possible event is to occur. Such a model can be used as a
classifier to identify the most probable sense of an ambiguous word
given the context in which it appears. 

For example, suppose a sentence that contains an ambiguous word is 
represented by the following feature vector: 
\begin{equation}
(C = c, V=v, R=r, T=t, S = ?)
\label{eq:fv1}
\end{equation}

The variable $S$ represents the sense of an  ambiguous word.  Variables 
$C$, $V$, $R$, and $T$ are the features that represent the context in 
which the ambiguous word occurs. The values of  the feature variables are 
known while the value of $S$ is unknown.  Given $x$ possible values of 
$S$, where each possible sense is notated $s_x$,  there are $x$ possible 
events associated with the incomplete feature  vector in Equation 
\ref{eq:fv1}. A probabilistic classifier determines  which of  these 
possible events has the highest associated probability  according to the
probabilistic model, i.e., it maximizes the probability of $S$
conditioned on the values of the observed feature variables. 

Thus, disambiguation is performed via a simple maximization function. 
Given values for the observed contextual features, a probabilistic 
classifier determines which value of $S$ is associated with the most 
probable event: 

\begin{equation}
S = \stackrel{argmax}{s_x} p(s_x |c,v,r,t) 
 = \stackrel{argmax}{s_x} 
\frac{p(c,v,r,t,s_x)}{p(c,v,r,t)}
\label{eq:classify1}
\end{equation}

The denominator in Equation \ref{eq:classify1} acts as a constant 
since it does not 
include $S$. As such it can be dropped and the maximization operation 
is simplified to finding the value of $S$ that  maximizes the joint 
probability distribution of the feature variables $C, V, R, T,$ and $S$:  
\begin{equation}
S = \stackrel{argmax}{s_x} p(c,v,r,t,s_x) 
= \stackrel{argmax}{s_x} \hat\theta_{c,v,r,t,s_x}^{C,V,R,T,S} 
\end{equation}

This chapter has shown how estimates of the joint probability
distribution  can be expressed in terms of more parsimonious decomposable
models. The following chapter shows how decomposable models can be 
automatically selected from sense--tagged text.


\chapter{SUPERVISED LEARNING FROM SENSE--TAGGED TEXT}

When applied to classification problems, supervised learning is a
methodology where examples of properly classified events 
are used to train an algorithm that will classify subsequent instances
of similar events. For word sense disambiguation, manually disambiguated
usages of an ambiguous word serve as training examples. This
sense--tagged text is used to learn a probabilistic model that
determines the most probable sense of an ambiguous word, given the
context in which it occurs.  

In this dissertation, the objective of supervised learning is to
select the parametric form of a decomposable model that represents the
important dependencies among a set of feature variables exhibited in a
particular sample of sense--tagged text. Given this
form, the joint probability distribution of this set of feature
variables can be expressed in terms of the marginal probability
distributions of the decomposable model. Once the values of these
parameters are estimated, the probabilistic model is complete and
can be used as a classifier to perform disambiguation.  

The challenge in learning probabilistic models is to locate a
parametric form that is both a specific representation of the
important dependencies in the training sample and yet general enough
to successfully disambiguate previously unobserved instances of the
ambiguous word.  A model is too complex if a substantial number of
parameters in the joint probability distribution have zero--valued
estimates; this indicates that the available data sample simply does
not contain enough information to support the estimates required by
the model. However, a model is too simple if relevant dependencies
among features are not represented. In  other words, the model should 
achieve an appropriate balance between model complexity and model fit.        

There are several supervised learning methodologies discussed in this
chapter. {\it Sequential  model selection} learns a single parametric
form that is judged to achieve the best balance between model
complexity and fit for a given sample of sense--tagged text.  This
methodology is extended by the {\it Naive Mix}, which learns an
averaged probabilistic model from the sequence of parametric forms
generated during a sequential model selection process. An alternative
to learning the parametric form is to simply assume one. In this case
no search of parametric forms is conducted; the form is specified by
the user and the sense--tagged text is only utilized to make parameter
estimates. This is the methodology of the {\it Naive Bayesian
classifier} \cite{DudaH73}, often simply referred to as Naive Bayes.

The degree to which a probabilistic  model successfully balances
complexity and fit is determined by its accuracy in disambiguating
previously unobserved  instances of an ambiguous word. The methods
discussed in this chapter are subjected to such an evaluation in
Chapter 6.  

\section{Sequential Model Selection}

Sequential model selection integrates a {\it search strategy} and an
{\it evaluation criterion}.  Since the number of possible parametric
forms is exponential in the number of features,  an exhaustive search
of the possible forms is usually not tractable.  A search strategy
determines which parametric forms, from the set of all possible
parametric forms, will be considered during the model selection
process. The evaluation criterion is the ultimate judge of which
parametric form achieves the most appropriate balance between
complexity and fit, where complexity is defined by the number of
dependencies in the model, i.e., the number of edges in its graphical 
representation.  

The search strategies employed here are greedy and result in the
evaluation of models of steadily increasing or decreasing levels of
complexity. A number of {\it candidate models} are generated at each
level of complexity.  The evaluation criterion determines which
candidate model results in the best fit to the training sample; this
model is designated as the {\it current model}. Another set of
candidate models is generated by increasing or decreasing the
complexity of the current model by one dependency.  The process of
evaluating candidates, selecting a current model, and generating new
candidate models from the current model is iterative and continues
until a model is found that achieves the best overall balance of
complexity and fit. This is the {\it selected model} and is the
ultimate result of the sequential model selection process. 

A selected model is parsimonious in that it has as few dependencies
as are necessary to characterize or fit the training sample. When
using maximum likelihood estimates, the saturated model exactly fits
the distribution of the observed events in the training sample.
However, the number of parameters is equal to the number of events in
the event space and obtaining non--zero estimates for large numbers of
parameters is usually difficult.  A parsimonious model should capture
the important dependencies among the features in a training sample and
yet allow the joint probability distribution to be expressed
relatively simply in terms of a smaller number of decomposable model
parameters.       

As formulated in this dissertation, the model selection process also
performs feature selection.  If a model is selected where there is no
dependency between a feature variable and the sense
variable, then that feature is removed from the model and
will not impact disambiguation. 

\subsection{Search Strategy}

Two sequential search strategies are employed in this dissertation:
{\it backward sequential search} \cite{Wermuth76} and {\it forward
sequential search} \cite{Dempster72}. These methods are also known as
{\it backward elimination} and {\it forward inclusion}. 
   
Backward sequential search for  probabilistic models of word sense
disambiguation was  introduced by Bruce and Wiebe (e.g.,
\cite{Bruce95},  \cite{BruceW94A}, \cite{BruceW94B}).  This
dissertation introduces forward sequential search. Forward searches
evaluate models of increasing  complexity based on how much candidate
models improve upon the fit of the current model, while backward searches 
evaluate candidate models based on how much they degrade the fit of the 
current model.   

A forward sequential search begins by designating the model of
independence as the current model. The level of complexity is
zero since there are no edges in the graphical representation of this
model.  The set of candidate models is generated from the model of
independence and consists of all possible one edge decomposable
models. These are individually evaluated for fit by an evaluation
criterion. The one edge model that exhibits the greatest improvement
in fit over the model of independence is designated as the new current
model. A new set of candidate models is generated by adding an  edge to
the current model and consists of all possible two edge decomposable
models. These models are evaluated for fit and the two
edge decomposable model that most improves on the fit of the one edge
current model becomes the new current model. A new set of three edge
candidate models is generated by adding one  edge at a time to the
two edge current model. This sequential search continues until:   
\begin{enumerate}
\item none of the candidate 
decomposable models of complexity level $i+1$ results in an
appreciable improvement in fit over the current model of complexity
level $i$, as defined by the evaluation criterion, or
\item the current model is the saturated model.
\end{enumerate}
In either case the current model becomes the selected model and the
search ends. 

In general then, during a forward search the current model is reset to
the decomposable model of complexity level $i$ that most improves on the
fit of the current decomposable model of complexity level $i-1$. 
All possible decomposable models of complexity level $i+1$ that
are generated from the current model of complexity level $i$ are
considered as candidate models and then evaluated for fit. The candidate model 
that most improves on the fit of the current model of complexity level $i$ 
is designated the new current model. This process continues until
either there is no decomposable model of complexity level $i+1$ that
results in an appreciable improvement in  fit over the current model
or the current  model of complexity level  $i$ is the saturated
model. In either case the current model is selected and the search
ends.    

For the sparse and skewed samples typical of natural language
data \cite{Zipf35}, forward sequential search is a natural
choice. Early in the search the models are of low complexity and the
number of parameters in the model is relatively small. This results in
few zero--valued estimates  and ensures that the model selection
process is based upon the best available information from the training
sample.  

A backwards sequential search begins by designating the
saturated model as the current model. If there are $n$ feature
variables then the number of edges in the saturated model is
$\frac{n(n-1)}{2}$. As an example, given 10 feature variables there
are 45 edges in a saturated model. The set of candidate models
consists of each possible decomposable model with 44 edges 
generated by removing a single edge from the saturated model. These
candidates are evaluated for fit  and the 44 edge model that results
in the least degradation in fit from the saturated model becomes the
new current model. Each possible 43 edge candidate decomposable model
is generated by removing a single edge from the 44 edge current model
and then evaluated for fit.  The 43 edge decomposable candidate
model that results in the least degradation in fit from the 44 edge
current model becomes the new current model. Each possible 42 edge
candidate decomposable  model is generated by removing a single edge from 
the current 43 edge model and then evaluated for fit.  This sequential
search  continues until:  
\begin{enumerate}
\item every candidate decomposable model of complexity level $i-1$
results in an appreciable degradation in fit from the current model of
complexity level $i$,  as defined by the evaluation criterion, or
\item the current model is the model of independence. 
\end{enumerate}
In either case the current model is selected and the search ends.       

In general then, during a backward search the current model is reset
to the decomposable model of complexity level $i$ that results in the
least degradation in fit from the current model of complexity level
$i+1$.  Each possible decomposable model of complexity level $i-1$ is
generated by removing a single edge from the current model of
complexity level $i$ and evaluated for fit. This process continues until 
either every decomposable model of complexity level $i-1$ results in an 
appreciable degradation in fit from the current model of complexity level 
$i$ or the current model has complexity level zero, i.e., the model of 
independence. In either  case the current model is selected and the search 
ends.

The backward search in this dissertation differs slightly from that of
Bruce and Wiebe. Their backward search begins with the saturated model
and generates a series of models of steadily decreasingly complexity
where the minimal or concluding model is Naive Bayes. All models in
this sequence are evaluated via a test of predictive accuracy; the
model that achieves the best balance between complexity and fit is the
model that achieves the highest disambiguation accuracy.  Here,
backward sequential search begins with the saturated model and
generates a series of models that concludes with the one that best
balances complexity and fit, as judged by an evaluation criterion. If
no such model is found then the model of independence is the
concluding model in the sequence.       

For sparse and skewed data samples, backward sequential search should
be used with care. Backward search begins with the saturated model
where the number of parameters equals the number of events in the event
space. Early in the search the models are of high
complexity. Parameter estimates based on the saturated model or other
complex models are often unreliable since many of the marginal events
required to make maximum likelihood estimates are not observed in the
training sample.   

\subsection{Evaluation Criteria}

The degradation and improvement in fit of candidate models relative to
the current model is assessed by an evaluation criterion. Two
different varieties of evaluation criteria  are employed in this
dissertation;  significance tests and  information  criteria. 

The use of significance tests as evaluation criteria during sequential
searches for probabilistic models of word sense disambiguation was
introduced by Bruce and Wiebe (e.g., \cite{Bruce95}, \cite{BruceW94A},
\cite{BruceW94B}). They employ the log--likelihood ratio $G^2$ and
assign significance values to this test statistic using an 
asymptotic distribution and an exact conditional distribution. 
This dissertation expands the range of evaluation criteria available
for model selection by introducing two information criteria, Akaike's
Information Criterion (AIC)  \cite{Akaike74} and the Bayesian
Information Criterion (BIC) \cite{Schwarz78}.  

\subsubsection{Significance Testing}

In significance testing, a model is a hypothesized representation of
the population from which a training sample was drawn. The adequacy of
this model is evaluated via a test statistic that measures the fit of
the model to the training sample.\footnote{When using maximum likelihood
estimates, the training sample is exactly characterized by the saturated
model. Thus the fit of the hypothesized model to the
training sample is assessed by measuring the fit of the model to the
saturated model.} The fit of the hypothesized model is judged 
acceptable if it differs from the training sample by an amount that is
consistent with sampling error, where that error is defined by the
distribution of the test statistic.  

The log--likelihood ratio $G^2$ is a frequently used test statistic:
\begin{equation}
G^2  = 2 \times \sum_{i=1}^q f_i \times log \frac{f_i}{e_i}
\label{eq:g2}
\end{equation}
where $f_i$ and $e_i$ are the observed and expected counts of the
$i^{th}$ feature vector. The observed count $f_i$ is calculated
directly from the training sample  while the expected count $e_i$ is
calculated assuming that the model under evaluation fits the sample,
i.e., that the null hypothesis is true. This statistic measures the
deviation between what is observed in the training sample and what
would be expected in that sample if the hypothesized model is an
accurate representation of the population.  

The distribution of $G^2$ is asymptotically approximated by the
$\chi^2$ distribution \cite{Wilks38} with adjusted degrees of freedom
(dof) equal to the number of parameters that have non--zero estimates
given the data in the sample. The degrees of freedom are adjusted to
remove those parameters in the hypothesized model that can not be
estimated from the training sample.  These are parameters whose
sufficient statistics have a value of zero  since the marginal events
they are associated with do not occur in the training  sample.   
The statistical significance of a model is equal to the probability of
observing its associated $G^2$ value in the $\chi^2$ distribution with
appropriate degrees of freedom. If this probability is less than a
pre--defined  cutoff value, $\alpha$, then the deviance of the
hypothesized model from the training sample is less than would be
expected due to sampling error. This suggests that the hypothesized
model is a reasonable representation of the population from which the
training sample was taken.  

During backward sequential search a significance test determines if a
candidate model results in a significantly worse fit than the current
model. During forward search a significance test determines if a
candidate model results in a significantly better fit than the current
model.  This is a different formulation than the significance
test described above, where the hypothesized or candidate model is
always fitted to the saturated model. This dissertation treats
sequential search as a series of local evaluations, where the fit of
candidate models is made relative to current models that have one more
or one less dependency, depending on the direction of the search. This
is in contrast to a global evaluation where the fit of candidate
models is always relative to the saturated model or some other fixed
model. 

The degree to which a candidate model improves upon or degrades the
fit of the current model is measured by the difference between the
$G^2$ values for the candidate and current model,  $\Delta G^2$.  Like
$G^2$, the distribution of $\Delta G^2$ is approximated by a $\chi^2$
distribution with adjusted degrees of freedom equal to the difference
in  the adjusted degrees of freedom of the candidate and current
model, $\Delta dof$ \cite{BishopFH75}.   

During backward search a candidate model does not result in a
significant degradation in fit from the current model if the
probability, i.e., significance, of its $\Delta G^2$ value is above
a pre--determined cutoff, $\alpha$, that defines the allowable
sampling error.  This error is defined by the asymptotic distribution
of $\Delta G^2$, which is in turn defined by the $\chi^2$ distribution
with degrees of freedom equal to $\Delta dof$.  If the error
is small then the candidate model is an adequate representations
of the population. 

A candidate model of complexity level $i-1$ inevitably results in a
degradation in fit from the current model of complexity level $i$. The
objective of backward search is to select the candidate model that
results in the least degradation in fit from the current model. Thus,
the candidate model of complexity level $i-1$ with the lowest
significance value less than $\alpha$ is selected as the current model
of complexity level $i-1$. The degradation in fit is judged acceptable
if the value of $\Delta G^2$ is statistically insignificant, according
to a $\chi^2$ distribution with degrees of freedom equal to $\Delta
dof$.  If the significance of $\Delta G^2$ is unacceptably large for
all candidate models the selection process stops and the current model
becomes the ultimately selected model. 

During forward search the candidate model has one more edge than the
current model. A candidate model of complexity level $i+1$ inevitably
improves upon the fit of the current model of complexity level $i$. 
The objective of forward search is  to select the candidate model that
results in the greatest increase in fit from the current model. The
candidate model of complexity level $i+1$  with the largest
significance value greater than $\alpha$ is selected as the current
model of complexity level $i+1$. This is the model that results in the
largest improvement in fit when moving from a model of complexity
level $i$ to one of $i+1$. The improvement in fit is judged acceptable
if a significance test shows that the value of  $\Delta G^2$ is
statistically significant. If all candidate models result in
insignificant levels of improvement in fit then model selection stops
and selects the current model of complexity level $i$.    

While it is standard to use a $\chi^2$ distribution to assess the
significance of  $G^2$ or $\Delta G^2$, it is known that this
approximation may not hold when the data is sparse and skewed
\cite{ReadC88}. An alternative to using an asymptotic approximation to
the distribution of test statistics such as $G^2$ and $\Delta G^2$ is
to define their exact distribution. There are two ways to define the
exact distribution of a test statistic:   
\begin{enumerate}
\item enumerate all elements of that distribution as in Fisher's Exact 
Test \cite{Fisher35} or 
\item sample from that distribution using a Monte Carlo sampling scheme 
\cite{Ripley87}.  
\end{enumerate}

The significance of $G^2$ and $\Delta G^2$ based on the exact
conditional distribution does not rely on an asymptotic approximation
and is accurate for sparse and skewed data samples. Sequential model
selection using the exact conditional test is developed for word sense
disambiguation in \cite{Bruce95}. The exact conditional test is also
applied to the identification of significant lexical relationships in
\cite{PedersenKB96}.   

This dissertation employs sequential model selection using both the
asymptotic approximation of the significance of $G^2$ values as well
as the exact conditional distribution. The forward and backward
sequential search procedures remain the same for both methods;
the distinction is in how significance is assigned. The asymptotic
assumption results in the assignment of significance values from a
$\chi^2$ distribution while the exact conditional test assigns
significance based upon a Monte Carlo sampling scheme.\footnote{The
freely available software package CoCo  \cite{Badsberg95} implements
the Monte Carlo sampling scheme described in  \cite{Kreiner87}.} 

\subsubsection{Information Criteria}

Two information criteria are employed as evaluation criteria in this
dissertation; Akaike's Information Criteria (AIC)  and the Bayesian
Information Criteria (BIC).  These criteria are formulated as follows
for use during sequential model selection:

\begin{equation}
AIC = \Delta G^2 - 2 \times \Delta dof
\label{eq:aic}
\end{equation}
\begin{equation}
BIC = \Delta G^2 - log(N) \times \Delta dof
\label{eq:bic}
\end{equation}
where $\Delta G^2$ again measures the deviation in fit
between the candidate model and the current model.  However, here
$\Delta G^2$ is treated as a raw score and not assigned significance. 
$\Delta{dof}$ represents the difference between the adjusted degrees
of freedom for the current and candidate models. Like $\Delta G^2$, it
is treated as a raw score and is not used to assign significance. In
Equation \ref{eq:bic}, $N$ represents the number of observations in
the training sample.  

The information criteria are alternatives to using a pre--defined
significance level, $\alpha$, to judge the acceptability of a model.
AIC and BIC explicitly balance model fit and complexity; fit is
determined by the value of $\Delta G^2$ while complexity is expressed
in terms of the difference in the adjusted degrees of freedom of the
two models, $\Delta dof$. Small values of $\Delta G^2$ imply that the
fit of the candidate model to the training data does not deviate
greatly from the fit obtained by the current model. Likewise, small
values for the adjusted degrees of freedom, $\Delta dof$, suggest that
the candidate and current models do not differ greatly in regards to
complexity.  

During backward search the candidate model with the lowest negative
AIC value is selected as the current model of complexity level
$i-1$. This is the  model that results in the least degradation in fit
when moving from a model of complexity level $i$ to one of $i-1$. This
degradation is judged acceptable if the AIC value for the candidate
model of complexity level $i-1$ is negative. If there are no such
candidate models then the degradation in fit is unacceptably large and
model selection stops and the current model of complexity level $i$
becomes the selected model.  

During forward search the candidate model with the largest positive
AIC value is selected as the current model of complexity level
$i+1$. This is the model that results in the largest improvement in
fit when moving from a model of complexity level $i$ to one of
$i+1$. This improvement is judged acceptable if the AIC value for the
model of complexity level $i+1$ is positive. If there are no such
models then the improvement in fit is unacceptably small and model
selection stops and the current model of complexity level $i$ becomes
the selected model. 

The information criteria have a number of appealing properties that
make them particularly well suited for sequential model
selection. First, they do not require that a pre--determined cutoff
point be specified to stop the model selection process; a mechanism  
to stop model selection is inherent in the formulation of the
statistic. Second, the balance between model complexity and fit is
explicit in the statistic and can be directly controlled by adjusting
the constant that precedes $\Delta dof$. As this value increases the
selection process results in models of decreasing
complexity.\footnote{In general, BIC selects models of lower
complexity than does AIC. This is discussed further in Chapter 6.} 

\subsection{Examples}

For clarity, the sequential model selection process is illustrated
with two simple examples; one using forward search in combination with
AIC and the other using backward search and AIC. These methodologies
are abbreviated as FSS AIC and BSS AIC, respectively. 
Both examples learn a parametric form from the 24 observation training
sample shown in Table \ref{tab:dataxx}, where the feature set consists
of three binary variables,  $A$, $B$, and $C$. There are eight
possible events in the event space. The frequency with which each
event occurs in the sample is shown by $freq(A,B,C)$. 

\begin{table}
\begin{center}
\caption{Model Selection Example Data}
\begin{tabular}{|ccc|c|} \hline
A & B & C & $freq(A,B,C)$ \\ \hline
0 & 0 & 0 & 0 \\
0 & 0 & 1 & 1 \\
0 & 1 & 0 & 5 \\
0 & 1 & 1 & 12 \\
1 & 0 & 0 & 0 \\
1 & 0 & 1 & 3 \\
1 & 1 & 0 & 2 \\
1 & 1 & 1 & 1 \\ \hline
\end{tabular} 
\label{tab:dataxx}
\end{center}
\myendtab

\subsubsection{FSS AIC}

During forward search, the candidate models are evaluated relative to
how much they improve upon the fit of the current model. Such an
improvement is expected since the candidate model has one more
dependency than the current model.  

The value of $\Delta G^2$ measures the amount of deviance between the
candidate model and the current model; a large value implies that the
candidate model greatly increases the fit of the model.  
Only candidate models that have positive AIC values improve upon the
fit of the current model sufficiently to merit designation as the new
current model. A negative value for AIC during forward search
indicates that the increase in fit is outweighed by the resulting
increase in complexity and will not result in a model that attains an
appropriate balance of complexity and fit. 

The steps in sequential model selection using FSS AIC are shown
in Table \ref{tab:fssexample}. The $G^2$ values for the current and
candidate models are shown, as is their difference, $\Delta G^2$. 
The steps of the sequential search using forward  search and AIC are
shown in Table \ref{tab:fssexample}. The value of $\Delta G^2$ 
measures the improvement in the fit when a dependency is added to the
current model. During forward search, $\Delta G^2$ is calculated by
subtracting the $G^2$ value associated with the candidate model from
the $G^2$ associated with the current model:
\begin{equation}
\Delta G^2 = G^2_{current} - G^2_{candidate}
\end{equation}
This difference shows the degree to which the candidate model improves
upon the fit of the current model. A large value of $\Delta G^2$ shows
that the fit of the candidate model to the training sample is
considerably better than that of the current model. 

The degree to which complexity is increased by the addition of a
dependency to the candidate model is measured by the difference in the
adjusted degrees of freedom for the two models, $\Delta dof$:   
\begin{equation}
\Delta dof = dof_{candidate} - dof_{current}
\end{equation}

\begin{table}
\begin{center}
\caption{Model Selection Example: FSS AIC}
\begin{tabular}{|c|lc|lc|ccc|} \hline
& Current & $G^2$ & Candidate  & $G^2$ & $\Delta G^2$ & $\Delta dof$ & AIC \\ \hline  
Step 1 & (A)(B)(C) & 10.14 & (AC)(B)  & 10.08  &  0.06  & 1 & -1.94 \\  
  & (A)(B)(C) & 10.14 & (A)(BC)  &  7.07  &  3.07  & 1 &  1.07 \\ 
  & (A)(B)(C) & 10.14 & (AB)(C)  &  4.56  &  5.58  & 1 &  3.58 \\ \hline 
Step 2 & (AB)(C) & 4.56 & (AB)(AC) &  4.50  & 0.06 & 1 & -1.94 \\  
  & (AB)(C) & 4.56 & (AB)(BC) &  1.48  &  3.08  & 1 & 1.08  \\ \hline 
Step 3 & (AB)(BC) & 1.48 & (ABC)  &  0.00  & 1.48  & 1 & -0.52
\\ \hline
\multicolumn{8}{|c|}{Selected: {\bf (AB)(BC)}} \\ \hline
\end{tabular} 
\label{tab:fssexample}
\end{center}
\myendtab

Step 1: Forward search begins with the model of independence,
$(A)(B)(C)$, as the current model. The set of one edge decomposable
candidate models is generated by adding an edge to the model of
independence.  The candidate models include $(AC)(B)$, $(A)(BC)$, and
$(AB)(C)$. These are all evaluated via AIC with the result that
$(AB)(C)$ has the greatest positive AIC value. This candidate model
exhibits the greatest deviance from the current model and therefore
most increases the fit. Thus, $(AB)(C)$ becomes the new current model.  

Step 2: A new set of candidate models is generated by adding an edge to
the current model, $(AB)(C)$. The candidate models are $(AB)(AC)$ and
$(AB)(BC)$. These are each evaluated relative to the current model
$(AB)(C)$. The model $(AB)(BC)$ has the greatest positive AIC value
associated with it and thus most increases the fit over the current
model. The new current model is now $(AB)(BC)$. 

Step 3: The set of candidate models is generated by adding an edge to
the current model $(AB)(BC)$. The only resulting candidate is 
$(ABC)$, the saturated model. However, when evaluated relative to the
current model it has a negative AIC value associated with it; this
suggests that the increase in fit is not sufficient to merit 
further increases in the complexity of the model. Thus, the current
model $(AB)(BC)$ becomes the selected model and is the ultimate
result of the selection process.    

\subsubsection{BSS AIC}

During backward search, candidate models are evaluated based upon how
much they degrade the fit of the current model. Since the candidate
models have one fewer dependency than the current model, it is
inevitable that there will be some degradation in fit. 

During backward search only candidate models that have negative AIC
values are eligible to be designated current models. A positive AIC
suggests that the degradation in model fit that occurs due to removal
of a dependency is too large and offsets the benefits of reducing the
complexity of the current model.  

The steps of the sequential search using backward search and AIC are
shown in Table \ref{tab:bssexample}. The value of $\Delta G^2$ 
measures the degradation in fit when a dependency is removed from the
current model: 
\begin{equation}
\Delta G^2 = G^2_{candidate} - G^2_{current}
\end{equation}
The degree to which complexity is decreased by the removal of a
dependency from the candidate model is shown by the difference in the
degrees of freedom for  the two models, $\Delta dof$:   
\begin{equation}
\Delta dof = dof_{current} - dof_{candidate}
\end{equation}

\begin{table}
\begin{center}
\caption{Model Selection Example: BSS AIC}
\begin{tabular}{|c|lc|lc|ccc|} \hline
& Candidate  & $G^2$  &Current & $G^2$ & $\Delta G^2$ &  $\Delta dof$ & AIC \\ \hline 
Step 1 & (AC)(BC) & 7.00   &  (ABC)    & 0.00  & 7.00 & 1 &  5.00 \\ 
& (AB)(AC)   & 4.49   &  (ABC)    & 0.00  & 4.49 & 1 &  2.49  \\ 
& (AB)(BC)   & 1.48   &  (ABC)    & 0.00  & 1.48 & 1 &  -0.52 \\ \hline 
Step 2 & (A)(BC)  & 7.07   &  (AB)(BC) & 1.48  &  5.59  & 1 &  3.59 \\  
& (AB)(C)    & 4.56   &  (AB)(BC) & 1.48  &  3.08  & 1 &  1.08 \\ \hline
\multicolumn{8}{|c|}{Selected: {\bf (AB)(BC)}} \\ \hline
\end{tabular} 
\label{tab:bssexample}
\end{center}
\myendtab

Step 1: A backward search begins with the saturated model, $(ABC)$, as
the current model. The set of candidate models consists of all two edge
models that are generated by removing a single edge from the saturated 
model. The models $(AC)(BC)$, $(AB)(AC)$, and $(AB)(BC)$ are
evaluated relative to the saturated model. $(AC)(BC)$ has the lowest
negative AIC value and becomes the current model.  

Step 2: The candidate models are all the one edge models generated by
removing a single edge from the current model. The models $(A)(BC)$
and $(AB)(C)$ are evaluated and both have positive AIC values. Both
result in a degradation in fit that is not offset by an appropriate
reduction in model complexity. Thus, model selection stops and  
the current model, $(AB)(BC)$, becomes the selected model.

\section{Naive Mix}
\label{sec:nmb}
\label{sec:naivemix}

This dissertation introduces the Naive Mix, a new supervised learning
algorithm that extends the sequential model selection  
methodology.
The usual objective of model selection is to find a single 
model that achieves the best representation of the training sample both in 
terms of complexity and fit.  However, empirical results described in 
Chapter 6 show that it is often the case that very different models can 
result in nearly identical levels of disambiguation accuracy. This 
suggests that there is an element of uncertainty in model selection and 
that a single best model may not always exist.  

The Naive Mix is based on the premise that each of the models that
serve as a current model during a sequential search have important
information that can be exploited for word sense disambiguation. 
The Naive Mix is an averaged probabilistic model based upon the average of 
the parameter estimates for all of the current models generated during a 
sequential model selection process. 

Sequential methods of model selection result in a sequence of 
decomposable models $(m_1,m_2$,$\ldots$, $m_{r-1}, 
m_r)$ where $m_1$ is the initial current model and  $m_r$ is the selected
model. Each model $m_i$ is designated as the current model at the
$i^{th}$ step in the search process. During forward search $m_1$ is
the model of independence and during backward search $m_1$ is the  
saturated model. 

Each model $m_i$ has a parametric form that expresses the dependencies
among the feature variables $(F_1, F_2,$ $\ldots$, $F_{n-1}, S)$,
where the sense of the ambiguous word is represented by $S$ and 
$(F_1, F_2$, $\ldots$,$F_{n-1})$ represent the feature variables.  
The joint probability distribution of this set of feature variables
can be expressed in terms of the marginal probability distributions
defined by each decomposable model $m_i$. 

Given a sequence of current models found during a sequential search,
the parameters of the joint probability distribution of the set of
feature variables are estimated based upon the marginal distributions
of each of these models. In other words, $r$ different estimates for
the joint probability distribution of a set of feature variables are
made. These $r$ estimates are averaged and the resulting joint
probability distribution is a Naive Mix:  

\begin{equation}
\hat\theta^{{(F_1,F_2,\ldots,F_{n-1},S)}_{average}} \ \ \ = \ \ \
        \frac{1}{r} \times \sum_{i=1}^r
\hat\theta^{{(F_1,F_2,\ldots,F_{n-1},S)}_{m_i}}
\end{equation}
where $\hat\theta^{{(F_1,F_2,\ldots,F_{n-1},S)}_{m_i}}$ represents the
parameter estimates given that the parametric form is $m_i$. 

A Naive Mix can be created using either forward or backward search. 
However, there are a number of advantages to formulating a Naive Mix
with a forward search. First, the inclusion of
very simple models in the mix eliminates the problem of zero--valued
parameter estimates in the averaged probabilistic model. The first
model in the Naive Mix is the model of independence which acts as a
majority classifier and has estimates associated with it for 
for every event in the event space. Second, forward
search incrementally builds on the strongest dependencies among
features while backward search incrementally removes the weakest
dependencies. Thus a Naive Mix formulated with backward search can
potentially contain many irrelevant dependencies while a forward
search only includes the most important dependencies.  

Consider an example that formulates a Naive Mix with a forward search;
this example follows the notation of the earlier {\it bill} example.  
Suppose that the sequence of  models shown in Table
\ref{tab:nmexample} are found to be the best fitting models by some
evaluation criterion at each step of the forward search.  These
represent the set of current models. 

Any marginal distributions of the current models that do not include
$S$, the sense variable, can be eliminated from the model included in the
mix. Such marginal distributions simply act as constants in a
probabilistic classifier and can be removed from the mix without
affecting the final result. In Table \ref{tab:nmexample}, the models
in the column {\it mixed model} are used to make the parameter
estimates of the joint probability distributions that are included in
the Naive Mix.  
 
\begin{table}
\begin{center}
\caption{Sequence of Models for Naive Mix created with FSS}
\begin{tabular}{|c|l|l|} \hline
      &  current model  & mixed model \\ \hline
$m_1$ & (C)(V)(R)(T)(S) & (S) \\
$m_2$ & (CS)(V)(R)(T)   & (CS) \\
$m_3$ & (CS)(ST)(V)(R)  & (CS)(ST) \\
$m_4$ & (CS)(ST)(SV)(R) & (CS)(ST)(SV) \\
$m_5$ & (CSV)(ST)(R)    & (CSV)(ST) \\
$m_6$ & (CSV)(ST)(TR)   & (CSV)(ST) \\
$m_7$ & (CSV)(RST)      & (CSV)(RST) \\ \hline
\end{tabular}
\label{tab:nmexample}
\end{center}
\end{table}

The parameters of the joint probability distribution of each
decomposable model $m_i$ are expressed as the product of the
marginal distributions of each current model. The parameters of the
joint distributions are averaged across all of the models to create
the Naive Mix. For example, the averaged parameter
$\theta_{c,v,r,t,s}^{(C,V,R,T,S)_{average}}$ is estimated as follows: 

\begin{eqnarray}
\theta_{c,v,r,t,s}^{{(C,V,R,T,S)_{average}}} \ \ \ = \ \ \  
\frac{1}{7} 
(\hat\theta_{c,v,r,t,s}^{{(C,V,R,T,S)_{m_1}}} +
\hat\theta_{c,v,r,t,s}^{{(C,V,R,T,S)_{m_2}}} + 
\hat\theta_{c,v,r,t,s}^{{(C,V,R,T,S)_{m_3}}} + \nonumber \\
\hat\theta_{c,v,r,t,s}^{{(C,V,R,T,S)_{m_4}}} +
\hat\theta_{c,v,r,t,s}^{{(C,V,R,T,S)_{m_5}}} +
\hat\theta_{c,v,r,t,s}^{{(C,V,R,T,S)_{m_6}}} +
\hat\theta_{c,v,r,t,s}^{{(C,V,R,T,S)_{m_7}}})
\end{eqnarray}
where 
\begin{eqnarray*}
\hat\theta_{c,v,r,t,s}^{{(C,V,R,T,S)_{m_1}}}  \ = \ 
\hat\theta_s^S \  ,  \ 
\hat\theta_{c,v,r,t,s}^{{(C,V,R,T,S)_{m_2}}}  \ = \
\hat\theta_{c,s}^{C,S} \ \ \ , \ \ \ 
\hat\theta_{c,v,r,t,s}^{{(C,V,R,T,S)_{m_3}}}  \ = \
\frac{\hat\theta_{c,s}^{C,S} \times
\hat\theta_{s,t}^{S,T}}{\hat\theta_s^S} 
\end{eqnarray*}

\begin{eqnarray*}
\theta_{c,v,r,t,s}^{{(C,V,R,T,S)_{m_4}}}  \ \ \ =  \ \ \ 
\frac{\hat\theta_{c,s}^{C,S} \times \hat\theta_{s,t}^{S,T} \times
\hat\theta_{s,V}^{S,V}}
{\hat\theta_s^S \times \hat\theta_s^S}  \ \ \ ,  \ \ \ 
\hat\theta_{c,v,r,t,s}^{{(C,V,R,T,S)_{m_5}}}  \ \ \ = \ \ \ 
\frac{\hat\theta_{c,s,v}^{C,S,V} \times
\hat\theta_{s,t}^{S,T}}{\hat\theta_s^S}  
\end{eqnarray*}

\begin{eqnarray*}
\hat\theta_{c,v,r,t,s}^{{(C,V,R,T,S)_{m_6}}} \ \ \ = \ \ \
\frac{\hat\theta_{c,s,v}^{C,S,V} \times
\hat\theta_{s,t}^{S,T}}{\hat\theta_s^S} \ \ \ , \ \ \ 
\theta_{c,v,r,t,s}^{{(C,V,R,T,S)_{m_7}}} \ \ \ = \ \ \
\frac{\hat\theta_{c,s,v}^{C,S,V} \times
\hat\theta_{r,s,t}^{R,S,T}}{\hat\theta_s^S} 
\end{eqnarray*}

Once the parameter estimates are made and averaged, the resulting
probabilistic model can be used as a classifier to perform
disambiguation. Suppose that the following feature vector represents a
sentence containing an ambiguous use of {\it bill}. $S$ represents the
sense of the ambiguous word and the other variables represent
observed features in the sentence:
\begin{equation}  
(C=c,V=v,R=r,T=t,S=?)
\end{equation}

The value of $S$ that maximizes
$\hat\theta_{c,v,r,t,s_x}^{{(C,V,R,T,S)_{average}}}$  is determined to 
be the 
sense of {\it bill}. Here again disambiguation reduces to finding the
value of $S$ that is most probable in a particular context as defined
by the observed values of the feature variables. 

\begin{equation}
S \ \ \ = \ \ \  \stackrel{argmax}{s_x}
\hat\theta_{c,v,r,t,s_x}^{(C,V,R,T,S)_{average}} \ \ \ = \ \ \  
\stackrel{argmax}{s_x} p(s_x|c,v,r,t)
\end{equation}

The Naive Mix addresses the uncertainty that exists in model
selection. Similar difficulties in selecting single best models have
been noted elsewhere; in fact, there is a general trend in model
selection research away from the selection of such models (e.g.,
\cite{MadiganR94}). A similar movement exists in machine learning,
based on the premise that no learning algorithm is superior for all
tasks \cite{Schaffer94}. This  has lead to hybrid approaches that
combine diverse learning paradigms (e.g., \cite{Domingos96}) and
approaches that select the most appropriate learning algorithm based
on the characteristics of the training data (e.g., \cite{Brodley95}). 

\section{Naive Bayes} \label{sec:naivebayes}

Naive Bayes differs from the models learned via sequential model 
selection and the Naive Mix since the parametric form of Naive Bayes
is always the same and does not have to be learned.  Naive Bayes
assumes that all feature variables are conditionally  independent
given the value of the classification variable.  Examples of both the
graphical representation of this parametric form and the associated
parameter estimates are shown in Chapter 2. 

In disambiguation, feature variables represent  contextual
properties of the sentence in which an ambiguous word occurs. The
classification variable represents the sense of the ambiguous
word. Thus, Naive Bayes assumes that the values of any two contextual 
features in a sentence do not directly affect each other. In general
this is not a likely representation of the dependencies among features
in language. For example, one kind of feature used in this dissertation
represents the part--of--speech of words that surround the ambiguous
word. It is  typically  the case that the part--of--speech of the
${i+1}^{th}$ word in a sentence is dependent on the part--of--speech
of the $i^{th}$ word.  When an article  occurs in the $i^{th}$
position, one can predict that a noun or  adjective is more likely to
occur at the $i+1^{th}$ position than is a  verb, for example.
However, despite the fact that Naive Bayes does not correspond to 
intuitions regarding the dependencies among features, experimental
results in Chapter 6 show that Naive Bayes  performs at levels
comparable to models with learned parametric  forms.

If the contextual features of a sentence are represented by variables
$(F_1, F_2$, $\ldots$, $F_{n-1})$ and the sense of the ambiguous word is 
represented by $S$, then the parameter estimates of Naive Bayes are
calculated as follows: 

\begin{equation}
\hat\theta_{f_1,f_2, \ldots, f_{n-1},s}^{F_1, F_2, \ldots, F_{n-1},S}
= \theta_s^S \times \prod_{i=1}^{n-1} 
\frac{\hat\theta_{f_i,s}^{F_i,S}}{\hat\theta_s^S} 
\end{equation}
Several alternative but equivalent formulations are shown in Chapter 2. 

Even with a large number of features, the number of parameters in
Naive Bayes is relatively small. For a problem with $n$ feature
variables, each having  $l$ possible values, and a classification
variable with $s$ possible values, the number of parameters in Naive
Bayes is $n*l*s$. More complicated models often require that an
exponential  number of parameters be learned. For example, a saturated
model given the same scenario will have $n^l*s$ parameters. 


\chapter{UNSUPERVISED LEARNING FROM RAW TEXT} 

The main limitation of the supervised learning methods presented in
Chapter 3 is the need for sense--tagged text to serve as training
examples. The creation of such text is time--consuming and proves to
be a significant bottleneck in porting and scaling the supervised
approaches to new and larger domains of text. 

Unsupervised learning presents an alternative that eliminates this
dependence on sense--tagged text. The object of unsupervised
learning is to determine the classification of each instance in a
sample without using training examples. In word sense disambiguation,
this corresponds to grouping instances of an ambiguous word into some
pre--specified number of sense groups, where each group corresponds to
a distinct sense of the ambiguous word. This is done strictly based on
information obtained from raw untagged text; no external
knowledge sources are employed. While this increases the  portability
of these approaches, it also imposes an important limitation. Since no
knowledge beyond the raw text is employed, the unsupervised learning
algorithms do not have access to the sense inventory for a word. Thus,
while they create sense groups based on the features observed in the
text, these groups are not labeled with a definition or any other
meaningful tag. If such labels are desired, they must be attached
after unsupervised learning has created the sense groups. One means of
attaching such labels is discussed in Chapter 7. 

This chapter describes a methodology by which probabilistic models can
be learned from raw text. It requires that the variable values associated
with the sense of an ambiguous word be treated as missing or 
unobserved data in the sample. Given that these values
are never present in the data sample, it is not possible to conduct a
systematic search for the parametric form of a model; one must simply
be assumed. In this framework, the Expectation Maximization (EM) algorithm
\cite{DempsterLR77} and Gibbs Sampling \cite{GemanG84} are used to
estimate the parameters of a probabilistic model. As a part of this 
process, values are imputed, i.e., filled--in, for the sense variable.
This effectively assigns instances of an ambiguous word to a
particular sense group.  

An alternative to this probabilistic methodology is to use an
agglomerative clustering algorithm that forms sense groups of untagged
instances of an ambiguous word by minimizing a distance measure
between the instances of an ambiguous word in each sense group. Two
agglomerative algorithms are explored here,  McQuitty's similarity
analysis \cite{Mcquitty66} and  Ward's minimum--variance  method
\cite{Ward63}.

\section{Probabilistic Models}

In supervised learning, given the parametric form of a decomposable
model, maximum likelihood estimates of parameters are simple to compute.  The
sufficient statistics of these parameters are the frequency counts of
marginal events that are defined by the marginal distributions of 
the model. These counts are
obtained directly from the training data. However, in unsupervised
learning, parameter estimation is more difficult since direct estimates
from the sample are not possible given that data is missing.  

To illustrate the problem, the {\it bill} example from the previous
chapter is recast as a problem in unsupervised learning. Suppose 
that  $(CVS)(RTS)$ is the  parametric form of a decomposable model. 
In supervised
learning, maximum likelihood estimates of the parameters of the joint
distribution are made by observing the frequency of the marginal
events $(CVS)$ and $(RTS)$.  However, when sense--tagged text is not
available this estimate can not be computed directly since the value
of $S$ is unknown.  There is no way, for example, to directly count
the occurrences of the marginal events $(C=yes, S=1, V=no)$ and
$(C=yes, S=2, V=no)$ in an untagged sample of text; the only observed
marginal event is $(C=yes, S=?, V=no)$. However, both the EM algorithm
and Gibbs Sampling impute values for this missing data and thereby
make parameter estimation possible.  

Here the assumption is made that the parametric form of the model is
Naive Bayes. In this model, all features are conditionally independent
given the value of the classification feature, i.e., the sense of the
ambiguous word.  This assumption is based on the success of the Naive
Bayes model when applied to supervised word--sense disambiguation 
(e.g.  \cite{BruceW94A}, \cite{GaleCY92}, \cite{LeacockTV93},
\cite{Mooney96},  \cite{PedersenB97A}, \cite{PedersenBW97}).  

In these discussions, the sense of an ambiguous word is  
represented  by a feature variable, $S$,  whose value is missing. The
observed  contextual features are represented by $Y = (F_1, F_2,
\ldots, F_n)$.   The complete data sample is then $D = (Y,S)$ and the
parameters of the model are represented by the vector $\Theta$.  

\subsection{EM Algorithm} \label{sec:em}

The EM algorithm  is an iterative estimation procedure in which a
problem with missing data is recast to make use of complete data
estimation techniques.  The EM algorithm formalizes a long--standing 
method of making estimates for the parameters of a model, $\Theta$,
when data is missing.  A high--level description of the algorithm
is as follows:

\begin{enumerate}
\item Randomly estimate initial values for the parameters 
$\Theta$. Call this set of estimates $\Theta^{old}$. 
\item Replace the missing values of $S$ by their expected values given
the parameter estimates $\Theta^{old}$.   
\item Re--estimate parameters based on the filled--in  values for the
missing variable $S$. Call these parameter estimates $\Theta^{new}$. 
\item Have $\Theta^{old}$ and $\Theta^{new}$ converged? If not,
rename $\Theta^{new}$ as $\Theta^{old}$ and go to step 2. 
\end{enumerate}

\subsubsection{General Description}

At the heart of the EM Algorithm lies the {\it Q-function}. This  is
the expected value of the log of the likelihood function for the
complete data sample, $D=(Y,S)$, where $Y$ is the observed data and $S$ is 
the missing sense value:
\begin{equation}
Q(\Theta^{new}|\Theta^{old}) \ \ \ = \ \ \
E[\ln p(Y,S|\Theta^{new})|\Theta^{old},Y)]
\label{eq:em1}
\end{equation}
Here, $\Theta^{old}$ is the previous value of the maximum likelihood
estimates of the parameters and $\Theta^{new}$ is the improved
estimate; $p(Y,S|\Theta^{new})$ is the likelihood of observing the
complete data given the improved estimate of the model parameters.

When approximating the maximum of the likelihood function, the EM
algorithm starts from a randomly generated initial estimate of the
model parameters and then replaces $\Theta^{old}$ by the
$\Theta^{new}$ which maximizes  $Q(\Theta^{new}|\Theta^{old})$. This
is a two step process, where the first step is known as the
expectation step, i.e., the E--step, and the second is the
maximization step, i.e., the M--step. The E--step finds the expected
values of the sufficient statistics of the complete model using the
current estimates of the model parameters. For decomposable models
these sufficient statistics are the frequency counts of events defined
by the marginal distributions of the model.  The M--step makes maximum
likelihood estimates of the model parameters using the sufficient
statistics from the E--step. These steps iterate until the parameter
estimates $\Theta^{old}$ and $\Theta^{new}$ converge.    

The M--step is usually easy, assuming it is easy for the complete
data problem. As shown in Chapter 2, making parameter estimates for
decomposable models is straightforward. 
In the general case the E--step may be complex. However, for
decomposable models the E--step simplifies to the calculation of the
expected marginal event counts defined by a decomposable model, where
the expectation is with respect to $\Theta^{old}$. The M--step
simplifies to the calculation of new parameter estimates from these
counts.  Further, these expected counts can be calculated by
multiplying the sample size $N$ by the probability of the complete
data within each marginal distribution, given $\Theta^{old}$ and the
observed data within each marginal $Y_{m}$.  This simplifies to: 
\begin{equation}
freq^{new}(S_{m},Y_{m}) = p(S_{m}|Y_{m}) \times  freq(Y_{m})
\label{eq:freqnew}
\end{equation}
where ${freq}^{new}$ is the current estimate of the expected 
count and $p(S_{m}|Y_{m})$ is formulated using $\Theta^{old}$. 

\subsubsection{Naive Bayes description} 

The expectation and maximization steps of the EM algorithm are
outlined here. It is assumed that the parametric form is
Naive Bayes, although this discussion extends easily  to any
decomposable model \cite{Lauritzen95}. Given that the parametric form
is Naive Bayes, it follows that:
\begin{equation}
 p(F_1,F_2,\ldots,F_n,S) = 
 p(S) \times \prod_{i=1}^n p(F_i|S)
\label{eq:naivebayes}
\end{equation}
where $p(S)$ and $p(F_i|S)$ are the model parameters. This is
equivalent to treating $p(F_i,S)$ as the model parameters since
$p(F_i,S) = \frac{p(F_i|S)}{p(S)}$. However, the conditional
representation lends itself to  developing certain analogies between
the EM algorithm and Gibbs Sampling.  

{\bf E--step}: The expected values of the sufficient statistics
of the Naive Bayes model are computed. These are the frequency counts 
of marginal events of the form $(F_i,S)$ and are notated
$freq(F_i,S)$.  Since $S$ is unobserved, values for it must be imputed
before the marginal events can be counted. During the first iteration
of the EM algorithm, values for $S$ are imputed by random
initialization. Thereafter, $S$ is imputed with values that maximize
the probability of observing a particular sense for an ambiguous word
in a given context:   
\begin{equation}
S = \stackrel{argmax}{s_x} \hat p(S | f_1, f_2, \ldots, f_{n-1}, f_n)  
\end{equation}
From $p(a|b) = \frac{p(a,b)}{p(b)}$ it follows that:
\begin{equation}
\hat p(S | f_1, f_2, \ldots, f_{n-1}, f_n) = 
\frac{\hat p(f_1, f_2, \ldots, f_{n-1}, f_n,S)}
{\hat p(f_1,f_2,\ldots,f_{n-1}, f_n)}
\label{eq:em1a}
\end{equation}
And from $p(a,b) = \sum_c p(a,b,c)$ it follows that:
\begin{eqnarray}
\hat p(f_1, f_2, \ldots, f_{n-1}, f_n) = 
\sum_S  \hat p(f_1, f_2, \ldots, f_n,S)
\label{eq:em4a}
\end{eqnarray}
Thus, 
\begin{equation}
S = \stackrel{argmax}{s_x} 
\frac{\hat p(S) \times \prod_{i=1}^n \hat p(f_i|S)}
{\sum_S \hat p(f_1,f_2,\ldots,f_n,S)}
\end{equation}

This calculation determines the value of $S$ to impute for each
possible combination of observed feature values. Given imputed
values for $S$, the expected values of the marginal event counts,
$freq(F_i,S)$, are determined directly from the data sample following
Equation \ref{eq:freqnew}. These counts are the sufficient statistics
for the Naive Bayes model.     

{\bf M--Step}: The sufficient statistics from the E--step are used to 
re--estimate the model parameters. This new set of estimates is
designated $\Theta^{new}$ while the previous set of parameter
estimates is called $\Theta^{old}$. The model parameters  $p(S)$ and
$p(F_i|S)$ are estimated as follows:  

\begin{equation}
\hat p(S) = \frac{freq(S)}{N} \ \ \ \ \ \ 
\hat p(F_i|S) = \frac{freq(F_i,S)}{freq(S)}
\label{eq:mstep}
\end{equation}

{\bf Convergence?}: If the difference between the parameter estimates 
obtained in the previous and current iteration is less than
some pre--specified value $\epsilon$, i.e.,:
\begin{equation}
|| \Theta^{old} - \Theta^{new}|| < \epsilon
\end{equation}
then the parameter estimates have converged and the EM algorithm
stops. If this difference is greater than $\epsilon$, $\Theta^{new}$
is renamed $\Theta^{old}$ and the EM algorithm continues. 

The EM algorithm is guaranteed to converge \cite{DempsterLR77},
however if the  likelihood function is very irregular it may converge
to a local maxima and not find the global maximum.  In this case, an
alternative is to use the more computationally  expensive method of
Gibbs Sampling which is guaranteed to converge to a global maximum. 

\subsubsection{Naive Bayes example}

The step by step operation of the EM algorithm is illustrated with a
simple example where the parametric form is assumed to be Naive
Bayes. Suppose that there is a data sample where events are described
by 3 random variables. The variables $F_1$ and $F_2$ are observed and
have two  possible values. The variable $S$ represents the class of
the event but is unobserved. Given this formulation, the model
parameters are $p(S)$, $p(F_1|S)$ and $p(F_2|S)$, following Equation
\ref{eq:naivebayes}. The data sample used in this example has  ten
observations and is shown in Table \ref{tab:data}.     
\begin{table}
\begin{center}
\caption{Unsupervised Learning Example Data}
\label{tab:data}
\begin{tabular}{cc|c} 
$F_1$ & $F_2$ & S \\ \hline
1 & 2 & ?  \\   
1 & 2 & ?  \\
2 & 2 & ?  \\ 
2 & 2 & ?  \\
1 & 2 & ?  \\
1 & 1 & ?  \\
1 & 1 & ?  \\
1 & 1 & ?  \\
1 & 2 & ?  \\
2 & 2 & ?  \\ 
\end{tabular}
\end{center}
\myendtab

{\bf E--Step Iteration 1}: The EM algorithm begins by randomly
assigning values to $S$. Such an assignment is shown on the left side
of Figure \ref{fig:estep1}. Given these random assignments, expected
values for the sufficient statistics of the Naive Bayes model are
determined by counting the marginal events defined by Naive Bayes, 
i.e., $freq(F_1,S)$ and $freq(F_2,S)$. These marginal event counts are
conveniently represented in a cross--classification or contingency
table, as appears in the center and right of Figure
\ref{fig:estep1}. For example, the center table shows that:
\begin{eqnarray*}
freq(F_1=1,S=1) = 3 \ \ \ \ \ \  freq(F_1=2,S=1) = 1 \\
freq(F_1=1,S=2) = 2 \ \ \ \ \ \  freq(F_1=2,S=2) = 2 \\
freq(F_1=1,S=3) = 2 \ \ \ \ \ \  freq(F_1=2,S=3) = 0 \\
\end{eqnarray*}

\begin{figure}
\begin{center}
\begin{tabular}{cc|ccc}
$F_1$ & $F_2$ & S  & & \\  
\cline{1-3}
1 & 2 & 1 & &  \\
1 & 2 & 3 & &  \\
2 & 2 & 2 & &  \\
2 & 2 & 2 & &  \\
1 & 2 & 1 & &  \\
1 & 1 & 3 & &  \\
1 & 1 & 1 & &  \\
1 & 1 & 2 & &  \\
1 & 2 & 2 & &  \\
2 & 2 & 1 & &  \\ 
\end{tabular}
\begin{tabular}{cc|cc|ccc} 
\multicolumn{2}{c}{} & 
\multicolumn{2}{c}{$F_1$} & 
\multicolumn{3}{c}{} \\
  &   &  1  &  2 & & &   \\ 
\cline{2-5}
  & 1 &  3  &  1 & 4 & &  \\
S & 2 &  2  &  2 & 4 & & \\
  & 3 &  2  &  0 & 2 & & \\
\cline{2-5}
  &   &  7  &  3 & 10 & & \\ 
\end{tabular}
\begin{tabular}{cc|cc|ccc} 
\multicolumn{2}{c}{} & 
\multicolumn{2}{c}{$F_2$} & 
\multicolumn{3}{c}{} \\
  &   &  1  &  2 &  & &   \\ 
\cline{2-5}
  & 1 &  1  &  3 & 4 & &  \\
S & 2 &  1  &  3 & 4 & &  \\
  & 3 &  1  &  1 & 2 & &  \\
\cline{2-5}
  &   &  7  &  3 & 10 & & \\ 
\end{tabular}
\caption{E--Step Iteration 1}
\label{fig:estep1}
\end{center}
\myendfig   

{\bf M--Step Iteration 1:}
Maximum likelihood estimates for the parameters of Naive Bayes are
made from the marginal event counts found during the E--step. Given the
marginal event counts in Figure \ref{fig:estep1}, the parameter estimates 
are computed following Equation \ref{eq:mstep}. The values for these
estimates found during iteration 1 are shown in the contingency
tables in Figure \ref{fig:mstep1}.   For example, the center table
shows that:
\begin{eqnarray*}
\hat p(F_1=1|S=1) = 0.75 \ \ \ \ \ \  \hat p(F_1=2|S=1) = 0.25 \\
\hat p(F_1=1|S=2) = 0.50 \ \ \ \ \ \  \hat p(F_1=2|S=2) = 0.50 \\
\hat p(F_1=1|S=3) = 1.00 \ \ \ \ \ \  \hat p(F_1=2|S=3) = 0.00 \\
\end{eqnarray*}

\begin{figure}
\begin{center}
\begin{tabular}{cc|c|ccc} 
  &   &         & & & \\
\cline{2-4}
  & 1 &  0.4    & & &  \\
S & 2 &  0.4    & & & \\
  & 3 &  0.2    & & & \\
\cline{2-4} 
  &   &         & & & \\ 
\end{tabular}
\begin{tabular}{cc|cc|cccc} 
\multicolumn{2}{c}{} & 
\multicolumn{2}{c}{$F_1$} & 
\multicolumn{4}{c}{} \\ 
  &   &  1     &  2    &      & & & \\
\cline{2-5}
  & 1 &  0.75  &  0.25 & 1.0 & & &  \\
S & 2 &  0.50  &  0.50 & 1.0 & & &  \\
  & 3 &  1.00  &  0.00 & 1.0 & & &  \\
\cline{2-5} 
  &   &        &       &      & & &  \\ 
\end{tabular}
\begin{tabular}{cc|cc|cc} 
\multicolumn{2}{c}{} & 
\multicolumn{2}{c}{$F_2$} & 
\multicolumn{2}{c}{} \\ 
  &   &  1     &  2   &   &  \\
\cline{2-5}
  &1  & 0.25  & 0.75 & 1.0 &  \\
S &2  & 0.25  & 0.75 & 1.0 &  \\
  &3  & 0.50  & 0.50 & 1.0 &  \\
\cline{2-5} 
  &   &       &      &     & \\ 
\end{tabular}
\caption{M--Step Iteration 1:  $\hat p(S)$, $\hat p(F_1|S)$, $\hat p(F_2|S)$} 
\label{fig:mstep1}
\end{center}
\myendfig   

{\bf E--Step Iteration 2:}
After the first iteration of the EM algorithm, all subsequent
iterations find the expected values of the marginal event counts by
imputing new values for $S$ that maximize the following conditional
distribution: 
\begin{equation}
S = \stackrel{argmax}{s} \hat p(S | F_1, F_2) = \frac{\hat p(S) \times
\hat p(F_1|S) \times \hat p(F_2|S)}{\hat p(F_1,F_2)} 
\label{eq:finds}
\end{equation}

The estimates of the parameters required by Equation \ref{eq:finds}
are the estimates made in the M--step of the previous iteration, shown 
in Figure \ref{fig:mstep1}. The computation in Equation \ref{eq:finds}
results in the value of $S$ that maximize the conditional probability
distribution where $S$ is conditioned on the values of the observed
features $F_1$ and $F_2$.  The values of $\hat p(S|F_1,F_2)$
are shown in Figure \ref{fig:estep2}. The maximum estimate for each
given pair of values for the features $(F_1, F_2)$ are shown
in bold face. The value of $S$ associated with each of these maximum 
probabilities is imputed for each observation in the data sample that
shares the same values for the observed feature values.  For example,
if $(F_1 = 1, F_2 = 1, S = ?)$ is an observation in the data sample,
then the value of 3 is imputed for $S$ since $\hat p(S=3|F_1=1,
F_2=1)$ is greater than both  $\hat p(S=2|F_1=1, F_2=1)$ and
 $\hat p(S=1|F_1=1, F_2=1)$. 

\begin{figure}
\begin{center}
\begin{tabular}{ccc|c}
$F_1$ & $F_2$ & $S$ & $\hat p(S|F_1,F_2)$ \\  \hline
 1    &   1   &  1  &  .333  \\
 1    &   1   &  2  &  .222  \\
 1    &   1   &  3  &  {\bf .444}  \\ \hline
 1    &   2   &  1  &  {\bf .474}  \\
 1    &   2   &  2  &  .316  \\
 1    &   2   &  3  &  .211  \\ \hline
 2    &   1   &  1  &  .333  \\
 2    &   1   &  2  &  {\bf .667}  \\
 2    &   1   &  3  &  .000  \\  \hline
 2    &   2   &  1  &  .333  \\
 2    &   2   &  2  &  {\bf .667}  \\
 2    &   2   &  3  &  .000  \\
\end{tabular}
\caption{E--Step Iteration 2}
\label{fig:estep2}
\end{center}
\myendfig   

\begin{figure}
\begin{center}
\begin{tabular}{cc|ccc}
$F_1$ & $F_2$ & S \\  
\cline{1-3}
1 & 2 & 1 & &  \\
1 & 2 & 1 & &  \\
2 & 2 & 2 & &  \\
2 & 2 & 2 & &  \\
1 & 2 & 1 & &  \\
1 & 1 & 3 & &  \\
1 & 1 & 3 & &  \\
1 & 1 & 3 & &  \\
1 & 2 & 1 & &  \\
2 & 2 & 2 & &  \\ 
\end{tabular}
\begin{tabular}{cc|cc|ccc} 
\multicolumn{2}{c}{} & 
\multicolumn{2}{c}{$F_1$} & 
\multicolumn{3}{c}{} \\
  &   &  1  &  2 &  & &   \\ 
\cline{2-5}
  & 1 &  4  &  0 & 4 & &  \\
S & 2 &  0  &  3 & 3 & &  \\
  & 3 &  3  &  0 & 3 & &  \\
\cline{2-5}
  &   &  7  &  3 & 10 & & \\ 
\end{tabular}
\begin{tabular}{cc|cc|ccc} 
\multicolumn{2}{c}{} & 
\multicolumn{2}{c}{$F_2$} & 
\multicolumn{3}{c}{} \\
  &   &  1  &  2 &  & &   \\ 
\cline{2-5}
  & 1 &  0  &  4 & 4 & &  \\
S & 2 &  0  &  3 & 3 & &  \\
  & 3 &  3  &  0 & 3 & &  \\
\cline{2-5}
  &   &  3  &  7 & 10 & &  \\ 
\end{tabular}
\caption{E--Step Iteration 2}
\label{fig:estep2a}
\end{center}
\myendfig   

The data sample that results from these imputations for 
$S$ is shown on the left of Figure \ref{fig:estep2a}. 
The expected counts of the marginal events in that updated 
data sample are shown in contingency table form in the center 
and right of this same figure. 

{\bf M--Step Iteration 2:}
Given the expected values of the  marginal event counts from the
previous E--step, values for the model parameters are re--estimated.
Figure  \ref{fig:mstep2} shows the values for the parameter
estimates $\hat p(S)$, $\hat p(F_1|S)$,
and  $\hat p(F_2|S)$. 

\begin{figure}
\begin{center}
\begin{tabular}{cc|c|ccc} 
  &   &         & & & \\
\cline{2-4}
  & 1 &  0.4    & & &  \\
S & 2 &  0.3    & & & \\
  & 3 &  0.3    & & & \\
\cline{2-4} 
  &   &         & & & \\ 
\end{tabular}
\begin{tabular}{cc|cc|cccc} 
\multicolumn{2}{c}{} & 
\multicolumn{2}{c}{$F_1$} & 
\multicolumn{4}{c}{} \\ 
  &   &  1     &  2    &      & & & \\
\cline{2-5}
  & 1 &  1.0  &  0.0 & 1.0 & & &  \\
S & 2 &  0.0  &  1.0 & 1.0 & & &  \\
  & 3 &  1.0  &  0.0 & 1.0 & & &  \\
\cline{2-5} 
  &   &        &       &   & & &  \\ 
\end{tabular}
\begin{tabular}{cc|cc|cc} 
\multicolumn{2}{c}{} & 
\multicolumn{2}{c}{$F_2$} & 
\multicolumn{2}{c}{} \\ 
  &   &  1     &  2   &   &  \\
\cline{2-5}
  &1  & 0.0   & 1.0 & 1.0 &  \\
S &2  & 0.0   & 1.0 & 1.0 &  \\
  &3  & 1.0   & 0.0 & 1.0 &  \\
\cline{2-5} 
  &   &       &      &     & \\ 
\end{tabular}
\caption{M--Step Iteration 2:  $\hat p(S)$, $\hat p(F_1|S)$, $\hat p(F_2|S)$} 
\label{fig:mstep2}
\end{center}
\myendfig   

At this point, two iterations of the EM algorithm have been performed. 
From this point forward, at the conclusion of each iteration a check for
convergence is made. The parameters estimated during the previous
iteration are $\Theta^{old}$ and those estimated during the current
iteration are $\Theta^{new}$. For example, during iterations 1 and 2 the
following estimates have been made:
\begin{eqnarray*}
\Theta^{old} = \{.4, .4, .2, .75, .25, .5, .5, 1.0, 0.0, .25, .75, .25, .75,
.5, .5\}  \\
\Theta^{new} = \{.4, .3, .3, 1.0, 0.0, 0.0, 1.0, 1.0, 0.0, 0.0, 1.0, 0.0,
1.0, 1.0, 0.0\} 
\end{eqnarray*}

The difference between $\Theta^{old}$ and $\Theta^{old}$ is
considerable and certainly more than a typical value of $\epsilon$,
i.e., .01 or .001.  Thus, the EM algorithm continues for at least
one more iteration.  

{\bf E--step Iteration 3:}
The expected values for the marginal event counts are determined as in
the previous iteration. First, the  values of $S$ that maximize the
conditional probability distribution of $\hat p(S|F_1, F_2)$ are
determined. This maximization is based on  the parameter estimates,
shown in Figure \ref{fig:mstep2}, which were determined during the
M--step of the previous iteration. 

\begin{figure}
\begin{center}
\begin{tabular}{ccc|c}
$F_1$ & $F_2$ & $S$ & $\hat p(S|F_1,F_2)$ \\  \hline
 1    &   1   &  1  &  0.0  \\
 1    &   1   &  2  &  0.0  \\
 1    &   1   &  3  &  {\bf 1.0}  \\ \hline
 1    &   2   &  1  &  {\bf 1.0}  \\
 1    &   2   &  2  &  0.0  \\
 1    &   2   &  3  &  0.0  \\ \hline
 2    &   1   &  1  &  0.0  \\
 2    &   1   &  2  &  0.0  \\
 2    &   1   &  3  &  0.0  \\  \hline
 2    &   2   &  1  &  0.0  \\
 2    &   2   &  2  &  {\bf 1.0}  \\
 2    &   2   &  3  &  0.0  \\
\end{tabular}
\caption{E--Step Iteration 3}
\label{fig:estep3}
\end{center}
\myendfig   

Figure \ref{fig:estep3} shows the estimated values for $\hat
p(S|F_1,F_2)$, the conditional distribution of $S$ given the values of
the observed feature values. The maximum probability for each
given pair of feature values is shown in bold face. 
The values of $S$ that result in this maximized conditional
distribution are imputed for the missing data of observations in
the sample that share the same observed feature values.  The data set 
that results from this 
imputation is shown in Figure \ref{fig:estep3a}. However, note that
the values found during the third iteration prove to be identical to
those found during the second iteration.   

\begin{figure}
\begin{center}
\begin{tabular}{cc|c}
$F_1$ & $F_2$ & S \\  \hline
1 & 2 & 1  \\
1 & 2 & 1  \\
2 & 2 & 2  \\
2 & 2 & 2  \\
1 & 2 & 1  \\
1 & 1 & 3  \\
1 & 1 & 3  \\
1 & 1 & 3  \\
1 & 2 & 1  \\
2 & 2 & 2  \\ 
\end{tabular}
\caption{E--Step Iteration 3}
\label{fig:estep3a}
\end{center}
\myendfig   

The expected counts of marginal events and the parameter estimates
found during iteration 3 are identical to those found during iteration
2.  Given this, the difference between $\Theta^{old}$ and
$\Theta^{new}$ is zero and the parameter estimates have converged. The
values for the missing variable $S$ are assigned as shown in Figure
\ref{fig:estep3a}. This is an intuitively appealing result that can be
interpreted in terms of assigning events to classes. The event
$(F_1=1,F_2=1)$ belongs to class 3, event  $(F_1=2,F_2=2)$ belongs to
class 2, and event $(F_1=1,F_2=2)$ belongs to class 1.    

\subsection{Gibbs Sampling} \label{sec:gibbs}

Gibbs Sampling is a more general tool than the EM algorithm in that 
it is not restricted to handling missing data; it is a special case 
of Markov Chain Monte Carlo methods for approximate inference. These
methods were first used for applications in statistical physics in 
the 1950's; perhaps the most notable example being the Metropolis
algorithm \cite{Metropolis53}.  Gibbs Sampling was originally
presented in the context of an image restoration problem but has since
been applied to a wide range of applications.     

In general, Gibbs Sampling provides a means of approximating complex
probabilistic models. In unsupervised learning probabilistic models
are complex because there is missing data, i.e., the sense of the
ambiguous word is unknown. Gibbs Sampling approximates the
distribution of the parameters of a model as if the missing data were
observed. By contrast, the EM algorithm simply maximizes the estimated
values for the parameters of a model, again by acting as if the
missing data were observed.

\subsubsection{General Description}

Gibbs Sampling has a Bayesian orientation in that it naturally 
incorporates prior distributions, $p(\Theta)$,  into the sampling
process.  When  a prior distribution is specified in conjunction with
an observed data sample, Gibbs Sampling approximates  the posterior
probability function, $p(\Theta|D)$, by taking a large number of
samples from it. If a prior distribution is not utilized then Gibbs
Sampling still takes a large number of samples, however, they are
drawn from the likelihood function $p(D|\Theta)$. In this
dissertation, non--informative prior distributions are employed and
the sampling is from the posterior distribution function.   

A Gibbs Sampler creates Markov Chains of parameter estimates and
values for missing data whose stationary distributions approximate the
posterior distribution, $p(\Theta|D)$,  by simulating a random walk in
the space of $\Theta$. A Markov Chain  is a series of random variables
$(X^0, X^1, \ldots)$ in which the influence of the values of
$(X^0,\ldots, X^n)$ on the distribution of $X^{n+1}$ is mediated
entirely by the value of $X^{n}$. 

Let the values of the observed contextual feature variables be
represented by $Y = (F_1, F_2,$ $\cdots,$ $F_{n-1}, F_n)$ and let $S$
represent the unknown sense of an ambiguous word. Given that the
parametric form of the model is known, random initial values are
generated for the missing data $S^0 = (S_1^0, S_2^0, \ldots, S_N^0)$
and the unknown parameter estimates of the assumed model $\Theta^0 =
(\theta_1^0, \theta_2^0, \ldots, \theta_q^0)$.   $S^0$ is a vector
containing a value for each instance of the missing sense data,
$\Theta^0$ is a vector containing the parameters of the model, $N$ is
the number of observations in the data sample, and $q$ is the number
of parameters in the model.   

A Gibbs Sampler performs the following loop, where $j$ is the
iteration counter,  until convergence is detected:  
\begin{eqnarray*}
S^{j+1} \sim p(S|\theta_1^j,\theta_2^{j},\cdots,\theta_q^{j},Y)  \\
\theta_1^{j+1} \sim 
p(\theta_1|\theta_2^{j},\cdots,\theta_q^{j},Y,S^{j+1}) \\
\theta_2^{j+1} \sim 
p(\theta_2|\theta_1^{j+1},\theta_3^j,\cdots,\theta_q^{j},Y,S^{j+1}) \\
\vdots \\
\theta_q^{j+1} \sim 
p(\theta_q|\theta_1^{j+1},\cdots,\theta_{q-1}^{j+1},Y,S^{j+1}) \\
\end{eqnarray*}

Each iteration of the Gibbs Sampler samples values for the
missing data and for the unknown parameter estimates. The values for
the missing data are conditioned on the values of the parameters and
the observed data. The parameter estimates are conditioned
on the previously estimated values of the other parameters and the
missing data as well as the observed data.    

A chain of values is constructed for each missing value and
parameter estimate via this sampling loop. Each chain is monitored for
convergence. A range of  techniques for monitoring convergence 
are discussed in  \cite{Tanner93}; this dissertation uses Geweke's
method \cite{Geweke92}.  In this approach, a chain is divided into two
windows, one at the beginning and the other at the end.
Each window contains about 10\% of the total number of
iterations in the chain. If the entire chain has reached a stationary
distribution,  then these two windows, one early in the chain and the
other late, will have approximately the same mean values. If they do
not then the parameters have not yet converged to a stationary
distribution.  

The early iterations of Gibbs Sampling produce chains of values with
very high variance. It is standard to discard  some portion of the
early iterations; this process is commonly known as a {\it
burn--in}. The general procedure followed here is to have a 500
iteration burn--in followed by 1000 iterations that are monitored for
convergence. If the  chains  do not converge after 1000 iterations
then additional iterations in increments of 500 are performed until
they do. This procedure was designed following recommendations by
\cite{RafteryL92}.  

A proof that convergence on the posterior probability
distribution  is guaranteed during Gibbs Sampling is given in
\cite{GemanG84}.  Once convergence occurs, the approximation to  the
posterior probability  function can be summarized like any other
probability function. Also, the median value in each chain of sampled
values for missing data becomes the sense group to which an instance
of an ambiguous word is ultimately assigned.  

\subsubsection{Naive Bayes description}

Gibbs Sampling is developed in further detail, given the assumption
that the parametric form is Naive Bayes. However, this discussion is
easily extended to any decomposable model. As in the previous example, 
the parameters of the model are $p(S)$ and $p(F_i|S)$, following 
Equation \ref{eq:naivebayes}. 

A Gibbs Sampler generates chains of values for each missing instance
of $S$ in the data sample and also for each of the parameters
$\hat p(S)$ and $\hat p(F_i|S)$. Each of these chains will eventually 
converge to a stationary distribution. 

In this dissertation the observed data sample is multinomial, i.e.,
each instance in the data sample is described by a combination of
discrete feature values.  As was shown in Chapter 2, such a sample can
be formally defined by a multinomial distribution with parameters
$(N;\theta_1, \theta_2, \ldots, \theta_q)$. However, here 
the distribution of multinomial data is represented by the
frequency counts for each possible event. This is notated as $M(f_1,
f_2,$ $\ldots, f_q)$, where $q$ is the number of  possible events and
$f_i$ is the frequency of  the $i^{th}$ event. For example,
$M(f_1,f_2,$ $\dots,f_q)$ represents a multinomial distribution with
$q$ possible events where the $i^{th}$ event occurs $f_i$
times.

The conjugate prior to the multinomial distribution  is the Dirichlet
distribution, described by $D(\alpha_1, \alpha_2,
\ldots, \alpha_q)$, where $\alpha_i$ represents the prior frequency of
the $i^{th}$ event. If all $\alpha_i$ are set to 1 then a
non--informative prior has been properly specified
\cite{Gelman95}. For example, if $q=3$, $D(1,1,1)$ describes a
non--informative Dirichlet prior distribution. 

Following Bayes Rule, the product of a prior distribution and the
likelihood distribution results in a posterior probability
distribution. If the prior distribution and the likelihood function
are conjugate, then the posterior distribution has the same
distribution as the prior. Here, since the observed data is
described by a multinomial distribution and the prior is specified
in terms of a Dirichlet distribution, the resulting posterior
distribution is Dirichlet. 

A multinomial distribution and a Dirichlet distribution 
are multiplied by adding the frequency counts associated with each
possible event in the multinomial with the prior frequency count
as specified by the Dirichlet. The resulting sums specify the
parameters that describe a posterior Dirichlet distribution
\cite{Gelman95}.  
\begin{eqnarray*}
D(f_1+g_1,  \ldots, f_{q-1}+g_{q-1}, f_q+g_q) =
M(f_1,  \ldots, f_{q-1}, f_q) +
D(g_1,  \ldots, g_{q-1}, g_q)
\end{eqnarray*}
This defines a distribution from which values can be sampled to
approximate the posterior distribution of the parameter estimates. 

In this discussion, Gibbs Sampling is cast as a non--deterministic
version of the EM algorithm. This treatment is similar in spirit to
that of \cite{Buntine94}, where the EM algorithm is treated as a
deterministic version of Gibbs Sampling.

\begin{enumerate}
\item {\bf Stochastic E--Step}: The expected values of the sufficient 
statistics are calculated. These are the counts of the marginal events
defined by the marginal distributions of the Naive Bayes model, $(F_i, S)$. 
However, before the marginal events can be counted, values for $S$
must be imputed for each instance in the data sample. In the EM
algorithm these values are obtained by finding the value of $S$ that
maximizes $\hat p(S|F_1, F_2,$ $\ldots, F_{n-1}, F_n)$. In Gibbs
Sampling, these values are imputed via sampling from that same
conditional distribution:
\begin{eqnarray}
S \sim \hat p(S | f_1, f_2, \ldots, f_{n-1}, f_n)  = 
\frac{\hat p(S) \times \prod_i^n \hat p(F_i|S)}
{\hat p(f_1, f_2, \ldots, f_{n-1}, f_n)}
\end{eqnarray}
This conditional distribution is based upon values for
$\hat p(S)$ and $\hat p(F_i|S)$ that are arrived at via sampling
during the previous iteration of the stochastic M--step. If this is
the first iteration of the Gibbs Sampler, then these values come about
as the result of random initialization.  After values for $S$ are
imputed via sampling, the marginal events are counted and the
stochastic E--step concludes.   

\item {\bf Stochastic M--Step}: The expected values of the sufficient
statistics found during the stochastic E-step are now used to
re--estimate the parameters of the model. The EM algorithm makes 
maximum likelihood estimates directly from these marginal event
counts. However, in Gibbs Sampling these marginal event counts are
used to describe a multinomial distribution that is multiplied by a
Dirichlet prior distribution to create a Dirichlet posterior
distribution from which values of the model parameters are sampled. 

The observed frequency counts of marginal events are used
to describe a multinomial distribution from which samples for the
model parameters can be drawn. To approximate the conditional distribution
of $\hat p(F_i|S)$ via sampling, suppose there are 2 possible events
when the value of S is fixed. The frequency count of each event is
represented by $f_1$ and $f_2$ and the multinomial distribution of
this data can be described by $M(f_1,f_2)$. Further suppose that a
non--information prior Dirichlet distribution is specified, i.e.,
$D(1,1)$. These two distributions are multiplied to create a posterior
Dirichlet distribution from which values for $\hat p(F_i|S)$ are
sampled:   
\begin{equation}
\hat p(F_i | S) \sim D(f_1 +1, f_2+1) = M(f_1,f_2) \times D(1,1)
\end{equation}
Values for $\hat p(S)$ are sampled along similar lines.  After values
for the model parameters have been sampled, the stochastic M--step
concludes.  

\item {\bf Convergence?}: 
After $j$ iterations there are chains of length $j$ for both the
sampled values for each model parameter and for each of the $N$
missing sense values in the data sample. After some set number of
iterations, these chains are checked for convergence. 

Once convergence is detected,  the median values in the chains
created during sampling are regarded as the estimates of parameters
and missing data.  For example, suppose that (1, 1, 1, 2, 2, 2, 2, 2,
2, 3) is a chain that represents the values for a missing sense value
sampled for a particular  instance in the data sample. The median
sense value is 2 and this value is imputed for $S$ for that
observation in the sample.   
\end{enumerate} 

\subsubsection{Naive Bayes example}

The same example used to demonstrate the EM algorithm is employed here 
with Gibbs Sampling. The data sample is shown in Table
\ref{tab:data} and the parametric form of the model is Naive Bayes
with model parameters $p(S)$, $p(F_1|S)$, and $p(F_2|S)$.  

{\bf Stochastic E--Step Iteration 1}: Like the EM algorithm, Gibbs 
Sampling begins by randomly assigning values to $S$. Assume that the 
random assignment of values to $S$ is as shown on the left of Figure
\ref{fig:gibbe1} and the expected counts of marginal events, as
represented in the contingency tables, are as shown in the center and
right of that figure. 

\begin{figure}
\begin{center}
\begin{tabular}{cc|ccc}
$F_1$ & $F_2$ & S  & & \\  
\cline{1-3}
1 & 2 & 1 & &  \\
1 & 2 & 3 & &  \\
2 & 2 & 2 & &  \\
2 & 2 & 2 & &  \\
1 & 2 & 1 & &  \\
1 & 1 & 3 & &  \\
1 & 1 & 1 & &  \\
1 & 1 & 2 & &  \\
1 & 2 & 2 & &  \\
2 & 2 & 1 & &  \\ 
\end{tabular}
\begin{tabular}{cc|cc|ccc} 
\multicolumn{2}{c}{} & 
\multicolumn{2}{c}{$F_1$} & 
\multicolumn{3}{c}{} \\
  &   &  1  &  2 & & &   \\ 
\cline{2-5}
  & 1 &  3  &  1 & 4 & &  \\
S & 2 &  2  &  2 & 4 & & \\
  & 3 &  2  &  0 & 2 & & \\
\cline{2-5}
  &   &  7  &  3 & 10 & & \\ 
\end{tabular}
\begin{tabular}{cc|cc|ccc} 
\multicolumn{2}{c}{} & 
\multicolumn{2}{c}{$F_2$} & 
\multicolumn{3}{c}{} \\
  &   &  1  &  2 &  & &   \\ 
\cline{2-5}
  & 1 &  1  &  3 & 4 & &  \\
S & 2 &  1  &  3 & 4 & &  \\
  & 3 &  1  &  1 & 2 & &  \\
\cline{2-5}
  &   &  7  &  3 & 10 & & \\ 
\end{tabular}
\caption{Stochastic E--Step Iteration 1}
\label{fig:gibbe1}
\end{center}
\myendfig   

{\bf Stochastic M--Step Iteration 1:}
In the EM algorithm the marginal event counts are used directly to
make maximum likelihood estimates for the model parameters. However,
in Gibbs Sampling no maximum likelihood estimates are computed;
instead, the frequency counts of observed marginal events combine with
a specified prior distribution to describe a posterior distribution
from which values for the model parameters are sampled. 

Each row in the contingency tables shown in Figure \ref{fig:gibbe1}
represents counts of marginal events where the value of $S$ is
fixed. These counts can be thought of as describing a multinomial
distribution that represents a conditional probability of the form 
$\hat p(F_i|S)$. This conditional distribution is multiplied by a
prior distribution to define a posterior distribution from which
estimated values for the model parameters are sampled.  
For example, given that $S$ has a fixed value of 1, the distribution
of $\hat p(F_1|S=1)$ is described by $M(3,1)$. 
Thus, based on the expected counts of the marginal events in Figure
\ref{fig:gibbe1} and the assumption that all priors are
non--informative,  the following sampling scheme is devised: 

\begin{eqnarray*}
\hat p(F_1 | S=1) \sim D(4,2) = M(3,1) \times  D(1,1) \\
\hat p(F_1 | S=2) \sim D(3,3) = M(2,2) \times  D(1,1) \\
\hat p(F_1 | S=3) \sim D(3,1) = M(2,0) \times  D(1,1) \\
\hat p(F_2 | S=1) \sim D(2,4) = M(1,3) \times  D(1,1) \\
\hat p(F_2 | S=2) \sim D(2,4) = M(1,3) \times  D(1,1) \\
\hat p(F_2 | S=3) \sim D(2,2) = M(1,1) \times  D(1,1) \\
\hat p(S) \sim D(5,5,3) = M(4,4,2) \times D(1,1,1)  
\end{eqnarray*}

\begin{figure}
\begin{center}
\begin{tabular}{cc|c|ccc} 
  &   &         & & & \\
\cline{2-4}
  & 1 &  0.03    & & &  \\
S & 2 &  0.51    & & & \\
  & 3 &  0.47    & & & \\
\cline{2-4} 
  &   &         & & & \\
\end{tabular}
\begin{tabular}{cc|cc|cccc} 
\multicolumn{2}{c}{} & 
\multicolumn{2}{c}{$F_1$} & 
\multicolumn{4}{c}{} \\ 
  &   &  1     &  2    &      & & & \\
\cline{2-5}
  & 1 &  0.64  &  0.36 & 1.0 & & &  \\
S & 2 &  0.54  &  0.46 & 1.0 & & &  \\
  & 3 &  0.76  &  0.24 & 1.0 & & &  \\
\cline{2-5} 
  &   &        &       &      & & &  \\ 
\end{tabular}
\begin{tabular}{cc|cc|cc} 
\multicolumn{2}{c}{} & 
\multicolumn{2}{c}{$F_2$} & 
\multicolumn{2}{c}{} \\ 
  &   &  1     &  2   &   &  \\
\cline{2-5}
  &1  & 0.37  & 0.63 & 1.0 &  \\
S &2  & 0.09  & 0.91 & 1.0 &  \\
  &3  & 0.73  & 0.27 & 1.0 &  \\
\cline{2-5} 
  &   &       &      &     & \\ 
\end{tabular}
\caption{Stochastic M--step Iteration 1: 
$\hat p(S)$, $\hat p(F_1|S)$, $\hat p(F_2|S)$}  
\label{fig:gibbm1}
\end{center}
\myendfig   
The parameter estimates shown in Figure \ref{fig:gibbm1} are the
result of this sampling plan. 

{\bf Stochastic E--Step Iteration 2:}
After the first iteration of Gibbs Sampling, all subsequent
iterations arrive at new values for the marginal event counts by
sampling new values for $S$ from the following conditional
distribution:  
\begin{equation}
S  \sim \hat p(S | F_1, F_2) = \frac{\hat p(S) \times \hat p(F_1|S)
\times \hat p(F_2|S)}{\hat p(F_1,F_2)} 
\label{eq:conditional}
\end{equation}

The estimates of this  conditional distribution are based upon the
estimates of the model parameters shown in Figure \ref{fig:gibbm1};
these were obtained via sampling during the stochastic M--step of the 
previous iteration. The estimated values of $\hat p(S|F_1,F_2)$ 
are shown in Figure \ref{fig:gibbe2}.  

\begin{figure}
\begin{center}
\begin{tabular}{ccc|c}
$F_1$ & $F_2$ & $S$ & $\hat p(S|F_1,F_2)$ \\  \hline
 1    &   1   &  1  &  .024  \\
 1    &   1   &  2  &  .085  \\
 1    &   1   &  3  &  .891  \\ \hline
 1    &   2   &  1  &  .033  \\
 1    &   2   &  2  &  .699  \\
 1    &   2   &  3  &  .267 \\ \hline
 2    &   1   &  1  &  .100  \\
 2    &   1   &  2  &  .525  \\
 2    &   1   &  3  &  .375  \\  \hline
 2    &   2   &  1  &  .028  \\
 2    &   2   &  2  &  .852  \\
 2    &   2   &  3  &  .120  \\
\end{tabular}
\caption{E--Step Iteration 2}
\label{fig:gibbe2}
\end{center}
\myendfig   

Rather than simply imputing the value of $S$ that maximizes 
$\hat p(S|F_1,F_2)$, as is the case in the EM algorithm, values of 
$S$ are sampled from the conditional distributions 
$\hat p(S|F_1,F_2)$. The result of this sampling process is an
imputed value for $S$ for a given pair of feature values. The
resulting data sample after imputation of $S$ is shown on the left in
Figure  \ref{fig:gibbe2a}. As an example of how Gibbs Sampling differs from
the EM algorithm, note that the first two observations in the data
sample have the same observed feature values,  $(F_1=1,
F_2=2)$. However, the stochastic E--step imputes different values of
$S$ for these observations. This occurs because a value of 2 is
imputed for $S$ with a probability of 70\% while a value of 1 is
imputed with a probability of 27\%. In the E--step of the EM
algorithm, only the value of $S$ that maximizes the conditional
probability is imputed.   

\begin{figure}
\begin{center}
\begin{tabular}{cc|ccc}
$F_1$ & $F_2$ & S \\  
\cline{1-3}
1 & 2 & 2 & &  \\
1 & 2 & 3 & &  \\
2 & 2 & 2 & &  \\
2 & 2 & 2 & &  \\
1 & 2 & 3 & &  \\
1 & 1 & 3 & &  \\
1 & 1 & 3 & &  \\
1 & 1 & 3 & &  \\
1 & 2 & 2 & &  \\
2 & 2 & 2 & &  \\ 
\end{tabular}
\begin{tabular}{cc|cc|ccc} 
\multicolumn{2}{c}{} & 
\multicolumn{2}{c}{$F_1$} & 
\multicolumn{3}{c}{} \\
  &   &  1  &  2 &  & &   \\ 
\cline{2-5}
  & 1 &  0  &  0 & 0 & &  \\
S & 2 &  2  &  3 & 5 & &  \\
  & 3 &  5  &  0 & 5 & &  \\
\cline{2-5}
  &   &  7  &  3 & 10 & & \\ 
\end{tabular}
\begin{tabular}{cc|cc|ccc} 
\multicolumn{2}{c}{} & 
\multicolumn{2}{c}{$F_2$} & 
\multicolumn{3}{c}{} \\
  &   &  1  &  2 &  & &   \\ 
\cline{2-5}
  & 1 &  0  &  0 & 0 & &  \\
S & 2 &  0  &  5 & 5 & &  \\
  & 3 &  3  &  2 & 5 & &  \\
\cline{2-5}
  &   &  3  &  7 & 10 & &  \\ 
\end{tabular}
\caption{Stochastic E--Step Iteration 2}
\label{fig:gibbe2a}
\end{center}
\myendfig   

Figure \ref{fig:gibbe2a} shows the contingency tables of marginal
event counts that result after values for $S$ are imputed. 

{\bf Stochastic M--Step Iteration 2}
Given the marginal event counts found during the stochastic E--step,
shown in Figure \ref{fig:gibbe2a}, sampling from the Dirichlet
posterior of the model parameters is performed according to the
following scheme:  
\begin{eqnarray*}
\hat p(F_1|S=1) \sim D(1,1) =  M(0,0) \times D(1,1) \\
\hat p(F_1|S=2) \sim D(3,4) =  M(2,3) \times D(1,1) \\
\hat p(F_1|S=3) \sim D(6,1) =  M(5,0) \times D(1,1) \\
\hat p(F_2|S=1) \sim D(1,1) =  M(0,0) \times D(1,1) \\
\hat p(F_2|S=2) \sim D(1,6) =  M(0,5) \times D(1,1) \\
\hat p(F_2|S=3) \sim D(4,3) =  M(3,2) \times D(1,1) \\
\hat p(S) \sim D(1,6,6) =  M(0,5,5) \times D(1,1,1) 
\end{eqnarray*}

Figure \ref{fig:gibbm2} shows the sampled estimates for
$\hat p(S)$, $\hat p(F_1|S)$, and $\hat p(F_2|S)$. This concludes the second
iteration of the Gibbs Sampler. 

\begin{figure}
\begin{center}
\begin{tabular}{cc|c|ccc} 
  &   &         & & & \\
\cline{2-4}
  & 1 &  0.06    & & &  \\
S & 2 &  0.54    & & & \\
  & 3 &  0.40    & & & \\
\cline{2-4} 
  &   &         & & & \\ 
\end{tabular}
\begin{tabular}{cc|cc|cccc} 
\multicolumn{2}{c}{} & 
\multicolumn{2}{c}{$F_1$} & 
\multicolumn{4}{c}{} \\ 
  &   &  1     &  2    &      & & & \\
\cline{2-5}
  & 1 &  0.25  &  0.75 & 1.0 & & &  \\
S & 2 &  0.68  &  0.32 & 1.0 & & &  \\
  & 3 &  0.93  &  0.06 & 1.0 & & &  \\
\cline{2-5} 
  &   &        &       &   & & &  \\ 
\end{tabular}
\begin{tabular}{cc|cc|cc} 
\multicolumn{2}{c}{} & 
\multicolumn{2}{c}{$F_2$} & 
\multicolumn{2}{c}{} \\ 
  &   &  1     &  2   &   &  \\
\cline{2-5}
  &1  & 0.16   & 0.84 & 1.0 &  \\
S &2  & 0.05   & 0.95 & 1.0 &  \\
  &3  & 0.52   & 0.48 & 1.0 &  \\
\cline{2-5} 
  &   &       &      &     & \\ 
\end{tabular}
\caption{Stochastic M--step Iteration 2:  
$\hat p(S)$, $\hat p(F_1|S)$, $\hat p(F_2|S)$} 
\label{fig:gibbm2}
\end{center}
\myendfig   

{\bf Stochastic E--step Iteration 3}
Given the parameter estimates from the previous stochastic M--step,
shown in Figure \ref{fig:gibbm2}, conditional distributions for $S$
given the values of the observed features are defined,  per Equation
\ref{eq:conditional}.  

The resulting distributions are shown in Figure \ref{fig:gibbe3}. From
those distributions imputed values for $S$ are obtained via
sampling. The updated data sample is shown in Figure \ref{fig:gibbe3a}.   

\begin{figure}
\begin{center}
\begin{tabular}{ccc|c}
$F_1$ & $F_2$ & $S$ & $\hat p(S|F_1,F_2)$ \\  \hline
 1    &   1   &  1  &  0.009  \\
 1    &   1   &  2  &  0.085  \\
 1    &   1   &  3  &  0.906  \\ \hline
 1    &   2   &  1  &  0.024  \\
 1    &   2   &  2  &  0.645  \\
 1    &   2   &  3  &  0.331  \\ \hline
 2    &   1   &  1  &  0.250  \\
 2    &   1   &  2  &  0.321  \\
 2    &   1   &  3  &  0.429  \\  \hline
 2    &   2   &  1  &  0.181  \\
 2    &   2   &  2  &  0.759  \\
 2    &   2   &  3  &  0.056  \\
\end{tabular}
\caption{Stochastic E--Step Iteration 3}
\label{fig:gibbe3}
\end{center}
\myendfig   

\begin{figure}
\begin{center}
\begin{tabular}{cc|c}
$F_1$ & $F_2$ & S \\  \hline
1 & 2 & 2  \\
1 & 2 & 2  \\
2 & 2 & 2  \\
2 & 2 & 2  \\
1 & 2 & 3  \\
1 & 1 & 3  \\
1 & 1 & 3  \\
1 & 1 & 3  \\
1 & 2 & 2  \\
2 & 2 & 2  \\ 
\end{tabular}
\caption{Stochastic E--Step Iteration 3}
\label{fig:gibbe3a}
\end{center}
\myendfig   

Once again, the expected counts of marginal events are represented in
contingency tables. These will be used to describe multinomial
distributions that will be used in conjunction with a non--informative
prior to create the posterior distributions from which new estimates
of the model parameters will be sampled.    

Normally Gibbs Sampling performs hundreds of iterations before it
is checked for convergence. However, in the interest of brevity no
further calculations or sampling operations will be shown. Unlike the
EM algorithm, Gibbs Sampling does not stop itself. It must be told how
many iterations to perform and then the resulting chains of parameter
estimates and chains of missing values are checked for convergence. If
convergence does not occur then some fixed number of additional
iterations must be performed and then, once again, the resulting
chains must be checked for convergence. 

\newpage
\section{Agglomerative Clustering} \label{sec:cluster}

In general, clustering methods rely on the assumption that classes  of
events occupy distinct regions in a feature space. The distance
between two points in a multi--dimensional space can be measured using
any of a wide variety of metrics (see, e.g. \cite{DevijverK82}).
Observations are grouped in the manner that minimizes the distance
between the members of each cluster. When applied to word sense
disambiguation, each cluster represents a particular sense group of an
ambiguous word. 

Ward's minimum--variance clustering and McQuitty's similarity analysis
are agglomerative clustering algorithms that only differ in regards to
their distance measures. All agglomerative algorithms begin by placing
each observation  in a unique cluster, i.e. a cluster of one. The two
closest clusters are  merged to form a new cluster that replaces the
two merged clusters.  Merging of the two closest clusters continues
until some pre--specified  number of clusters remain. 

However, natural language data does not immediately lend itself to a
distance--based interpretation.  Typical features represent 
part--of--speech  tags, morphological characteristics, and word 
co-occurrence; such features are nominal and their values do not have 
scale.  However, suppose that the values of a part--of--speech feature 
are represented numerically such that $noun = 1$, $verb = 2$, $adjective = 
3$, and $adverb = 4$. While distance measures could be computed using this
representation, they would be meaningless since the fact that a noun has 
a smaller value than an adverb is purely arbitrary and 
reflects nothing about the relationship between nouns and adverbs. 

Thus, before a clustering algorithm is employed, the data must be
converted into a form where spatial distances actually convey
a meaningful relationship between observations. In this dissertation
this is done by representing the data sample as a {\it dissimilarity
matrix}.  Given $N$ observations in a data sample,  
this can be represented in a $N \times N$ dissimilarity matrix such that 
the value in cell $(i,j)$, where $i$ represents the row number and $j$ 
represents the column,  is equal to the number of features in observations 
$i$ and $j$ that do not match.

For example, in Figure \ref{fig:similar} the matrix on the left
represents a data sample consisting of four observations, where each
observation has three nominal features. This sample is converted into 
a $4 \times 4$ dissimilarity matrix that is shown on the left in this
figure. In the dissimilarity matrix, cells $(1,2)$ and $(2,1)$ have
the value 2, indicating that the first and second observations in the
matrix of feature values have different values for two  of the three
features. A value of 0 indicates that observations $i$ and  $j$ are
identical.  

\begin{figure}
\begin{center}
\begin{tabular}{cccccc} 
noun & verb & car & & &\\
adjective & verb & defeat & & & \\
adverb & verb & car & & & \\
noun & verb & car & & & 
\end{tabular} 
\begin{tabular}{cccc} 
0 & 2 & 1 & 0 \\
2 & 0 & 2 & 2 \\
1 & 2 & 0 & 1 \\
0 & 2 & 1 & 0 
\end{tabular} 
\caption{Matrix of Feature Values, Dissimilarity Matrix}
\label{fig:similar}
\end{center}
\myendfig

When clustering this data, each observation is represented by its 
corresponding row (or column) in the dissimilarity  matrix.  Using this 
representation, observations that fall close together in feature space
are likely to belong to the same class and are grouped together into
clusters.  In this dissertation, Ward's and McQuitty's methods are used 
to  form clusters of observations; each cluster corresponds to a sense
group of related instances of an ambiguous word.  

\subsection{Ward's minimum--variance method}

In Ward's method, the internal variance of a cluster
is the sum of squared distances between each
observation in the cluster and the mean observation for that cluster, 
i.e., the average of all the observations in the cluster. 
At each step in Ward's method, a new cluster, $C_{KL}$, with the smallest 
possible internal variance, is created by merging the two
clusters, $C_K$ and $C_L$, that have the minimum variance between
them. The variance between $C_K$ and $C_L$ is computed
as follows:

\begin{equation}
V_{KL} = \frac{{|| \overline{x}_K - \overline{x}_L||}^2}{\frac{1}{N_K} +
\frac{1}{N_L}}
\end{equation}
where $\overline{x}_K$ is the mean observation for cluster $C_K$,
$N_K$ is the number of observations in $C_K$, and $\overline{x}_L$ and 
$N_L$ are defined similarly for $C_L$.

Implicit in Ward's method is the assumption that the sample comes from a
mixture of normal distributions \cite{Ward63}.  Natural language
data is  typically not well characterized by a normal 
distribution. However, when such data is converted into a
dissimilarity matrix there is reason to believe that a normal
approximation is adequate. The number of features employed here is
relatively small, thus the number of possible feature mismatches
between observations is limited. This tends to have a smoothing effect
on data that may be quite sparse and skewed when represented strictly
as a matrix of feature values.

\subsection{McQuitty's similarity analysis}

In McQuitty's method, clusters are based on a simple averaging of the
number of dissimilar features as represented in the dissimilarity
matrix. 

At each step in McQuitty's method, a new cluster, $C_{KL}$, is formed by 
merging the clusters $C_K$ and $C_L$ that have the fewest number of 
dissimilar features between them. Put another way, these are the
clusters that have the most number of features in common. 
The clusters to be merged, $C_K$ and $C_L$, are identified by finding the 
cell $(l,k)$ (or $(k,l)$), where $k \neq l$, that has the minimum value in 
the dissimilarity matrix.  

Once the new cluster $C_{KL}$ is created, the dissimilarity matrix is 
updated to reflect the number of dissimilar features between  
$C_{KL}$ and all other existing clusters. The dissimilarity between
any existing cluster
$C_I$ and $C_{KL}$ is computed as:
\begin{equation}
D_{KL-I} = \frac{D_{KI} + D_{LI}}{2}
\end{equation}
where $D_{KI}$ is the number of dissimilar features between clusters
$C_K$ and $C_I$ and $D_{LI}$ is similarly defined for clusters
$C_L$ and $C_I$. This is simply the average number of mismatches
between each component of the new cluster and the components of
the existing cluster.

Unlike Ward's method, McQuitty's method makes no assumptions concerning
the underlying distribution of the data sample \cite{Mcquitty66}.


\chapter{EXPERIMENTAL DATA}

\section{Words} \label{sec:words}

In addition to having many possible meanings, words are also ambiguous
syntactically in that they can serve as multiple possible 
parts--of--speech. For instance, {\it
line} can be used as a noun, {\it Cut the telephone line}, or as a verb, 
{\it I line my pockets with cash.}   This dissertation does not address 
syntactic ambiguity; it is assumed that this has been resolved for
each of the 13 words studied here. Those words and their
part--of--speech are as follows:
\begin{itemize}
\item {Adjectives:} {\it chief}, {\it common}, {\it last}, and {\it public}. 
\item {Nouns:} {\it bill}, {\it concern}, {\it drug}, {\it interest}, 
and {\it line}.  
\item {Verbs:} {\it agree}, {\it close}, {\it help},  and {\it include}.
\end{itemize}

The {\it line} data \cite{LeacockTV93} is from the ACL/DCI Wall
Street Journal corpus \cite{MarcusSM93} and the American Printing
House for the Blind corpus and tagged with WordNet \cite{Miller95}
senses. The remaining twelve words \cite{BruceWP96} are from the
ACL/DCI Wall Street Journal corpus and tagged with senses from the
Longman  Dictionary of Contemporary English \cite{Procter78}.  The text 
that occurs with these twelve words is tagged with part--of--speech 
information using the Penn TreeBank tag set\footnote{The {\it line} data is 
excluded from the supervised experiments since the text from the 
American Printing House for the Blind is not part--of--speech tagged.}.  

The possible senses for each word are shown in Tables \ref{tab:adj},
\ref{tab:nouns}, and \ref{tab:verbs}. The distribution of senses in
the supervised and unsupervised learning experiments is also
shown. The sense inventories for the latter are reduced in order to 
eliminate very small minority senses. Making fine grained sense
distinctions using unsupervised techniques is not considered in this
dissertation and remains a challenging problem for future work.  

In the supervised and unsupervised experiments a separate model
is learned for each word. Only sentences that contain the ambiguous
word for which a model is being constructed are included in the
learning process. This group of sentences is referred to as a
``word--corpus''. The number of sentences in each word--corpus is
shown in Table \ref{tab:adj}, \ref{tab:nouns}, and \ref{tab:verbs}
in the row ``total count''.

\begin{table}
\begin{center}
\caption{Adjective Senses}
\label{tab:adj}
\begin{tabular}{|lrr|} \hline
\multicolumn{1}{|c}{} & 
\multicolumn{1}{c}{supervised} & 
\multicolumn{1}{c|}{unsupervised} \\ \hline
{\it chief}: & &   \\ 
highest in rank: & 86\% & 86\%  \\
most important; main: & 14\% & 14\%   \\ \hline
total count:  & 1048 & 1048 \\ \hline
{\it common}: & &  \\ 
as in the phrase `common stock': & 80\% &  84\%  \\
belonging to or shared by 2 or more: & 7\% & 8\%  \\
happening often; usual: & 8\% & 8\%  \\ 
widely known; general; ordinary: & 3\% &   \\ 
of no special quality; ordinary: & 1\% &   \\
same relationship to 2 more or quantities: &$<$ 1\% & \\ \hline
total count:  & 1113 & 1060 \\ \hline
{\it last}: & & \\ 	
on the occasion nearest in the past: & 93\% & 94\%   \\
after all others: & 6\% & 6\%  \\ 
least desirable: & $<$ 1\%& \\ \hline
total count:  & 3187 & 3154 \\ \hline
{\it public}: & &   \\ 
concerning people in general: & 56\% & 68\% \\
concerning the government and people: & 16\%& 19\% \\
not secret or private: & 11\%  & 13\%  \\ 
for the use of everyone: & 8\%  &   \\ 
to become a company: & 6\%  &   \\ 
known to all or many: & 3\%  &   \\ 
as in public TV or public radio & 1\%  &   \\  \hline
total count:  & 871 & 715 \\ \hline
\end{tabular}
\end{center}
\vskip -0.01in
\end{table}

\begin{table}
\begin{center}
\caption{Noun Senses}
\label{tab:nouns}
\begin{tabular}{|lrr|} \hline
\multicolumn{1}{|c}{} & 
\multicolumn{1}{c}{supervised} & 
\multicolumn{1}{c|}{unsupervised} \\ \hline
{\it bill}: & &  \\ 
a proposed law under consideration: &68\% & 68\% \\
a piece of paper money or treasury bill:& 22\% & 22\% \\
a list of things bought and their price: &10\% & 10\% \\ \hline
total count: & 1341 & 1341 \\ \hline
{\it concern}: & &  \\ 
a business; firm:  & 64\%   & 67\% \\
worry; anxiety: & 32\%  & 33\% \\ 
a matter of interest or importance & 3\% & \\
serious care or interest & 2\% & \\ \hline
total count: & 1490  & 1429 \\ \hline  
{\it drug}:  & & \\ 
a medicine; used to make medicine:& 57\% & 57\% \\
a habit-forming substance:& 43\% & 43\% \\ \hline
total count: & 1217  & 1217 \\ \hline  
{\it interest}: & &  \\ 
money paid for the use of money:  &53\% & 59\% \\
a share in a company or business: &21\% & 24\% \\
readiness to give attention: & 15\% & 17\% \\ 
advantage, advancement or favor: & 8\% &  \\ 
activity, etc. that one gives attention to: & 3\% &  \\ 
quality of causing attention to be given to: &  $<$ 1\% &  \\ \hline
total count: & 2367 & 2113 \\ \hline
{\it line}: & & \\ 
a wire connecting telephones: &   & 37\% \\
a cord; cable: &  & 32\% \\
an orderly series: & & 30\% \\ \hline
total count: & 0  & 1149 \\ \hline
\end{tabular}
\end{center}
\vskip -0.01in
\end{table}

\begin{table}
\begin{center}
\caption{Verb Senses}
\label{tab:verbs}
\begin{tabular}{|lrr|} \hline
\multicolumn{1}{|c}{} & 
\multicolumn{1}{c}{supervised} & 
\multicolumn{1}{c|}{unsupervised} \\ \hline
{\it agree}: & & \\ 
to concede after disagreement:& 74\% & 74\% \\
to share the same opinion: & 26\%  & 26\% \\
to be happy together; get on well together: & $<$ 1\% &  \\ \hline
total count: & 1115 & 1109 \\ \hline
{\it close}: & &  \\ 
to (cause to) end: & 68\% & 77\% \\
to (cause to) stop operation: & 20\% & 23\% \\
to close a deal: & 6\% &  \\
to (cause to) shut: & 2\% &  \\
to (cause to) be not open to the public: & 2\% & \\
to come together by making less space between: & 2\% & \\ \hline
total count: & 1535 & 1354 \\ \hline
{\it help}: & & \\ 
to enhance - inanimate object: & 75\%& 79\% \\
to assist - human object: & 20\% & 21\% \\
to make better - human object: & 4\% &  \\
to avoid; prevent; change - inanimate object: & 1\% &  \\ \hline
total count: & 1398 & 1328 \\ \hline
{\it include}: & &  \\ 
to contain in addition to other parts: & 91\% & 91\% \\
to be a part of - human subject: & 9\% & 9\% \\ \hline
total count: & 1526 & 1526 \\ \hline
\end{tabular}
\end{center}
\vskip -0.01in
\end{table}

\section{Feature Sets} \label{sec:feature}

Each sentence containing an ambiguous word is reduced to a vector of 
feature values. One set of features is employed in the supervised
learning experiments and three are used in the unsupervised. All of
these features occur within the sentence in
which the ambiguous word occurs. Extending the features beyond 
sentence boundaries is a potential area for future work. 

\subsection{Supervised Learning Feature Set}

The feature set used in the supervised experiments was developed by
Bruce and Wiebe and is described in  \cite{Bruce95}, \cite{BruceW94A}, 
\cite{BruceW94B}, and \cite{BruceWP96}.  In subsequent discussion this is 
referred to as feature set BW.  This feature set has 
one morphological feature describing the ambiguous word, four 
part--of-speech features describing the surrounding words, and three 
co--occurrence features that indicate if certain key words occur
anywhere within the sentence. 

{\bf Morphology:} This feature represents the morphology of the ambiguous
word.  It is binary for an ambiguous noun and indicates if it is
singular or plural. It shows the tense of an ambiguous verb and has up
to 7 possible  values.  This feature is not used for adjectives. It is
represented by variable $M$.  

{\bf Part of Speech:}
These features represent the part--of--speech of words within $\pm i$ 
positions of the ambiguous word.  Feature set BW contains features
that indicate the part of speech of words 1 and 2 positions to the
left (--) and right (+) of the ambiguous word. Each feature has one of
25 possible values which are derived from the first letter of the Penn
TreeBank tag contained in the ACL/DCI WSJ corpus.  These features are
represented by variables $P_{-2}$, $P_{-1}$, $P_{+1}$, and $P_{+2}$.  

{\bf Co--occurrences:}
These are binary  features that indicate whether or not a
particular word occurs in the sentence with the ambiguous word. 
The values of these features are selected from among the 400 words
that occur most frequently in each word--corpus. The three words
chosen  are most indicative of the sense of the ambiguous word as
judged by a test for independence. These features are represented by
variables $C_1$, $C_2$, and $C_3$ and the words whose occurrence they
represent are shown in Table \ref{tab:cooccurrence}. 

\begin{table}
\begin{center}
\caption{Supervised Co--occurrence features}
\label{tab:cooccurrence}
\begin{tabular}{|l|l|l|l|} \hline
       & $C_1$   & $C_2$ & $C_3$ \\ \hline
agree  & million & that  & to  \\
bill   & auction & discount & treasury  \\
chief  & economist & executive  & officer  \\
close  & at & cents  & trading  \\
common & million & sense  & share  \\
concern& about & million  & that  \\
drug   & company & FDA  & generic  \\
help   & him & not  & then  \\
include& are & be  &  in \\
interest & in & percent & rate  \\
last   & month & week  & year  \\
public & going & offering  & school  \\ \hline
\end{tabular} 
\end{center}
\vskip -0.01in
\end{table}

\newpage
\subsection{Unsupervised Learning Feature Sets}

There are three different feature sets employed in the unsupervised
experiments.  This dissertation evaluates the effect that different
types of features  have on the accuracy of unsupervised learning
algorithms; particular attention is paid to features that occur in
close proximity to the ambiguous word, i.e.,  ``local context''
features.  As the amount of context is increased the size of the
associated event space grows and unsupervised methods require
increasing amounts of computational time and space. 

The unsupervised learning feature sets are  designated  A, B, and C. 
They are composed of combinations of the following five types of features.    

{\bf Morphology:} This feature represents the morphology of ambiguous
nouns and verbs. It is the same as the morphology feature in set BW. 

{\bf Part of Speech:}
As in feature set BW, these features represent the part--of--speech of 
words that occur within 1 and 2 positions of the ambiguous word. 
However, in the unsupervised experiments the range of possible values for 
these features is reduced to five: noun, verb, adjective,  adverb, or 
other.  These crude distinctions are made with the rule--based
part--of--speech tagger incorporated in the Unix command  {\tt  style} 
\cite{Cherry78}.  The tags available in the ACL/DCI WSJ
corpus are not used since such high--quality, detailed
tagging is not generally available for raw text.  These features are 
represented by variables $P5_{-2}$, $P5_{-1}$,  $P5_{+1}$, and $P5_{+2}$. 

{\bf Co--occurrences:} These binary features represent whether or not certain
high frequency words in the sentence with the ambiguous word.
These features differ from the co--occurrence features in set BW since
sense--tagged text is not available to select their values via a test of
independence. Rather,  the words whose occurrences are represented
are determined by the most frequent content words\footnote{Content words 
are defined here to include nouns, pronouns, verbs, adjectives and 
adverbs.} that occur in each word--corpus. Three such features are used. 
$CF_1$ represents the most frequent content word, $CF_2$ the second most 
frequent,  and  $CF_3$ the third.  The words represented by these features 
are shown in Table \ref{tab:surround}. 

\begin{table}
\begin{center}
\caption{Unsupervised Co--occurrence Features}
\label{tab:surround}
\begin{tabular}{|l|l|l|l|} \hline
\multicolumn{1}{|c|}{word} &   
\multicolumn{1}{|c|}{$CF_1$} &   
\multicolumn{1}{|c|}{$CF_2$} &   
\multicolumn{1}{|c|}{$CF_3$} \\ \hline
chief &   officer & executive & president \\ 
common &   share & million & stock \\ 
last  &   year & week & million \\ 
public  &   offering & million & company \\ 
bill &   treasury & billion & house \\ 
concern  &   million &  company & market\\ 
drug &   fda & company & generic \\ 
interest &   rate & million & company \\ 
line &   he & it & telephone \\ 
agree &   million & company & pay \\ 
close &   trading & exchange & stock \\ 
help &   it & say & he \\ 
include &  million & company & year \\ \hline
\end{tabular} 
\end{center}
\vskip -0.01in
\end{table}

{\bf Unrestricted Collocations:}
These features represent the most frequent words that occur within $\pm 2$
positions of the ambiguous word.  These features have 21 possible values. 
Nineteen correspond to the 19 most frequent words that occur in that fixed 
position in the word--corpus.  There is also a value, (none), that 
indicates when the position $i$  to the left or right is occupied by a 
word that is not among the 19 most  frequent, and a value, (null), 
indicating that the position $\pm i$  falls outside the sentence boundary. 
These features are represented by variables $UC_{-2}$, $UC_{-1}$, 
$UC_{+1}$, and $UC_{+2}$. For example, the values of the unrestricted
collocation features for {\it concern} are as follows:
\begin{itemize}
\item $UC_{-2}$: and, the, a, of, to, financial, have, because, an, 
's, real, cause, calif., york, u.s., other, mass., german, jersey, 
(null), (none)
\item $UC_{-1}$ : the, services, of, products, banking, 's,
pharmaceutical, 
energy, their, expressed, electronics, some, biotechnology, 
aerospace, environmental, such, japanese, gas, investment, (null), (none)
\item $UC_{+1}$: about, said, that, over, 's, in, with, had, are, based,
and,
is, has, was, to, for, among, will, did, (null), (none) 
\item $UC_{+2}$: the, said, a, it, in, that, to, n't, is, which, by, 
and, was, has, its, possible, net, but, annual, (null), (none) 
\end{itemize}

{\bf Content Collocations:}
These features represent high frequency content words that occur
within 1 position of the ambiguous word.  The values of these features are
determined by the most frequent content words that occur on either side of
the ambiguous word in the word--corpus. These features are represented by
variables $CC_{-1}$ and $CC_{+1}$. The content collocations 
associated with {\it concern} are as follows: 
\begin{itemize}
\item $CC_{-1}$: services, products, banking, pharmaceutical, energy, 
expressed, electronics, biotechnology, aerospace, environmental, japanese, 
gas, investment, food, chemical, broadcasting, u.s., industrial, growing, 
(null), (none)
\item $CC_{+1}$: said, had, are, based, has, was, did, owned, were,
regarding, 
have, declined, expressed, currently, controlled, bought, announced, 
reported, posted, (null), (none)
\end{itemize}

There is a limitation to frequency based features such as the 
co--occurrences and collocations previously described;  they contain
little information  about low frequency minority senses and  are
skewed towards the majority  sense.  Consider the values of the
co--occurrence features associated with  {\it chief}: {\it officer},
{\it executive} and {\it president}. {\it  Chief}  has a majority
class distribution of 86\% and, not surprisingly,  these three content
words are all indicative of ``highest in rank'', the  majority
sense. However, when using raw text it isn't clear how features  that
are indicative of minority senses can be identified. This remains an
interesting question for future work.  

\newpage
\subsection{Feature Sets and Event Distributions} 

The 4 feature sets used in this dissertation are designated BW, A, B, 
and C. The supervised experiments are conducted with feature set BW
and the unsupervised with A, B, and C. 
Each of these feature sets results in a different event space, i.e., the
set of possible marginal events. The formulation of each feature set as 
well as  the maximum size of the event spaces associated with the
saturated model and the Naive Bayes model are as follows:

\begin{itemize}
\item BW: $M, P_{-2}, P_{-1}, P_{+1}, P_{+2}, C_1, C_2, C_3$ \\
Saturated Event Space:  15,857,856 \\
Naive Bayes Event Space:  534
\item A: $M, P5_{-2}, P5_{-1}, P5_{+1}, P5_{+2}, CF_1, CF_2, CF_3$ \\
Saturated Event Space:  105,000 \\
Naive Bayes Event Space:  99
\item B: $M, UC_{-2}, UC_{-1}, UC_{+1}, UC_{+2}$ \\
Saturated Event Space:  4,084,101\\
Naive Bayes Event Space: 273 
\item C: $M, P5_{-2}, P5_{-1}, P5_{+1}, P5_{+2}, CC_{-1}, CC_{+1}$  \\
Saturated Event Space: 5,788,125\\
Naive Bayes Event Space: 207
\end{itemize}

The minimum size of the event space depends on the number of possible
senses and the value of the morphological feature. It also varies if 
possible values of a feature variable do not occur in the training data. 
For example, if there are 20 possible values for a feature and only 
5 are observed in the training data  the parameter estimates associated
with the 15 non--occurring events will be zero. The degrees of freedom of
models are adjusted to eliminate zero estimates and reduce the size of the
event space further. 

Tables \ref{tab:hchief} through \ref{tab:hinclude} contrast the
size of the event spaces associated with the saturated model and the 
Naive Bayes model. This illustrates that event distributions are very
skewed under the saturated model and that this skewness is reduced,
but not eliminated, with the Naive Bayes model.  The reduction in  model 
complexity results in a smaller number of marginal events that must 
be observed to make parameter estimates. The number of marginal events  
given the saturated model and Naive Bayes are shown in 
the row ``total events''. 

For example, as shown in Table \ref{tab:hinterest}, the number of
marginal events for {\it interest} under the saturated model 
and feature set BW  is approximately 16,000,000
while under Naive Bayes it is 534. Given such a large number of
marginal events under the saturated model it is inevitable that most
of these will not be observed and parameter estimates will be zero since 
the training sample sizes are so small by comparison. However, when the 
model is simplified the number of marginal events is reduced and the 
percentage of marginal events that are observed in the training data 
increases. For {\it interest},  99.9\% of the marginal events under the 
saturated model for feature set BW  are unobserved while under Naive Bayes 
only 36.9\% are never observed. The distribution of the event space for 
all words is smoothed and results in more reliable parameter estimates 
since the majority of possible marginal events are observed in the 
training data.

\begin{table}[t]
\begin{center}
\caption{Event Distribution for Adjective {\it chief}}
\label{tab:hchief}
\begin{tabular}{|l|rrrr|rrrr|} \hline
\multicolumn{1}{|l|}{event} & 
\multicolumn{4}{|c|}{Saturated} & 
\multicolumn{4}{|c|}{Naive Bayes} \\  \cline{2-9}
count   & BW &  A     & B    & C    & BW & A  & B & C\\ \hline
0       & 99.9 & 99.1  & 99.9  & 99.9 & 19.4 & 13.0 & 29.5 & 26.3\\
1--5    &$\ll$0.1&  0.8  & 0.1  & 0.1 & 32.7 & 22.2 & 39.2 & 36.8\\
6--10   &$\ll$0.1&  0.1 &  0.0 & 0.0 & 7.1 & 5.6 & 9.0 & 6.1\\
11--50  &$\ll$0.1&  0.1  &  0.0 & 0.0 & 16.3 & 14.8 & 11.4 & 11.4\\
51--100 & $\ll$0.1 &  0.0  &  0.0 & 0.0 & 9.2 & 5.6 & 4.2 & 6.1\\
101--1000 & 0.0 & 0.0 &  0.0 & 0.0 & 15.3 & 38.9 & 6.6 & 13.2\\ 
1000+  &0.0 &  0.0   &  0.0 &  0.0 & 0.0 & 0.0 & 0.0 & 0.0\\ \hline
total events & 1.4$\times 10^5$ & 1.6$\times 10^4$ &6.7$\times 10^5$ &
6.4$\times 10^5$ & 98 & 54 & 166 & 114 \\
\hline
\end{tabular}
\end{center}
\vskip -0.01in
\end{table}

\begin{table}
\begin{center}
\caption{Event Distribution for Adjective {\it common}}
\label{tab:hcommon}
\begin{tabular}{|l|rrrr|rrrr|} \hline
\multicolumn{1}{|l|}{event} & 
\multicolumn{4}{|c|}{Saturated} & 
\multicolumn{4}{|c|}{Naive Bayes} \\  \cline{2-9}
count   & BW & A     & B    & C  & BW  & A  & B & C\\ \hline
0       & 99.9 & 99.1 &  99.9 & 99.9 & 42.1 & 7.1 & 40.5 & 42.2\\
1--5    & $\ll$0.1 & 0.8  &  $\ll$0.1 &  $\ll$0.1 & 30.6 & 28.6 & 27.8 &
25.5\\
6--10   & $\ll$0.1 & 0.1  &  0.0 &  0.0 & 8.5 & 6.0 & 10.7 & 4.7\\
11--50  & $\ll$0.1 & 0.1  &  0.0 &  0.0 & 11.7 & 26.2 & 12.3 & 15.1\\
51--100 &  0.0 & 0.0  &  0.0 &  0.0 & 2.7 & 15.5 & 5.2 & 6.3\\
101--1000 & 0.0 & 0.0 &  0.0 &  0.0 & 4.4 & 16.7 & 3.6 & 6.3\\ 
1000+  & 0.0  & 0.0   &  0.0 &  0.0 & 0.0 & 0.0 & 0.0 & 0.0\\ \hline
total events& 1.6$\times 10^6$ & 3.0$\times 10^4$ & 1.1$\times
10^6$ & 1.7$\times 10^6$ & 366 & 84  & 252  & 192 \\ \hline
\end{tabular}
\end{center}
\vskip -0.01in
\end{table}

\begin{table}
\begin{center}
\caption{Event Distribution for Adjective {\it last}}
\label{tab:hlast}
\begin{tabular}{|l|rrrr|rrrr|} \hline
\multicolumn{1}{|l|}{event} & 
\multicolumn{4}{|c|}{Saturated} & 
\multicolumn{4}{|c|}{Naive Bayes} \\  \cline{2-9}
count   & BW & A     & B    & C & BW   & A  & B & C\\ \hline
0       & 99.9 & 95.5  & 99.8 & 99.9 & 33.3 & 1.9 & 27.9 & 30.5\\
1--5    & $\ll$0.1 & 3.3  &  0.2 &  0.1 & 23.4 &  17.3 & 12.2 & 10.2\\
6--10   & $\ll$0.1 & 0.4  &  0.0 &  0.0 & 4.7 & 3.8 & 6.4 & 7.8\\
11--50  & $\ll$0.1 & 0.7  &  0.0 &  0.0 & 16.7 & 11.5 & 35.5 & 22.7\\
51--100 &  0.0 & 0.1  &  0.0 &  0.0 & 4.2 & 15.4 & 7.0 & 10.2\\
101--1000 & 0.0  & 0.0 &  0.0 &  0.0 & 14.1 & 32.7 & 8.7 & 14.1\\ 
1000+  & 0.0  & 0.0   &  0.0 &  0.0 & 3.6 & 17.3 & 2.3 & 4.7\\ \hline
total events & 1.0$\times 10^6$ & 8.0$\times 10^3$  & 4.1$\times
10^5$ & 4.8$\times 10^5$ & 192 & 52  & 172  & 128
\\ \hline
\end{tabular}
\end{center}
\vskip -0.01in
\end{table}

\begin{table}
\begin{center}
\caption{Event Distribution for Adjective {\it public}}
\label{tab:hpublic}
\begin{tabular}{|l|rrrr|rrrr|} \hline
\multicolumn{1}{|l|}{event} & 
\multicolumn{4}{|c|}{Saturated} & 
\multicolumn{4}{|c|}{Naive Bayes} \\  \cline{2-9}
count   & BW & A   & B    & C & BW   & A  & B & C\\ \hline
0       & 99.9 &  99.2&  99.9 &  99.9 & 38.3 & 8.3 & 28.5 & 38.3\\
1--5    & $\ll$0.1 &  0.8  &  $\ll$0.1 &  $\ll$0.1 & 32.0 & 20.2 & 37.8 &
25.9\\
6--10   & $\ll$0.1 &  0.0  &  0.0 &  0.0 & 10.7 & 9.5 & 15.7 & 9.0\\
11--50  & $\ll$0.1 &  0.0  &  0.0 &  0.0 & 12.9 & 25.0 & 10.8 & 13.9\\
51--100 & 0.0 &  0.0  &  0.0 &  0.0 & 3.2 & 17.9 & 4.8 & 6.5\\
101--1000 & 0.0 & 0.0 &  0.0 &  0.0 & 2.9 & 19.0 & 2.4 & 6.5\\ 
1000+  & 0.0 & 0.0  &  0.0 &  0.0 & 0.0 & 0.0 & 0.0 & 0.0\\ \hline
total events & 2.3$\times 10^6$ & 3.0$\times 10^4$  & 1.0$\times
10^6$ & 1.9$\times 10^6$ & 441 & 84  & 249  & 201
\\ \hline
\end{tabular}
\end{center}
\vskip -0.01in
\end{table}

\begin{table}
\begin{center}
\caption{Event Distribution for Noun {\it bill}}
\label{tab:hbill}
\begin{tabular}{|l|rrrr|rrrr|} \hline
\multicolumn{1}{|l|}{event} & 
\multicolumn{4}{|c|}{Saturated} & 
\multicolumn{4}{|c|}{Naive Bayes} \\  \cline{2-9}
count & BW  & A     & B    & C & BW   & A  & B & C\\ \hline
0     & 99.9 &  99.2 &  99.9 &  99.9& 15.4 & 10.0 & 30.3 & 38.3\\
1--5  & $\ll$0.1 &  0.8  &  $\ll$0.1 &  $\ll$0.1 & 24.4& 12.2 & 25.7 &
13.4\\
6--10 & $\ll$0.1 &  0.0  &  0.0 &  0.0 & 9.5& 6.7 & 10.7 & 10.9\\
11--50 &$\ll$0.1 &  0.0  &  0.0 &  0.0 &27.4 & 21.1 & 23.4 & 17.9\\
51--100 & 0.0 &  0.0  &  0.0 &  0.0 &11.4 & 17.8 & 5.7 & 8.0\\
101--1000 & 0.0 & 0.0 &  0.0 &  0.0 &11.9 & 32.2 & 4.2 & 11.4\\ 
1000+  & 0.0 &  0.0   &  0.0 &  0.0 &0.0 & 0.0 & 0.0 & 0.0\\ \hline
total events & 2.2$\times 10^6$ & 6.0$\times 10^4$   & 2.6$\times
10^6$ & 3.5$\times 10^6$ & 201 & 90  & 261  & 201
\\ \hline
\end{tabular}
\end{center}
\vskip -0.01in
\end{table}

\begin{table}
\begin{center}
\caption{Event Distribution for Noun {\it concern}}
\label{tab:hconcern}
\begin{tabular}{|l|rrrr|rrrr|} \hline
\multicolumn{1}{|l|}{event} & 
\multicolumn{4}{|c|}{Saturated} & 
\multicolumn{4}{|c|}{Naive Bayes} \\  \cline{2-9}
count & BW  & A     & B    & C & BW & A  & B & C\\ \hline
0     & 99.9 &  97.7  & 99.9 & 99.9 & 25.4 & 3.6 & 22.8 & 29.0\\
1--5  &  $\ll$0.1 &  2.1  &  0.1 &  $\ll$0.1 & 30.6 & 3.6 & 19.1 & 13.7\\
6--10 &  $\ll$0.1 &  0.1  &  0.0 &  0.0 &  8.6 & 7.1 & 17.9 & 16.1\\
11--50 & $\ll$0.1  &  0.1  &  0.0 &  0.0 & 19.4 & 25.0 & 27.8 & 17.7\\
51--100 & 0.0 &  0.0  &  0.0 &  0.0 & 4.1 & 7.1 & 2.5 & 2.4\\
101--1000 & 0.0 & 0.0 &  0.0 &  0.0 & 11.9 & 53.6 & 9.9 & 21.0\\ 
1000+  & 0.0 &  0.0   &  0.0 &  0.0 & 0.0 & 0.0 & 0.0 & 0.0\\ \hline
total events & 2.9$\times 10^6$ & 2.0$\times 10^4$ & 6.1$\times
10^5$ & 1.0$\times 10^6$ & 268 & 56  & 162  & 124
\\ \hline
\end{tabular}
\end{center}
\vskip -0.01in
\end{table}

\begin{table}
\begin{center}
\caption{Event Distribution for Noun {\it drug}}
\label{tab:hdrug}
\begin{tabular}{|l|rrrr|rrrr|} \hline
\multicolumn{1}{|l|}{event} & 
\multicolumn{4}{|c|}{Saturated} & 
\multicolumn{4}{|c|}{Naive Bayes} \\  \cline{2-9}
count  & BW & A  & B    & C & BW   & A  & B & C\\ \hline
0      & 99.9 &  98.6  &  99.9 &  99.9 & 6.6 & 3.4 & 15.1 & 25.8\\
1--5   & $\ll$0.1 &  1.3  &  0.1 &  $\ll$0.1 & 17.6& 10.3 & 21.1 & 18.0\\
6--10  & $\ll$0.1 &  0.1  &  0.0 &  0.0 & 11.0& 5.2 & 19.9 & 12.5\\
11--50 & $\ll$0.1 &  0.1  &  0.0 &  0.0 & 33.1& 19.0 & 34.3 & 19.5\\
51--100& 0.0 &  0.0  &  0.0 &  0.0 & 14.0& 15.5 & 1.8 & 5.5\\
101--1000& 0.0 & 0.0 &  0.0 &  0.0 & 17.6& 46.6 & 7.8 & 18.8\\ 
1000+  & 0.0 &  0.0   &  0.0 &  0.0 & 0.0& 0.0 & 0.0 & 0.0\\ \hline
total events & 2.3$\times 10^6$ & 3.0$\times 10^4$ & 9.6$\times
10^5$ & 1.6$\times 10^6$ & 136 & 58  & 166  & 128 \\
\hline
\end{tabular}
\end{center}
\vskip -0.01in
\end{table}

\begin{table}
\begin{center}
\caption{Event Distribution for Noun {\it interest}}
\label{tab:hinterest}
\begin{tabular}{|l|rrrr|rrrr|} \hline
\multicolumn{1}{|l|}{event} & 
\multicolumn{4}{|c|}{Saturated} & 
\multicolumn{4}{|c|}{Naive Bayes} \\  \cline{2-9}
count  & BW & A     & B    & C  & BW  & A  & B & C\\ \hline
0      & 99.9 & 98.7  & 99.9 & 99.9 & 36.9& 6.9  & 22.5 & 33.3\\
1--5   &  $\ll$0.1&  1.1  &  $\ll$0.1 &  0.1 & 26.2& 8.0  & 25.7 &  5.2\\
6--10  &  $\ll$0.1 &  0.1  &  0.0 &  0.0 & 6.0& 4.6  & 11.6 &  3.0\\
11--50 &  $\ll$0.1 &  0.1  &  0.0 &  0.0 & 16.7& 16.1 & 26.1 & 20.0\\
51--100 & 0.0 &  0.0  &  0.0 &  0.0 & 6.2& 17.2 &  5.6 & 11.1\\
101--1000 & 0.0 & 0.0 &  0.0 &  0.0 & 7.5& 42.5 &  8.0 & 25.9\\ 
1000+  & 0.0 &  0.0   &  0.0 &  0.0 & 0.6&  4.6 &  0.4 &  1.5\\ \hline
total events & 1.6$\times 10^7$ & 4.5$\times 10^4$ & 1.4$\times
10^6$ & 6.8$\times 10^5$ & 534 & 87 & 249 &135 \\ \hline
\end{tabular}
\end{center}
\vskip -0.01in
\end{table}

\begin{table}
\begin{center}
\caption{Event Distribution for Noun {\it line}}
\label{tab:hline}
\begin{tabular}{|l|rrrr|rrrr|} \hline
\multicolumn{1}{|l|}{event} & 
\multicolumn{4}{|c|}{Saturated} & 
\multicolumn{4}{|c|}{Naive Bayes} \\  \cline{2-9}
count  & BW & A     & B    & C & BW   & A  & B & C\\ \hline
0      & na & 97.9  & 99.9 & 99.9 & na & 2.4  & 22.4 & 34.4\\
1--5   & na &  2.0  &  0.1 &  $\ll$0.1 & na & 9.5  & 34.6 & 20.4\\
6--10  & na &  0.1  &  0.0 &  0.0 & na & 4.8  & 14.2 &  7.5\\
11--50 & na &  0.0  &  0.0 &  0.0 & na & 19.0 & 18.7 & 12.9\\
51--100 & na &  0.0  &  0.0 &  0.0 & na & 22.6 &  2.0 & 9.7\\
101--1000 & na & 0.0 &  0.0 &  0.0 & na & 41.7 &  8.1 & 15.1\\ 
1000+  & na &  0.0   &  0.0 &  0.0 & na &  0.0 &  0.0 &  0.0\\ \hline
total events & na & 3.0$\times 10^4$ & 9.6$\times 10^5$ & 1.5
$\times 10^6$ & na & 84  & 246  &186 \\ \hline
\end{tabular}
\end{center}
\vskip -0.01in
\end{table}

\begin{table}
\begin{center}
\caption{Event Distribution for Verb {\it agree}}
\label{tab:hagree}
\begin{tabular}{|l|rrrr|rrrr|} \hline
\multicolumn{1}{|l|}{event} & 
\multicolumn{4}{|c|}{Saturated} & 
\multicolumn{4}{|c|}{Naive Bayes} \\  \cline{2-9}
count  & BW & A     & B    & C & BW   & A  & B & C\\ \hline
0      & 99.9 & 99.4  & 99.9 & 99.9 & 36.2& 3.1  & 16.5 & 12.3\\
1--5   &  $\ll$0.1 &  0.5  &  $\ll$0.1 &  $\ll$0.1 & 23.2& 10.9  & 29.4 &
34.6\\
6--10  &  $\ll$0.1 &  0.1  &  0.0 &  0.0 & 6.8 & 6.3  & 16.5 &  7.7\\
11--50 &  $\ll$0.1 &  0.0  &  0.0 &  0.0 & 18.4 & 28.1 & 24.7 & 21.5\\
51--100 & 0.0 &  0.0  &  0.0 &  0.0 & 3.9 & 14.1 &  7.1 & 8.5\\
101--1000 & 0.0 & 0.0 &  0.0 &  0.0 & 11.1 & 37.5 &  5.9 & 15.4\\ 
1000+  & 0.0 &  0.0   &  0.0 &  0.0 & 0.5 &  0.0 &  0.0 &  0.0\\ \hline
total events & 5.1$\times 10^6$ & 2.8$\times 10^6$ & 1.8$\times
10^6$  & 2.8$\times 10^6$ & 207
& 64 & 170 &
130 \\ \hline
\end{tabular}
\end{center}
\vskip -0.01in
\end{table}

\begin{table}
\begin{center}
\caption{Event Distribution for Verb {\it close}}
\label{tab:hclose}
\begin{tabular}{|l|rrrr|rrrr|} \hline
\multicolumn{1}{|l|}{event} & 
\multicolumn{4}{|c|}{Saturated} & 
\multicolumn{4}{|c|}{Naive Bayes} \\  \cline{2-9}
count & BW  & A     & B    & C & BW   & A  & B & C\\ \hline
0     & 99.9 & 99.4 & 99.9 & 99.9 & 28.6 & 1.5  & 25.3 & 22.8\\
1--5  & $\ll$0.1 &  0.6  &  $\ll$0.1 &  $\ll$0.1 & 36.2 & 16.7  & 23.0 &
17.6\\
6--10 & $\ll$0.1 &  0.0  &  0.0 &  0.0 & 7.4 & 6.1  & 9.6 &  11.0\\
11--50 & $\ll$0.1 &  0.0 &  0.0 &  0.0 & 18.6 & 24.2 & 29.8 & 25.0\\
51--100 & 0.0  &  0.0 &  0.0 &  0.0 & 3.6 & 15.2 &  4.5 & 8.8\\
101--1000 & 0.0 & 0.0 &  0.0 &  0.0 & 5.7 & 36.4 &  7.9 & 14.7\\ 
1000+  & 0.0 &  0.0   &  0.0 &  0.0 & 0.0 &  0.0 &  0.0 &  0.0\\ \hline
total events & 9.5$\times 10^6$ & 7.0$\times 10^4$ & 2.5$\times
10^6$ & 3.7$\times 10^6$ & 420 & 66 & 178 &136 \\ \hline
\end{tabular}
\end{center}
\vskip -0.01in
\end{table}

\begin{table}
\begin{center}
\caption{Event Distribution for Verb {\it help}}
\label{tab:hhelp}
\begin{tabular}{|l|rrrr|rrrr|} \hline
\multicolumn{1}{|l|}{event} & 
\multicolumn{4}{|c|}{Saturated} & 
\multicolumn{4}{|c|}{Naive Bayes} \\  \cline{2-9}
count   & BW & A     & B    & C  & BW  & A  & B & C\\ \hline
0       &99.9 & 98.6  & 99.9 & 99.9 & 28.7 & 0.0  & 8.6 & 11.2\\
1--5    & $\ll$0.1 &  1.4  &  $\ll$0.1 &  $\ll$0.1 & 27.6 & 1.6  & 31.6 &
28.4\\
6--10   & $\ll$0.1 &  0.0  &  0.0 &  0.0 & 8.1 &  3.2  & 23.0 &  13.4\\
11--50  & $\ll$0.1 &  0.0  &  0.0 &  0.0 & 21.3& 33.9 & 25.9 & 23.1\\
51--100 & 0.0 &  0.0  &  0.0 &  0.0 & 6.6 & 17.7 &  3.4 & 7.5\\
101--1000 & 0.0 & 0.0 &  0.0 &  0.0 & 7.0 & 43.5 &  7.5 & 16.4\\ 
1000+  &0.0 &  0.0   &  0.0 &  0.0 & 0.7 &  0.0 &  0.0 &  0.0\\ \hline
total events & 6.9$\times 10^6$ & 4.8$\times 10^4$ & 2.0$\times
10^6$ & 2.6$\times 10^6$ & 272 & 62 & 174 & 134 \\ \hline
\end{tabular}
\end{center}
\vskip -0.01in
\end{table}

\begin{table}
\begin{center}
\caption{Event Distribution for Verb {\it include}}
\label{tab:hinclude}
\begin{tabular}{|l|rrrr|rrrr|} \hline
\multicolumn{1}{|l|}{event} & 
\multicolumn{4}{|c|}{Saturated} & 
\multicolumn{4}{|c|}{Naive Bayes} \\  \cline{2-9}
count  & BW & A     & B    & C  & BW  & A  & B & C\\ \hline
0      & 99.9 & 98.8  & 99.9 & 99.9 & 13.7 & 1.6  & 25.9 & 23.1 \\
1--5   &  $\ll$0.1&  1.1  &  $\ll$0.1 &  $\ll$0.1 & 26.0 & 12.5  & 18.4 &
23.1 \\
6--10  &  $\ll$0.1&  0.1  &  0.0 &  0.0 & 6.8 & 6.3  & 15.5 &  14.9\\
11--50 &  0.0&  0.0  &  0.0 &  0.0 & 26.7 & 28.1 & 26.4 & 17.9\\
51--100 & 0.0&  0.0  &  0.0 &  0.0 & 10.3 & 7.8 &  5.2 & 3.7\\
101--1000 & 0.0& 0.0 &  0.0 &  0.0 & 15.1 & 39.1 &  8.0 & 15.7 \\ 
1000+  & 0.0 &  0.0   &  0.0 &  0.0 & 1.4 &  0.0 &  0.6 &  1.5 \\ \hline
total events & 3.9$\times 10^6$ & 6.0$\times 10^4$ & 2.0$\times
10^6$ & 3.2$\times 10^6$ & 146 & 61 & 174 & 134 \\ \hline
\end{tabular}
\end{center}
\vskip -0.01in
\end{table}


\chapter{SUPERVISED LEARNING EXPERIMENTAL RESULTS}

The theoretical foundations of sequential model selection and the
Naive Mix are introduced in Chapter 3. This chapter
discusses four experiments that evaluate these
methods.\footnote{The freely available software package CoCo
\cite{Badsberg95} was used in conjunction with the Class.3.0 
classifier \cite{Pedersen97C} for all the model selection
experiments.}
The principal measure is disambiguation accuracy,
the percentage of ambiguous words in a held--out test sample that are
disambiguated correctly.  

The first experiment measures the accuracy of models selected using
various combinations of search strategy and evaluation criterion. The
second compares the accuracy of the Naive Mix to several leading
machine learning algorithms. The third experiment studies the learning
rate of the most accurate methods from the first two experiments. The
final experiment decomposes the overall classification error of two of
the most accurate methods into more fundamental components. 

\section{Experiment 1: Sequential Model Selection} \label{sec:ms}

In the first experiment, each of the eight possible combinations
of search strategy and evaluation criterion as described in Chapter
3 are utilized to select a probabilistic model of
disambiguation for each word. 

The accuracy of each model is evaluated via {\it 10--fold cross
validation}. All of the sense--tagged examples for a word  are
randomly shuffled  and divided into 10 equal folds. Nine
folds  are used as the training sample and the remaining fold acts as
a held--out test set. This process is repeated 10 times so that each
fold serves as the test set once. The disambiguation accuracy 
for each word is the average accuracy across all 10 test sets. 

Within this first experiment there are four separate analyses
performed. The first examines the overall disambiguation accuracy of
the selected models for each word. The second evaluates model
complexity by comparing the number of interactions in the various
models. The third assesses the robustness of the selection process
relative to changes in the search strategy. The fourth and final
analysis is a case study of the model selection process for a single
word.   

All of these evaluations assume that the overall objective of
sequential model selection is to automatically stop the search process
at an accurate model of disambiguation. However, there are
alternatives to this orientation. For example, Bruce and
Wiebe (e.g., \cite{Bruce95}, \cite{BruceW94A}, and \cite{BruceW94B})
use backward search and the exact conditional test to generate a
sequence of models beginning with the saturated model and concluding
with Naive Bayes. The most accurate model is 
selected from this sequence using a test of predictive accuracy. 

\subsection{Overall Accuracy}

Table \ref{tab:result1} shows the accuracy and standard deviation of 
models selected using each evaluation criterion with forward (F)
and backward (S) sequential search. The accuracy of Naive Bayes and
the majority classifier are also reported since they serve as simple
benchmarks; neither performs a model search but rather rely upon
assumed parametric forms. 
When averaged over all twelve words, Naive Bayes and FSS AIC are the
most accurate approaches.\footnote{Each combination of strategy and
criterion is sometimes referred to in an abbreviated form. For
example, the combination of a forward sequential search and Akaike's
Information Criteria is called FSS AIC.}  However, the differences
between Naive Bayes, FSS AIC, and BSS AIC are not statistically
significant for any word. Throughout this evaluation, judgments
regarding the significance of differences are made using a 
two--sided pairwise t--test where $p= .01$.   

\begin{table}
\begin{center}
\caption{Sequential Model Selection Accuracy}
\label{tab:result1}
\begin{tabular}{|l|c|c|c|c|c|c|c|} \hline
        & Majority  & Naive     & & $G^2 \sim \chi^2$ & exact & AIC & BIC\\
     &      & Bayes     & &  &  & &\\ \hline
agree   & .777 .032 & .930 .026 & B   & .914 .018   & .915 .019   & .909 .026 & .924 .023  \\ 
        &           &           & F   & .916 .026   & .896 .030   & .911 .026 & .921 .024  \\ \hline
bill    & .681 .044 & .865 .026 & B   & {\it .634 .073}   & {\it .612
.022} & .836 .036 & .850 .034  \\ 
        &           &           & F   & .778 .048   & {\it .625 .039}   &
.851 .029 & .851 .041 \\ \hline
chief   & .862 .026 & .943 .015 & B   & .945 .020   & .895 .033   & .945 .020 & .936 .020 \\
        &           &           & F   & .926 .027   & .891 .043   & .939 .020 & .943 .021 \\
\hline
\end{tabular}
\end{center}
\vskip -0.01in
\end{table}

A reasonable lower bound on supervised disambiguation algorithms is the
accuracy attained by the majority classifier. The majority classifier
is based on the model of independence and classifies each instance of
an ambiguous word with the most frequent sense in the training
data. 

Neither AIC nor BIC ever selects a model that results in accuracy
significantly less than the lower bound. This is true for both 
forward and backward searches.
However, FSS exact conditional has accuracy significantly less than
the lower bound for four words and BSS exact conditional has accuracy
below the lower bound for two words. FSS $G^2 \sim \chi^2$ and BSS
$G^2 \sim \chi^2$ are significantly less accurate than the lower bound
for one and two words respectively. These cases are italicized in
Table \ref{tab:result1}. 

This behavior is suggestive of a difficulty in using significance
tests as evaluation criteria when the objective of model selection is
to automatically stop the search process at an accurate model of
disambiguation. The value of $\alpha$ determines when the selection
process will stop; unfortunately there is no single value of $\alpha$
that leads to consistent results. In this experiment the $\alpha$
values .01, .05, .001, and .0001 are evaluated and .0001 is found to
select the most accurate models overall. However, there is
considerable variation from word to word and improved results are
likely if $\alpha$ is adjusted for each word. 

It is generally expected that the accuracy of the majority classifier
will be improved upon by more sophisticated models. The majority
classifier does not take into account any of the available contextual
information when performing disambiguation; presumably a model that
does will prove to be more accurate. 

However, there are four words where no method ever significantly
improves upon the majority classifier; {\it help}, {\it include}, {\it
last}, and  {\it public}. Two of these words, {\it last} and {\it
include}, have majority senses of over 90\% so significant improvement
over the lower bound is not likely. But the majority senses of {\it public}
and {\it help} are 56\% and 75\% so there is certainly
room for improvement. However, the most accurate models for these
words have 4 and 6 interactions and disregard most of the features in
set BW. For these two words the set of features may need to be
modified to disambiguate at higher levels of accuracy. 

\newpage
\subsection{Model Complexity}

The number of interactions in a model is a general indicator of 
the complexity of the model. Given $n$ features,  the saturated model
has $\frac{n^2 - n}{2}$ interactions, Naive Bayes has $n$
interactions, and the model of independence has 0. Table
\ref{tab:result1a} shows the number of interactions in the models
selected by each combination of evaluation criterion and search
strategy. The number of interactions in Naive Bayes is included as a
point of comparison although this is not a selected model.

This table shows that BIC and $G^2 \sim \chi^2$ often select models
with fewer interactions than either AIC or the exact conditional
test. However, these models also result in reduced accuracy when
compared to AIC. Since BIC assesses a greater penalty on complexity
than AIC, it has a stronger bias towards less complex models.  As a
result, BSS BIC is more aggressive in removing interactions than BSS
AIC; similarly FSS BIC is more conservative than FSS AIC in adding
them.    

\begin{table}
\begin{center}
\caption{Complexity of Selected Models}
\label{tab:result1a}
\begin{tabular}{|l|c|c|c|c|c|c|} \hline
        & Naive     & & $G^2 \sim \chi^2$ & exact & AIC & BIC\\
        & Bayes     & & .0001 & .0001 & &\\ \hline
agree   &  9 & B   & 8   & 10   & 15 & 9  \\ 
        &    & F   & 12  & 15   &  13 & 7  \\ \hline
bill    &  9 & B   & 22  & 25   &  26 & 7   \\ 
        &    & F   & 20  & 28   &  20 & 11 \\ \hline
chief   &  8 & B   & 6   & 17   &  14 & 6 \\
        &    & F   & 6   & 18   &  14 & 7 \\ \hline
close   &  9 & B   & 12  & 13   &  13 & 3 \\
        &    & F   & 13  & 19   &  10 & 3 \\ \hline
common  &  8 & B   & 4   & 10   &  7  & 2 \\ 
        &    & F   & 4   & 16   &  7  & 2 \\ \hline
concern &  9 & B   & 5   & 15   & 16  & 6 \\
        &    & F   & 17  & 24   & 13  & 9 \\ \hline
drug    &  9 & B   & 10  & 7    & 14  & 9 \\ 
        &    & F   & 10  & 19   & 12  & 9 \\ \hline
help    &  9 & B   & 7   & 6    & 6   & 4 \\ 
        &    & F   & 3   & 9    & 4   & 4 \\ \hline
include &  9 & B   & 6   & 3    & 16  & 8 \\ 
        &    & F   & 6   & 22   & 9   & 9 \\ \hline
interest&  9 & B   & 24  & 24   & 21  & 6  \\ 
        &    & F   & 22  & 32   & 15  & 4 \\ \hline
last    &  8 & B   & 8   & 9    & 14  & 9 \\ 
        &    & F   & 15  & 18   & 14  & 2 \\ \hline
public  &  8 & B   & 7   & 9    & 8   & 3 \\ 
        &    & F   & 6   & 11   & 6   & 3 \\ \hline \hline
average  &  9 & B  & 10  & 12   & 14  & 6 \\ 
         &    & F  & 11  & 19   & 11  & 6 \\ \hline
\end{tabular}
\end{center}
\vskip -0.01in
\end{table}

The comparable levels of accuracy among models selected with the
information criteria and Naive Bayes are curious since Table
\ref{tab:result1a} shows that these methods select models of
differing complexity. For example, the model selected for {\it bill}
by FSS AIC has 20 interactions, the model selected by FSS BIC  has
11, and Naive Bayes has only 9. However, the accuracy of these 3
models is nearly identical. 

The fact that models of differing levels of complexity  yield similar
levels of accuracy demonstrates that model selection is an uncertain
enterprise. In other words, there is not a single model for a word
that will result in overall superior disambiguation accuracy. This
motivates the development of the Naive Mix, an extension to the
sequential model selection process that is evaluated later in this
chapter.  

\subsection{Model Selection as a Robust Process}

A model selection process is robust when models of similar accuracy
are selected as a result of both a forward and a backward search
using the same evaluation criterion.  In general the information
criteria result in a robust selection process while the significance
tests do not.      

\begin{figure}
\begin{center}
\setlength{\unitlength}{0.240900pt}
\ifx\plotpoint\undefined\newsavebox{\plotpoint}\fi
\sbox{\plotpoint}{\rule[-0.200pt]{0.400pt}{0.400pt}}%
\begin{picture}(1500,900)(0,0)
\font\gnuplot=cmr10 at 10pt
\gnuplot
\sbox{\plotpoint}{\rule[-0.200pt]{0.400pt}{0.400pt}}%
\put(220.0,113.0){\rule[-0.200pt]{4.818pt}{0.400pt}}
\put(198,113){\makebox(0,0)[r]{$0.4$}}
\put(1416.0,113.0){\rule[-0.200pt]{4.818pt}{0.400pt}}
\put(220.0,240.0){\rule[-0.200pt]{4.818pt}{0.400pt}}
\put(198,240){\makebox(0,0)[r]{$0.5$}}
\put(1416.0,240.0){\rule[-0.200pt]{4.818pt}{0.400pt}}
\put(220.0,368.0){\rule[-0.200pt]{4.818pt}{0.400pt}}
\put(198,368){\makebox(0,0)[r]{$0.6$}}
\put(1416.0,368.0){\rule[-0.200pt]{4.818pt}{0.400pt}}
\put(220.0,495.0){\rule[-0.200pt]{4.818pt}{0.400pt}}
\put(198,495){\makebox(0,0)[r]{$0.7$}}
\put(1416.0,495.0){\rule[-0.200pt]{4.818pt}{0.400pt}}
\put(220.0,622.0){\rule[-0.200pt]{4.818pt}{0.400pt}}
\put(198,622){\makebox(0,0)[r]{$0.8$}}
\put(1416.0,622.0){\rule[-0.200pt]{4.818pt}{0.400pt}}
\put(220.0,750.0){\rule[-0.200pt]{4.818pt}{0.400pt}}
\put(198,750){\makebox(0,0)[r]{$0.9$}}
\put(1416.0,750.0){\rule[-0.200pt]{4.818pt}{0.400pt}}
\put(220.0,877.0){\rule[-0.200pt]{4.818pt}{0.400pt}}
\put(198,877){\makebox(0,0)[r]{$1$}}
\put(1416.0,877.0){\rule[-0.200pt]{4.818pt}{0.400pt}}
\put(220.0,113.0){\rule[-0.200pt]{0.400pt}{4.818pt}}
\put(220,68){\makebox(0,0){$0.4$}}
\put(220.0,857.0){\rule[-0.200pt]{0.400pt}{4.818pt}}
\put(423.0,113.0){\rule[-0.200pt]{0.400pt}{4.818pt}}
\put(423,68){\makebox(0,0){$0.5$}}
\put(423.0,857.0){\rule[-0.200pt]{0.400pt}{4.818pt}}
\put(625.0,113.0){\rule[-0.200pt]{0.400pt}{4.818pt}}
\put(625,68){\makebox(0,0){$0.6$}}
\put(625.0,857.0){\rule[-0.200pt]{0.400pt}{4.818pt}}
\put(828.0,113.0){\rule[-0.200pt]{0.400pt}{4.818pt}}
\put(828,68){\makebox(0,0){$0.7$}}
\put(828.0,857.0){\rule[-0.200pt]{0.400pt}{4.818pt}}
\put(1031.0,113.0){\rule[-0.200pt]{0.400pt}{4.818pt}}
\put(1031,68){\makebox(0,0){$0.8$}}
\put(1031.0,857.0){\rule[-0.200pt]{0.400pt}{4.818pt}}
\put(1233.0,113.0){\rule[-0.200pt]{0.400pt}{4.818pt}}
\put(1233,68){\makebox(0,0){$0.9$}}
\put(1233.0,857.0){\rule[-0.200pt]{0.400pt}{4.818pt}}
\put(1436.0,113.0){\rule[-0.200pt]{0.400pt}{4.818pt}}
\put(1436,68){\makebox(0,0){$1$}}
\put(1436.0,857.0){\rule[-0.200pt]{0.400pt}{4.818pt}}
\put(220.0,113.0){\rule[-0.200pt]{292.934pt}{0.400pt}}
\put(1436.0,113.0){\rule[-0.200pt]{0.400pt}{184.048pt}}
\put(220.0,877.0){\rule[-0.200pt]{292.934pt}{0.400pt}}
\put(45,495){\makebox(0,0){FSS}}
\put(828,23){\makebox(0,0){BSS}}
\put(220.0,113.0){\rule[-0.200pt]{0.400pt}{184.048pt}}
\put(625,813){\makebox(0,0)[r]{AIC}}
\put(669,813){\raisebox{-.8pt}{\makebox(0,0){$\Diamond$}}}
\put(1278,796){\raisebox{-.8pt}{\makebox(0,0){$\Diamond$}}}
\put(1133,696){\raisebox{-.8pt}{\makebox(0,0){$\Diamond$}}}
\put(1364,832){\raisebox{-.8pt}{\makebox(0,0){$\Diamond$}}}
\put(1113,674){\raisebox{-.8pt}{\makebox(0,0){$\Diamond$}}}
\put(1225,744){\raisebox{-.8pt}{\makebox(0,0){$\Diamond$}}}
\put(1110,655){\raisebox{-.8pt}{\makebox(0,0){$\Diamond$}}}
\put(1120,637){\raisebox{-.8pt}{\makebox(0,0){$\Diamond$}}}
\put(1013,611){\raisebox{-.8pt}{\makebox(0,0){$\Diamond$}}}
\put(1312,807){\raisebox{-.8pt}{\makebox(0,0){$\Diamond$}}}
\put(913,538){\raisebox{-.8pt}{\makebox(0,0){$\Diamond$}}}
\put(1318,803){\raisebox{-.8pt}{\makebox(0,0){$\Diamond$}}}
\put(457,276){\raisebox{-.8pt}{\makebox(0,0){$\Diamond$}}}
\sbox{\plotpoint}{\rule[-0.500pt]{1.000pt}{1.000pt}}%
\put(625,768){\makebox(0,0)[r]{BIC}}
\put(669,768){\makebox(0,0){$+$}}
\put(1321,805){\makebox(0,0){$+$}}
\put(1194,677){\makebox(0,0){$+$}}
\put(1346,832){\makebox(0,0){$+$}}
\put(946,569){\makebox(0,0){$+$}}
\put(1189,722){\makebox(0,0){$+$}}
\put(865,672){\makebox(0,0){$+$}}
\put(987,595){\makebox(0,0){$+$}}
\put(940,566){\makebox(0,0){$+$}}
\put(1312,791){\makebox(0,0){$+$}}
\put(688,486){\makebox(0,0){$+$}}
\put(1318,799){\makebox(0,0){$+$}}
\put(525,305){\makebox(0,0){$+$}}
\put(625,723){\makebox(0,0)[r]{BSS=FSS}}
\multiput(647,723)(20.756,0.000){4}{\usebox{\plotpoint}}
\put(713,723){\usebox{\plotpoint}}
\put(220,113){\usebox{\plotpoint}}
\put(220.00,113.00){\usebox{\plotpoint}}
\put(237.58,124.00){\usebox{\plotpoint}}
\put(255.25,134.84){\usebox{\plotpoint}}
\multiput(257,136)(17.270,11.513){0}{\usebox{\plotpoint}}
\put(272.52,146.35){\usebox{\plotpoint}}
\put(290.31,157.01){\usebox{\plotpoint}}
\multiput(294,159)(17.270,11.513){0}{\usebox{\plotpoint}}
\put(307.78,168.19){\usebox{\plotpoint}}
\put(325.46,179.02){\usebox{\plotpoint}}
\multiput(331,182)(17.270,11.513){0}{\usebox{\plotpoint}}
\put(343.03,190.02){\usebox{\plotpoint}}
\put(360.30,201.53){\usebox{\plotpoint}}
\put(378.19,212.02){\usebox{\plotpoint}}
\multiput(380,213)(17.270,11.513){0}{\usebox{\plotpoint}}
\put(395.56,223.37){\usebox{\plotpoint}}
\put(413.34,234.03){\usebox{\plotpoint}}
\multiput(417,236)(17.270,11.513){0}{\usebox{\plotpoint}}
\put(430.81,245.21){\usebox{\plotpoint}}
\put(448.08,256.72){\usebox{\plotpoint}}
\multiput(453,260)(18.275,9.840){0}{\usebox{\plotpoint}}
\put(466.06,267.04){\usebox{\plotpoint}}
\put(483.33,278.56){\usebox{\plotpoint}}
\put(501.22,289.04){\usebox{\plotpoint}}
\multiput(503,290)(17.270,11.513){0}{\usebox{\plotpoint}}
\put(518.59,300.39){\usebox{\plotpoint}}
\put(535.86,311.91){\usebox{\plotpoint}}
\multiput(539,314)(18.275,9.840){0}{\usebox{\plotpoint}}
\put(553.84,322.23){\usebox{\plotpoint}}
\put(571.11,333.74){\usebox{\plotpoint}}
\multiput(576,337)(17.270,11.513){0}{\usebox{\plotpoint}}
\put(588.40,345.22){\usebox{\plotpoint}}
\put(606.37,355.58){\usebox{\plotpoint}}
\put(623.64,367.09){\usebox{\plotpoint}}
\multiput(625,368)(18.275,9.840){0}{\usebox{\plotpoint}}
\put(641.62,377.41){\usebox{\plotpoint}}
\put(658.89,388.93){\usebox{\plotpoint}}
\multiput(662,391)(17.270,11.513){0}{\usebox{\plotpoint}}
\put(676.28,400.23){\usebox{\plotpoint}}
\put(694.14,410.76){\usebox{\plotpoint}}
\multiput(699,414)(17.270,11.513){0}{\usebox{\plotpoint}}
\put(711.44,422.24){\usebox{\plotpoint}}
\put(729.40,432.60){\usebox{\plotpoint}}
\put(746.67,444.11){\usebox{\plotpoint}}
\multiput(748,445)(17.270,11.513){0}{\usebox{\plotpoint}}
\put(764.17,455.24){\usebox{\plotpoint}}
\put(781.92,465.95){\usebox{\plotpoint}}
\multiput(785,468)(17.270,11.513){0}{\usebox{\plotpoint}}
\put(799.32,477.25){\usebox{\plotpoint}}
\put(817.18,487.78){\usebox{\plotpoint}}
\multiput(822,491)(17.270,11.513){0}{\usebox{\plotpoint}}
\put(834.45,499.30){\usebox{\plotpoint}}
\put(852.05,510.26){\usebox{\plotpoint}}
\put(869.70,521.13){\usebox{\plotpoint}}
\multiput(871,522)(17.270,11.513){0}{\usebox{\plotpoint}}
\put(887.20,532.26){\usebox{\plotpoint}}
\put(904.95,542.97){\usebox{\plotpoint}}
\multiput(908,545)(17.270,11.513){0}{\usebox{\plotpoint}}
\put(922.22,554.48){\usebox{\plotpoint}}
\put(939.93,565.27){\usebox{\plotpoint}}
\multiput(945,568)(17.270,11.513){0}{\usebox{\plotpoint}}
\put(957.48,576.32){\usebox{\plotpoint}}
\put(975.08,587.27){\usebox{\plotpoint}}
\put(992.73,598.15){\usebox{\plotpoint}}
\multiput(994,599)(17.270,11.513){0}{\usebox{\plotpoint}}
\put(1010.00,609.67){\usebox{\plotpoint}}
\put(1027.81,620.28){\usebox{\plotpoint}}
\multiput(1031,622)(17.270,11.513){0}{\usebox{\plotpoint}}
\put(1045.26,631.50){\usebox{\plotpoint}}
\put(1062.96,642.29){\usebox{\plotpoint}}
\multiput(1068,645)(17.270,11.513){0}{\usebox{\plotpoint}}
\put(1080.51,653.34){\usebox{\plotpoint}}
\put(1097.78,664.85){\usebox{\plotpoint}}
\put(1115.69,675.30){\usebox{\plotpoint}}
\multiput(1117,676)(17.270,11.513){0}{\usebox{\plotpoint}}
\put(1133.03,686.69){\usebox{\plotpoint}}
\put(1150.30,698.20){\usebox{\plotpoint}}
\multiput(1153,700)(18.275,9.840){0}{\usebox{\plotpoint}}
\put(1168.29,708.53){\usebox{\plotpoint}}
\put(1185.56,720.04){\usebox{\plotpoint}}
\multiput(1190,723)(18.275,9.840){0}{\usebox{\plotpoint}}
\put(1203.54,730.36){\usebox{\plotpoint}}
\put(1220.81,741.87){\usebox{\plotpoint}}
\put(1238.08,753.39){\usebox{\plotpoint}}
\multiput(1239,754)(18.275,9.840){0}{\usebox{\plotpoint}}
\put(1256.07,763.71){\usebox{\plotpoint}}
\put(1273.34,775.22){\usebox{\plotpoint}}
\multiput(1276,777)(18.275,9.840){0}{\usebox{\plotpoint}}
\put(1291.32,785.55){\usebox{\plotpoint}}
\put(1308.59,797.06){\usebox{\plotpoint}}
\multiput(1313,800)(17.270,11.513){0}{\usebox{\plotpoint}}
\put(1325.91,808.49){\usebox{\plotpoint}}
\put(1343.84,818.90){\usebox{\plotpoint}}
\put(1361.11,830.41){\usebox{\plotpoint}}
\multiput(1362,831)(18.275,9.840){0}{\usebox{\plotpoint}}
\put(1379.10,840.73){\usebox{\plotpoint}}
\put(1396.37,852.24){\usebox{\plotpoint}}
\multiput(1399,854)(17.270,11.513){0}{\usebox{\plotpoint}}
\put(1413.79,863.50){\usebox{\plotpoint}}
\put(1431.62,874.08){\usebox{\plotpoint}}
\put(1436,877){\usebox{\plotpoint}}
\end{picture}
\bigskip
\bigskip
\bigskip
\bigskip
\setlength{\unitlength}{0.240900pt}
\ifx\plotpoint\undefined\newsavebox{\plotpoint}\fi
\sbox{\plotpoint}{\rule[-0.200pt]{0.400pt}{0.400pt}}%
\begin{picture}(1500,900)(0,0)
\font\gnuplot=cmr10 at 10pt
\gnuplot
\sbox{\plotpoint}{\rule[-0.200pt]{0.400pt}{0.400pt}}%
\put(220.0,113.0){\rule[-0.200pt]{4.818pt}{0.400pt}}
\put(198,113){\makebox(0,0)[r]{$0.4$}}
\put(1416.0,113.0){\rule[-0.200pt]{4.818pt}{0.400pt}}
\put(220.0,240.0){\rule[-0.200pt]{4.818pt}{0.400pt}}
\put(198,240){\makebox(0,0)[r]{$0.5$}}
\put(1416.0,240.0){\rule[-0.200pt]{4.818pt}{0.400pt}}
\put(220.0,368.0){\rule[-0.200pt]{4.818pt}{0.400pt}}
\put(198,368){\makebox(0,0)[r]{$0.6$}}
\put(1416.0,368.0){\rule[-0.200pt]{4.818pt}{0.400pt}}
\put(220.0,495.0){\rule[-0.200pt]{4.818pt}{0.400pt}}
\put(198,495){\makebox(0,0)[r]{$0.7$}}
\put(1416.0,495.0){\rule[-0.200pt]{4.818pt}{0.400pt}}
\put(220.0,622.0){\rule[-0.200pt]{4.818pt}{0.400pt}}
\put(198,622){\makebox(0,0)[r]{$0.8$}}
\put(1416.0,622.0){\rule[-0.200pt]{4.818pt}{0.400pt}}
\put(220.0,750.0){\rule[-0.200pt]{4.818pt}{0.400pt}}
\put(198,750){\makebox(0,0)[r]{$0.9$}}
\put(1416.0,750.0){\rule[-0.200pt]{4.818pt}{0.400pt}}
\put(220.0,877.0){\rule[-0.200pt]{4.818pt}{0.400pt}}
\put(198,877){\makebox(0,0)[r]{$1$}}
\put(1416.0,877.0){\rule[-0.200pt]{4.818pt}{0.400pt}}
\put(220.0,113.0){\rule[-0.200pt]{0.400pt}{4.818pt}}
\put(220,68){\makebox(0,0){$0.4$}}
\put(220.0,857.0){\rule[-0.200pt]{0.400pt}{4.818pt}}
\put(423.0,113.0){\rule[-0.200pt]{0.400pt}{4.818pt}}
\put(423,68){\makebox(0,0){$0.5$}}
\put(423.0,857.0){\rule[-0.200pt]{0.400pt}{4.818pt}}
\put(625.0,113.0){\rule[-0.200pt]{0.400pt}{4.818pt}}
\put(625,68){\makebox(0,0){$0.6$}}
\put(625.0,857.0){\rule[-0.200pt]{0.400pt}{4.818pt}}
\put(828.0,113.0){\rule[-0.200pt]{0.400pt}{4.818pt}}
\put(828,68){\makebox(0,0){$0.7$}}
\put(828.0,857.0){\rule[-0.200pt]{0.400pt}{4.818pt}}
\put(1031.0,113.0){\rule[-0.200pt]{0.400pt}{4.818pt}}
\put(1031,68){\makebox(0,0){$0.8$}}
\put(1031.0,857.0){\rule[-0.200pt]{0.400pt}{4.818pt}}
\put(1233.0,113.0){\rule[-0.200pt]{0.400pt}{4.818pt}}
\put(1233,68){\makebox(0,0){$0.9$}}
\put(1233.0,857.0){\rule[-0.200pt]{0.400pt}{4.818pt}}
\put(1436.0,113.0){\rule[-0.200pt]{0.400pt}{4.818pt}}
\put(1436,68){\makebox(0,0){$1$}}
\put(1436.0,857.0){\rule[-0.200pt]{0.400pt}{4.818pt}}
\put(220.0,113.0){\rule[-0.200pt]{292.934pt}{0.400pt}}
\put(1436.0,113.0){\rule[-0.200pt]{0.400pt}{184.048pt}}
\put(220.0,877.0){\rule[-0.200pt]{292.934pt}{0.400pt}}
\put(45,495){\makebox(0,0){FSS}}
\put(828,23){\makebox(0,0){BSS}}
\put(220.0,113.0){\rule[-0.200pt]{0.400pt}{184.048pt}}
\sbox{\plotpoint}{\rule[-0.400pt]{0.800pt}{0.800pt}}%
\put(625,813){\makebox(0,0)[r]{exact}}
\put(669,813){\raisebox{-.8pt}{\makebox(0,0){$\Diamond$}}}
\put(1264,751){\raisebox{-.8pt}{\makebox(0,0){$\Diamond$}}}
\put(665,478){\raisebox{-.8pt}{\makebox(0,0){$\Diamond$}}}
\put(1273,775){\raisebox{-.8pt}{\makebox(0,0){$\Diamond$}}}
\put(933,496){\raisebox{-.8pt}{\makebox(0,0){$\Diamond$}}}
\put(1172,545){\raisebox{-.8pt}{\makebox(0,0){$\Diamond$}}}
\put(960,501){\raisebox{-.8pt}{\makebox(0,0){$\Diamond$}}}
\put(1120,261){\raisebox{-.8pt}{\makebox(0,0){$\Diamond$}}}
\put(969,593){\raisebox{-.8pt}{\makebox(0,0){$\Diamond$}}}
\put(1349,549){\raisebox{-.8pt}{\makebox(0,0){$\Diamond$}}}
\put(414,188){\raisebox{-.8pt}{\makebox(0,0){$\Diamond$}}}
\put(1206,682){\raisebox{-.8pt}{\makebox(0,0){$\Diamond$}}}
\put(502,305){\raisebox{-.8pt}{\makebox(0,0){$\Diamond$}}}
\sbox{\plotpoint}{\rule[-0.600pt]{1.200pt}{1.200pt}}%
\put(625,768){\makebox(0,0)[r]{$G^2 \sim \chi^2$}}
\put(669,768){\makebox(0,0){$+$}}
\put(1220,787){\makebox(0,0){$+$}}
\put(740,620){\makebox(0,0){$+$}}
\put(1327,809){\makebox(0,0){$+$}}
\put(946,609){\makebox(0,0){$+$}}
\put(1277,777){\makebox(0,0){$+$}}
\put(1082,655){\makebox(0,0){$+$}}
\put(1054,637){\makebox(0,0){$+$}}
\put(911,566){\makebox(0,0){$+$}}
\put(1349,822){\makebox(0,0){$+$}}
\put(805,517){\makebox(0,0){$+$}}
\put(1250,725){\makebox(0,0){$+$}}
\put(502,276){\makebox(0,0){$+$}}
\sbox{\plotpoint}{\rule[-0.500pt]{1.000pt}{1.000pt}}%
\put(625,723){\makebox(0,0)[r]{BSS=FSS}}
\multiput(647,723)(20.756,0.000){4}{\usebox{\plotpoint}}
\put(713,723){\usebox{\plotpoint}}
\put(220,113){\usebox{\plotpoint}}
\put(220.00,113.00){\usebox{\plotpoint}}
\put(237.58,124.00){\usebox{\plotpoint}}
\put(255.25,134.84){\usebox{\plotpoint}}
\multiput(257,136)(17.270,11.513){0}{\usebox{\plotpoint}}
\put(272.52,146.35){\usebox{\plotpoint}}
\put(290.31,157.01){\usebox{\plotpoint}}
\multiput(294,159)(17.270,11.513){0}{\usebox{\plotpoint}}
\put(307.78,168.19){\usebox{\plotpoint}}
\put(325.46,179.02){\usebox{\plotpoint}}
\multiput(331,182)(17.270,11.513){0}{\usebox{\plotpoint}}
\put(343.03,190.02){\usebox{\plotpoint}}
\put(360.30,201.53){\usebox{\plotpoint}}
\put(378.19,212.02){\usebox{\plotpoint}}
\multiput(380,213)(17.270,11.513){0}{\usebox{\plotpoint}}
\put(395.56,223.37){\usebox{\plotpoint}}
\put(413.34,234.03){\usebox{\plotpoint}}
\multiput(417,236)(17.270,11.513){0}{\usebox{\plotpoint}}
\put(430.81,245.21){\usebox{\plotpoint}}
\put(448.08,256.72){\usebox{\plotpoint}}
\multiput(453,260)(18.275,9.840){0}{\usebox{\plotpoint}}
\put(466.06,267.04){\usebox{\plotpoint}}
\put(483.33,278.56){\usebox{\plotpoint}}
\put(501.22,289.04){\usebox{\plotpoint}}
\multiput(503,290)(17.270,11.513){0}{\usebox{\plotpoint}}
\put(518.59,300.39){\usebox{\plotpoint}}
\put(535.86,311.91){\usebox{\plotpoint}}
\multiput(539,314)(18.275,9.840){0}{\usebox{\plotpoint}}
\put(553.84,322.23){\usebox{\plotpoint}}
\put(571.11,333.74){\usebox{\plotpoint}}
\multiput(576,337)(17.270,11.513){0}{\usebox{\plotpoint}}
\put(588.40,345.22){\usebox{\plotpoint}}
\put(606.37,355.58){\usebox{\plotpoint}}
\put(623.64,367.09){\usebox{\plotpoint}}
\multiput(625,368)(18.275,9.840){0}{\usebox{\plotpoint}}
\put(641.62,377.41){\usebox{\plotpoint}}
\put(658.89,388.93){\usebox{\plotpoint}}
\multiput(662,391)(17.270,11.513){0}{\usebox{\plotpoint}}
\put(676.28,400.23){\usebox{\plotpoint}}
\put(694.14,410.76){\usebox{\plotpoint}}
\multiput(699,414)(17.270,11.513){0}{\usebox{\plotpoint}}
\put(711.44,422.24){\usebox{\plotpoint}}
\put(729.40,432.60){\usebox{\plotpoint}}
\put(746.67,444.11){\usebox{\plotpoint}}
\multiput(748,445)(17.270,11.513){0}{\usebox{\plotpoint}}
\put(764.17,455.24){\usebox{\plotpoint}}
\put(781.92,465.95){\usebox{\plotpoint}}
\multiput(785,468)(17.270,11.513){0}{\usebox{\plotpoint}}
\put(799.32,477.25){\usebox{\plotpoint}}
\put(817.18,487.78){\usebox{\plotpoint}}
\multiput(822,491)(17.270,11.513){0}{\usebox{\plotpoint}}
\put(834.45,499.30){\usebox{\plotpoint}}
\put(852.05,510.26){\usebox{\plotpoint}}
\put(869.70,521.13){\usebox{\plotpoint}}
\multiput(871,522)(17.270,11.513){0}{\usebox{\plotpoint}}
\put(887.20,532.26){\usebox{\plotpoint}}
\put(904.95,542.97){\usebox{\plotpoint}}
\multiput(908,545)(17.270,11.513){0}{\usebox{\plotpoint}}
\put(922.22,554.48){\usebox{\plotpoint}}
\put(939.93,565.27){\usebox{\plotpoint}}
\multiput(945,568)(17.270,11.513){0}{\usebox{\plotpoint}}
\put(957.48,576.32){\usebox{\plotpoint}}
\put(975.08,587.27){\usebox{\plotpoint}}
\put(992.73,598.15){\usebox{\plotpoint}}
\multiput(994,599)(17.270,11.513){0}{\usebox{\plotpoint}}
\put(1010.00,609.67){\usebox{\plotpoint}}
\put(1027.81,620.28){\usebox{\plotpoint}}
\multiput(1031,622)(17.270,11.513){0}{\usebox{\plotpoint}}
\put(1045.26,631.50){\usebox{\plotpoint}}
\put(1062.96,642.29){\usebox{\plotpoint}}
\multiput(1068,645)(17.270,11.513){0}{\usebox{\plotpoint}}
\put(1080.51,653.34){\usebox{\plotpoint}}
\put(1097.78,664.85){\usebox{\plotpoint}}
\put(1115.69,675.30){\usebox{\plotpoint}}
\multiput(1117,676)(17.270,11.513){0}{\usebox{\plotpoint}}
\put(1133.03,686.69){\usebox{\plotpoint}}
\put(1150.30,698.20){\usebox{\plotpoint}}
\multiput(1153,700)(18.275,9.840){0}{\usebox{\plotpoint}}
\put(1168.29,708.53){\usebox{\plotpoint}}
\put(1185.56,720.04){\usebox{\plotpoint}}
\multiput(1190,723)(18.275,9.840){0}{\usebox{\plotpoint}}
\put(1203.54,730.36){\usebox{\plotpoint}}
\put(1220.81,741.87){\usebox{\plotpoint}}
\put(1238.08,753.39){\usebox{\plotpoint}}
\multiput(1239,754)(18.275,9.840){0}{\usebox{\plotpoint}}
\put(1256.07,763.71){\usebox{\plotpoint}}
\put(1273.34,775.22){\usebox{\plotpoint}}
\multiput(1276,777)(18.275,9.840){0}{\usebox{\plotpoint}}
\put(1291.32,785.55){\usebox{\plotpoint}}
\put(1308.59,797.06){\usebox{\plotpoint}}
\multiput(1313,800)(17.270,11.513){0}{\usebox{\plotpoint}}
\put(1325.91,808.49){\usebox{\plotpoint}}
\put(1343.84,818.90){\usebox{\plotpoint}}
\put(1361.11,830.41){\usebox{\plotpoint}}
\multiput(1362,831)(18.275,9.840){0}{\usebox{\plotpoint}}
\put(1379.10,840.73){\usebox{\plotpoint}}
\put(1396.37,852.24){\usebox{\plotpoint}}
\multiput(1399,854)(17.270,11.513){0}{\usebox{\plotpoint}}
\put(1413.79,863.50){\usebox{\plotpoint}}
\put(1431.62,874.08){\usebox{\plotpoint}}
\put(1436,877){\usebox{\plotpoint}}
\end{picture}
\caption{Robustness of Selection Process}
\label{fig:fb1}
\end{center}
\myendfig

In Figure \ref{fig:fb1} each point represents
the accuracy of models selected using the specified evaluation
criterion with backward and forward search. The $BSS$ coordinate is
the accuracy attained by the model selected during backward search
while the $FSS$ coordinate is the accuracy resulting from forward
search. Points that fall close to the line $BSS=FSS$ represent an
evaluation criterion that selects models of similar accuracy
regardless of search strategy. Figure \ref{fig:fb1} (top) shows that,
in general, the information criteria select models of similar accuracy
using either forward or backward search.  

However, Figure \ref{fig:fb1} (bottom) shows that the significance
tests are sensitive to changes in search strategy. For example, BSS
exact conditional is more accurate than FSS exact conditional.
FSS $G^2 \sim \chi^2$ is slightly more accurate than BSS $G^2 \sim
\chi^2$. This suggests that the value of $\alpha$ may need to be
adjusted depending on the direction of the search strategy. The
following section shows that this is due to changes in the bias of 
the evaluation criteria that are caused by changing the search 
strategy.  

\subsection{Model selection for Noun {\it interest} }

The model selection process for {\it interest} is  discussed in some 
detail here. Figures \ref{fig:bssacc} and \ref{fig:bssrec} show the 
accuracy and recall\footnote{The percentage of ambiguous words in a held 
out test sample that are disambiguated, correctly or not. A word is not 
disambiguated if any of the model parameters needed to assign a sense tag 
cannot be estimated from the training sample.} of the model at each level
of complexity during the selection process. The rightmost point on
each plot is the measure associated with  the model ultimately
selected by the evaluation criterion and search strategy.  

\begin{figure}
\begin{center}
\setlength{\unitlength}{0.240900pt}
\ifx\plotpoint\undefined\newsavebox{\plotpoint}\fi
\sbox{\plotpoint}{\rule[-0.200pt]{0.400pt}{0.400pt}}%
\begin{picture}(1500,900)(0,0)
\font\gnuplot=cmr10 at 10pt
\gnuplot
\sbox{\plotpoint}{\rule[-0.200pt]{0.400pt}{0.400pt}}%
\put(220.0,113.0){\rule[-0.200pt]{4.818pt}{0.400pt}}
\put(198,113){\makebox(0,0)[r]{$0.3$}}
\put(1416.0,113.0){\rule[-0.200pt]{4.818pt}{0.400pt}}
\put(220.0,215.0){\rule[-0.200pt]{4.818pt}{0.400pt}}
\put(198,215){\makebox(0,0)[r]{$0.4$}}
\put(1416.0,215.0){\rule[-0.200pt]{4.818pt}{0.400pt}}
\put(220.0,317.0){\rule[-0.200pt]{4.818pt}{0.400pt}}
\put(198,317){\makebox(0,0)[r]{$0.5$}}
\put(1416.0,317.0){\rule[-0.200pt]{4.818pt}{0.400pt}}
\put(220.0,419.0){\rule[-0.200pt]{4.818pt}{0.400pt}}
\put(198,419){\makebox(0,0)[r]{$0.6$}}
\put(1416.0,419.0){\rule[-0.200pt]{4.818pt}{0.400pt}}
\put(220.0,520.0){\rule[-0.200pt]{4.818pt}{0.400pt}}
\put(198,520){\makebox(0,0)[r]{$0.7$}}
\put(1416.0,520.0){\rule[-0.200pt]{4.818pt}{0.400pt}}
\put(220.0,622.0){\rule[-0.200pt]{4.818pt}{0.400pt}}
\put(198,622){\makebox(0,0)[r]{$0.8$}}
\put(1416.0,622.0){\rule[-0.200pt]{4.818pt}{0.400pt}}
\put(220.0,724.0){\rule[-0.200pt]{4.818pt}{0.400pt}}
\put(198,724){\makebox(0,0)[r]{$0.9$}}
\put(1416.0,724.0){\rule[-0.200pt]{4.818pt}{0.400pt}}
\put(220.0,826.0){\rule[-0.200pt]{4.818pt}{0.400pt}}
\put(198,826){\makebox(0,0)[r]{$1$}}
\put(1416.0,826.0){\rule[-0.200pt]{4.818pt}{0.400pt}}
\put(1436.0,113.0){\rule[-0.200pt]{0.400pt}{4.818pt}}
\put(1436,68){\makebox(0,0){$0$}}
\put(1436.0,857.0){\rule[-0.200pt]{0.400pt}{4.818pt}}
\put(1267.0,113.0){\rule[-0.200pt]{0.400pt}{4.818pt}}
\put(1267,68){\makebox(0,0){$5$}}
\put(1267.0,857.0){\rule[-0.200pt]{0.400pt}{4.818pt}}
\put(1098.0,113.0){\rule[-0.200pt]{0.400pt}{4.818pt}}
\put(1098,68){\makebox(0,0){$10$}}
\put(1098.0,857.0){\rule[-0.200pt]{0.400pt}{4.818pt}}
\put(929.0,113.0){\rule[-0.200pt]{0.400pt}{4.818pt}}
\put(929,68){\makebox(0,0){$15$}}
\put(929.0,857.0){\rule[-0.200pt]{0.400pt}{4.818pt}}
\put(760.0,113.0){\rule[-0.200pt]{0.400pt}{4.818pt}}
\put(760,68){\makebox(0,0){$20$}}
\put(760.0,857.0){\rule[-0.200pt]{0.400pt}{4.818pt}}
\put(592.0,113.0){\rule[-0.200pt]{0.400pt}{4.818pt}}
\put(592,68){\makebox(0,0){$25$}}
\put(592.0,857.0){\rule[-0.200pt]{0.400pt}{4.818pt}}
\put(423.0,113.0){\rule[-0.200pt]{0.400pt}{4.818pt}}
\put(423,68){\makebox(0,0){$30$}}
\put(423.0,857.0){\rule[-0.200pt]{0.400pt}{4.818pt}}
\put(254.0,113.0){\rule[-0.200pt]{0.400pt}{4.818pt}}
\put(254,68){\makebox(0,0){$35$}}
\put(254.0,857.0){\rule[-0.200pt]{0.400pt}{4.818pt}}
\put(220.0,113.0){\rule[-0.200pt]{292.934pt}{0.400pt}}
\put(1436.0,113.0){\rule[-0.200pt]{0.400pt}{184.048pt}}
\put(220.0,877.0){\rule[-0.200pt]{292.934pt}{0.400pt}}
\put(45,495){\makebox(0,0){\%}}
\put(828,23){\makebox(0,0){\# of interactions in model}}
\put(220.0,113.0){\rule[-0.200pt]{0.400pt}{184.048pt}}
\sbox{\plotpoint}{\rule[-0.400pt]{0.800pt}{0.800pt}}%
\put(1098,317){\makebox(0,0)[r]{AIC}}
\put(1120.0,317.0){\rule[-0.400pt]{15.899pt}{0.800pt}}
\put(727,563){\usebox{\plotpoint}}
\put(693,559.34){\rule{7.000pt}{0.800pt}}
\multiput(712.47,561.34)(-19.471,-4.000){2}{\rule{3.500pt}{0.800pt}}
\multiput(683.48,557.08)(-1.360,-0.509){19}{\rule{2.292pt}{0.123pt}}
\multiput(688.24,557.34)(-29.242,-13.000){2}{\rule{1.146pt}{0.800pt}}
\put(490,542.34){\rule{34.000pt}{0.800pt}}
\multiput(588.43,544.34)(-98.431,-4.000){2}{\rule{17.000pt}{0.800pt}}
\multiput(479.76,540.08)(-1.485,-0.511){17}{\rule{2.467pt}{0.123pt}}
\multiput(484.88,540.34)(-28.880,-12.000){2}{\rule{1.233pt}{0.800pt}}
\multiput(451.52,528.09)(-0.548,-0.503){53}{\rule{1.080pt}{0.121pt}}
\multiput(453.76,528.34)(-30.758,-30.000){2}{\rule{0.540pt}{0.800pt}}
\multiput(408.06,498.08)(-2.382,-0.520){9}{\rule{3.600pt}{0.125pt}}
\multiput(415.53,498.34)(-26.528,-8.000){2}{\rule{1.800pt}{0.800pt}}
\multiput(381.53,490.09)(-1.020,-0.507){27}{\rule{1.800pt}{0.122pt}}
\multiput(385.26,490.34)(-30.264,-17.000){2}{\rule{0.900pt}{0.800pt}}
\multiput(353.09,466.45)(-0.503,-1.174){61}{\rule{0.121pt}{2.059pt}}
\multiput(353.34,470.73)(-34.000,-74.727){2}{\rule{0.800pt}{1.029pt}}
\multiput(306.47,394.08)(-2.309,-0.520){9}{\rule{3.500pt}{0.125pt}}
\multiput(313.74,394.34)(-25.736,-8.000){2}{\rule{1.750pt}{0.800pt}}
\multiput(286.09,378.97)(-0.503,-1.249){61}{\rule{0.121pt}{2.176pt}}
\multiput(286.34,383.48)(-34.000,-79.483){2}{\rule{0.800pt}{1.088pt}}
\multiput(252.09,296.63)(-0.503,-0.994){61}{\rule{0.121pt}{1.776pt}}
\multiput(252.34,300.31)(-34.000,-63.313){2}{\rule{0.800pt}{0.888pt}}
\put(1142,317){\makebox(0,0){$\star$}}
\put(727,563){\makebox(0,0){$\star$}}
\put(693,559){\makebox(0,0){$\star$}}
\put(659,546){\makebox(0,0){$\star$}}
\put(490,542){\makebox(0,0){$\star$}}
\put(456,530){\makebox(0,0){$\star$}}
\put(423,500){\makebox(0,0){$\star$}}
\put(389,492){\makebox(0,0){$\star$}}
\put(355,475){\makebox(0,0){$\star$}}
\put(321,396){\makebox(0,0){$\star$}}
\put(288,388){\makebox(0,0){$\star$}}
\put(254,304){\makebox(0,0){$\star$}}
\put(220,237){\makebox(0,0){$\star$}}
\sbox{\plotpoint}{\rule[-0.200pt]{0.400pt}{0.400pt}}%
\put(1098,272){\makebox(0,0)[r]{BIC}}
\put(1120.0,272.0){\rule[-0.200pt]{15.899pt}{0.400pt}}
\put(1233,450){\usebox{\plotpoint}}
\multiput(1094.53,450.58)(-0.920,0.499){181}{\rule{0.835pt}{0.120pt}}
\multiput(1096.27,449.17)(-167.267,92.000){2}{\rule{0.417pt}{0.400pt}}
\multiput(907.00,542.58)(-6.682,0.493){23}{\rule{5.300pt}{0.119pt}}
\multiput(918.00,541.17)(-158.000,13.000){2}{\rule{2.650pt}{0.400pt}}
\multiput(750.29,553.92)(-2.827,-0.497){57}{\rule{2.340pt}{0.120pt}}
\multiput(755.14,554.17)(-163.143,-30.000){2}{\rule{1.170pt}{0.400pt}}
\multiput(580.36,523.92)(-3.421,-0.497){47}{\rule{2.804pt}{0.120pt}}
\multiput(586.18,524.17)(-163.180,-25.000){2}{\rule{1.402pt}{0.400pt}}
\multiput(417.88,498.92)(-1.444,-0.492){21}{\rule{1.233pt}{0.119pt}}
\multiput(420.44,499.17)(-31.440,-12.000){2}{\rule{0.617pt}{0.400pt}}
\multiput(381.53,486.93)(-2.211,-0.488){13}{\rule{1.800pt}{0.117pt}}
\multiput(385.26,487.17)(-30.264,-8.000){2}{\rule{0.900pt}{0.400pt}}
\multiput(353.92,476.12)(-0.498,-1.049){65}{\rule{0.120pt}{0.935pt}}
\multiput(354.17,478.06)(-34.000,-69.059){2}{\rule{0.400pt}{0.468pt}}
\multiput(317.98,407.92)(-0.789,-0.496){39}{\rule{0.729pt}{0.119pt}}
\multiput(319.49,408.17)(-31.488,-21.000){2}{\rule{0.364pt}{0.400pt}}
\multiput(286.92,383.48)(-0.498,-1.242){65}{\rule{0.120pt}{1.088pt}}
\multiput(287.17,385.74)(-34.000,-81.741){2}{\rule{0.400pt}{0.544pt}}
\multiput(252.92,300.31)(-0.498,-0.989){65}{\rule{0.120pt}{0.888pt}}
\multiput(253.17,302.16)(-34.000,-65.156){2}{\rule{0.400pt}{0.444pt}}
\put(1142,272){\raisebox{-.8pt}{\makebox(0,0){$\Diamond$}}}
\put(1233,450){\raisebox{-.8pt}{\makebox(0,0){$\Diamond$}}}
\put(1200,450){\raisebox{-.8pt}{\makebox(0,0){$\Diamond$}}}
\put(1166,450){\raisebox{-.8pt}{\makebox(0,0){$\Diamond$}}}
\put(1132,450){\raisebox{-.8pt}{\makebox(0,0){$\Diamond$}}}
\put(1098,450){\raisebox{-.8pt}{\makebox(0,0){$\Diamond$}}}
\put(929,542){\raisebox{-.8pt}{\makebox(0,0){$\Diamond$}}}
\put(760,555){\raisebox{-.8pt}{\makebox(0,0){$\Diamond$}}}
\put(592,525){\raisebox{-.8pt}{\makebox(0,0){$\Diamond$}}}
\put(423,500){\raisebox{-.8pt}{\makebox(0,0){$\Diamond$}}}
\put(389,488){\raisebox{-.8pt}{\makebox(0,0){$\Diamond$}}}
\put(355,480){\raisebox{-.8pt}{\makebox(0,0){$\Diamond$}}}
\put(321,409){\raisebox{-.8pt}{\makebox(0,0){$\Diamond$}}}
\put(288,388){\raisebox{-.8pt}{\makebox(0,0){$\Diamond$}}}
\put(254,304){\raisebox{-.8pt}{\makebox(0,0){$\Diamond$}}}
\put(220,237){\raisebox{-.8pt}{\makebox(0,0){$\Diamond$}}}
\put(1098.0,450.0){\rule[-0.200pt]{32.521pt}{0.400pt}}
\sbox{\plotpoint}{\rule[-0.500pt]{1.000pt}{1.000pt}}%
\put(1098,227){\makebox(0,0)[r]{Exact $\alpha =.0001$}}
\multiput(1120,227)(41.511,0.000){2}{\usebox{\plotpoint}}
\put(1186,227){\usebox{\plotpoint}}
\put(625,313){\usebox{\plotpoint}}
\multiput(625,313)(-40.935,-6.890){5}{\usebox{\plotpoint}}
\put(420.57,277.79){\usebox{\plotpoint}}
\put(383.38,259.36){\usebox{\plotpoint}}
\put(345.52,242.38){\usebox{\plotpoint}}
\put(305.85,231.16){\usebox{\plotpoint}}
\put(264.64,231.75){\usebox{\plotpoint}}
\put(284.59,229.40){\usebox{\plotpoint}}
\put(250.18,233.45){\usebox{\plotpoint}}
\put(220,237){\usebox{\plotpoint}}
\put(1142,227){\makebox(0,0){$\triangle$}}
\put(625,313){\makebox(0,0){$\triangle$}}
\put(423,279){\makebox(0,0){$\triangle$}}
\put(389,262){\makebox(0,0){$\triangle$}}
\put(355,246){\makebox(0,0){$\triangle$}}
\put(321,233){\makebox(0,0){$\triangle$}}
\put(288,229){\makebox(0,0){$\triangle$}}
\put(254,233){\makebox(0,0){$\triangle$}}
\put(288,229){\makebox(0,0){$\triangle$}}
\put(220,237){\makebox(0,0){$\triangle$}}
\put(1098,182){\makebox(0,0)[r]{$G^2 \sim \chi^2$ $\alpha =.0001$}}
\multiput(1120,182)(20.756,0.000){4}{\usebox{\plotpoint}}
\put(1186,182){\usebox{\plotpoint}}
\put(625,509){\usebox{\plotpoint}}
\multiput(625,509)(-20.605,2.498){2}{\usebox{\plotpoint}}
\multiput(592,513)(-20.613,-2.425){2}{\usebox{\plotpoint}}
\multiput(558,509)(-20.470,-3.429){10}{\usebox{\plotpoint}}
\multiput(355,475)(-12.337,-16.691){2}{\usebox{\plotpoint}}
\multiput(321,429)(-10.823,-17.710){3}{\usebox{\plotpoint}}
\multiput(288,375)(-14.894,-14.456){3}{\usebox{\plotpoint}}
\multiput(254,342)(-6.394,-19.746){5}{\usebox{\plotpoint}}
\put(220,237){\usebox{\plotpoint}}
\put(1142,182){\circle{12}}
\put(625,509){\circle{12}}
\put(592,513){\circle{12}}
\put(558,509){\circle{12}}
\put(355,475){\circle{12}}
\put(321,429){\circle{12}}
\put(288,375){\circle{12}}
\put(254,342){\circle{12}}
\put(220,237){\circle{12}}
\end{picture}
\bigskip
\bigskip
\bigskip
\bigskip
\input{figs/interest2a}
\caption{BSS (top) and FSS (bottom) accuracy for Noun {\it interest} }
\label{fig:bssacc}
\end{center}
\end{figure}

\begin{figure}
\begin{center}
\setlength{\unitlength}{0.240900pt}
\ifx\plotpoint\undefined\newsavebox{\plotpoint}\fi
\sbox{\plotpoint}{\rule[-0.200pt]{0.400pt}{0.400pt}}%
\begin{picture}(1500,900)(0,0)
\font\gnuplot=cmr10 at 10pt
\gnuplot
\sbox{\plotpoint}{\rule[-0.200pt]{0.400pt}{0.400pt}}%
\put(220.0,113.0){\rule[-0.200pt]{4.818pt}{0.400pt}}
\put(198,113){\makebox(0,0)[r]{$0.3$}}
\put(1416.0,113.0){\rule[-0.200pt]{4.818pt}{0.400pt}}
\put(220.0,215.0){\rule[-0.200pt]{4.818pt}{0.400pt}}
\put(198,215){\makebox(0,0)[r]{$0.4$}}
\put(1416.0,215.0){\rule[-0.200pt]{4.818pt}{0.400pt}}
\put(220.0,317.0){\rule[-0.200pt]{4.818pt}{0.400pt}}
\put(198,317){\makebox(0,0)[r]{$0.5$}}
\put(1416.0,317.0){\rule[-0.200pt]{4.818pt}{0.400pt}}
\put(220.0,419.0){\rule[-0.200pt]{4.818pt}{0.400pt}}
\put(198,419){\makebox(0,0)[r]{$0.6$}}
\put(1416.0,419.0){\rule[-0.200pt]{4.818pt}{0.400pt}}
\put(220.0,520.0){\rule[-0.200pt]{4.818pt}{0.400pt}}
\put(198,520){\makebox(0,0)[r]{$0.7$}}
\put(1416.0,520.0){\rule[-0.200pt]{4.818pt}{0.400pt}}
\put(220.0,622.0){\rule[-0.200pt]{4.818pt}{0.400pt}}
\put(198,622){\makebox(0,0)[r]{$0.8$}}
\put(1416.0,622.0){\rule[-0.200pt]{4.818pt}{0.400pt}}
\put(220.0,724.0){\rule[-0.200pt]{4.818pt}{0.400pt}}
\put(198,724){\makebox(0,0)[r]{$0.9$}}
\put(1416.0,724.0){\rule[-0.200pt]{4.818pt}{0.400pt}}
\put(220.0,826.0){\rule[-0.200pt]{4.818pt}{0.400pt}}
\put(198,826){\makebox(0,0)[r]{$1$}}
\put(1416.0,826.0){\rule[-0.200pt]{4.818pt}{0.400pt}}
\put(1436.0,113.0){\rule[-0.200pt]{0.400pt}{4.818pt}}
\put(1436,68){\makebox(0,0){$0$}}
\put(1436.0,857.0){\rule[-0.200pt]{0.400pt}{4.818pt}}
\put(1267.0,113.0){\rule[-0.200pt]{0.400pt}{4.818pt}}
\put(1267,68){\makebox(0,0){$5$}}
\put(1267.0,857.0){\rule[-0.200pt]{0.400pt}{4.818pt}}
\put(1098.0,113.0){\rule[-0.200pt]{0.400pt}{4.818pt}}
\put(1098,68){\makebox(0,0){$10$}}
\put(1098.0,857.0){\rule[-0.200pt]{0.400pt}{4.818pt}}
\put(929.0,113.0){\rule[-0.200pt]{0.400pt}{4.818pt}}
\put(929,68){\makebox(0,0){$15$}}
\put(929.0,857.0){\rule[-0.200pt]{0.400pt}{4.818pt}}
\put(760.0,113.0){\rule[-0.200pt]{0.400pt}{4.818pt}}
\put(760,68){\makebox(0,0){$20$}}
\put(760.0,857.0){\rule[-0.200pt]{0.400pt}{4.818pt}}
\put(592.0,113.0){\rule[-0.200pt]{0.400pt}{4.818pt}}
\put(592,68){\makebox(0,0){$25$}}
\put(592.0,857.0){\rule[-0.200pt]{0.400pt}{4.818pt}}
\put(423.0,113.0){\rule[-0.200pt]{0.400pt}{4.818pt}}
\put(423,68){\makebox(0,0){$30$}}
\put(423.0,857.0){\rule[-0.200pt]{0.400pt}{4.818pt}}
\put(254.0,113.0){\rule[-0.200pt]{0.400pt}{4.818pt}}
\put(254,68){\makebox(0,0){$35$}}
\put(254.0,857.0){\rule[-0.200pt]{0.400pt}{4.818pt}}
\put(220.0,113.0){\rule[-0.200pt]{292.934pt}{0.400pt}}
\put(1436.0,113.0){\rule[-0.200pt]{0.400pt}{184.048pt}}
\put(220.0,877.0){\rule[-0.200pt]{292.934pt}{0.400pt}}
\put(45,495){\makebox(0,0){\%}}
\put(828,23){\makebox(0,0){\# of interactions in model}}
\put(220.0,113.0){\rule[-0.200pt]{0.400pt}{184.048pt}}
\sbox{\plotpoint}{\rule[-0.400pt]{0.800pt}{0.800pt}}%
\put(1098,317){\makebox(0,0)[r]{AIC}}
\put(1120.0,317.0){\rule[-0.400pt]{15.899pt}{0.800pt}}
\put(727,814){\usebox{\plotpoint}}
\multiput(713.62,812.08)(-2.062,-0.516){11}{\rule{3.222pt}{0.124pt}}
\multiput(720.31,812.34)(-27.312,-9.000){2}{\rule{1.611pt}{0.800pt}}
\multiput(625.16,803.09)(-5.198,-0.507){27}{\rule{8.153pt}{0.122pt}}
\multiput(642.08,803.34)(-152.078,-17.000){2}{\rule{4.076pt}{0.800pt}}
\put(456,784.34){\rule{7.000pt}{0.800pt}}
\multiput(475.47,786.34)(-19.471,-4.000){2}{\rule{3.500pt}{0.800pt}}
\put(423,780.34){\rule{6.800pt}{0.800pt}}
\multiput(441.89,782.34)(-18.886,-4.000){2}{\rule{3.400pt}{0.800pt}}
\multiput(421.09,774.68)(-0.503,-0.678){61}{\rule{0.121pt}{1.282pt}}
\multiput(421.34,777.34)(-34.000,-43.338){2}{\rule{0.800pt}{0.641pt}}
\multiput(387.09,726.63)(-0.503,-0.994){61}{\rule{0.121pt}{1.776pt}}
\multiput(387.34,730.31)(-34.000,-63.313){2}{\rule{0.800pt}{0.888pt}}
\multiput(353.09,653.57)(-0.503,-1.926){61}{\rule{0.121pt}{3.235pt}}
\multiput(353.34,660.28)(-34.000,-122.285){2}{\rule{0.800pt}{1.618pt}}
\multiput(314.95,536.09)(-0.793,-0.505){35}{\rule{1.457pt}{0.122pt}}
\multiput(317.98,536.34)(-29.976,-21.000){2}{\rule{0.729pt}{0.800pt}}
\multiput(286.09,506.79)(-0.503,-1.430){61}{\rule{0.121pt}{2.459pt}}
\multiput(286.34,511.90)(-34.000,-90.897){2}{\rule{0.800pt}{1.229pt}}
\multiput(252.09,410.40)(-0.503,-1.490){61}{\rule{0.121pt}{2.553pt}}
\multiput(252.34,415.70)(-34.000,-94.701){2}{\rule{0.800pt}{1.276pt}}
\put(1142,317){\makebox(0,0){$\star$}}
\put(727,814){\makebox(0,0){$\star$}}
\put(693,805){\makebox(0,0){$\star$}}
\put(659,805){\makebox(0,0){$\star$}}
\put(490,788){\makebox(0,0){$\star$}}
\put(456,784){\makebox(0,0){$\star$}}
\put(423,780){\makebox(0,0){$\star$}}
\put(389,734){\makebox(0,0){$\star$}}
\put(355,667){\makebox(0,0){$\star$}}
\put(321,538){\makebox(0,0){$\star$}}
\put(288,517){\makebox(0,0){$\star$}}
\put(254,421){\makebox(0,0){$\star$}}
\put(220,321){\makebox(0,0){$\star$}}
\put(659.0,805.0){\rule[-0.400pt]{8.191pt}{0.800pt}}
\sbox{\plotpoint}{\rule[-0.200pt]{0.400pt}{0.400pt}}%
\put(1098,272){\makebox(0,0)[r]{BIC}}
\put(1120.0,272.0){\rule[-0.200pt]{15.899pt}{0.400pt}}
\put(1233,826){\usebox{\plotpoint}}
\multiput(1074.20,824.92)(-7.260,-0.492){21}{\rule{5.733pt}{0.119pt}}
\multiput(1086.10,825.17)(-157.100,-12.000){2}{\rule{2.867pt}{0.400pt}}
\multiput(872.46,812.93)(-18.744,-0.477){7}{\rule{13.620pt}{0.115pt}}
\multiput(900.73,813.17)(-140.731,-5.000){2}{\rule{6.810pt}{0.400pt}}
\multiput(742.15,807.92)(-5.363,-0.494){29}{\rule{4.300pt}{0.119pt}}
\multiput(751.08,808.17)(-159.075,-16.000){2}{\rule{2.150pt}{0.400pt}}
\multiput(570.00,791.92)(-6.682,-0.493){23}{\rule{5.300pt}{0.119pt}}
\multiput(581.00,792.17)(-158.000,-13.000){2}{\rule{2.650pt}{0.400pt}}
\multiput(421.92,775.92)(-0.498,-1.108){65}{\rule{0.120pt}{0.982pt}}
\multiput(422.17,777.96)(-34.000,-72.961){2}{\rule{0.400pt}{0.491pt}}
\multiput(387.92,701.51)(-0.498,-0.930){65}{\rule{0.120pt}{0.841pt}}
\multiput(388.17,703.25)(-34.000,-61.254){2}{\rule{0.400pt}{0.421pt}}
\multiput(353.92,636.90)(-0.498,-1.420){65}{\rule{0.120pt}{1.229pt}}
\multiput(354.17,639.45)(-34.000,-93.448){2}{\rule{0.400pt}{0.615pt}}
\multiput(318.70,544.92)(-0.568,-0.497){55}{\rule{0.555pt}{0.120pt}}
\multiput(319.85,545.17)(-31.848,-29.000){2}{\rule{0.278pt}{0.400pt}}
\multiput(286.92,511.90)(-0.498,-1.420){65}{\rule{0.120pt}{1.229pt}}
\multiput(287.17,514.45)(-34.000,-93.448){2}{\rule{0.400pt}{0.615pt}}
\multiput(252.92,415.70)(-0.498,-1.480){65}{\rule{0.120pt}{1.276pt}}
\multiput(253.17,418.35)(-34.000,-97.351){2}{\rule{0.400pt}{0.638pt}}
\put(1142,272){\raisebox{-.8pt}{\makebox(0,0){$\Diamond$}}}
\put(1233,826){\raisebox{-.8pt}{\makebox(0,0){$\Diamond$}}}
\put(1200,826){\raisebox{-.8pt}{\makebox(0,0){$\Diamond$}}}
\put(1166,826){\raisebox{-.8pt}{\makebox(0,0){$\Diamond$}}}
\put(1132,826){\raisebox{-.8pt}{\makebox(0,0){$\Diamond$}}}
\put(1098,826){\raisebox{-.8pt}{\makebox(0,0){$\Diamond$}}}
\put(929,814){\raisebox{-.8pt}{\makebox(0,0){$\Diamond$}}}
\put(760,809){\raisebox{-.8pt}{\makebox(0,0){$\Diamond$}}}
\put(592,793){\raisebox{-.8pt}{\makebox(0,0){$\Diamond$}}}
\put(423,780){\raisebox{-.8pt}{\makebox(0,0){$\Diamond$}}}
\put(389,705){\raisebox{-.8pt}{\makebox(0,0){$\Diamond$}}}
\put(355,642){\raisebox{-.8pt}{\makebox(0,0){$\Diamond$}}}
\put(321,546){\raisebox{-.8pt}{\makebox(0,0){$\Diamond$}}}
\put(288,517){\raisebox{-.8pt}{\makebox(0,0){$\Diamond$}}}
\put(254,421){\raisebox{-.8pt}{\makebox(0,0){$\Diamond$}}}
\put(220,321){\raisebox{-.8pt}{\makebox(0,0){$\Diamond$}}}
\put(1098.0,826.0){\rule[-0.200pt]{32.521pt}{0.400pt}}
\sbox{\plotpoint}{\rule[-0.500pt]{1.000pt}{1.000pt}}%
\put(1098,227){\makebox(0,0)[r]{Exact $\alpha =.0001$}}
\multiput(1120,227)(41.511,0.000){2}{\usebox{\plotpoint}}
\put(1186,227){\usebox{\plotpoint}}
\put(625,413){\usebox{\plotpoint}}
\multiput(625,413)(-40.642,-8.450){5}{\usebox{\plotpoint}}
\put(421.95,370.35){\usebox{\plotpoint}}
\put(386.51,348.76){\usebox{\plotpoint}}
\put(349.08,330.91){\usebox{\plotpoint}}
\put(309.26,321.00){\usebox{\plotpoint}}
\put(267.75,321.00){\usebox{\plotpoint}}
\put(281.76,321.00){\usebox{\plotpoint}}
\put(252.73,321.00){\usebox{\plotpoint}}
\put(220,321){\usebox{\plotpoint}}
\put(1142,227){\makebox(0,0){$\triangle$}}
\put(625,413){\makebox(0,0){$\triangle$}}
\put(423,371){\makebox(0,0){$\triangle$}}
\put(389,350){\makebox(0,0){$\triangle$}}
\put(355,333){\makebox(0,0){$\triangle$}}
\put(321,321){\makebox(0,0){$\triangle$}}
\put(288,321){\makebox(0,0){$\triangle$}}
\put(254,321){\makebox(0,0){$\triangle$}}
\put(288,321){\makebox(0,0){$\triangle$}}
\put(220,321){\makebox(0,0){$\triangle$}}
\put(1098,182){\makebox(0,0)[r]{$G^2 \sim \chi^2$ $\alpha =.0001$}}
\multiput(1120,182)(20.756,0.000){4}{\usebox{\plotpoint}}
\put(1186,182){\usebox{\plotpoint}}
\put(625,722){\usebox{\plotpoint}}
\multiput(625,722)(-20.756,0.000){2}{\usebox{\plotpoint}}
\multiput(592,722)(-20.064,-5.311){2}{\usebox{\plotpoint}}
\multiput(558,713)(-19.738,-6.417){10}{\usebox{\plotpoint}}
\multiput(355,647)(-11.513,-17.270){3}{\usebox{\plotpoint}}
\multiput(321,596)(-6.065,-19.850){5}{\usebox{\plotpoint}}
\multiput(288,488)(-13.840,-15.468){3}{\usebox{\plotpoint}}
\multiput(254,450)(-5.290,-20.070){6}{\usebox{\plotpoint}}
\put(220,321){\usebox{\plotpoint}}
\put(1142,182){\circle{12}}
\put(625,722){\circle{12}}
\put(592,722){\circle{12}}
\put(558,713){\circle{12}}
\put(355,647){\circle{12}}
\put(321,596){\circle{12}}
\put(288,488){\circle{12}}
\put(254,450){\circle{12}}
\put(220,321){\circle{12}}
\end{picture}
\bigskip
\bigskip
\bigskip
\bigskip
\setlength{\unitlength}{0.240900pt}
\ifx\plotpoint\undefined\newsavebox{\plotpoint}\fi
\sbox{\plotpoint}{\rule[-0.200pt]{0.400pt}{0.400pt}}%
\begin{picture}(1500,900)(0,0)
\font\gnuplot=cmr10 at 10pt
\gnuplot
\sbox{\plotpoint}{\rule[-0.200pt]{0.400pt}{0.400pt}}%
\put(220.0,113.0){\rule[-0.200pt]{0.400pt}{184.048pt}}
\put(220.0,113.0){\rule[-0.200pt]{4.818pt}{0.400pt}}
\put(198,113){\makebox(0,0)[r]{$0.3$}}
\put(1416.0,113.0){\rule[-0.200pt]{4.818pt}{0.400pt}}
\put(220.0,215.0){\rule[-0.200pt]{4.818pt}{0.400pt}}
\put(198,215){\makebox(0,0)[r]{$0.4$}}
\put(1416.0,215.0){\rule[-0.200pt]{4.818pt}{0.400pt}}
\put(220.0,317.0){\rule[-0.200pt]{4.818pt}{0.400pt}}
\put(198,317){\makebox(0,0)[r]{$0.5$}}
\put(1416.0,317.0){\rule[-0.200pt]{4.818pt}{0.400pt}}
\put(220.0,419.0){\rule[-0.200pt]{4.818pt}{0.400pt}}
\put(198,419){\makebox(0,0)[r]{$0.6$}}
\put(1416.0,419.0){\rule[-0.200pt]{4.818pt}{0.400pt}}
\put(220.0,520.0){\rule[-0.200pt]{4.818pt}{0.400pt}}
\put(198,520){\makebox(0,0)[r]{$0.7$}}
\put(1416.0,520.0){\rule[-0.200pt]{4.818pt}{0.400pt}}
\put(220.0,622.0){\rule[-0.200pt]{4.818pt}{0.400pt}}
\put(198,622){\makebox(0,0)[r]{$0.8$}}
\put(1416.0,622.0){\rule[-0.200pt]{4.818pt}{0.400pt}}
\put(220.0,724.0){\rule[-0.200pt]{4.818pt}{0.400pt}}
\put(198,724){\makebox(0,0)[r]{$0.9$}}
\put(1416.0,724.0){\rule[-0.200pt]{4.818pt}{0.400pt}}
\put(220.0,826.0){\rule[-0.200pt]{4.818pt}{0.400pt}}
\put(198,826){\makebox(0,0)[r]{$1$}}
\put(1416.0,826.0){\rule[-0.200pt]{4.818pt}{0.400pt}}
\put(220.0,113.0){\rule[-0.200pt]{0.400pt}{4.818pt}}
\put(220,68){\makebox(0,0){$0$}}
\put(220.0,857.0){\rule[-0.200pt]{0.400pt}{4.818pt}}
\put(389.0,113.0){\rule[-0.200pt]{0.400pt}{4.818pt}}
\put(389,68){\makebox(0,0){$5$}}
\put(389.0,857.0){\rule[-0.200pt]{0.400pt}{4.818pt}}
\put(558.0,113.0){\rule[-0.200pt]{0.400pt}{4.818pt}}
\put(558,68){\makebox(0,0){$10$}}
\put(558.0,857.0){\rule[-0.200pt]{0.400pt}{4.818pt}}
\put(727.0,113.0){\rule[-0.200pt]{0.400pt}{4.818pt}}
\put(727,68){\makebox(0,0){$15$}}
\put(727.0,857.0){\rule[-0.200pt]{0.400pt}{4.818pt}}
\put(896.0,113.0){\rule[-0.200pt]{0.400pt}{4.818pt}}
\put(896,68){\makebox(0,0){$20$}}
\put(896.0,857.0){\rule[-0.200pt]{0.400pt}{4.818pt}}
\put(1064.0,113.0){\rule[-0.200pt]{0.400pt}{4.818pt}}
\put(1064,68){\makebox(0,0){$25$}}
\put(1064.0,857.0){\rule[-0.200pt]{0.400pt}{4.818pt}}
\put(1233.0,113.0){\rule[-0.200pt]{0.400pt}{4.818pt}}
\put(1233,68){\makebox(0,0){$30$}}
\put(1233.0,857.0){\rule[-0.200pt]{0.400pt}{4.818pt}}
\put(1402.0,113.0){\rule[-0.200pt]{0.400pt}{4.818pt}}
\put(1402,68){\makebox(0,0){$35$}}
\put(1402.0,857.0){\rule[-0.200pt]{0.400pt}{4.818pt}}
\put(220.0,113.0){\rule[-0.200pt]{292.934pt}{0.400pt}}
\put(1436.0,113.0){\rule[-0.200pt]{0.400pt}{184.048pt}}
\put(220.0,877.0){\rule[-0.200pt]{292.934pt}{0.400pt}}
\put(45,495){\makebox(0,0){\%}}
\put(828,23){\makebox(0,0){\# of interactions in model}}
\put(220.0,113.0){\rule[-0.200pt]{0.400pt}{184.048pt}}
\sbox{\plotpoint}{\rule[-0.400pt]{0.800pt}{0.800pt}}%
\put(963,317){\makebox(0,0)[r]{AIC}}
\put(985.0,317.0){\rule[-0.400pt]{15.899pt}{0.800pt}}
\put(727,801){\usebox{\plotpoint}}
\put(558,801.34){\rule{7.000pt}{0.800pt}}
\multiput(577.47,799.34)(-19.471,4.000){2}{\rule{3.500pt}{0.800pt}}
\put(592.0,801.0){\rule[-0.400pt]{32.521pt}{0.800pt}}
\put(423,805.34){\rule{6.800pt}{0.800pt}}
\multiput(441.89,803.34)(-18.886,4.000){2}{\rule{3.400pt}{0.800pt}}
\put(456.0,805.0){\rule[-0.400pt]{24.572pt}{0.800pt}}
\multiput(374.89,810.41)(-2.073,0.507){27}{\rule{3.400pt}{0.122pt}}
\multiput(381.94,807.34)(-60.943,17.000){2}{\rule{1.700pt}{0.800pt}}
\put(389.0,809.0){\rule[-0.400pt]{8.191pt}{0.800pt}}
\put(1007,317){\makebox(0,0){$\star$}}
\put(727,801){\makebox(0,0){$\star$}}
\put(693,801){\makebox(0,0){$\star$}}
\put(659,801){\makebox(0,0){$\star$}}
\put(592,801){\makebox(0,0){$\star$}}
\put(558,805){\makebox(0,0){$\star$}}
\put(490,805){\makebox(0,0){$\star$}}
\put(456,805){\makebox(0,0){$\star$}}
\put(423,809){\makebox(0,0){$\star$}}
\put(389,809){\makebox(0,0){$\star$}}
\put(321,826){\makebox(0,0){$\star$}}
\put(288,826){\makebox(0,0){$\star$}}
\put(254,826){\makebox(0,0){$\star$}}
\put(220,826){\makebox(0,0){$\star$}}
\put(220.0,826.0){\rule[-0.400pt]{24.331pt}{0.800pt}}
\sbox{\plotpoint}{\rule[-0.200pt]{0.400pt}{0.400pt}}%
\put(963,272){\makebox(0,0)[r]{BIC}}
\put(985.0,272.0){\rule[-0.200pt]{15.899pt}{0.400pt}}
\put(355,826){\usebox{\plotpoint}}
\put(1007,272){\raisebox{-.8pt}{\makebox(0,0){$\Diamond$}}}
\put(355,826){\raisebox{-.8pt}{\makebox(0,0){$\Diamond$}}}
\put(321,826){\raisebox{-.8pt}{\makebox(0,0){$\Diamond$}}}
\put(288,826){\raisebox{-.8pt}{\makebox(0,0){$\Diamond$}}}
\put(254,826){\raisebox{-.8pt}{\makebox(0,0){$\Diamond$}}}
\put(220,826){\raisebox{-.8pt}{\makebox(0,0){$\Diamond$}}}
\put(220.0,826.0){\rule[-0.200pt]{32.521pt}{0.400pt}}
\sbox{\plotpoint}{\rule[-0.500pt]{1.000pt}{1.000pt}}%
\put(963,227){\makebox(0,0)[r]{Exact $\alpha =.0001$}}
\multiput(985,227)(41.511,0.000){2}{\usebox{\plotpoint}}
\put(1051,227){\usebox{\plotpoint}}
\put(1301,363){\usebox{\plotpoint}}
\put(1301.00,363.00){\usebox{\plotpoint}}
\put(1261.90,376.95){\usebox{\plotpoint}}
\put(1223.49,392.61){\usebox{\plotpoint}}
\multiput(1200,404)(-19.715,36.531){2}{\usebox{\plotpoint}}
\put(1150.05,488.58){\usebox{\plotpoint}}
\multiput(1132,513)(-38.542,15.417){4}{\usebox{\plotpoint}}
\multiput(997,567)(-39.473,12.846){4}{\usebox{\plotpoint}}
\put(813.57,636.01){\usebox{\plotpoint}}
\multiput(794,655)(-17.139,37.807){2}{\usebox{\plotpoint}}
\multiput(760,730)(-20.528,36.080){2}{\usebox{\plotpoint}}
\put(695.88,790.29){\usebox{\plotpoint}}
\put(654.47,793.00){\usebox{\plotpoint}}
\put(612.96,793.00){\usebox{\plotpoint}}
\multiput(592,793)(-41.511,0.000){2}{\usebox{\plotpoint}}
\put(490.45,804.84){\usebox{\plotpoint}}
\multiput(490,805)(-37.129,18.564){0}{\usebox{\plotpoint}}
\multiput(456,822)(-41.511,0.000){2}{\usebox{\plotpoint}}
\put(369.96,822.00){\usebox{\plotpoint}}
\put(328.45,822.00){\usebox{\plotpoint}}
\multiput(321,822)(-41.511,0.000){0}{\usebox{\plotpoint}}
\put(286.94,822.12){\usebox{\plotpoint}}
\put(245.66,826.00){\usebox{\plotpoint}}
\put(220,826){\usebox{\plotpoint}}
\put(1007,227){\makebox(0,0){$\triangle$}}
\put(1301,363){\makebox(0,0){$\triangle$}}
\put(1267,375){\makebox(0,0){$\triangle$}}
\put(1233,388){\makebox(0,0){$\triangle$}}
\put(1200,404){\makebox(0,0){$\triangle$}}
\put(1166,467){\makebox(0,0){$\triangle$}}
\put(1132,513){\makebox(0,0){$\triangle$}}
\put(997,567){\makebox(0,0){$\triangle$}}
\put(828,622){\makebox(0,0){$\triangle$}}
\put(794,655){\makebox(0,0){$\triangle$}}
\put(760,730){\makebox(0,0){$\triangle$}}
\put(727,788){\makebox(0,0){$\triangle$}}
\put(659,793){\makebox(0,0){$\triangle$}}
\put(625,793){\makebox(0,0){$\triangle$}}
\put(592,793){\makebox(0,0){$\triangle$}}
\put(524,793){\makebox(0,0){$\triangle$}}
\put(490,805){\makebox(0,0){$\triangle$}}
\put(456,822){\makebox(0,0){$\triangle$}}
\put(389,822){\makebox(0,0){$\triangle$}}
\put(355,822){\makebox(0,0){$\triangle$}}
\put(321,822){\makebox(0,0){$\triangle$}}
\put(288,822){\makebox(0,0){$\triangle$}}
\put(254,826){\makebox(0,0){$\triangle$}}
\put(220,826){\makebox(0,0){$\triangle$}}
\put(963,182){\makebox(0,0)[r]{$G^2 \sim \chi^2$ $\alpha =.0001$}}
\multiput(985,182)(20.756,0.000){4}{\usebox{\plotpoint}}
\put(1051,182){\usebox{\plotpoint}}
\put(963,717){\usebox{\plotpoint}}
\multiput(963,717)(-20.756,0.000){2}{\usebox{\plotpoint}}
\multiput(929,717)(-15.358,13.962){2}{\usebox{\plotpoint}}
\multiput(896,747)(-20.756,0.000){2}{\usebox{\plotpoint}}
\multiput(862,747)(-20.204,4.754){2}{\usebox{\plotpoint}}
\put(811.44,763.28){\usebox{\plotpoint}}
\multiput(794,772)(-20.756,0.000){2}{\usebox{\plotpoint}}
\multiput(760,772)(-19.446,7.256){4}{\usebox{\plotpoint}}
\multiput(693,797)(-20.720,1.219){3}{\usebox{\plotpoint}}
\put(610.57,801.00){\usebox{\plotpoint}}
\multiput(592,801)(-20.613,2.425){2}{\usebox{\plotpoint}}
\multiput(558,805)(-20.756,0.000){3}{\usebox{\plotpoint}}
\multiput(490,805)(-20.756,0.000){2}{\usebox{\plotpoint}}
\multiput(456,805)(-20.605,2.498){2}{\usebox{\plotpoint}}
\put(403.49,809.00){\usebox{\plotpoint}}
\multiput(389,809)(-20.136,5.034){4}{\usebox{\plotpoint}}
\put(301.81,826.00){\usebox{\plotpoint}}
\multiput(288,826)(-20.756,0.000){2}{\usebox{\plotpoint}}
\put(239.54,826.00){\usebox{\plotpoint}}
\put(220,826){\usebox{\plotpoint}}
\put(1007,182){\circle{12}}
\put(963,717){\circle{12}}
\put(929,717){\circle{12}}
\put(896,747){\circle{12}}
\put(862,747){\circle{12}}
\put(828,755){\circle{12}}
\put(794,772){\circle{12}}
\put(760,772){\circle{12}}
\put(693,797){\circle{12}}
\put(625,801){\circle{12}}
\put(592,801){\circle{12}}
\put(558,805){\circle{12}}
\put(490,805){\circle{12}}
\put(456,805){\circle{12}}
\put(423,809){\circle{12}}
\put(389,809){\circle{12}}
\put(321,826){\circle{12}}
\put(288,826){\circle{12}}
\put(254,826){\circle{12}}
\put(220,826){\circle{12}}
\end{picture}
\caption{BSS (top) and FSS (bottom) recall for Noun {\it interest} }
\label{fig:bssrec}
\end{center}
\end{figure}

As shown in Table \ref{tab:result1a}, BIC selects models that
have fewer interactions than the other criteria. Table
\ref{tab:result1} indicates that this often results in less accurate
classification than AIC. In Figure \ref{fig:bssacc}, BSS BIC
(top) removes too many interactions and goes past the more accurate model
selected by AIC, while FSS BIC (bottom) does not add enough
interactions and stops short of selecting a highly accurate model. For
{\it interest},  BSS BIC selects a model of 6 interactions  while FSS
BIC selects a model of 4 interactions. The other combinations of
strategy and criterion select models with between 15 and 32
interactions and result in higher levels of accuracy. As mentioned
previously, the bias of BIC towards  models with small numbers of
interactions is not surprising given the large penalty that it
assesses to complexity.   

The exact conditional test suffers from the reverse problem in that it
selects models with too many interactions. BSS exact
conditional removes a small number of interactions while FSS exact
conditional adds a great many; in both cases the resulting
models have lower accuracy than the other approaches. 

Figure \ref{fig:bssrec} shows that for both forward and backward search,
models of relatively low recall are selected by the exact conditional
test. This suggests that the selected models are overly complex
and contain many parameters that can not be estimated from the training 
data. 

The contrast between the exact conditional test and the other criteria is 
stark during backward search. Figures \ref{fig:bssacc} (top) 
and
\ref{fig:bssrec} (top) 
show that the exact conditional test remains at
low levels of accuracy and recall while the other criteria
rapidly increase both recall and accuracy. 
However, during forward search Figures \ref{fig:bssacc} (bottom) and  
\ref{fig:bssrec} (bottom) show there is little difference among the 
criteria; all select high  recall models  that achieve high accuracy
early in the search. 

A backward search begins with the saturated model. For feature 
set BW saturated models have millions of parameters to estimate. 
The information criteria remove the interactions that result in models
with the highest degrees  of freedom in the early stages of the
search. This is a consequence of the complexity penalty that
both AIC and BIC  assess. The significance test $G^2 \sim \chi^2$ also
targets interactions with high degrees of freedom for removal. When
$G^2$ values are assigned significance there is an implicit weighting
for complexity. Larger degrees of freedom result  in smaller
significance values due to the nature of the $\chi^2$
distribution. Interactions that result in models with very high
degrees of freedom and are more likely to be removed early in the
course of a backward search since they have smaller significance values.   

The bias of the information criteria and $G^2 \sim \chi^2$
towards removing interactions with high degrees of freedom results in a 
rapid reduction in the number of model  parameters to estimate. This
in turn increases the percentage of model parameters that can
be estimated from the training data. In Figure \ref{fig:bssrec} (top),
recall increases rapidly as  interactions with high degrees of freedom
are removed by the information criteria. Most of the model parameters
can be estimated from the training data early in the search
process and this results in a rapid increase in accuracy. 

On the other hand, the exact conditional test does not take degrees of
freedom into account when evaluating models. Interactions are removed or 
added via significance values that are based upon the distribution of 
randomly generated values of $G^2$ relative to the observed value of 
$G^2$.  Thus, the exact conditional test may result in the removal of
interactions with relatively small degrees of freedom in cases where
the information  criteria and $G^2 \sim \chi^2$ would remove
interactions with higher degrees of freedom. During a
backward search the exact conditional test has a different bias; 
the other criteria tend to remove interactions that result in models with 
large degrees of freedom while the exact conditional test removes
interactions that maintain the fit of the model to the training data. 

However, during forward search the exact conditional test achieves
approximately the same levels of recall and accuracy as do the other
criteria. Forward search begins with the model of independence. The
information criteria add those interactions that most increase the fit of
the model; in the early stages of forward search this results in a bias
towards interactions that have lower degrees of freedom. The same
occurs with $G^2 \sim \chi^2$ since it seeks to add those interactions
that have the largest significance value. Interactions with smaller
degrees of freedom will tend to have higher significance values due to
the nature of the $\chi^2$ distribution. Thus, during a forward search all 
of the criteria are biased towards the inclusion of interactions with 
lower degrees of freedom. This results in models that have relatively
small numbers of model parameters; both recall and accuracy are likely
to be high.  

The only difference among the criteria during forward search is that
the significance tests tend to add too many interactions to the model. 
During a forward search where the evaluation criteria is a
significance test,  the value of $\alpha$ should be larger than in the
backward search in order to stop the selection process sooner. This
has the added benefit of reducing the computational complexity of the
exact conditional test since the number of random tables that must be
generated  to assess the significance of each interaction is $1/\alpha$
\cite{Kreiner87}. 

\section{Experiment 2: Naive Mix} \label{sec:nm}

The Naive Mix is defined in Section \ref{sec:nmb} and extends sequential 
model selection methods by allowing
for the incorporation of uncertainty in the development of the model.
Rather than attempting to find a single most accurate model, the Naive
Mix forms an averaged probabilistic model from the sequence of models
generated during FSS AIC. 
In this experiment the Naive Mix is compared to the following machine 
learning algorithms: 

{\bf PEBLS} \cite{CostS93}: A $k$ nearest--neighbor algorithm where
classification is performed by assigning an ambiguous word to the
majority class of the $k$--nearest training examples.  In these
experiments each ambiguous word is assigned the sense of the single
most similar training example, i.e., $k=1$.   

{\bf C4.5} \cite{Quinlan92}: A decision tree learner in which
classification rules are formulated by recursively partitioning the
training sample.  Each nested partition is based on the feature value
that provides the greatest increase in the information gain ratio for
the current partition.  

{\bf CN2} \cite{ClarkN89}: A rule induction algorithm that selects 
classification rules that cover the largest possible subsets of the
training sample as measured by the Laplace error estimate.

The Naive Mix, C4.5, CN2, and any model selection method using forward
sequential search  all perform general--to--specific searches that
add features to the learned representation of the training sample
based on some measure of information content  increase. These methods
all perform  feature selection and have a bias towards simpler models.
All of these methods can suffer from {\it fragmentation} when learning
from sparse training data. Fragmentation occurs when the model is
complex, incorporating a large number of feature  values to describe a
small number  of training instances.  When this occurs, there is
inadequate support in the training data for the inference being
specified by the model.  The Naive Mix is designed to reduce the
effects of fragmentation in a general--to--specific search by
averaging the distributions of high complexity models with those of
low complexity models that include only the most relevant features. 

The nearest--neighbor algorithm PEBLS shares a number of traits  with
Naive Bayes.  Neither perform a search to create a representation of the
training sample. Naive Bayes assumes the form of a model in which all
features are  regarded as relevant to disambiguation but, as in PEBLS,
their   interdependencies are not considered.  Weights are assigned to
features via parameter estimates from  the training sample. These weights
allow some discounting of less relevant features. Here, PEBLS stores all 
instances of the training sample and treats each feature independently and 
equally, making it susceptible to irrelevant features.  

Table \ref{tab:result2} shows the accuracy of the Naive Mix, Naive Bayes, 
the majority classifier, C4.5, CN2, and PEBLS.  The accuracies of the
sequential model selection methods from the first experiment are
directly comparable to these since both are evaluated via 10--fold
cross validation using all of the sense--tagged examples for each word. 

\begin{table}
\begin{center}
\caption{Naive Mix and Machine Learning Accuracy}
\label{tab:result2}
\begin{tabular}{|l|c|c|c|c|c|c|} \hline
     &   Majority   & Naive &PEBLS  &  &   &  Naive \\
     &   classifier  & Bayes & k=1&C4.5& CN2 & Mix \\ 
\hline
agree &  .777 .032 & .930 .026  & .928 .030 & .947 .031 & .947
.031
& .948 .017\\
bill &  .681 .044 & .865 .026  & .855 .034  & .878 .029 & .873
.035
& {\bf .897 .026} \\ 
chief &  .862 .026 & .943 .015  & .945 .018  & .947 .020 & .945
.013 
& .951 .016 \\
close &  .680 .033 & .817 .023  & .843 .042  & {\bf .853 .021} 
& .834 .036  & .831 .033 \\
common &  .802 .029 & .832 .034  & .853 .019  & .871 .030 & .803
.029
& .853 .024 \\ 
concern & .639 .054 & .859 .037  & .840 .036  & .852 .042 & .859
.033
& .846 .039 \\ 
drug   & .575 .033 & .807 .036  & .778 .034  & .798 .038 & .777
.069 
& .815 .041 \\  
help  & .753 .032 & .780 .033  & .710 .047  & .790 .039 & .779
.045 
& .796 .038 \\
include  & .912 .024 & .944 .021  & .939 .015  & .954 .019 & .951
.018
& .956 .018 \\
interest & .529 .026 & .763 .016  & .768 .020  & {\bf .793 .019} 
& .729 .034 & {\bf .800 .019} \\ 
last   & .933 .014 & .919 .011 & {\bf .947 .012}  & {\bf .945 .008} 
& .935 .013 & {\bf .940 .016} \\ 
public & .560 .055 & .593 .054  & .536 .039 & .598 .047  & .579
.057
& .615 .055 \\
\hline \hline
\multicolumn{1}{|l}{average} &
\multicolumn{1}{|l}{.725} &
\multicolumn{1}{|l}{.838} &
\multicolumn{1}{|l}{.829} &
\multicolumn{1}{|l}{.852} &
\multicolumn{1}{|l}{.834} &
\multicolumn{1}{|l|}{.854} \\
\hline
\end{tabular} 
\end{center}
\vskip -0.01in
\end{table}

Based on a word by word comparison, this experiment shows that the
Naive Mix improves upon the accuracy of models selected by FSS
AIC. However, in general there prove to be few significant differences
between the accuracy of the Naive Mix, Naive Bayes, C4.5, PEBLS, and CN2. 

The success of Naive Bayes in the first two experiments is a bit
surprising since it neither performs feature selection nor a
model search. Despite this, it is among the most accurate of the methods
considered in this study. 

Table \ref{tab:nb} summarizes the accuracy of Naive Bayes relative to
the other methods employed in the first two experiments. The 
accuracy reported is based on 10--fold cross validation. 
The {\it win--tie--loss} measure is also shown; this indicates the
number of  times Naive Bayes is significantly more--equally--less
accurate than the competing method. The win--tie--loss record 1--10--1
associated with PEBLS means that Naive Bayes is significantly more
accurate than PEBLS for 1 word, not significantly different than PEBLS
for 10 words, and significantly less accurate than PEBLS for 1
word. This measure shows that there are only 7 out of a possible 144
cases where Naive Bayes is significantly less accurate than a competing
method. The cases where a competing method is significantly more
accurate than Naive Bayes are shown in Table \ref{tab:result2} in bold
face. There is no such case in Table \ref{tab:result1}; Naive Bayes is
never significantly less accurate than a sequential model selection
method in the first experiment.

\begin{table}
\begin{center}
\caption{Naive Bayes Comparison}
\label{tab:nb}
\begin{tabular}{|l|c|c|} \hline
            &   Average   & win--tie--loss \\
            &   Accuracy  & \\ \hline
Naive Mix   & .854 & 0--9--3 \\
C4.5        & .852 & 0--9--3 \\
Naive Bayes & .838 & \\
FSS AIC     & .836 & 0--12--0 \\
BSS AIC     & .829 & 0--12--0 \\
PEBLS       & .829 & 1--10--1 \\
FSS BIC     & .828 & 1--11--0 \\
FSS $G^2$   & .819 & 1--11--0 \\
BSS BIC     & .813 & 3--9--0  \\
BSS exact   & .761 & 6--6--0  \\
BSS $G^2$   & .746 & 5--7--0  \\
Majority    & .725 & 8--4--0  \\
FSS exact   & .712 & 9--3--0  \\ \hline
\end{tabular} 
\end{center}
\vskip -0.01in
\end{table}

The success of Naive Bayes in these experiments confirms
the results of previous studies of disambiguation. For instance,
\cite{LeacockTV93} compares a neural network, Naive Bayes,
and a content vector when disambiguating six senses of
{\it line}.\footnote{This data is described in Chapter 5, Experimental
Data.} They report that all three methods are equally accurate. The
{\it line} data is utilized again in \cite{Mooney96} with an even
wider range of methods. Naive Bayes, a  perceptron, a
decision tree learner, a nearest--neighbor classifier, a logic  based
disjunctive normal form learner, a logic based conjunctive normal form
learner, and a decision list learner are compared.  Naive Bayes and
the perceptron are found to be the most accurate approaches. Finally,
\cite{Ng97} compare PEBLS and Naive Bayes and finds them to be of
comparable accuracy when disambiguating the Defence Science
Organization sense--tagged corpus \cite{NgL96}. However, all of these
studies differ from  this dissertation in that they employ a feature
set that consists of thousands of binary co--occurrence features, each
of which represents the  occurrence of  a particular word within some
fixed distance of the  ambiguous word. This feature set is commonly
known as {\it bag--of--words}.       

The relatively high accuracy achieved by Naive Bayes in disambiguation
is sometimes explained as a consequence of the bag--of--words feature
set, e.g., \cite{PedersenB97A}. Given so many features, 
the assumptions of conditional independence made by Naive Bayes are
potentially valid  and may result in a model that fits the training data
reasonably well. However, this explanation does not apply to feature
set BW since a previous study \cite{BruceWP96} shows that all of these
features are good indicators of the sense of the ambiguous word. 
In addition, Naive Bayes is successful in a number of other domains
where the bag--of--words explanation is not relevant. 

For example, \cite{ClarkN89} compare  Naive Bayes, a rule induction 
system, and a decision tree learner. They  find that  Naive Bayes performs 
as accurately as these more sophisticated  methods in various medical 
diagnosis problems.   

A more extensive study of Naive Bayes appears in \cite{LangleyIT92}. They 
compare Naive Bayes and a decision tree learner using data from the
University of California at Irvine (UCI) Machine Learning repository
\cite{MerzMA97}. For 4 of 5 naturally  occurring data sets they report
that Naive Bayes is the more accurate.  They also present an average
case analysis of Naive Bayes that is verified  empirically using
artificial data.   

Naive Bayes and more elaborate Bayesian networks that diagnose the cause 
of acute abdominal pain are compared in \cite{ProvanS96}. They argue that 
simple classification models will often outperform more detailed
ones if the domain is complex and the amount of data available is
relatively small. Their experiment consists of 1270 cases, 
each of which has 169 features. They find that the Naive Bayes model
with 169 interactions is more accurate than a Bayesian network that
has 590 interactions.    

A software agent that learns to rate Web pages according to a user's level 
of interest is discussed in \cite{PazzaniMB96}. They construct a  profile 
using examples of pages that a user likes and dislikes. They compare
Naive Bayes,  a nearest--neighbor algorithm, a term--weighting method
from information retrieval, a perceptron, and a multi--layer neural
network  and find that Naive Bayes is most accurate at predicting Web
pages a user will find interesting.   

Finally, \cite{DomingosP97} compare the accuracy of Naive Bayes with
a decision tree learner, a nearest--neighbor algorithm, and a rule
induction system.  They report that Naive Bayes is at least as accurate as 
the rule induction system and nearest--neighbor algorithm for 22 of 28
UCI data sets and at least as accurate as the  decision tree learner 
for 20 of 28 data sets. They also present an extensive analysis of
the  conditions under which Naive Bayes is an optimal classifier
even when the conditional independence assumptions are not valid.  

\section{Experiment 3: Learning Rate} \label{sec:lr}

The first two experiments suggest that Naive Bayes may be an effective
general purpose method of disambiguation. However, these experiments
only study the disambiguation accuracy of models that are learned from
relatively large amounts of training data, i.e, 90\% of the total
available sense--tagged text for a word. 

The third experiment differentiates among the most accurate
methods in the first two experiments by training each algorithm with
steadily increasing amounts of training data and studying the {\it
learning rate}.  This shows the relationship between accuracy 
and the number of training examples. A ``slow'' learning rate implies that
accuracy gradually increases with the number of training examples. 
A ``fast'' learning rate suggests an algorithm that
attains high accuracy with a very small number of examples. This
experiment compares the learning rates of the Naive Mix, Naive Bayes,
C4.5, and FSS AIC.   

A variant of 10--fold cross validation is employed. Each word--corpus is  
divided into 10 folds; the desired number of training examples are
sampled from 9 folds and the remaining fold is held out as the test
set. Each algorithm learns a model from the training data and uses
this to disambiguate the test set.  This is repeated until
each fold serves as the test set once. The accuracy reported is
averaged over all 10 folds.         

This procedure is repeated with increasing quantities of training
data. In this experiment the number of training examples is
first 10, then 50, and then 100. Thereafter the number of examples is 
incremented 100 at a time until all the available training data is
used. For each amount of training data the accuracy attained for all
the words belonging to a particular part--of--speech are
averaged. These values are plotted in Figures \ref{fig:lradj} through
\ref{fig:lrverb} and show the learning rate for each method for each
part--of--speech.  Also included is the learning rate of the majority
classifier.  This proves to be constant since it classifies every
held--out instance of an ambiguous word with the most frequent sense
in the training data. This is easy to correctly identify even  with
very small amounts of training data due to the skewed sense
distributions.  

\begin{figure}
\begin{center}
\setlength{\unitlength}{0.240900pt}
\ifx\plotpoint\undefined\newsavebox{\plotpoint}\fi
\sbox{\plotpoint}{\rule[-0.200pt]{0.400pt}{0.400pt}}%
\begin{picture}(1500,900)(0,0)
\font\gnuplot=cmr10 at 10pt
\gnuplot
\sbox{\plotpoint}{\rule[-0.200pt]{0.400pt}{0.400pt}}%
\put(176.0,68.0){\rule[-0.200pt]{303.534pt}{0.400pt}}
\put(176.0,68.0){\rule[-0.200pt]{0.400pt}{194.888pt}}
\put(176.0,68.0){\rule[-0.200pt]{4.818pt}{0.400pt}}
\put(154,68){\makebox(0,0)[r]{$0$}}
\put(1416.0,68.0){\rule[-0.200pt]{4.818pt}{0.400pt}}
\put(176.0,222.0){\rule[-0.200pt]{4.818pt}{0.400pt}}
\put(154,222){\makebox(0,0)[r]{$0.2$}}
\put(1416.0,222.0){\rule[-0.200pt]{4.818pt}{0.400pt}}
\put(176.0,376.0){\rule[-0.200pt]{4.818pt}{0.400pt}}
\put(154,376){\makebox(0,0)[r]{$0.4$}}
\put(1416.0,376.0){\rule[-0.200pt]{4.818pt}{0.400pt}}
\put(176.0,530.0){\rule[-0.200pt]{4.818pt}{0.400pt}}
\put(154,530){\makebox(0,0)[r]{$0.6$}}
\put(1416.0,530.0){\rule[-0.200pt]{4.818pt}{0.400pt}}
\put(176.0,684.0){\rule[-0.200pt]{4.818pt}{0.400pt}}
\put(154,684){\makebox(0,0)[r]{$0.8$}}
\put(1416.0,684.0){\rule[-0.200pt]{4.818pt}{0.400pt}}
\put(176.0,838.0){\rule[-0.200pt]{4.818pt}{0.400pt}}
\put(154,838){\makebox(0,0)[r]{$1$}}
\put(1416.0,838.0){\rule[-0.200pt]{4.818pt}{0.400pt}}
\put(176.0,68.0){\rule[-0.200pt]{0.400pt}{4.818pt}}
\put(176,23){\makebox(0,0){$0$}}
\put(176.0,857.0){\rule[-0.200pt]{0.400pt}{4.818pt}}
\put(316.0,68.0){\rule[-0.200pt]{0.400pt}{4.818pt}}
\put(316,23){\makebox(0,0){$100$}}
\put(316.0,857.0){\rule[-0.200pt]{0.400pt}{4.818pt}}
\put(456.0,68.0){\rule[-0.200pt]{0.400pt}{4.818pt}}
\put(456,23){\makebox(0,0){$200$}}
\put(456.0,857.0){\rule[-0.200pt]{0.400pt}{4.818pt}}
\put(596.0,68.0){\rule[-0.200pt]{0.400pt}{4.818pt}}
\put(596,23){\makebox(0,0){$300$}}
\put(596.0,857.0){\rule[-0.200pt]{0.400pt}{4.818pt}}
\put(736.0,68.0){\rule[-0.200pt]{0.400pt}{4.818pt}}
\put(736,23){\makebox(0,0){$400$}}
\put(736.0,857.0){\rule[-0.200pt]{0.400pt}{4.818pt}}
\put(876.0,68.0){\rule[-0.200pt]{0.400pt}{4.818pt}}
\put(876,23){\makebox(0,0){$500$}}
\put(876.0,857.0){\rule[-0.200pt]{0.400pt}{4.818pt}}
\put(1016.0,68.0){\rule[-0.200pt]{0.400pt}{4.818pt}}
\put(1016,23){\makebox(0,0){$600$}}
\put(1016.0,857.0){\rule[-0.200pt]{0.400pt}{4.818pt}}
\put(1156.0,68.0){\rule[-0.200pt]{0.400pt}{4.818pt}}
\put(1156,23){\makebox(0,0){$700$}}
\put(1156.0,857.0){\rule[-0.200pt]{0.400pt}{4.818pt}}
\put(1296.0,68.0){\rule[-0.200pt]{0.400pt}{4.818pt}}
\put(1296,23){\makebox(0,0){$800$}}
\put(1296.0,857.0){\rule[-0.200pt]{0.400pt}{4.818pt}}
\put(1436.0,68.0){\rule[-0.200pt]{0.400pt}{4.818pt}}
\put(1436,23){\makebox(0,0){$900$}}
\put(1436.0,857.0){\rule[-0.200pt]{0.400pt}{4.818pt}}
\put(176.0,68.0){\rule[-0.200pt]{303.534pt}{0.400pt}}
\put(1436.0,68.0){\rule[-0.200pt]{0.400pt}{194.888pt}}
\put(176.0,877.0){\rule[-0.200pt]{303.534pt}{0.400pt}}
\put(176.0,68.0){\rule[-0.200pt]{0.400pt}{194.888pt}}
\put(1156,376){\makebox(0,0)[r]{Naive Bayes}}
\put(1178.0,376.0){\rule[-0.200pt]{15.899pt}{0.400pt}}
\put(190,335){\usebox{\plotpoint}}
\multiput(190.58,335.00)(0.499,2.126){109}{\rule{0.120pt}{1.793pt}}
\multiput(189.17,335.00)(56.000,233.279){2}{\rule{0.400pt}{0.896pt}}
\multiput(246.00,572.58)(0.625,0.499){109}{\rule{0.600pt}{0.120pt}}
\multiput(246.00,571.17)(68.755,56.000){2}{\rule{0.300pt}{0.400pt}}
\multiput(316.00,628.58)(2.815,0.498){97}{\rule{2.340pt}{0.120pt}}
\multiput(316.00,627.17)(275.143,50.000){2}{\rule{1.170pt}{0.400pt}}
\multiput(596.00,678.61)(62.305,0.447){3}{\rule{37.433pt}{0.108pt}}
\multiput(596.00,677.17)(202.305,3.000){2}{\rule{18.717pt}{0.400pt}}
\multiput(876.00,681.61)(62.305,0.447){3}{\rule{37.433pt}{0.108pt}}
\multiput(876.00,680.17)(202.305,3.000){2}{\rule{18.717pt}{0.400pt}}
\multiput(1156.00,684.58)(7.121,0.496){37}{\rule{5.700pt}{0.119pt}}
\multiput(1156.00,683.17)(268.169,20.000){2}{\rule{2.850pt}{0.400pt}}
\sbox{\plotpoint}{\rule[-0.500pt]{1.000pt}{1.000pt}}%
\put(1156,331){\makebox(0,0)[r]{C4.5}}
\multiput(1178,331)(20.756,0.000){4}{\usebox{\plotpoint}}
\put(1244,331){\usebox{\plotpoint}}
\put(190,684){\usebox{\plotpoint}}
\multiput(190,684)(19.546,-6.981){3}{\usebox{\plotpoint}}
\multiput(246,664)(18.241,9.902){4}{\usebox{\plotpoint}}
\multiput(316,702)(20.752,-0.371){14}{\usebox{\plotpoint}}
\multiput(596,697)(20.722,1.184){13}{\usebox{\plotpoint}}
\multiput(876,713)(20.745,-0.667){14}{\usebox{\plotpoint}}
\multiput(1156,704)(20.752,0.371){13}{\usebox{\plotpoint}}
\put(1436,709){\usebox{\plotpoint}}
\sbox{\plotpoint}{\rule[-0.400pt]{0.800pt}{0.800pt}}%
\put(1156,286){\makebox(0,0)[r]{Naive Mix}}
\put(1178.0,286.0){\rule[-0.400pt]{15.899pt}{0.800pt}}
\put(190,686){\usebox{\plotpoint}}
\put(190,683.84){\rule{13.490pt}{0.800pt}}
\multiput(190.00,684.34)(28.000,-1.000){2}{\rule{6.745pt}{0.800pt}}
\multiput(246.00,686.41)(2.277,0.507){25}{\rule{3.700pt}{0.122pt}}
\multiput(246.00,683.34)(62.320,16.000){2}{\rule{1.850pt}{0.800pt}}
\multiput(316.00,702.41)(10.638,0.509){21}{\rule{16.200pt}{0.123pt}}
\multiput(316.00,699.34)(246.376,14.000){2}{\rule{8.100pt}{0.800pt}}
\put(596,714.84){\rule{67.452pt}{0.800pt}}
\multiput(596.00,713.34)(140.000,3.000){2}{\rule{33.726pt}{0.800pt}}
\multiput(876.00,716.08)(24.389,-0.526){7}{\rule{32.200pt}{0.127pt}}
\multiput(876.00,716.34)(213.167,-7.000){2}{\rule{16.100pt}{0.800pt}}
\multiput(1156.00,712.41)(12.605,0.511){17}{\rule{18.867pt}{0.123pt}}
\multiput(1156.00,709.34)(240.841,12.000){2}{\rule{9.433pt}{0.800pt}}
\sbox{\plotpoint}{\rule[-0.500pt]{1.000pt}{1.000pt}}%
\put(1156,241){\makebox(0,0)[r]{FSS AIC}}
\multiput(1178,241)(41.511,0.000){2}{\usebox{\plotpoint}}
\put(1244,241){\usebox{\plotpoint}}
\put(190,671){\usebox{\plotpoint}}
\multiput(190,671)(40.985,6.587){2}{\usebox{\plotpoint}}
\multiput(246,680)(41.243,4.713){2}{\usebox{\plotpoint}}
\multiput(316,688)(41.372,3.398){6}{\usebox{\plotpoint}}
\multiput(596,711)(41.498,1.037){7}{\usebox{\plotpoint}}
\multiput(876,718)(41.452,-2.221){7}{\usebox{\plotpoint}}
\multiput(1156,703)(41.443,2.368){7}{\usebox{\plotpoint}}
\put(1436,719){\usebox{\plotpoint}}
\sbox{\plotpoint}{\rule[-0.200pt]{0.400pt}{0.400pt}}%
\put(1156,196){\makebox(0,0)[r]{Majority}}
\multiput(1178,196)(20.756,0.000){4}{\usebox{\plotpoint}}
\put(1244,196){\usebox{\plotpoint}}
\put(190,684){\usebox{\plotpoint}}
\multiput(190,684)(20.756,0.000){3}{\usebox{\plotpoint}}
\multiput(246,684)(20.756,0.000){4}{\usebox{\plotpoint}}
\multiput(316,684)(20.756,0.000){13}{\usebox{\plotpoint}}
\multiput(596,684)(20.756,0.000){14}{\usebox{\plotpoint}}
\multiput(876,684)(20.756,0.000){13}{\usebox{\plotpoint}}
\multiput(1156,684)(20.756,0.000){14}{\usebox{\plotpoint}}
\put(1436,684){\usebox{\plotpoint}}
\end{picture}
\caption{Learning Rate for Adjectives}
\label{fig:lradj}
\end{center}
\myendfig

{\bf Adjectives} C4.5, FSS AIC, and the Naive Mix achieve nearly the
same level of accuracy learning from 10 examples as they do
900 examples. Naive Bayes has a slower learning rate; accuracy is
low with a small  number of examples but improves with the addition of
training data.  Naive Bayes achieves  approximately  the same accuracy
as C4.5, FSS AIC,  and  the Naive Mix after 300 training
examples. However, none of the methods significantly exceeds the
accuracy of the majority classifier.  

\begin{figure}
\begin{center}
\setlength{\unitlength}{0.240900pt}
\ifx\plotpoint\undefined\newsavebox{\plotpoint}\fi
\sbox{\plotpoint}{\rule[-0.200pt]{0.400pt}{0.400pt}}%
\begin{picture}(1500,900)(0,0)
\font\gnuplot=cmr10 at 10pt
\gnuplot
\sbox{\plotpoint}{\rule[-0.200pt]{0.400pt}{0.400pt}}%
\put(176.0,68.0){\rule[-0.200pt]{303.534pt}{0.400pt}}
\put(176.0,68.0){\rule[-0.200pt]{0.400pt}{194.888pt}}
\put(176.0,68.0){\rule[-0.200pt]{4.818pt}{0.400pt}}
\put(154,68){\makebox(0,0)[r]{$0$}}
\put(1416.0,68.0){\rule[-0.200pt]{4.818pt}{0.400pt}}
\put(176.0,222.0){\rule[-0.200pt]{4.818pt}{0.400pt}}
\put(154,222){\makebox(0,0)[r]{$0.2$}}
\put(1416.0,222.0){\rule[-0.200pt]{4.818pt}{0.400pt}}
\put(176.0,376.0){\rule[-0.200pt]{4.818pt}{0.400pt}}
\put(154,376){\makebox(0,0)[r]{$0.4$}}
\put(1416.0,376.0){\rule[-0.200pt]{4.818pt}{0.400pt}}
\put(176.0,530.0){\rule[-0.200pt]{4.818pt}{0.400pt}}
\put(154,530){\makebox(0,0)[r]{$0.6$}}
\put(1416.0,530.0){\rule[-0.200pt]{4.818pt}{0.400pt}}
\put(176.0,684.0){\rule[-0.200pt]{4.818pt}{0.400pt}}
\put(154,684){\makebox(0,0)[r]{$0.8$}}
\put(1416.0,684.0){\rule[-0.200pt]{4.818pt}{0.400pt}}
\put(176.0,838.0){\rule[-0.200pt]{4.818pt}{0.400pt}}
\put(154,838){\makebox(0,0)[r]{$1$}}
\put(1416.0,838.0){\rule[-0.200pt]{4.818pt}{0.400pt}}
\put(176.0,68.0){\rule[-0.200pt]{0.400pt}{4.818pt}}
\put(176,23){\makebox(0,0){$0$}}
\put(176.0,857.0){\rule[-0.200pt]{0.400pt}{4.818pt}}
\put(405.0,68.0){\rule[-0.200pt]{0.400pt}{4.818pt}}
\put(405,23){\makebox(0,0){$200$}}
\put(405.0,857.0){\rule[-0.200pt]{0.400pt}{4.818pt}}
\put(634.0,68.0){\rule[-0.200pt]{0.400pt}{4.818pt}}
\put(634,23){\makebox(0,0){$400$}}
\put(634.0,857.0){\rule[-0.200pt]{0.400pt}{4.818pt}}
\put(863.0,68.0){\rule[-0.200pt]{0.400pt}{4.818pt}}
\put(863,23){\makebox(0,0){$600$}}
\put(863.0,857.0){\rule[-0.200pt]{0.400pt}{4.818pt}}
\put(1092.0,68.0){\rule[-0.200pt]{0.400pt}{4.818pt}}
\put(1092,23){\makebox(0,0){$800$}}
\put(1092.0,857.0){\rule[-0.200pt]{0.400pt}{4.818pt}}
\put(1321.0,68.0){\rule[-0.200pt]{0.400pt}{4.818pt}}
\put(1321,23){\makebox(0,0){$1000$}}
\put(1321.0,857.0){\rule[-0.200pt]{0.400pt}{4.818pt}}
\put(176.0,68.0){\rule[-0.200pt]{303.534pt}{0.400pt}}
\put(1436.0,68.0){\rule[-0.200pt]{0.400pt}{194.888pt}}
\put(176.0,877.0){\rule[-0.200pt]{303.534pt}{0.400pt}}
\put(176.0,68.0){\rule[-0.200pt]{0.400pt}{194.888pt}}
\put(1207,376){\makebox(0,0)[r]{Naive Bayes}}
\put(1229.0,376.0){\rule[-0.200pt]{15.899pt}{0.400pt}}
\put(187,148){\usebox{\plotpoint}}
\multiput(187.58,148.00)(0.498,3.511){89}{\rule{0.120pt}{2.891pt}}
\multiput(186.17,148.00)(46.000,314.999){2}{\rule{0.400pt}{1.446pt}}
\multiput(233.58,469.00)(0.499,0.838){113}{\rule{0.120pt}{0.769pt}}
\multiput(232.17,469.00)(58.000,95.404){2}{\rule{0.400pt}{0.384pt}}
\multiput(291.00,566.58)(1.015,0.499){223}{\rule{0.911pt}{0.120pt}}
\multiput(291.00,565.17)(227.110,113.000){2}{\rule{0.455pt}{0.400pt}}
\multiput(520.00,679.58)(5.539,0.496){39}{\rule{4.462pt}{0.119pt}}
\multiput(520.00,678.17)(219.739,21.000){2}{\rule{2.231pt}{0.400pt}}
\multiput(749.00,700.59)(17.426,0.485){11}{\rule{13.186pt}{0.117pt}}
\multiput(749.00,699.17)(201.632,7.000){2}{\rule{6.593pt}{0.400pt}}
\multiput(978.00,705.93)(17.426,-0.485){11}{\rule{13.186pt}{0.117pt}}
\multiput(978.00,706.17)(201.632,-7.000){2}{\rule{6.593pt}{0.400pt}}
\put(1207,699.67){\rule{55.166pt}{0.400pt}}
\multiput(1207.00,699.17)(114.500,1.000){2}{\rule{27.583pt}{0.400pt}}
\sbox{\plotpoint}{\rule[-0.500pt]{1.000pt}{1.000pt}}%
\put(1207,331){\makebox(0,0)[r]{C4.5}}
\multiput(1229,331)(20.756,0.000){4}{\usebox{\plotpoint}}
\put(1295,331){\usebox{\plotpoint}}
\put(187,564){\usebox{\plotpoint}}
\multiput(187,564)(10.061,18.154){5}{\usebox{\plotpoint}}
\multiput(233,647)(20.392,3.867){3}{\usebox{\plotpoint}}
\multiput(291,658)(20.556,2.872){11}{\usebox{\plotpoint}}
\multiput(520,690)(20.732,0.996){11}{\usebox{\plotpoint}}
\multiput(749,701)(20.755,-0.091){11}{\usebox{\plotpoint}}
\multiput(978,700)(20.736,0.905){11}{\usebox{\plotpoint}}
\multiput(1207,710)(20.743,0.725){11}{\usebox{\plotpoint}}
\put(1436,718){\usebox{\plotpoint}}
\sbox{\plotpoint}{\rule[-0.400pt]{0.800pt}{0.800pt}}%
\put(1207,286){\makebox(0,0)[r]{Naive Mix}}
\put(1229.0,286.0){\rule[-0.400pt]{15.899pt}{0.800pt}}
\put(187,533){\usebox{\plotpoint}}
\multiput(188.41,533.00)(0.502,1.338){85}{\rule{0.121pt}{2.322pt}}
\multiput(185.34,533.00)(46.000,117.181){2}{\rule{0.800pt}{1.161pt}}
\multiput(233.00,653.08)(3.577,-0.516){11}{\rule{5.356pt}{0.124pt}}
\multiput(233.00,653.34)(46.884,-9.000){2}{\rule{2.678pt}{0.800pt}}
\multiput(291.00,647.41)(2.576,0.502){83}{\rule{4.271pt}{0.121pt}}
\multiput(291.00,644.34)(220.135,45.000){2}{\rule{2.136pt}{0.800pt}}
\multiput(520.00,692.41)(5.378,0.505){37}{\rule{8.527pt}{0.122pt}}
\multiput(520.00,689.34)(211.301,22.000){2}{\rule{4.264pt}{0.800pt}}
\multiput(749.00,714.40)(12.677,0.514){13}{\rule{18.520pt}{0.124pt}}
\multiput(749.00,711.34)(190.561,10.000){2}{\rule{9.260pt}{0.800pt}}
\multiput(978.00,721.09)(5.378,-0.505){37}{\rule{8.527pt}{0.122pt}}
\multiput(978.00,721.34)(211.301,-22.000){2}{\rule{4.264pt}{0.800pt}}
\multiput(1207.00,702.41)(6.640,0.506){29}{\rule{10.378pt}{0.122pt}}
\multiput(1207.00,699.34)(207.460,18.000){2}{\rule{5.189pt}{0.800pt}}
\sbox{\plotpoint}{\rule[-0.500pt]{1.000pt}{1.000pt}}%
\put(1207,241){\makebox(0,0)[r]{FSS AIC}}
\multiput(1229,241)(41.511,0.000){2}{\usebox{\plotpoint}}
\put(1295,241){\usebox{\plotpoint}}
\put(187,529){\usebox{\plotpoint}}
\multiput(187,529)(15.892,38.348){3}{\usebox{\plotpoint}}
\multiput(233,640)(41.486,-1.431){2}{\usebox{\plotpoint}}
\multiput(291,638)(40.798,7.661){5}{\usebox{\plotpoint}}
\multiput(520,681)(41.410,2.893){6}{\usebox{\plotpoint}}
\multiput(749,697)(41.471,1.811){5}{\usebox{\plotpoint}}
\multiput(978,707)(41.497,-1.087){6}{\usebox{\plotpoint}}
\multiput(1207,701)(41.471,1.811){5}{\usebox{\plotpoint}}
\put(1436,711){\usebox{\plotpoint}}
\sbox{\plotpoint}{\rule[-0.200pt]{0.400pt}{0.400pt}}%
\put(1207,196){\makebox(0,0)[r]{Majority}}
\multiput(1229,196)(20.756,0.000){4}{\usebox{\plotpoint}}
\put(1295,196){\usebox{\plotpoint}}
\put(187,552){\usebox{\plotpoint}}
\multiput(187,552)(20.756,0.000){3}{\usebox{\plotpoint}}
\multiput(233,552)(20.756,0.000){3}{\usebox{\plotpoint}}
\multiput(291,552)(20.756,0.000){11}{\usebox{\plotpoint}}
\multiput(520,552)(20.756,0.000){11}{\usebox{\plotpoint}}
\multiput(749,552)(20.756,0.000){11}{\usebox{\plotpoint}}
\multiput(978,552)(20.756,0.000){11}{\usebox{\plotpoint}}
\multiput(1207,552)(20.756,0.000){11}{\usebox{\plotpoint}}
\put(1436,552){\usebox{\plotpoint}}
\end{picture}
\caption{Learning Rate for Nouns}
\label{fig:lrnoun}
\end{center}
\myendfig

{\bf Nouns} C4.5, FSS AIC, and the Naive Mix are nearly as
accurate as the  majority classifier after only learning from 10
training examples. However, unlike the adjectives, accuracy increases
with additional training data and significantly exceeds the majority
classifier. Like the adjectives, Naive Bayes begins at very low
accuracy but reaches the same level as C4.5, FSS AIC, and the Naive
Mix  when approximately 300 training examples are available.   

\begin{figure}
\begin{center}
\setlength{\unitlength}{0.240900pt}
\ifx\plotpoint\undefined\newsavebox{\plotpoint}\fi
\sbox{\plotpoint}{\rule[-0.200pt]{0.400pt}{0.400pt}}%
\begin{picture}(1500,900)(0,0)
\font\gnuplot=cmr10 at 10pt
\gnuplot
\sbox{\plotpoint}{\rule[-0.200pt]{0.400pt}{0.400pt}}%
\put(176.0,68.0){\rule[-0.200pt]{303.534pt}{0.400pt}}
\put(176.0,68.0){\rule[-0.200pt]{0.400pt}{194.888pt}}
\put(176.0,68.0){\rule[-0.200pt]{4.818pt}{0.400pt}}
\put(154,68){\makebox(0,0)[r]{$0$}}
\put(1416.0,68.0){\rule[-0.200pt]{4.818pt}{0.400pt}}
\put(176.0,222.0){\rule[-0.200pt]{4.818pt}{0.400pt}}
\put(154,222){\makebox(0,0)[r]{$0.2$}}
\put(1416.0,222.0){\rule[-0.200pt]{4.818pt}{0.400pt}}
\put(176.0,376.0){\rule[-0.200pt]{4.818pt}{0.400pt}}
\put(154,376){\makebox(0,0)[r]{$0.4$}}
\put(1416.0,376.0){\rule[-0.200pt]{4.818pt}{0.400pt}}
\put(176.0,530.0){\rule[-0.200pt]{4.818pt}{0.400pt}}
\put(154,530){\makebox(0,0)[r]{$0.6$}}
\put(1416.0,530.0){\rule[-0.200pt]{4.818pt}{0.400pt}}
\put(176.0,684.0){\rule[-0.200pt]{4.818pt}{0.400pt}}
\put(154,684){\makebox(0,0)[r]{$0.8$}}
\put(1416.0,684.0){\rule[-0.200pt]{4.818pt}{0.400pt}}
\put(176.0,838.0){\rule[-0.200pt]{4.818pt}{0.400pt}}
\put(154,838){\makebox(0,0)[r]{$1$}}
\put(1416.0,838.0){\rule[-0.200pt]{4.818pt}{0.400pt}}
\put(176.0,68.0){\rule[-0.200pt]{0.400pt}{4.818pt}}
\put(176,23){\makebox(0,0){$0$}}
\put(176.0,857.0){\rule[-0.200pt]{0.400pt}{4.818pt}}
\put(370.0,68.0){\rule[-0.200pt]{0.400pt}{4.818pt}}
\put(370,23){\makebox(0,0){$200$}}
\put(370.0,857.0){\rule[-0.200pt]{0.400pt}{4.818pt}}
\put(564.0,68.0){\rule[-0.200pt]{0.400pt}{4.818pt}}
\put(564,23){\makebox(0,0){$400$}}
\put(564.0,857.0){\rule[-0.200pt]{0.400pt}{4.818pt}}
\put(758.0,68.0){\rule[-0.200pt]{0.400pt}{4.818pt}}
\put(758,23){\makebox(0,0){$600$}}
\put(758.0,857.0){\rule[-0.200pt]{0.400pt}{4.818pt}}
\put(951.0,68.0){\rule[-0.200pt]{0.400pt}{4.818pt}}
\put(951,23){\makebox(0,0){$800$}}
\put(951.0,857.0){\rule[-0.200pt]{0.400pt}{4.818pt}}
\put(1145.0,68.0){\rule[-0.200pt]{0.400pt}{4.818pt}}
\put(1145,23){\makebox(0,0){$1000$}}
\put(1145.0,857.0){\rule[-0.200pt]{0.400pt}{4.818pt}}
\put(1339.0,68.0){\rule[-0.200pt]{0.400pt}{4.818pt}}
\put(1339,23){\makebox(0,0){$1200$}}
\put(1339.0,857.0){\rule[-0.200pt]{0.400pt}{4.818pt}}
\put(176.0,68.0){\rule[-0.200pt]{303.534pt}{0.400pt}}
\put(1436.0,68.0){\rule[-0.200pt]{0.400pt}{194.888pt}}
\put(176.0,877.0){\rule[-0.200pt]{303.534pt}{0.400pt}}
\put(176.0,68.0){\rule[-0.200pt]{0.400pt}{194.888pt}}
\put(1242,376){\makebox(0,0)[r]{Naive Bayes}}
\put(1264.0,376.0){\rule[-0.200pt]{15.899pt}{0.400pt}}
\put(186,250){\usebox{\plotpoint}}
\multiput(186.58,250.00)(0.498,3.886){73}{\rule{0.120pt}{3.184pt}}
\multiput(185.17,250.00)(38.000,286.391){2}{\rule{0.400pt}{1.592pt}}
\multiput(224.58,543.00)(0.498,1.239){95}{\rule{0.120pt}{1.088pt}}
\multiput(223.17,543.00)(49.000,118.742){2}{\rule{0.400pt}{0.544pt}}
\multiput(273.00,664.58)(1.623,0.499){117}{\rule{1.393pt}{0.120pt}}
\multiput(273.00,663.17)(191.108,60.000){2}{\rule{0.697pt}{0.400pt}}
\multiput(467.00,724.58)(4.930,0.496){37}{\rule{3.980pt}{0.119pt}}
\multiput(467.00,723.17)(185.739,20.000){2}{\rule{1.990pt}{0.400pt}}
\multiput(661.00,744.60)(28.117,0.468){5}{\rule{19.400pt}{0.113pt}}
\multiput(661.00,743.17)(152.734,4.000){2}{\rule{9.700pt}{0.400pt}}
\multiput(854.00,748.59)(12.777,0.488){13}{\rule{9.800pt}{0.117pt}}
\multiput(854.00,747.17)(173.660,8.000){2}{\rule{4.900pt}{0.400pt}}
\multiput(1048.00,754.93)(21.527,-0.477){7}{\rule{15.620pt}{0.115pt}}
\multiput(1048.00,755.17)(161.580,-5.000){2}{\rule{7.810pt}{0.400pt}}
\multiput(1242.00,751.61)(43.105,0.447){3}{\rule{25.967pt}{0.108pt}}
\multiput(1242.00,750.17)(140.105,3.000){2}{\rule{12.983pt}{0.400pt}}
\sbox{\plotpoint}{\rule[-0.500pt]{1.000pt}{1.000pt}}%
\put(1242,331){\makebox(0,0)[r]{C4.5}}
\multiput(1264,331)(20.756,0.000){4}{\usebox{\plotpoint}}
\put(1330,331){\usebox{\plotpoint}}
\put(186,633){\usebox{\plotpoint}}
\multiput(186,633)(14.295,15.048){3}{\usebox{\plotpoint}}
\multiput(224,673)(15.600,13.690){3}{\usebox{\plotpoint}}
\multiput(273,716)(20.733,0.962){10}{\usebox{\plotpoint}}
\multiput(467,725)(20.685,1.706){9}{\usebox{\plotpoint}}
\multiput(661,741)(20.756,0.000){9}{\usebox{\plotpoint}}
\multiput(854,741)(20.751,0.428){10}{\usebox{\plotpoint}}
\multiput(1048,745)(20.753,-0.321){9}{\usebox{\plotpoint}}
\multiput(1242,742)(20.733,0.962){9}{\usebox{\plotpoint}}
\put(1436,751){\usebox{\plotpoint}}
\sbox{\plotpoint}{\rule[-0.400pt]{0.800pt}{0.800pt}}%
\put(1242,286){\makebox(0,0)[r]{Naive Mix}}
\put(1264.0,286.0){\rule[-0.400pt]{15.899pt}{0.800pt}}
\put(186,654){\usebox{\plotpoint}}
\multiput(186.00,655.41)(0.512,0.503){67}{\rule{1.022pt}{0.121pt}}
\multiput(186.00,652.34)(35.880,37.000){2}{\rule{0.511pt}{0.800pt}}
\multiput(224.00,692.41)(1.827,0.509){21}{\rule{3.000pt}{0.123pt}}
\multiput(224.00,689.34)(42.773,14.000){2}{\rule{1.500pt}{0.800pt}}
\multiput(273.00,706.41)(2.994,0.503){59}{\rule{4.903pt}{0.121pt}}
\multiput(273.00,703.34)(183.824,33.000){2}{\rule{2.452pt}{0.800pt}}
\multiput(467.00,739.40)(14.080,0.520){9}{\rule{19.600pt}{0.125pt}}
\multiput(467.00,736.34)(153.319,8.000){2}{\rule{9.800pt}{0.800pt}}
\multiput(661.00,747.40)(14.007,0.520){9}{\rule{19.500pt}{0.125pt}}
\multiput(661.00,744.34)(152.527,8.000){2}{\rule{9.750pt}{0.800pt}}
\put(1242,754.34){\rule{39.000pt}{0.800pt}}
\multiput(1242.00,752.34)(113.054,4.000){2}{\rule{19.500pt}{0.800pt}}
\put(854.0,754.0){\rule[-0.400pt]{93.469pt}{0.800pt}}
\sbox{\plotpoint}{\rule[-0.500pt]{1.000pt}{1.000pt}}%
\put(1242,241){\makebox(0,0)[r]{FSS AIC}}
\multiput(1264,241)(41.511,0.000){2}{\usebox{\plotpoint}}
\put(1330,241){\usebox{\plotpoint}}
\put(186,646){\usebox{\plotpoint}}
\multiput(186,646)(29.353,29.353){2}{\usebox{\plotpoint}}
\put(252.30,691.51){\usebox{\plotpoint}}
\multiput(273,697)(40.888,7.166){5}{\usebox{\plotpoint}}
\multiput(467,731)(41.484,1.497){4}{\usebox{\plotpoint}}
\multiput(661,738)(41.466,1.934){5}{\usebox{\plotpoint}}
\multiput(854,747)(41.510,-0.214){5}{\usebox{\plotpoint}}
\multiput(1048,746)(41.510,0.214){4}{\usebox{\plotpoint}}
\multiput(1242,747)(41.502,0.856){5}{\usebox{\plotpoint}}
\put(1436,751){\usebox{\plotpoint}}
\sbox{\plotpoint}{\rule[-0.200pt]{0.400pt}{0.400pt}}%
\put(1242,196){\makebox(0,0)[r]{Majority}}
\multiput(1264,196)(20.756,0.000){4}{\usebox{\plotpoint}}
\put(1330,196){\usebox{\plotpoint}}
\put(186,673){\usebox{\plotpoint}}
\multiput(186,673)(20.756,0.000){2}{\usebox{\plotpoint}}
\multiput(224,673)(20.756,0.000){3}{\usebox{\plotpoint}}
\multiput(273,673)(20.756,0.000){9}{\usebox{\plotpoint}}
\multiput(467,673)(20.756,0.000){9}{\usebox{\plotpoint}}
\multiput(661,673)(20.756,0.000){10}{\usebox{\plotpoint}}
\multiput(854,673)(20.756,0.000){9}{\usebox{\plotpoint}}
\multiput(1048,673)(20.756,0.000){9}{\usebox{\plotpoint}}
\multiput(1242,673)(20.756,0.000){10}{\usebox{\plotpoint}}
\put(1436,673){\usebox{\plotpoint}}
\end{picture}
\caption{Learning Rate for Verbs}
\label{fig:lrverb}
\end{center}
\myendfig

{\bf Verbs} As is the case with adjectives and nouns, Naive Bayes
begins at a very low level of accuracy while C4.5, FSS AIC, and the
Naive Mix nearly match the accuracy of the majority classifier after
only 10 training examples. All methods exceed the majority classifier
and perform at nearly exactly the same level of accuracy after
learning from approximately 600 examples.    

The main distinction among these approaches is that 
Naive Bayes has a slower learning rate; C4.5, FSS AIC, and the Naive Mix
achieve  at least the accuracy of the majority classifier after just 10
or 50 examples. However, after 300--600 examples all of the methods
perform at roughly the same level of accuracy and no method shows
significant improvement in accuracy when given more training
examples. This suggests that high levels of accuracy in word sense
disambiguation are attainable with relatively small quantities of
training data. 

The fast learning rates of C4.5, FSS AIC, and the Naive Mix are 
largely due to skewed sense distributions, especially for 
adjectives and verbs.  A small number of examples is sufficient to  
correctly determine the majority sense. With only 10 or 50 examples to
learn from, a decision tree or probabilistic model consists of a  few 
features and relies upon knowledge of the majority 
sense to perform disambiguation. Given the large majority
senses that exist, most models are able to attain high levels of
accuracy with very small numbers of examples. However, 
Naive Bayes estimates model parameters that involve all of the
contextual features even when there is only a
very small amount of training data.  In these cases it becomes an 
inaccurate classifier since  it is easily mislead by spurious 
relationships in the data that do not hold true in larger samples. 

\section{Experiment 4: Bias Variance Decomposition} \label{sec:bv}

The  success of Naive Bayes may seem a
bit mysterious. It is a simple approach that does not perform feature
selection nor does it engage in a systematic search for a model. 
It simply assumes a parametric form that is usually not an accurate 
representation of the interactions among contextual features. 
Despite this, it performs as accurately as any other method 
except when the amount of training data is  small. 

The decomposition of classification error, i.e., $(1 - accuracy)$,
into bias and variance components offers an explanation for this
behavior. This experiment shows that different representations of the
same training data attain similar levels of accuracy due to the
differing degrees with which bias and variance contribute to overall
classification error.\footnote{There are two different senses of {\it
bias} used in  this dissertation. One indicates a preference exhibited
by a learning algorithm. The other refers to a component of
classification error. Hopefully, the context will be sufficient to
allow for immediate disambiguation.}

The estimated bias of an algorithm reflects how often the average 
classifications, across multiple training samples, of a learned model fail 
to correspond to the actual classifications in the held--out test
data. Variance estimates the degree to which the classifications predicted by 
the learned model vary across multiple training samples. An algorithm that 
always makes the same classification regardless of the training data will 
have variance of 0.  

Decision tree learners are inherently unstable in that they produce very 
different models across multiple samples of training data, even if there 
are only minor differences in the samples \cite{Breiman96B}. These
models result in different levels of accuracy when
applied to a held--out test set. Such algorithms are said to have low
bias and high variance. 

Naive Bayes is more robust in that it is relatively unaffected by
minor changes in the training data. Naive Bayes is not  particularly
representative of the training data since the parametric form is
assumed rather than learned. Naive Bayes is an example of a high bias
and low variance algorithm. 

This experiment estimates the degree to which bias and variance contribute
to the classification error made by Naive Bayes and the decision tree 
learner  MC4\footnote{The MLC++\cite{KohaviSD96} version of
C4.5.}. Naive Bayes represents  a  high bias and low variance approach
while MC4 represents a  low bias and high variance algorithm.  
The estimates of bias and variance reported here  are made following the 
sampling procedure described in \cite{KohaviW96}:  

\begin{enumerate}
\item Randomly divide the data into two sets, $D$ and $E$. $D$ serves
as a super--set of the training sample while $E$ is the held--out test
set. 
\item Generate $T$ samples of size $m$ from $D$.  Let the size of $D=2m$
so that there are 
\begin{tiny}
 $\left(\begin{array}{c}
2m \\
m
\end{array} \right)$
\end{tiny}
possible training sets. For even a small value of $m$ this will ensure
that there are relatively few duplicate training sets sampled from $D$.
\item Run a learning algorithm on the $T$ training samples. Classify
each observation in the test--set $E$ using each of the $T$ learned
models. Store these results in an array called $R$.  
\item Array $R$ and the correct classifications of $E$ serve as the
input to the procedure described in \cite{KohaviW96} that
estimates bias and variance.  
\end{enumerate}
 
As discussed in \cite{KohaviW96}, the reliability of the bias
and variance estimates depends upon having as large a test set as
possible. Therefore, a training sample size of $m=400$ is employed
since this is generally the lowest number of examples where the
decision tree learner and Naive Bayes perform at comparable levels of
accuracy. The size of the $D$ is set to $800$ and $T=1000$ different
training samples are randomly selected. 

Table \ref{tab:bv2} shows the bias and variance estimates for MC4 and
Naive Bayes using a training sample size of 400. The size of the test
set is also listed. The results in this table are divided into
three groups. In the first group, Naive Bayes has higher bias and
lower variance than MC4. This corresponds to what would be expected; 
a decision tree learner should have lower bias since it learns a
more representative model of the training data than Naive Bayes.   The
second group has very similar bias and variance for Naive
Bayes and MC4. Only {\it drug} falls into this group; for this word
both Naive Bayes and a learned decision tree have approximately the
same representational power. In the third group of results MC4 has higher 
bias and lower variance than Naive Bayes. This is the reverse of what is 
expected and initially appears somewhat counter--intuitive. 

\begin{table}
\begin{center}
\caption{Bias Variance Estimates, m = 400}
\label{tab:bv2}
\begin{tabular}{|l|c|c|c|c|c|c|c|} \hline
\multicolumn{1}{|c|}{word} &
\multicolumn{1}{|c|}{test} &
\multicolumn{3}{|c|}{Naive Bayes} &
\multicolumn{3}{|c|}{MC4} \\ \cline{3-8}
  & size  & bias  & var   & error & bias  &  var & error  \\ \hline
agree & 550 & .063  & .014  & .077  & .048  & .021 & .069   \\
bill  & 540 & .133  & .027  & .160  & .110  & .039 & .149   \\
chief & 240 & .060  & .004  & .064  & .037  & .020 & .057   \\
close & 730 & .137  & .028  & .165  & .110  & .054 & .164   \\ 
common& 310 & .140  & .027  & .167  & .116  & .027 & .143  \\
concern& 790 & .112  & .027  & .139  & .101  & .035 & .136   \\ 
interest&1560 & .212  & .051  & .263  & .189  & .089 & .278  \\ \hline 
drug  & 420  & .192  & .027  & .219  & .192  & .017 & .208   \\ \hline
help  & 590 & .182  & .037  & .219  & .207  & .028 & .235   \\
last   & 2380 & .044  & .013  & .057  & .051  & .006 & .057  \\ 
include& 760 & .055  & .010  & .065  & .086  & .007 & .093  \\ 
public & 70 & .359  & .071  & .430  & .381  & .024 & .405   \\  \hline
\end{tabular} 
\end{center}
\vskip -0.01in
\end{table}

However, the words in this third group have appeared together in a 
previous experiment; they are the same words where no supervised learning 
method significantly improves upon the accuracy of the majority 
classifier: {\it help}, {\it last}, {\it include}, and {\it public}. 
For these words disambiguation accuracy is based almost entirely on
knowledge of the majority sense; the contextual features provide
little information helpful in disambiguation. However, Naive 
Bayes assumes a parametric form that includes all of the contextual
features. This accounts for the higher amounts of variance
since these irrelevant features are included in the model and affect
disambiguation. However, the decision tree learner disregards features
that are not relevant and essentially becomes a majority classifier
where variance is very low and bias is the main source of error. 

The data in Table \ref{tab:bv2} is presented again as 
correlation plots in Figures \ref{fig:correrr},  \ref{fig:corrbias}, and
\ref{fig:corrvar}.  The classification error,  bias, and  variance
for each word are represented by a point in each plot. The $x$
coordinate represents the estimate associated with Naive Bayes and
the $y$ coordinate is associated with MC4. Thus, points on or near $x=y$ 
are associated with measures that have nearly identical estimates for 
Naive Bayes and MC4 for a particular word. 

\begin{figure}
\begin{center}
\setlength{\unitlength}{0.240900pt}
\ifx\plotpoint\undefined\newsavebox{\plotpoint}\fi
\sbox{\plotpoint}{\rule[-0.200pt]{0.400pt}{0.400pt}}%
\begin{picture}(1500,900)(0,0)
\font\gnuplot=cmr10 at 10pt
\gnuplot
\sbox{\plotpoint}{\rule[-0.200pt]{0.400pt}{0.400pt}}%
\put(220.0,113.0){\rule[-0.200pt]{292.934pt}{0.400pt}}
\put(220.0,113.0){\rule[-0.200pt]{0.400pt}{184.048pt}}
\put(220.0,113.0){\rule[-0.200pt]{4.818pt}{0.400pt}}
\put(198,113){\makebox(0,0)[r]{$0$}}
\put(1416.0,113.0){\rule[-0.200pt]{4.818pt}{0.400pt}}
\put(220.0,283.0){\rule[-0.200pt]{4.818pt}{0.400pt}}
\put(198,283){\makebox(0,0)[r]{$0.1$}}
\put(1416.0,283.0){\rule[-0.200pt]{4.818pt}{0.400pt}}
\put(220.0,453.0){\rule[-0.200pt]{4.818pt}{0.400pt}}
\put(198,453){\makebox(0,0)[r]{$0.2$}}
\put(1416.0,453.0){\rule[-0.200pt]{4.818pt}{0.400pt}}
\put(220.0,622.0){\rule[-0.200pt]{4.818pt}{0.400pt}}
\put(198,622){\makebox(0,0)[r]{$0.3$}}
\put(1416.0,622.0){\rule[-0.200pt]{4.818pt}{0.400pt}}
\put(220.0,792.0){\rule[-0.200pt]{4.818pt}{0.400pt}}
\put(198,792){\makebox(0,0)[r]{$0.4$}}
\put(1416.0,792.0){\rule[-0.200pt]{4.818pt}{0.400pt}}
\put(220.0,113.0){\rule[-0.200pt]{0.400pt}{4.818pt}}
\put(220,68){\makebox(0,0){$0$}}
\put(220.0,857.0){\rule[-0.200pt]{0.400pt}{4.818pt}}
\put(490.0,113.0){\rule[-0.200pt]{0.400pt}{4.818pt}}
\put(490,68){\makebox(0,0){$0.1$}}
\put(490.0,857.0){\rule[-0.200pt]{0.400pt}{4.818pt}}
\put(760.0,113.0){\rule[-0.200pt]{0.400pt}{4.818pt}}
\put(760,68){\makebox(0,0){$0.2$}}
\put(760.0,857.0){\rule[-0.200pt]{0.400pt}{4.818pt}}
\put(1031.0,113.0){\rule[-0.200pt]{0.400pt}{4.818pt}}
\put(1031,68){\makebox(0,0){$0.3$}}
\put(1031.0,857.0){\rule[-0.200pt]{0.400pt}{4.818pt}}
\put(1301.0,113.0){\rule[-0.200pt]{0.400pt}{4.818pt}}
\put(1301,68){\makebox(0,0){$0.4$}}
\put(1301.0,857.0){\rule[-0.200pt]{0.400pt}{4.818pt}}
\put(220.0,113.0){\rule[-0.200pt]{292.934pt}{0.400pt}}
\put(1436.0,113.0){\rule[-0.200pt]{0.400pt}{184.048pt}}
\put(220.0,877.0){\rule[-0.200pt]{292.934pt}{0.400pt}}
\put(45,495){\makebox(0,0){MC4}}
\put(828,23){\makebox(0,0){Naive Bayes}}
\put(220.0,113.0){\rule[-0.200pt]{0.400pt}{184.048pt}}
\put(1166,325){\makebox(0,0)[r]{noun}}
\put(1210,325){\raisebox{-.8pt}{\makebox(0,0){$\Diamond$}}}
\put(652,366){\raisebox{-.8pt}{\makebox(0,0){$\Diamond$}}}
\put(596,344){\raisebox{-.8pt}{\makebox(0,0){$\Diamond$}}}
\put(812,466){\raisebox{-.8pt}{\makebox(0,0){$\Diamond$}}}
\put(931,585){\raisebox{-.8pt}{\makebox(0,0){$\Diamond$}}}
\sbox{\plotpoint}{\rule[-0.400pt]{0.800pt}{0.800pt}}%
\put(1166,280){\makebox(0,0)[r]{adjective}}
\put(1210,280){\makebox(0,0){$+$}}
\put(393,210){\makebox(0,0){$+$}}
\put(671,356){\makebox(0,0){$+$}}
\put(339,200){\makebox(0,0){$+$}}
\put(1382,801){\makebox(0,0){$+$}}
\sbox{\plotpoint}{\rule[-0.500pt]{1.000pt}{1.000pt}}%
\put(1166,235){\makebox(0,0)[r]{verb}}
\put(1210,235){\raisebox{-.8pt}{\makebox(0,0){$\Box$}}}
\put(428,230){\raisebox{-.8pt}{\makebox(0,0){$\Box$}}}
\put(666,391){\raisebox{-.8pt}{\makebox(0,0){$\Box$}}}
\put(812,512){\raisebox{-.8pt}{\makebox(0,0){$\Box$}}}
\put(396,271){\raisebox{-.8pt}{\makebox(0,0){$\Box$}}}
\put(220,113){\usebox{\plotpoint}}
\put(220.00,113.00){\usebox{\plotpoint}}
\put(237.58,124.00){\usebox{\plotpoint}}
\put(255.25,134.84){\usebox{\plotpoint}}
\multiput(257,136)(17.270,11.513){0}{\usebox{\plotpoint}}
\put(272.52,146.35){\usebox{\plotpoint}}
\put(290.31,157.01){\usebox{\plotpoint}}
\multiput(294,159)(17.270,11.513){0}{\usebox{\plotpoint}}
\put(307.78,168.19){\usebox{\plotpoint}}
\put(325.46,179.02){\usebox{\plotpoint}}
\multiput(331,182)(17.270,11.513){0}{\usebox{\plotpoint}}
\put(343.03,190.02){\usebox{\plotpoint}}
\put(360.30,201.53){\usebox{\plotpoint}}
\put(378.19,212.02){\usebox{\plotpoint}}
\multiput(380,213)(17.270,11.513){0}{\usebox{\plotpoint}}
\put(395.56,223.37){\usebox{\plotpoint}}
\put(413.34,234.03){\usebox{\plotpoint}}
\multiput(417,236)(17.270,11.513){0}{\usebox{\plotpoint}}
\put(430.81,245.21){\usebox{\plotpoint}}
\put(448.08,256.72){\usebox{\plotpoint}}
\multiput(453,260)(18.275,9.840){0}{\usebox{\plotpoint}}
\put(466.06,267.04){\usebox{\plotpoint}}
\put(483.33,278.56){\usebox{\plotpoint}}
\put(501.22,289.04){\usebox{\plotpoint}}
\multiput(503,290)(17.270,11.513){0}{\usebox{\plotpoint}}
\put(518.59,300.39){\usebox{\plotpoint}}
\put(535.86,311.91){\usebox{\plotpoint}}
\multiput(539,314)(18.275,9.840){0}{\usebox{\plotpoint}}
\put(553.84,322.23){\usebox{\plotpoint}}
\put(571.11,333.74){\usebox{\plotpoint}}
\multiput(576,337)(17.270,11.513){0}{\usebox{\plotpoint}}
\put(588.40,345.22){\usebox{\plotpoint}}
\put(606.37,355.58){\usebox{\plotpoint}}
\put(623.64,367.09){\usebox{\plotpoint}}
\multiput(625,368)(18.275,9.840){0}{\usebox{\plotpoint}}
\put(641.62,377.41){\usebox{\plotpoint}}
\put(658.89,388.93){\usebox{\plotpoint}}
\multiput(662,391)(17.270,11.513){0}{\usebox{\plotpoint}}
\put(676.28,400.23){\usebox{\plotpoint}}
\put(694.14,410.76){\usebox{\plotpoint}}
\multiput(699,414)(17.270,11.513){0}{\usebox{\plotpoint}}
\put(711.44,422.24){\usebox{\plotpoint}}
\put(729.40,432.60){\usebox{\plotpoint}}
\put(746.67,444.11){\usebox{\plotpoint}}
\multiput(748,445)(17.270,11.513){0}{\usebox{\plotpoint}}
\put(764.17,455.24){\usebox{\plotpoint}}
\put(781.92,465.95){\usebox{\plotpoint}}
\multiput(785,468)(17.270,11.513){0}{\usebox{\plotpoint}}
\put(799.32,477.25){\usebox{\plotpoint}}
\put(817.18,487.78){\usebox{\plotpoint}}
\multiput(822,491)(17.270,11.513){0}{\usebox{\plotpoint}}
\put(834.45,499.30){\usebox{\plotpoint}}
\put(852.05,510.26){\usebox{\plotpoint}}
\put(869.70,521.13){\usebox{\plotpoint}}
\multiput(871,522)(17.270,11.513){0}{\usebox{\plotpoint}}
\put(887.20,532.26){\usebox{\plotpoint}}
\put(904.95,542.97){\usebox{\plotpoint}}
\multiput(908,545)(17.270,11.513){0}{\usebox{\plotpoint}}
\put(922.22,554.48){\usebox{\plotpoint}}
\put(939.93,565.27){\usebox{\plotpoint}}
\multiput(945,568)(17.270,11.513){0}{\usebox{\plotpoint}}
\put(957.48,576.32){\usebox{\plotpoint}}
\put(975.08,587.27){\usebox{\plotpoint}}
\put(992.73,598.15){\usebox{\plotpoint}}
\multiput(994,599)(17.270,11.513){0}{\usebox{\plotpoint}}
\put(1010.00,609.67){\usebox{\plotpoint}}
\put(1027.81,620.28){\usebox{\plotpoint}}
\multiput(1031,622)(17.270,11.513){0}{\usebox{\plotpoint}}
\put(1045.26,631.50){\usebox{\plotpoint}}
\put(1062.96,642.29){\usebox{\plotpoint}}
\multiput(1068,645)(17.270,11.513){0}{\usebox{\plotpoint}}
\put(1080.51,653.34){\usebox{\plotpoint}}
\put(1097.78,664.85){\usebox{\plotpoint}}
\put(1115.69,675.30){\usebox{\plotpoint}}
\multiput(1117,676)(17.270,11.513){0}{\usebox{\plotpoint}}
\put(1133.03,686.69){\usebox{\plotpoint}}
\put(1150.30,698.20){\usebox{\plotpoint}}
\multiput(1153,700)(18.275,9.840){0}{\usebox{\plotpoint}}
\put(1168.29,708.53){\usebox{\plotpoint}}
\put(1185.56,720.04){\usebox{\plotpoint}}
\multiput(1190,723)(18.275,9.840){0}{\usebox{\plotpoint}}
\put(1203.54,730.36){\usebox{\plotpoint}}
\put(1220.81,741.87){\usebox{\plotpoint}}
\put(1238.08,753.39){\usebox{\plotpoint}}
\multiput(1239,754)(18.275,9.840){0}{\usebox{\plotpoint}}
\put(1256.07,763.71){\usebox{\plotpoint}}
\put(1273.34,775.22){\usebox{\plotpoint}}
\multiput(1276,777)(18.275,9.840){0}{\usebox{\plotpoint}}
\put(1291.32,785.55){\usebox{\plotpoint}}
\put(1308.59,797.06){\usebox{\plotpoint}}
\multiput(1313,800)(17.270,11.513){0}{\usebox{\plotpoint}}
\put(1325.91,808.49){\usebox{\plotpoint}}
\put(1343.84,818.90){\usebox{\plotpoint}}
\put(1361.11,830.41){\usebox{\plotpoint}}
\multiput(1362,831)(18.275,9.840){0}{\usebox{\plotpoint}}
\put(1379.10,840.73){\usebox{\plotpoint}}
\put(1396.37,852.24){\usebox{\plotpoint}}
\multiput(1399,854)(17.270,11.513){0}{\usebox{\plotpoint}}
\put(1413.79,863.50){\usebox{\plotpoint}}
\put(1431.62,874.08){\usebox{\plotpoint}}
\put(1436,877){\usebox{\plotpoint}}
\end{picture}
\caption{Classification Error Correlation, m=400}
\label{fig:correrr}
\end{center}
\vskip -0.14in
\end{figure}

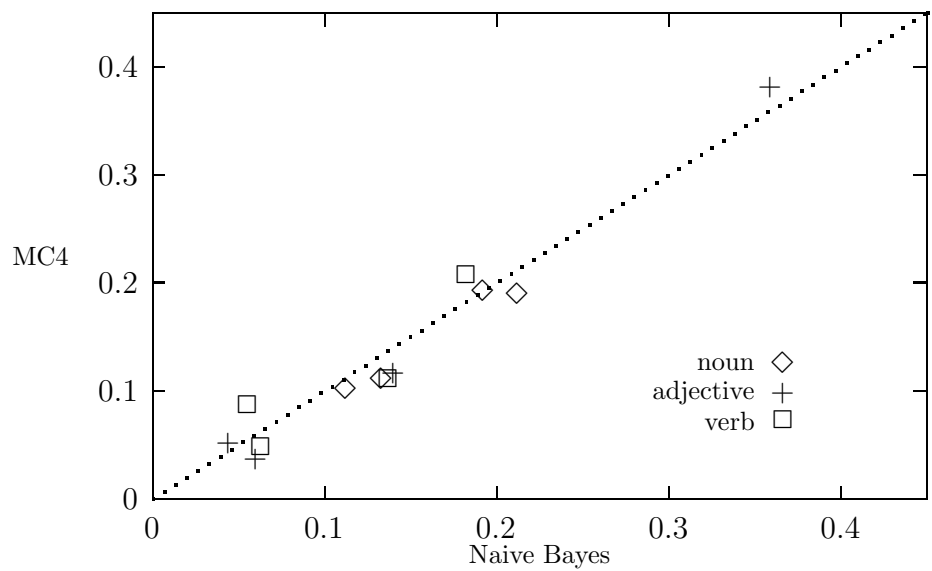
\begin{figure}
\begin{center}
\setlength{\unitlength}{0.240900pt}
\ifx\plotpoint\undefined\newsavebox{\plotpoint}\fi
\sbox{\plotpoint}{\rule[-0.200pt]{0.400pt}{0.400pt}}%
\begin{picture}(1500,900)(0,0)
\font\gnuplot=cmr10 at 10pt
\gnuplot
\sbox{\plotpoint}{\rule[-0.200pt]{0.400pt}{0.400pt}}%
\put(220.0,113.0){\rule[-0.200pt]{292.934pt}{0.400pt}}
\put(220.0,113.0){\rule[-0.200pt]{0.400pt}{184.048pt}}
\put(220.0,113.0){\rule[-0.200pt]{4.818pt}{0.400pt}}
\put(198,113){\makebox(0,0)[r]{$0$}}
\put(1416.0,113.0){\rule[-0.200pt]{4.818pt}{0.400pt}}
\put(220.0,283.0){\rule[-0.200pt]{4.818pt}{0.400pt}}
\put(198,283){\makebox(0,0)[r]{$0.1$}}
\put(1416.0,283.0){\rule[-0.200pt]{4.818pt}{0.400pt}}
\put(220.0,453.0){\rule[-0.200pt]{4.818pt}{0.400pt}}
\put(198,453){\makebox(0,0)[r]{$0.2$}}
\put(1416.0,453.0){\rule[-0.200pt]{4.818pt}{0.400pt}}
\put(220.0,622.0){\rule[-0.200pt]{4.818pt}{0.400pt}}
\put(198,622){\makebox(0,0)[r]{$0.3$}}
\put(1416.0,622.0){\rule[-0.200pt]{4.818pt}{0.400pt}}
\put(220.0,792.0){\rule[-0.200pt]{4.818pt}{0.400pt}}
\put(198,792){\makebox(0,0)[r]{$0.4$}}
\put(1416.0,792.0){\rule[-0.200pt]{4.818pt}{0.400pt}}
\put(220.0,113.0){\rule[-0.200pt]{0.400pt}{4.818pt}}
\put(220,68){\makebox(0,0){$0$}}
\put(220.0,857.0){\rule[-0.200pt]{0.400pt}{4.818pt}}
\put(490.0,113.0){\rule[-0.200pt]{0.400pt}{4.818pt}}
\put(490,68){\makebox(0,0){$0.1$}}
\put(490.0,857.0){\rule[-0.200pt]{0.400pt}{4.818pt}}
\put(760.0,113.0){\rule[-0.200pt]{0.400pt}{4.818pt}}
\put(760,68){\makebox(0,0){$0.2$}}
\put(760.0,857.0){\rule[-0.200pt]{0.400pt}{4.818pt}}
\put(1031.0,113.0){\rule[-0.200pt]{0.400pt}{4.818pt}}
\put(1031,68){\makebox(0,0){$0.3$}}
\put(1031.0,857.0){\rule[-0.200pt]{0.400pt}{4.818pt}}
\put(1301.0,113.0){\rule[-0.200pt]{0.400pt}{4.818pt}}
\put(1301,68){\makebox(0,0){$0.4$}}
\put(1301.0,857.0){\rule[-0.200pt]{0.400pt}{4.818pt}}
\put(220.0,113.0){\rule[-0.200pt]{292.934pt}{0.400pt}}
\put(1436.0,113.0){\rule[-0.200pt]{0.400pt}{184.048pt}}
\put(220.0,877.0){\rule[-0.200pt]{292.934pt}{0.400pt}}
\put(45,495){\makebox(0,0){MC4}}
\put(828,23){\makebox(0,0){Naive Bayes}}
\put(220.0,113.0){\rule[-0.200pt]{0.400pt}{184.048pt}}
\put(1166,325){\makebox(0,0)[r]{noun}}
\put(1210,325){\raisebox{-.8pt}{\makebox(0,0){$\Diamond$}}}
\put(579,300){\raisebox{-.8pt}{\makebox(0,0){$\Diamond$}}}
\put(523,284){\raisebox{-.8pt}{\makebox(0,0){$\Diamond$}}}
\put(739,439){\raisebox{-.8pt}{\makebox(0,0){$\Diamond$}}}
\put(793,434){\raisebox{-.8pt}{\makebox(0,0){$\Diamond$}}}
\sbox{\plotpoint}{\rule[-0.400pt]{0.800pt}{0.800pt}}%
\put(1166,280){\makebox(0,0)[r]{adjective}}
\put(1210,280){\makebox(0,0){$+$}}
\put(382,176){\makebox(0,0){$+$}}
\put(598,310){\makebox(0,0){$+$}}
\put(339,200){\makebox(0,0){$+$}}
\put(1190,760){\makebox(0,0){$+$}}
\sbox{\plotpoint}{\rule[-0.500pt]{1.000pt}{1.000pt}}%
\put(1166,235){\makebox(0,0)[r]{verb}}
\put(1210,235){\raisebox{-.8pt}{\makebox(0,0){$\Box$}}}
\put(390,194){\raisebox{-.8pt}{\makebox(0,0){$\Box$}}}
\put(590,300){\raisebox{-.8pt}{\makebox(0,0){$\Box$}}}
\put(712,464){\raisebox{-.8pt}{\makebox(0,0){$\Box$}}}
\put(369,259){\raisebox{-.8pt}{\makebox(0,0){$\Box$}}}
\put(220,113){\usebox{\plotpoint}}
\put(220.00,113.00){\usebox{\plotpoint}}
\put(237.58,124.00){\usebox{\plotpoint}}
\put(255.25,134.84){\usebox{\plotpoint}}
\multiput(257,136)(17.270,11.513){0}{\usebox{\plotpoint}}
\put(272.52,146.35){\usebox{\plotpoint}}
\put(290.31,157.01){\usebox{\plotpoint}}
\multiput(294,159)(17.270,11.513){0}{\usebox{\plotpoint}}
\put(307.78,168.19){\usebox{\plotpoint}}
\put(325.46,179.02){\usebox{\plotpoint}}
\multiput(331,182)(17.270,11.513){0}{\usebox{\plotpoint}}
\put(343.03,190.02){\usebox{\plotpoint}}
\put(360.30,201.53){\usebox{\plotpoint}}
\put(378.19,212.02){\usebox{\plotpoint}}
\multiput(380,213)(17.270,11.513){0}{\usebox{\plotpoint}}
\put(395.56,223.37){\usebox{\plotpoint}}
\put(413.34,234.03){\usebox{\plotpoint}}
\multiput(417,236)(17.270,11.513){0}{\usebox{\plotpoint}}
\put(430.81,245.21){\usebox{\plotpoint}}
\put(448.08,256.72){\usebox{\plotpoint}}
\multiput(453,260)(18.275,9.840){0}{\usebox{\plotpoint}}
\put(466.06,267.04){\usebox{\plotpoint}}
\put(483.33,278.56){\usebox{\plotpoint}}
\put(501.22,289.04){\usebox{\plotpoint}}
\multiput(503,290)(17.270,11.513){0}{\usebox{\plotpoint}}
\put(518.59,300.39){\usebox{\plotpoint}}
\put(535.86,311.91){\usebox{\plotpoint}}
\multiput(539,314)(18.275,9.840){0}{\usebox{\plotpoint}}
\put(553.84,322.23){\usebox{\plotpoint}}
\put(571.11,333.74){\usebox{\plotpoint}}
\multiput(576,337)(17.270,11.513){0}{\usebox{\plotpoint}}
\put(588.40,345.22){\usebox{\plotpoint}}
\put(606.37,355.58){\usebox{\plotpoint}}
\put(623.64,367.09){\usebox{\plotpoint}}
\multiput(625,368)(18.275,9.840){0}{\usebox{\plotpoint}}
\put(641.62,377.41){\usebox{\plotpoint}}
\put(658.89,388.93){\usebox{\plotpoint}}
\multiput(662,391)(17.270,11.513){0}{\usebox{\plotpoint}}
\put(676.28,400.23){\usebox{\plotpoint}}
\put(694.14,410.76){\usebox{\plotpoint}}
\multiput(699,414)(17.270,11.513){0}{\usebox{\plotpoint}}
\put(711.44,422.24){\usebox{\plotpoint}}
\put(729.40,432.60){\usebox{\plotpoint}}
\put(746.67,444.11){\usebox{\plotpoint}}
\multiput(748,445)(17.270,11.513){0}{\usebox{\plotpoint}}
\put(764.17,455.24){\usebox{\plotpoint}}
\put(781.92,465.95){\usebox{\plotpoint}}
\multiput(785,468)(17.270,11.513){0}{\usebox{\plotpoint}}
\put(799.32,477.25){\usebox{\plotpoint}}
\put(817.18,487.78){\usebox{\plotpoint}}
\multiput(822,491)(17.270,11.513){0}{\usebox{\plotpoint}}
\put(834.45,499.30){\usebox{\plotpoint}}
\put(852.05,510.26){\usebox{\plotpoint}}
\put(869.70,521.13){\usebox{\plotpoint}}
\multiput(871,522)(17.270,11.513){0}{\usebox{\plotpoint}}
\put(887.20,532.26){\usebox{\plotpoint}}
\put(904.95,542.97){\usebox{\plotpoint}}
\multiput(908,545)(17.270,11.513){0}{\usebox{\plotpoint}}
\put(922.22,554.48){\usebox{\plotpoint}}
\put(939.93,565.27){\usebox{\plotpoint}}
\multiput(945,568)(17.270,11.513){0}{\usebox{\plotpoint}}
\put(957.48,576.32){\usebox{\plotpoint}}
\put(975.08,587.27){\usebox{\plotpoint}}
\put(992.73,598.15){\usebox{\plotpoint}}
\multiput(994,599)(17.270,11.513){0}{\usebox{\plotpoint}}
\put(1010.00,609.67){\usebox{\plotpoint}}
\put(1027.81,620.28){\usebox{\plotpoint}}
\multiput(1031,622)(17.270,11.513){0}{\usebox{\plotpoint}}
\put(1045.26,631.50){\usebox{\plotpoint}}
\put(1062.96,642.29){\usebox{\plotpoint}}
\multiput(1068,645)(17.270,11.513){0}{\usebox{\plotpoint}}
\put(1080.51,653.34){\usebox{\plotpoint}}
\put(1097.78,664.85){\usebox{\plotpoint}}
\put(1115.69,675.30){\usebox{\plotpoint}}
\multiput(1117,676)(17.270,11.513){0}{\usebox{\plotpoint}}
\put(1133.03,686.69){\usebox{\plotpoint}}
\put(1150.30,698.20){\usebox{\plotpoint}}
\multiput(1153,700)(18.275,9.840){0}{\usebox{\plotpoint}}
\put(1168.29,708.53){\usebox{\plotpoint}}
\put(1185.56,720.04){\usebox{\plotpoint}}
\multiput(1190,723)(18.275,9.840){0}{\usebox{\plotpoint}}
\put(1203.54,730.36){\usebox{\plotpoint}}
\put(1220.81,741.87){\usebox{\plotpoint}}
\put(1238.08,753.39){\usebox{\plotpoint}}
\multiput(1239,754)(18.275,9.840){0}{\usebox{\plotpoint}}
\put(1256.07,763.71){\usebox{\plotpoint}}
\put(1273.34,775.22){\usebox{\plotpoint}}
\multiput(1276,777)(18.275,9.840){0}{\usebox{\plotpoint}}
\put(1291.32,785.55){\usebox{\plotpoint}}
\put(1308.59,797.06){\usebox{\plotpoint}}
\multiput(1313,800)(17.270,11.513){0}{\usebox{\plotpoint}}
\put(1325.91,808.49){\usebox{\plotpoint}}
\put(1343.84,818.90){\usebox{\plotpoint}}
\put(1361.11,830.41){\usebox{\plotpoint}}
\multiput(1362,831)(18.275,9.840){0}{\usebox{\plotpoint}}
\put(1379.10,840.73){\usebox{\plotpoint}}
\put(1396.37,852.24){\usebox{\plotpoint}}
\multiput(1399,854)(17.270,11.513){0}{\usebox{\plotpoint}}
\put(1413.79,863.50){\usebox{\plotpoint}}
\put(1431.62,874.08){\usebox{\plotpoint}}
\put(1436,877){\usebox{\plotpoint}}
\end{picture}
\caption{Bias Correlation,  m=400}
\label{fig:corrbias}
\end{center}
\vskip -0.14in
\end{figure}

\begin{figure}
\begin{center}
\setlength{\unitlength}{0.240900pt}
\ifx\plotpoint\undefined\newsavebox{\plotpoint}\fi
\sbox{\plotpoint}{\rule[-0.200pt]{0.400pt}{0.400pt}}%
\begin{picture}(1500,900)(0,0)
\font\gnuplot=cmr10 at 10pt
\gnuplot
\sbox{\plotpoint}{\rule[-0.200pt]{0.400pt}{0.400pt}}%
\put(220.0,113.0){\rule[-0.200pt]{292.934pt}{0.400pt}}
\put(220.0,113.0){\rule[-0.200pt]{0.400pt}{184.048pt}}
\put(220.0,113.0){\rule[-0.200pt]{4.818pt}{0.400pt}}
\put(198,113){\makebox(0,0)[r]{$0$}}
\put(1416.0,113.0){\rule[-0.200pt]{4.818pt}{0.400pt}}
\put(220.0,259.0){\rule[-0.200pt]{4.818pt}{0.400pt}}
\put(198,259){\makebox(0,0)[r]{$0.02$}}
\put(1416.0,259.0){\rule[-0.200pt]{4.818pt}{0.400pt}}
\put(220.0,404.0){\rule[-0.200pt]{4.818pt}{0.400pt}}
\put(198,404){\makebox(0,0)[r]{$0.04$}}
\put(1416.0,404.0){\rule[-0.200pt]{4.818pt}{0.400pt}}
\put(220.0,550.0){\rule[-0.200pt]{4.818pt}{0.400pt}}
\put(198,550){\makebox(0,0)[r]{$0.06$}}
\put(1416.0,550.0){\rule[-0.200pt]{4.818pt}{0.400pt}}
\put(220.0,695.0){\rule[-0.200pt]{4.818pt}{0.400pt}}
\put(198,695){\makebox(0,0)[r]{$0.08$}}
\put(1416.0,695.0){\rule[-0.200pt]{4.818pt}{0.400pt}}
\put(220.0,841.0){\rule[-0.200pt]{4.818pt}{0.400pt}}
\put(198,841){\makebox(0,0)[r]{$0.1$}}
\put(1416.0,841.0){\rule[-0.200pt]{4.818pt}{0.400pt}}
\put(220.0,113.0){\rule[-0.200pt]{0.400pt}{4.818pt}}
\put(220,68){\makebox(0,0){$0$}}
\put(220.0,857.0){\rule[-0.200pt]{0.400pt}{4.818pt}}
\put(452.0,113.0){\rule[-0.200pt]{0.400pt}{4.818pt}}
\put(452,68){\makebox(0,0){$0.02$}}
\put(452.0,857.0){\rule[-0.200pt]{0.400pt}{4.818pt}}
\put(683.0,113.0){\rule[-0.200pt]{0.400pt}{4.818pt}}
\put(683,68){\makebox(0,0){$0.04$}}
\put(683.0,857.0){\rule[-0.200pt]{0.400pt}{4.818pt}}
\put(915.0,113.0){\rule[-0.200pt]{0.400pt}{4.818pt}}
\put(915,68){\makebox(0,0){$0.06$}}
\put(915.0,857.0){\rule[-0.200pt]{0.400pt}{4.818pt}}
\put(1146.0,113.0){\rule[-0.200pt]{0.400pt}{4.818pt}}
\put(1146,68){\makebox(0,0){$0.08$}}
\put(1146.0,857.0){\rule[-0.200pt]{0.400pt}{4.818pt}}
\put(1378.0,113.0){\rule[-0.200pt]{0.400pt}{4.818pt}}
\put(1378,68){\makebox(0,0){$0.1$}}
\put(1378.0,857.0){\rule[-0.200pt]{0.400pt}{4.818pt}}
\put(220.0,113.0){\rule[-0.200pt]{292.934pt}{0.400pt}}
\put(1436.0,113.0){\rule[-0.200pt]{0.400pt}{184.048pt}}
\put(220.0,877.0){\rule[-0.200pt]{292.934pt}{0.400pt}}
\put(45,495){\makebox(0,0){MC4}}
\put(828,23){\makebox(0,0){Naive Bayes}}
\put(220.0,113.0){\rule[-0.200pt]{0.400pt}{184.048pt}}
\put(452,768){\makebox(0,0)[r]{noun}}
\put(496,768){\raisebox{-.8pt}{\makebox(0,0){$\Diamond$}}}
\put(533,397){\raisebox{-.8pt}{\makebox(0,0){$\Diamond$}}}
\put(533,368){\raisebox{-.8pt}{\makebox(0,0){$\Diamond$}}}
\put(533,237){\raisebox{-.8pt}{\makebox(0,0){$\Diamond$}}}
\put(811,761){\raisebox{-.8pt}{\makebox(0,0){$\Diamond$}}}
\sbox{\plotpoint}{\rule[-0.400pt]{0.800pt}{0.800pt}}%
\put(452,723){\makebox(0,0)[r]{adjective}}
\put(496,723){\makebox(0,0){$+$}}
\put(266,259){\makebox(0,0){$+$}}
\put(533,309){\makebox(0,0){$+$}}
\put(371,157){\makebox(0,0){$+$}}
\put(1042,288){\makebox(0,0){$+$}}
\sbox{\plotpoint}{\rule[-0.500pt]{1.000pt}{1.000pt}}%
\put(452,678){\makebox(0,0)[r]{verb}}
\put(496,678){\raisebox{-.8pt}{\makebox(0,0){$\Box$}}}
\put(382,266){\raisebox{-.8pt}{\makebox(0,0){$\Box$}}}
\put(544,506){\raisebox{-.8pt}{\makebox(0,0){$\Box$}}}
\put(648,317){\raisebox{-.8pt}{\makebox(0,0){$\Box$}}}
\put(336,164){\raisebox{-.8pt}{\makebox(0,0){$\Box$}}}
\put(220,113){\usebox{\plotpoint}}
\put(220.00,113.00){\usebox{\plotpoint}}
\put(237.58,124.00){\usebox{\plotpoint}}
\put(255.25,134.84){\usebox{\plotpoint}}
\multiput(257,136)(17.270,11.513){0}{\usebox{\plotpoint}}
\put(272.52,146.35){\usebox{\plotpoint}}
\put(290.31,157.01){\usebox{\plotpoint}}
\multiput(294,159)(17.270,11.513){0}{\usebox{\plotpoint}}
\put(307.78,168.19){\usebox{\plotpoint}}
\put(325.46,179.02){\usebox{\plotpoint}}
\multiput(331,182)(17.270,11.513){0}{\usebox{\plotpoint}}
\put(343.03,190.02){\usebox{\plotpoint}}
\put(360.30,201.53){\usebox{\plotpoint}}
\put(378.19,212.02){\usebox{\plotpoint}}
\multiput(380,213)(17.270,11.513){0}{\usebox{\plotpoint}}
\put(395.56,223.37){\usebox{\plotpoint}}
\put(413.34,234.03){\usebox{\plotpoint}}
\multiput(417,236)(17.270,11.513){0}{\usebox{\plotpoint}}
\put(430.81,245.21){\usebox{\plotpoint}}
\put(448.08,256.72){\usebox{\plotpoint}}
\multiput(453,260)(18.275,9.840){0}{\usebox{\plotpoint}}
\put(466.06,267.04){\usebox{\plotpoint}}
\put(483.33,278.56){\usebox{\plotpoint}}
\put(501.22,289.04){\usebox{\plotpoint}}
\multiput(503,290)(17.270,11.513){0}{\usebox{\plotpoint}}
\put(518.59,300.39){\usebox{\plotpoint}}
\put(535.86,311.91){\usebox{\plotpoint}}
\multiput(539,314)(18.275,9.840){0}{\usebox{\plotpoint}}
\put(553.84,322.23){\usebox{\plotpoint}}
\put(571.11,333.74){\usebox{\plotpoint}}
\multiput(576,337)(17.270,11.513){0}{\usebox{\plotpoint}}
\put(588.40,345.22){\usebox{\plotpoint}}
\put(606.37,355.58){\usebox{\plotpoint}}
\put(623.64,367.09){\usebox{\plotpoint}}
\multiput(625,368)(18.275,9.840){0}{\usebox{\plotpoint}}
\put(641.62,377.41){\usebox{\plotpoint}}
\put(658.89,388.93){\usebox{\plotpoint}}
\multiput(662,391)(17.270,11.513){0}{\usebox{\plotpoint}}
\put(676.28,400.23){\usebox{\plotpoint}}
\put(694.14,410.76){\usebox{\plotpoint}}
\multiput(699,414)(17.270,11.513){0}{\usebox{\plotpoint}}
\put(711.44,422.24){\usebox{\plotpoint}}
\put(729.40,432.60){\usebox{\plotpoint}}
\put(746.67,444.11){\usebox{\plotpoint}}
\multiput(748,445)(17.270,11.513){0}{\usebox{\plotpoint}}
\put(764.17,455.24){\usebox{\plotpoint}}
\put(781.92,465.95){\usebox{\plotpoint}}
\multiput(785,468)(17.270,11.513){0}{\usebox{\plotpoint}}
\put(799.32,477.25){\usebox{\plotpoint}}
\put(817.18,487.78){\usebox{\plotpoint}}
\multiput(822,491)(17.270,11.513){0}{\usebox{\plotpoint}}
\put(834.45,499.30){\usebox{\plotpoint}}
\put(852.05,510.26){\usebox{\plotpoint}}
\put(869.70,521.13){\usebox{\plotpoint}}
\multiput(871,522)(17.270,11.513){0}{\usebox{\plotpoint}}
\put(887.20,532.26){\usebox{\plotpoint}}
\put(904.95,542.97){\usebox{\plotpoint}}
\multiput(908,545)(17.270,11.513){0}{\usebox{\plotpoint}}
\put(922.22,554.48){\usebox{\plotpoint}}
\put(939.93,565.27){\usebox{\plotpoint}}
\multiput(945,568)(17.270,11.513){0}{\usebox{\plotpoint}}
\put(957.48,576.32){\usebox{\plotpoint}}
\put(975.08,587.27){\usebox{\plotpoint}}
\put(992.73,598.15){\usebox{\plotpoint}}
\multiput(994,599)(17.270,11.513){0}{\usebox{\plotpoint}}
\put(1010.00,609.67){\usebox{\plotpoint}}
\put(1027.81,620.28){\usebox{\plotpoint}}
\multiput(1031,622)(17.270,11.513){0}{\usebox{\plotpoint}}
\put(1045.26,631.50){\usebox{\plotpoint}}
\put(1062.96,642.29){\usebox{\plotpoint}}
\multiput(1068,645)(17.270,11.513){0}{\usebox{\plotpoint}}
\put(1080.51,653.34){\usebox{\plotpoint}}
\put(1097.78,664.85){\usebox{\plotpoint}}
\put(1115.69,675.30){\usebox{\plotpoint}}
\multiput(1117,676)(17.270,11.513){0}{\usebox{\plotpoint}}
\put(1133.03,686.69){\usebox{\plotpoint}}
\put(1150.30,698.20){\usebox{\plotpoint}}
\multiput(1153,700)(18.275,9.840){0}{\usebox{\plotpoint}}
\put(1168.29,708.53){\usebox{\plotpoint}}
\put(1185.56,720.04){\usebox{\plotpoint}}
\multiput(1190,723)(18.275,9.840){0}{\usebox{\plotpoint}}
\put(1203.54,730.36){\usebox{\plotpoint}}
\put(1220.81,741.87){\usebox{\plotpoint}}
\put(1238.08,753.39){\usebox{\plotpoint}}
\multiput(1239,754)(18.275,9.840){0}{\usebox{\plotpoint}}
\put(1256.07,763.71){\usebox{\plotpoint}}
\put(1273.34,775.22){\usebox{\plotpoint}}
\multiput(1276,777)(18.275,9.840){0}{\usebox{\plotpoint}}
\put(1291.32,785.55){\usebox{\plotpoint}}
\put(1308.59,797.06){\usebox{\plotpoint}}
\multiput(1313,800)(17.270,11.513){0}{\usebox{\plotpoint}}
\put(1325.91,808.49){\usebox{\plotpoint}}
\put(1343.84,818.90){\usebox{\plotpoint}}
\put(1361.11,830.41){\usebox{\plotpoint}}
\multiput(1362,831)(18.275,9.840){0}{\usebox{\plotpoint}}
\put(1379.10,840.73){\usebox{\plotpoint}}
\put(1396.37,852.24){\usebox{\plotpoint}}
\multiput(1399,854)(17.270,11.513){0}{\usebox{\plotpoint}}
\put(1413.79,863.50){\usebox{\plotpoint}}
\put(1431.62,874.08){\usebox{\plotpoint}}
\put(1436,877){\usebox{\plotpoint}}
\end{picture}
\caption{Variance Correlation, m=400}
\label{fig:corrvar}
\end{center}
\vskip -0.14in
\end{figure}
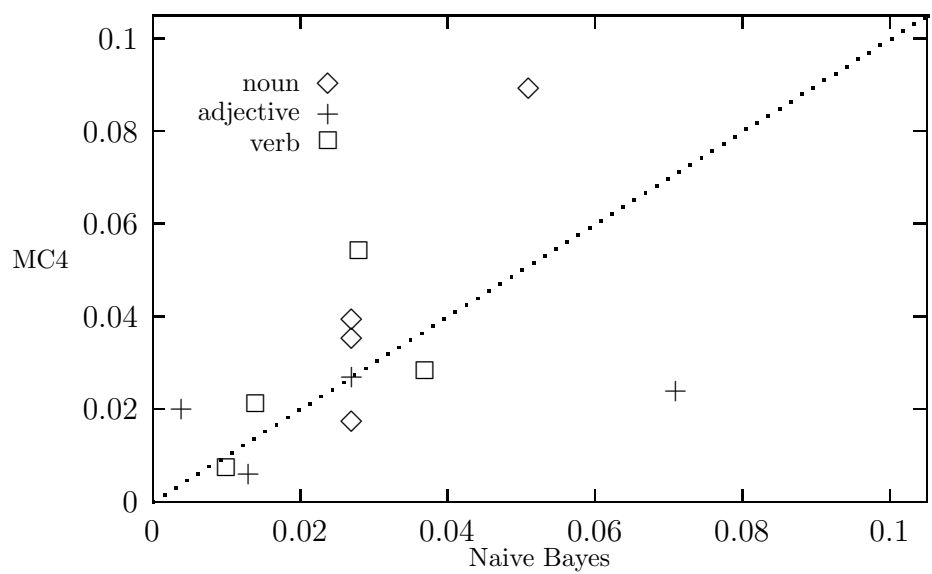

Figure \ref{fig:correrr} confirms that the two methods result in
approximately the same level of classification error. 
Figure \ref{fig:corrbias} shows that the bias error of Naive Bayes is
at least slightly higher for 8 of 12 words. And Figure
\ref{fig:corrvar} confirms that the variance error tends to be greater
for MC4.


\chapter{UNSUPERVISED LEARNING EXPERIMENTAL RESULTS} 

This chapter contains an experimental evaluation of the unsupervised
learning methodologies described in Chapter 4. These approaches differ
from the supervised learning methods in that no sense--tagged text is
required; only raw untagged text is used to perform disambiguation. 

A probabilistic model of disambiguation is learned from raw text by
treating the sense of an ambiguous word as missing data. As discussed
in Chapter 4, the learning is restricted to parameter estimation since the
parametric form must be specified rather than learned. These
experiments assume that the parametric form is Naive Bayes and use two
different methods to estimate parameter values; the EM algorithm and
Gibbs Sampling. Two agglomerative clustering algorithms are also
considered, Ward's minimum--variance method and McQuitty's similarity
analysis. As discussed in Chapter 4, these methods perform
disambiguation based upon measures of distance that are derived from a
dissimilarity matrix representation of the raw untagged text.\footnote{The
freely available software package Bugs \cite{GilksTS94}  was used for all
Gibbs Sampling  experiments. Various freely available
implementations of the EM algorithm were  employed; AutoClass
\cite{CheesemanS96}, GAMES \cite{Thiesson96}, and  CoCo \cite{Badsberg95}.
The commercial software package SAS \cite{Sas90}, was used to perform the 
agglomerative clustering methods.}  

There is one experiment discussed in this chapter. Thirteen words are
disambiguated using four unsupervised methodologies where the
context in which the ambiguous word occurs is represented by three
different feature sets. This results in disambiguation by 156 possible
combinations of word, method, and feature set. Each possible
combination is repeated 25 times in order to measure the deviation
introduced by randomly selecting initial parameter estimates
for the EM algorithm and Gibbs Sampling, and randomly selecting among
equally distant clusters when using Ward's and McQuitty's algorithms. 
The results of this experiment are evaluated in terms of disambiguation
accuracy. However, as will be outlined in Section \ref{sec:unsupacc},
the evaluation methodology for unsupervised learning is somewhat
different than in the supervised case. 

There are three pairwise analyses of the four unsupervised algorithms
presented.  First, the accuracy of the probabilistic models where the
parameter estimates are learned with the EM algorithm and Gibbs
Sampling are compared. Second, the accuracy of agglomerative
clustering performed by Ward's and McQuitty's methods are
compared. The final analysis compares the two most accurate methods
from the first two pairwise comparisons, McQuitty's similarity
analysis and Gibbs Sampling.   

Each analysis contains two discussions. First, a methodological
comparison is made that highlights any significant differences in
accuracy between two  unsupervised learning algorithms when both use
the same feature set. Second, a feature set comparison is made that
focuses on the variations in accuracy for an unsupervised learner 
as different feature sets are used. 

\section{Assessing Accuracy in Unsupervised Learning}
\label{sec:unsupacc}

In the supervised learning experiments, accuracy is the rate of
agreement between the sense tags automatically assigned by a learned
model and the sense tags assigned by a human judge. This 
definition of accuracy is also employed in the unsupervised
experiment. However, the means of arriving at that evaluation measure
are different and in fact point to some important differences between
supervised and unsupervised learning. 

In supervised learning accuracy is fairly easy to
measure. The supervised algorithm learns from examples where a human
judge has assigned sense tags to ambiguous words that refer to
specific entries in the sense inventory of a word.  For example,
given multiple instances of {\it line} and the sense inventory {\it
(telephone, series, cable)},  a human judge tags some instances with
the {\it telephone} sense, others with the {\it cable} sense, and
still others with the {\it series} sense.  From these examples, the
supervised algorithm learns how to assign these same meaningful tags
to other instances of the ambiguous word.  

In the supervised framework, the act of sense--tagging richly
augments the knowledge contained in a text by creating a link from the
manually disambiguated instances of a word to a sense inventory
provided by a dictionary or other lexical resource.  These links are
critical for evaluation relative to a human judge since they connect
the text to the same sense inventory that the human tagger used. 
However, in unsupervised learning no such links exist. The text is
disconnected from the sense inventory. No human tagger is involved and
the only information available to the unsupervised learner is the raw
text which has no links to a sense inventory or any other external
knowledge source.  

An unsupervised algorithm is limited to creating {\it sense
groups}. A sense group is simply a number of instances of an ambiguous
word that are considered to belong to the same sense. However, there
is no link from the members of the sense group to a sense
inventory. The sense group is labeled  by the unsupervised learner but
this label has no relation to the sense inventory nor does it describe
the contents of the group; it essentially meaningless. 

In order to evaluate the accuracy of the unsupervised algorithm
relative to a human judge, a mapping between the uninformative labels
attached to sense groups and the sense inventory for a word must be
established. In supervised learning these mappings are automatic since
the human tagger provides the link between the text and the sense
inventory. In unsupervised learning this mapping must be made as a
second step after the sense groups are learned.\footnote{The 
separation of disambiguation into a two step process is discussed in
\cite{Schutze98}. There the act of creating sense groups is termed
{\it sense discrimination} and the process of attaching meaningful
labels to these groups is called {\it sense labeling}.} 

Consider the following example. Suppose there are 10 instances of {\it
line} to be disambiguated. A human tagger is told that there are three
entries in the sense inventory,  {\it (telephone, series, cable)}. The
human assigns these sense tags as shown in Figure
\ref{fig:eval4}. An unsupervised learner is given these same 10
instances and told that there are three possible senses. It divides
the usages of the ambiguous word into the 3 sense groups  shown in
Figure \ref{fig:eval1}.   

At this point there is not an immediately apparent means to assess the
agreement between the sense group assignments made by the unsupervised
learner and the sense tags assigned by the human judge.  In order to
determine accuracy, the sense groups must be linked to the entries
in the sense inventory. For this example there are 6 possible mappings
between {\it (1,2,3)} and   {\it (telephone, series,
cable)}.\footnote{Given $n$ sense groups to be assigned $n$ meaningful
sense tags, there are $n!$ possible mappings to be considered.} Each
possible mapping is examined to determine which results in the closest
agreement with the human judge.  

One possible mapping is shown in Figure \ref{fig:eval3}. Sense group 1
is assigned sense tag {\it series}, sense group 2 is assigned {\it
cable}, and sense group 3 is assigned {\it telephone}. The shaded
instances show where the sense group label matches the sense tag
assigned by the human judge. In this figure,  3 of 10 instances agree
and result in unsupervised accuracy of 30\%.    

\begin{figure}[t]
\centerline{\epsfbox{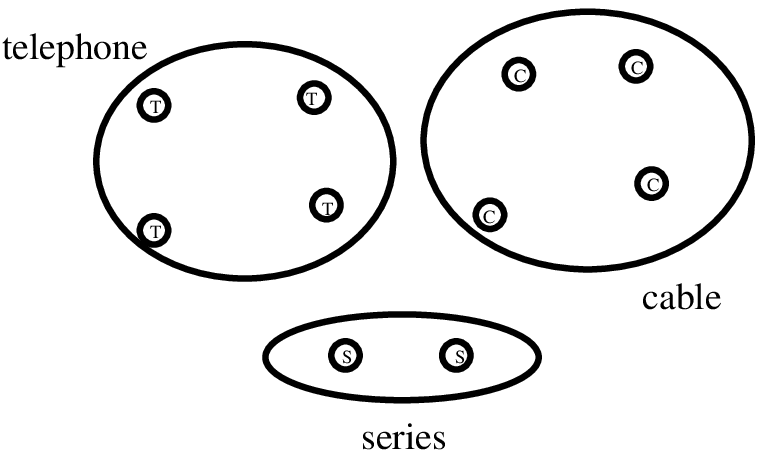}}
\caption{Human Labeled Senses for {\it line}}
\label{fig:eval4}
\myendfig

\begin{figure}
\centerline{\epsfbox{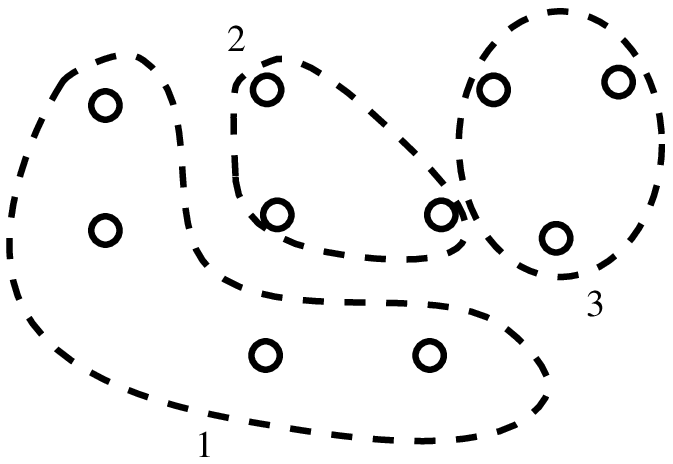}}
\caption{Unlabeled Sense Groups for {\it line}}
\label{fig:eval1}
\myendfig

\begin{figure}
\centerline{\epsfbox{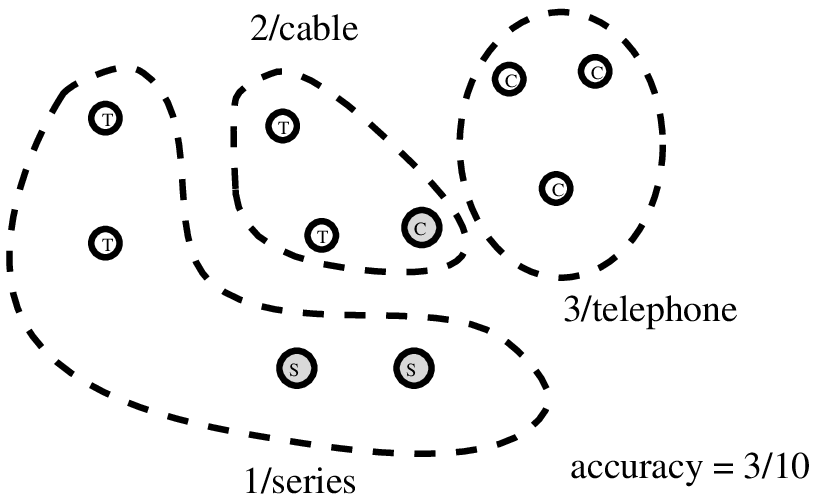}}
\caption{Thirty Percent Accuracy Mapping of {\it line} }
\label{fig:eval3}
\myendfig

\begin{figure}
\centerline{\epsfbox{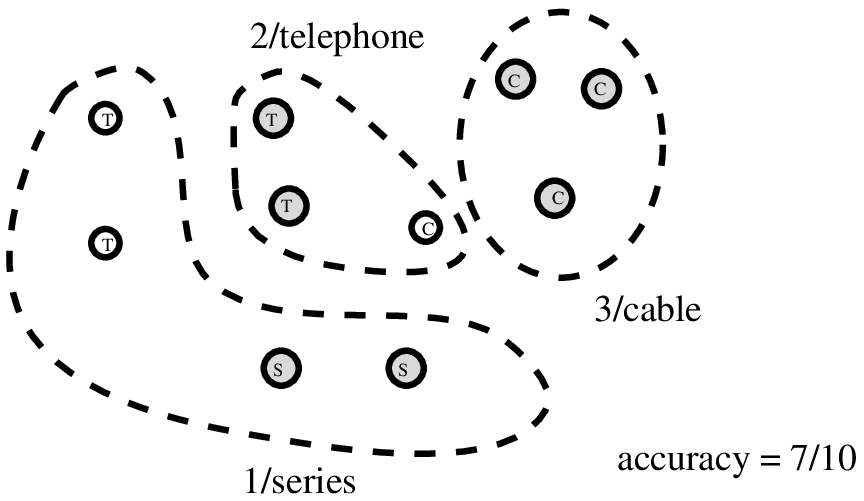}}
\caption{Seventy Percent Accuracy Mapping of {\it line}}
\label{fig:eval2}
\myendfig

A second possible mapping is shown in Figure \ref{fig:eval2}. Sense
group 1 is assigned {\it series}, 2 is assigned {\it telephone}, and 3
is assigned {\it cable}. Here 7 of 10 instances agree with the human
judge so accuracy is 70\%.  The other four possible mappings are
evaluated and the maximum accuracy is
reported as the unsupervised accuracy. 

Given this evaluation methodology, a convenient means of determining a
lower bound for unsupervised disambiguation accuracy emerges. An
unsupervised learner can achieve accuracy equal to the percentage of 
the majority sense by not performing any disambiguation at all. In
other words, a lower bound classifier for unsupervised learning simply
assigns every instance of an ambiguous word to the same sense
group.  This is somewhat analogous to supervised learning where the
lower bound is established by assigning every instance of an ambiguous
word to the most frequent sense in the training data, i.e., the
majority sense. However, in supervised  learning it is relatively easy
to exceed this lower bound. The same does not prove to be true for
unsupervised learning.   

\section{Analysis 1: Probabilistic Models}
\label{sec:anal1}

The first analysis of this experiment compares the accuracy of 
a probabilistic model where the parametric form is assumed to be Naive
Bayes and the parameter estimates are learned by the EM algorithm and
Gibbs Sampling. Table \ref{tab:emgibbs} shows the average
unsupervised disambiguation accuracy and standard deviation 
for each combination of word, feature set, and parameter estimation
method  over 25 trials, where each trial begins with a different
random initialization of the parameter estimates. 

In this table, significant differences in the
disambiguation accuracy of a word using the EM algorithm and Gibbs
Sampling for a given feature set are shown in bold face. These
differences are discussed in Section \ref{sec:method1}, the
methodological comparison. The highest overall accuracy for a word
using either the EM algorithm or Gibbs Sampling and any of the three
feature sets is shown in parenthesis. Any other values that are
not significantly less than the maximum accuracy are
underlined. These results are discussed in Section \ref{sec:feature1},
the feature set comparison.\footnote{As in the supervised learning
experiments, judgments as to the significance of differences in
accuracy are made by a two tailed  t--test where $p=.01$.}  

\begin{table}
\begin{center}
\caption{Unsupervised Accuracy of EM and Gibbs}
\label{tab:emgibbs}
\begin{footnotesize}
\begin{tabular}{|l|c|c|c||c|c||c|c|} \hline
\multicolumn{1}{|l}{} & 
\multicolumn{1}{|l|}{} & 
\multicolumn{2}{|c||}{Feature Set A} & 
\multicolumn{2}{|c||}{Feature Set B} &
\multicolumn{2}{|c|}{Feature Set C} \\ \cline{3-8} 
         & Maj.  &  Gibbs  & EM      & Gibbs   & EM      & Gibbs   & EM      \\ \hline
chief    &.861   & \underline{.719 .01} & \underline{(.729 .06)} & .648
.00 & .646 .01 & \underline{.728 .04} & \underline{.697 .06} \\ 
common   &.842   & .522 .00 & .521 .00 & .507 .07 & .464 .06 & {\bf
\underline{(.670 .01)}} & .543 .09 \\ 
last     &.940   & .900 .00 & .903 .00 & \underline{(.912 .00)} & .909 .00
& .908 .00 & .874 .07 \\
public   &.683   & {\bf .514 .00} & .473 .03 & {\bf .478 .04} & .411 .03 &
{\bf \underline{(.578 .00)}} & .507 .03 \\ \hline 
\multicolumn{1}{|l|}{adjectives} &
\multicolumn{1}{|l|}{.832} &
\multicolumn{1}{|l|}{.663} &
\multicolumn{1}{|l||}{.657} &
\multicolumn{1}{|l|}{.636} &
\multicolumn{1}{|l||}{.608} &
\multicolumn{1}{|l|}{.721} &
\multicolumn{1}{|l|}{.655} \\ \hline \hline
bill     & .681  & .590 .04 & .537 .05 & \underline{(.705 .10)} &
\underline{.624 .08} & .592 .04
& .569 .04 \\
\hline
\end{tabular} 
\end{footnotesize}
\end{center}
\vskip -0.01in
\end{table}

\begin{figure}
\begin{center}
\setlength{\unitlength}{0.240900pt}
\ifx\plotpoint\undefined\newsavebox{\plotpoint}\fi
\sbox{\plotpoint}{\rule[-0.200pt]{0.400pt}{0.400pt}}%
\begin{picture}(1500,900)(0,0)
\font\gnuplot=cmr10 at 10pt
\gnuplot
\sbox{\plotpoint}{\rule[-0.200pt]{0.400pt}{0.400pt}}%
\put(220.0,172.0){\rule[-0.200pt]{4.818pt}{0.400pt}}
\put(198,172){\makebox(0,0)[r]{$0.4$}}
\put(1416.0,172.0){\rule[-0.200pt]{4.818pt}{0.400pt}}
\put(220.0,289.0){\rule[-0.200pt]{4.818pt}{0.400pt}}
\put(198,289){\makebox(0,0)[r]{$0.5$}}
\put(1416.0,289.0){\rule[-0.200pt]{4.818pt}{0.400pt}}
\put(220.0,407.0){\rule[-0.200pt]{4.818pt}{0.400pt}}
\put(198,407){\makebox(0,0)[r]{$0.6$}}
\put(1416.0,407.0){\rule[-0.200pt]{4.818pt}{0.400pt}}
\put(220.0,524.0){\rule[-0.200pt]{4.818pt}{0.400pt}}
\put(198,524){\makebox(0,0)[r]{$0.7$}}
\put(1416.0,524.0){\rule[-0.200pt]{4.818pt}{0.400pt}}
\put(220.0,642.0){\rule[-0.200pt]{4.818pt}{0.400pt}}
\put(198,642){\makebox(0,0)[r]{$0.8$}}
\put(1416.0,642.0){\rule[-0.200pt]{4.818pt}{0.400pt}}
\put(220.0,759.0){\rule[-0.200pt]{4.818pt}{0.400pt}}
\put(198,759){\makebox(0,0)[r]{$0.9$}}
\put(1416.0,759.0){\rule[-0.200pt]{4.818pt}{0.400pt}}
\put(220.0,877.0){\rule[-0.200pt]{4.818pt}{0.400pt}}
\put(198,877){\makebox(0,0)[r]{$1$}}
\put(1416.0,877.0){\rule[-0.200pt]{4.818pt}{0.400pt}}
\put(314.0,113.0){\rule[-0.200pt]{0.400pt}{4.818pt}}
\put(314,68){\makebox(0,0){$0.4$}}
\put(314.0,857.0){\rule[-0.200pt]{0.400pt}{4.818pt}}
\put(501.0,113.0){\rule[-0.200pt]{0.400pt}{4.818pt}}
\put(501,68){\makebox(0,0){$0.5$}}
\put(501.0,857.0){\rule[-0.200pt]{0.400pt}{4.818pt}}
\put(688.0,113.0){\rule[-0.200pt]{0.400pt}{4.818pt}}
\put(688,68){\makebox(0,0){$0.6$}}
\put(688.0,857.0){\rule[-0.200pt]{0.400pt}{4.818pt}}
\put(875.0,113.0){\rule[-0.200pt]{0.400pt}{4.818pt}}
\put(875,68){\makebox(0,0){$0.7$}}
\put(875.0,857.0){\rule[-0.200pt]{0.400pt}{4.818pt}}
\put(1062.0,113.0){\rule[-0.200pt]{0.400pt}{4.818pt}}
\put(1062,68){\makebox(0,0){$0.8$}}
\put(1062.0,857.0){\rule[-0.200pt]{0.400pt}{4.818pt}}
\put(1249.0,113.0){\rule[-0.200pt]{0.400pt}{4.818pt}}
\put(1249,68){\makebox(0,0){$0.9$}}
\put(1249.0,857.0){\rule[-0.200pt]{0.400pt}{4.818pt}}
\put(1436.0,113.0){\rule[-0.200pt]{0.400pt}{4.818pt}}
\put(1436,68){\makebox(0,0){$1$}}
\put(1436.0,857.0){\rule[-0.200pt]{0.400pt}{4.818pt}}
\put(220.0,113.0){\rule[-0.200pt]{292.934pt}{0.400pt}}
\put(1436.0,113.0){\rule[-0.200pt]{0.400pt}{184.048pt}}
\put(220.0,877.0){\rule[-0.200pt]{292.934pt}{0.400pt}}
\put(45,495){\makebox(0,0){EM}}
\put(828,23){\makebox(0,0){Gibbs Sampling}}
\put(220.0,113.0){\rule[-0.200pt]{0.400pt}{184.048pt}}
\put(501,818){\makebox(0,0)[r]{Feature Set A}}
\put(545,818){\raisebox{-.8pt}{\makebox(0,0){$\Diamond$}}}
\put(910,558){\raisebox{-.8pt}{\makebox(0,0){$\Diamond$}}}
\put(542,314){\raisebox{-.8pt}{\makebox(0,0){$\Diamond$}}}
\put(1249,763){\raisebox{-.8pt}{\makebox(0,0){$\Diamond$}}}
\put(527,258){\raisebox{-.8pt}{\makebox(0,0){$\Diamond$}}}
\put(669,333){\raisebox{-.8pt}{\makebox(0,0){$\Diamond$}}}
\put(1140,691){\raisebox{-.8pt}{\makebox(0,0){$\Diamond$}}}
\put(830,475){\raisebox{-.8pt}{\makebox(0,0){$\Diamond$}}}
\put(738,426){\raisebox{-.8pt}{\makebox(0,0){$\Diamond$}}}
\put(400,239){\raisebox{-.8pt}{\makebox(0,0){$\Diamond$}}}
\put(705,443){\raisebox{-.8pt}{\makebox(0,0){$\Diamond$}}}
\put(620,360){\raisebox{-.8pt}{\makebox(0,0){$\Diamond$}}}
\put(796,390){\raisebox{-.8pt}{\makebox(0,0){$\Diamond$}}}
\put(938,554){\raisebox{-.8pt}{\makebox(0,0){$\Diamond$}}}
\put(501,773){\makebox(0,0)[r]{Feature Set B}}
\put(545,773){\makebox(0,0){$+$}}
\put(777,461){\makebox(0,0){$+$}}
\put(514,247){\makebox(0,0){$+$}}
\put(1271,770){\makebox(0,0){$+$}}
\put(459,185){\makebox(0,0){$+$}}
\put(884,435){\makebox(0,0){$+$}}
\put(1097,689){\makebox(0,0){$+$}}
\put(581,349){\makebox(0,0){$+$}}
\put(785,424){\makebox(0,0){$+$}}
\put(458,259){\makebox(0,0){$+$}}
\put(901,504){\makebox(0,0){$+$}}
\put(901,491){\makebox(0,0){$+$}}
\put(546,320){\makebox(0,0){$+$}}
\put(1124,622){\makebox(0,0){$+$}}
\put(501,728){\makebox(0,0)[r]{Feature Set C}}
\put(545,728){\raisebox{-.8pt}{\makebox(0,0){$\Box$}}}
\put(927,521){\raisebox{-.8pt}{\makebox(0,0){$\Box$}}}
\put(819,340){\raisebox{-.8pt}{\makebox(0,0){$\Box$}}}
\put(1264,729){\raisebox{-.8pt}{\makebox(0,0){$\Box$}}}
\put(647,298){\raisebox{-.8pt}{\makebox(0,0){$\Box$}}}
\put(673,370){\raisebox{-.8pt}{\makebox(0,0){$\Box$}}}
\put(1034,593){\raisebox{-.8pt}{\makebox(0,0){$\Box$}}}
\put(826,468){\raisebox{-.8pt}{\makebox(0,0){$\Box$}}}
\put(719,464){\raisebox{-.8pt}{\makebox(0,0){$\Box$}}}
\put(420,240){\raisebox{-.8pt}{\makebox(0,0){$\Box$}}}
\put(847,507){\raisebox{-.8pt}{\makebox(0,0){$\Box$}}}
\put(755,463){\raisebox{-.8pt}{\makebox(0,0){$\Box$}}}
\put(867,409){\raisebox{-.8pt}{\makebox(0,0){$\Box$}}}
\put(596,330){\raisebox{-.8pt}{\makebox(0,0){$\Box$}}}
\sbox{\plotpoint}{\rule[-0.500pt]{1.000pt}{1.000pt}}%
\put(220,113){\usebox{\plotpoint}}
\put(220.00,113.00){\usebox{\plotpoint}}
\put(237.58,124.00){\usebox{\plotpoint}}
\put(255.25,134.84){\usebox{\plotpoint}}
\multiput(257,136)(17.270,11.513){0}{\usebox{\plotpoint}}
\put(272.52,146.35){\usebox{\plotpoint}}
\put(290.31,157.01){\usebox{\plotpoint}}
\multiput(294,159)(17.270,11.513){0}{\usebox{\plotpoint}}
\put(307.78,168.19){\usebox{\plotpoint}}
\put(325.46,179.02){\usebox{\plotpoint}}
\multiput(331,182)(17.270,11.513){0}{\usebox{\plotpoint}}
\put(343.03,190.02){\usebox{\plotpoint}}
\put(360.30,201.53){\usebox{\plotpoint}}
\put(378.19,212.02){\usebox{\plotpoint}}
\multiput(380,213)(17.270,11.513){0}{\usebox{\plotpoint}}
\put(395.56,223.37){\usebox{\plotpoint}}
\put(413.34,234.03){\usebox{\plotpoint}}
\multiput(417,236)(17.270,11.513){0}{\usebox{\plotpoint}}
\put(430.81,245.21){\usebox{\plotpoint}}
\put(448.08,256.72){\usebox{\plotpoint}}
\multiput(453,260)(18.275,9.840){0}{\usebox{\plotpoint}}
\put(466.06,267.04){\usebox{\plotpoint}}
\put(483.33,278.56){\usebox{\plotpoint}}
\put(501.22,289.04){\usebox{\plotpoint}}
\multiput(503,290)(17.270,11.513){0}{\usebox{\plotpoint}}
\put(518.59,300.39){\usebox{\plotpoint}}
\put(535.86,311.91){\usebox{\plotpoint}}
\multiput(539,314)(18.275,9.840){0}{\usebox{\plotpoint}}
\put(553.84,322.23){\usebox{\plotpoint}}
\put(571.11,333.74){\usebox{\plotpoint}}
\multiput(576,337)(17.270,11.513){0}{\usebox{\plotpoint}}
\put(588.40,345.22){\usebox{\plotpoint}}
\put(606.37,355.58){\usebox{\plotpoint}}
\put(623.64,367.09){\usebox{\plotpoint}}
\multiput(625,368)(18.275,9.840){0}{\usebox{\plotpoint}}
\put(641.62,377.41){\usebox{\plotpoint}}
\put(658.89,388.93){\usebox{\plotpoint}}
\multiput(662,391)(17.270,11.513){0}{\usebox{\plotpoint}}
\put(676.28,400.23){\usebox{\plotpoint}}
\put(694.14,410.76){\usebox{\plotpoint}}
\multiput(699,414)(17.270,11.513){0}{\usebox{\plotpoint}}
\put(711.44,422.24){\usebox{\plotpoint}}
\put(729.40,432.60){\usebox{\plotpoint}}
\put(746.67,444.11){\usebox{\plotpoint}}
\multiput(748,445)(17.270,11.513){0}{\usebox{\plotpoint}}
\put(764.17,455.24){\usebox{\plotpoint}}
\put(781.92,465.95){\usebox{\plotpoint}}
\multiput(785,468)(17.270,11.513){0}{\usebox{\plotpoint}}
\put(799.32,477.25){\usebox{\plotpoint}}
\put(817.18,487.78){\usebox{\plotpoint}}
\multiput(822,491)(17.270,11.513){0}{\usebox{\plotpoint}}
\put(834.45,499.30){\usebox{\plotpoint}}
\put(852.05,510.26){\usebox{\plotpoint}}
\put(869.70,521.13){\usebox{\plotpoint}}
\multiput(871,522)(17.270,11.513){0}{\usebox{\plotpoint}}
\put(887.20,532.26){\usebox{\plotpoint}}
\put(904.95,542.97){\usebox{\plotpoint}}
\multiput(908,545)(17.270,11.513){0}{\usebox{\plotpoint}}
\put(922.22,554.48){\usebox{\plotpoint}}
\put(939.93,565.27){\usebox{\plotpoint}}
\multiput(945,568)(17.270,11.513){0}{\usebox{\plotpoint}}
\put(957.48,576.32){\usebox{\plotpoint}}
\put(975.08,587.27){\usebox{\plotpoint}}
\put(992.73,598.15){\usebox{\plotpoint}}
\multiput(994,599)(17.270,11.513){0}{\usebox{\plotpoint}}
\put(1010.00,609.67){\usebox{\plotpoint}}
\put(1027.81,620.28){\usebox{\plotpoint}}
\multiput(1031,622)(17.270,11.513){0}{\usebox{\plotpoint}}
\put(1045.26,631.50){\usebox{\plotpoint}}
\put(1062.96,642.29){\usebox{\plotpoint}}
\multiput(1068,645)(17.270,11.513){0}{\usebox{\plotpoint}}
\put(1080.51,653.34){\usebox{\plotpoint}}
\put(1097.78,664.85){\usebox{\plotpoint}}
\put(1115.69,675.30){\usebox{\plotpoint}}
\multiput(1117,676)(17.270,11.513){0}{\usebox{\plotpoint}}
\put(1133.03,686.69){\usebox{\plotpoint}}
\put(1150.30,698.20){\usebox{\plotpoint}}
\multiput(1153,700)(18.275,9.840){0}{\usebox{\plotpoint}}
\put(1168.29,708.53){\usebox{\plotpoint}}
\put(1185.56,720.04){\usebox{\plotpoint}}
\multiput(1190,723)(18.275,9.840){0}{\usebox{\plotpoint}}
\put(1203.54,730.36){\usebox{\plotpoint}}
\put(1220.81,741.87){\usebox{\plotpoint}}
\put(1238.08,753.39){\usebox{\plotpoint}}
\multiput(1239,754)(18.275,9.840){0}{\usebox{\plotpoint}}
\put(1256.07,763.71){\usebox{\plotpoint}}
\put(1273.34,775.22){\usebox{\plotpoint}}
\multiput(1276,777)(18.275,9.840){0}{\usebox{\plotpoint}}
\put(1291.32,785.55){\usebox{\plotpoint}}
\put(1308.59,797.06){\usebox{\plotpoint}}
\multiput(1313,800)(17.270,11.513){0}{\usebox{\plotpoint}}
\put(1325.91,808.49){\usebox{\plotpoint}}
\put(1343.84,818.90){\usebox{\plotpoint}}
\put(1361.11,830.41){\usebox{\plotpoint}}
\multiput(1362,831)(18.275,9.840){0}{\usebox{\plotpoint}}
\put(1379.10,840.73){\usebox{\plotpoint}}
\put(1396.37,852.24){\usebox{\plotpoint}}
\multiput(1399,854)(17.270,11.513){0}{\usebox{\plotpoint}}
\put(1413.79,863.50){\usebox{\plotpoint}}
\put(1431.62,874.08){\usebox{\plotpoint}}
\put(1436,877){\usebox{\plotpoint}}
\end{picture}
\caption{Probabilistic Model Correlation of Accuracy for all words}
\label{fig:accemgibb}
\end{center}
\vskip -0.14in
\end{figure}

\begin{figure}
\begin{center}
\setlength{\unitlength}{0.240900pt}
\ifx\plotpoint\undefined\newsavebox{\plotpoint}\fi
\sbox{\plotpoint}{\rule[-0.200pt]{0.400pt}{0.400pt}}%
\begin{picture}(1500,900)(0,0)
\font\gnuplot=cmr10 at 10pt
\gnuplot
\sbox{\plotpoint}{\rule[-0.200pt]{0.400pt}{0.400pt}}%
\put(220.0,172.0){\rule[-0.200pt]{4.818pt}{0.400pt}}
\put(198,172){\makebox(0,0)[r]{$0.4$}}
\put(1416.0,172.0){\rule[-0.200pt]{4.818pt}{0.400pt}}
\put(220.0,289.0){\rule[-0.200pt]{4.818pt}{0.400pt}}
\put(198,289){\makebox(0,0)[r]{$0.5$}}
\put(1416.0,289.0){\rule[-0.200pt]{4.818pt}{0.400pt}}
\put(220.0,407.0){\rule[-0.200pt]{4.818pt}{0.400pt}}
\put(198,407){\makebox(0,0)[r]{$0.6$}}
\put(1416.0,407.0){\rule[-0.200pt]{4.818pt}{0.400pt}}
\put(220.0,524.0){\rule[-0.200pt]{4.818pt}{0.400pt}}
\put(198,524){\makebox(0,0)[r]{$0.7$}}
\put(1416.0,524.0){\rule[-0.200pt]{4.818pt}{0.400pt}}
\put(220.0,642.0){\rule[-0.200pt]{4.818pt}{0.400pt}}
\put(198,642){\makebox(0,0)[r]{$0.8$}}
\put(1416.0,642.0){\rule[-0.200pt]{4.818pt}{0.400pt}}
\put(220.0,759.0){\rule[-0.200pt]{4.818pt}{0.400pt}}
\put(198,759){\makebox(0,0)[r]{$0.9$}}
\put(1416.0,759.0){\rule[-0.200pt]{4.818pt}{0.400pt}}
\put(220.0,877.0){\rule[-0.200pt]{4.818pt}{0.400pt}}
\put(198,877){\makebox(0,0)[r]{$1$}}
\put(1416.0,877.0){\rule[-0.200pt]{4.818pt}{0.400pt}}
\put(314.0,113.0){\rule[-0.200pt]{0.400pt}{4.818pt}}
\put(314,68){\makebox(0,0){$0.4$}}
\put(314.0,857.0){\rule[-0.200pt]{0.400pt}{4.818pt}}
\put(501.0,113.0){\rule[-0.200pt]{0.400pt}{4.818pt}}
\put(501,68){\makebox(0,0){$0.5$}}
\put(501.0,857.0){\rule[-0.200pt]{0.400pt}{4.818pt}}
\put(688.0,113.0){\rule[-0.200pt]{0.400pt}{4.818pt}}
\put(688,68){\makebox(0,0){$0.6$}}
\put(688.0,857.0){\rule[-0.200pt]{0.400pt}{4.818pt}}
\put(875.0,113.0){\rule[-0.200pt]{0.400pt}{4.818pt}}
\put(875,68){\makebox(0,0){$0.7$}}
\put(875.0,857.0){\rule[-0.200pt]{0.400pt}{4.818pt}}
\put(1062.0,113.0){\rule[-0.200pt]{0.400pt}{4.818pt}}
\put(1062,68){\makebox(0,0){$0.8$}}
\put(1062.0,857.0){\rule[-0.200pt]{0.400pt}{4.818pt}}
\put(1249.0,113.0){\rule[-0.200pt]{0.400pt}{4.818pt}}
\put(1249,68){\makebox(0,0){$0.9$}}
\put(1249.0,857.0){\rule[-0.200pt]{0.400pt}{4.818pt}}
\put(1436.0,113.0){\rule[-0.200pt]{0.400pt}{4.818pt}}
\put(1436,68){\makebox(0,0){$1$}}
\put(1436.0,857.0){\rule[-0.200pt]{0.400pt}{4.818pt}}
\put(220.0,113.0){\rule[-0.200pt]{292.934pt}{0.400pt}}
\put(1436.0,113.0){\rule[-0.200pt]{0.400pt}{184.048pt}}
\put(220.0,877.0){\rule[-0.200pt]{292.934pt}{0.400pt}}
\put(45,495){\makebox(0,0){EM}}
\put(828,23){\makebox(0,0){Gibbs Sampling}}
\put(220.0,113.0){\rule[-0.200pt]{0.400pt}{184.048pt}}
\put(501,818){\makebox(0,0)[r]{Feature Set A}}
\put(545,818){\raisebox{-.8pt}{\makebox(0,0){$\Diamond$}}}
\put(669,333){\raisebox{-.8pt}{\makebox(0,0){$\Diamond$}}}
\put(1140,691){\raisebox{-.8pt}{\makebox(0,0){$\Diamond$}}}
\put(830,475){\raisebox{-.8pt}{\makebox(0,0){$\Diamond$}}}
\put(738,426){\raisebox{-.8pt}{\makebox(0,0){$\Diamond$}}}
\put(400,239){\raisebox{-.8pt}{\makebox(0,0){$\Diamond$}}}
\put(501,773){\makebox(0,0)[r]{Feature Set B}}
\put(545,773){\makebox(0,0){$+$}}
\put(884,435){\makebox(0,0){$+$}}
\put(1097,689){\makebox(0,0){$+$}}
\put(581,349){\makebox(0,0){$+$}}
\put(785,424){\makebox(0,0){$+$}}
\put(458,259){\makebox(0,0){$+$}}
\put(501,728){\makebox(0,0)[r]{Feature Set C}}
\put(545,728){\raisebox{-.8pt}{\makebox(0,0){$\Box$}}}
\put(673,370){\raisebox{-.8pt}{\makebox(0,0){$\Box$}}}
\put(1034,593){\raisebox{-.8pt}{\makebox(0,0){$\Box$}}}
\put(826,468){\raisebox{-.8pt}{\makebox(0,0){$\Box$}}}
\put(719,464){\raisebox{-.8pt}{\makebox(0,0){$\Box$}}}
\put(420,240){\raisebox{-.8pt}{\makebox(0,0){$\Box$}}}
\sbox{\plotpoint}{\rule[-0.500pt]{1.000pt}{1.000pt}}%
\put(220,113){\usebox{\plotpoint}}
\put(220.00,113.00){\usebox{\plotpoint}}
\put(237.58,124.00){\usebox{\plotpoint}}
\put(255.25,134.84){\usebox{\plotpoint}}
\multiput(257,136)(17.270,11.513){0}{\usebox{\plotpoint}}
\put(272.52,146.35){\usebox{\plotpoint}}
\put(290.31,157.01){\usebox{\plotpoint}}
\multiput(294,159)(17.270,11.513){0}{\usebox{\plotpoint}}
\put(307.78,168.19){\usebox{\plotpoint}}
\put(325.46,179.02){\usebox{\plotpoint}}
\multiput(331,182)(17.270,11.513){0}{\usebox{\plotpoint}}
\put(343.03,190.02){\usebox{\plotpoint}}
\put(360.30,201.53){\usebox{\plotpoint}}
\put(378.19,212.02){\usebox{\plotpoint}}
\multiput(380,213)(17.270,11.513){0}{\usebox{\plotpoint}}
\put(395.56,223.37){\usebox{\plotpoint}}
\put(413.34,234.03){\usebox{\plotpoint}}
\multiput(417,236)(17.270,11.513){0}{\usebox{\plotpoint}}
\put(430.81,245.21){\usebox{\plotpoint}}
\put(448.08,256.72){\usebox{\plotpoint}}
\multiput(453,260)(18.275,9.840){0}{\usebox{\plotpoint}}
\put(466.06,267.04){\usebox{\plotpoint}}
\put(483.33,278.56){\usebox{\plotpoint}}
\put(501.22,289.04){\usebox{\plotpoint}}
\multiput(503,290)(17.270,11.513){0}{\usebox{\plotpoint}}
\put(518.59,300.39){\usebox{\plotpoint}}
\put(535.86,311.91){\usebox{\plotpoint}}
\multiput(539,314)(18.275,9.840){0}{\usebox{\plotpoint}}
\put(553.84,322.23){\usebox{\plotpoint}}
\put(571.11,333.74){\usebox{\plotpoint}}
\multiput(576,337)(17.270,11.513){0}{\usebox{\plotpoint}}
\put(588.40,345.22){\usebox{\plotpoint}}
\put(606.37,355.58){\usebox{\plotpoint}}
\put(623.64,367.09){\usebox{\plotpoint}}
\multiput(625,368)(18.275,9.840){0}{\usebox{\plotpoint}}
\put(641.62,377.41){\usebox{\plotpoint}}
\put(658.89,388.93){\usebox{\plotpoint}}
\multiput(662,391)(17.270,11.513){0}{\usebox{\plotpoint}}
\put(676.28,400.23){\usebox{\plotpoint}}
\put(694.14,410.76){\usebox{\plotpoint}}
\multiput(699,414)(17.270,11.513){0}{\usebox{\plotpoint}}
\put(711.44,422.24){\usebox{\plotpoint}}
\put(729.40,432.60){\usebox{\plotpoint}}
\put(746.67,444.11){\usebox{\plotpoint}}
\multiput(748,445)(17.270,11.513){0}{\usebox{\plotpoint}}
\put(764.17,455.24){\usebox{\plotpoint}}
\put(781.92,465.95){\usebox{\plotpoint}}
\multiput(785,468)(17.270,11.513){0}{\usebox{\plotpoint}}
\put(799.32,477.25){\usebox{\plotpoint}}
\put(817.18,487.78){\usebox{\plotpoint}}
\multiput(822,491)(17.270,11.513){0}{\usebox{\plotpoint}}
\put(834.45,499.30){\usebox{\plotpoint}}
\put(852.05,510.26){\usebox{\plotpoint}}
\put(869.70,521.13){\usebox{\plotpoint}}
\multiput(871,522)(17.270,11.513){0}{\usebox{\plotpoint}}
\put(887.20,532.26){\usebox{\plotpoint}}
\put(904.95,542.97){\usebox{\plotpoint}}
\multiput(908,545)(17.270,11.513){0}{\usebox{\plotpoint}}
\put(922.22,554.48){\usebox{\plotpoint}}
\put(939.93,565.27){\usebox{\plotpoint}}
\multiput(945,568)(17.270,11.513){0}{\usebox{\plotpoint}}
\put(957.48,576.32){\usebox{\plotpoint}}
\put(975.08,587.27){\usebox{\plotpoint}}
\put(992.73,598.15){\usebox{\plotpoint}}
\multiput(994,599)(17.270,11.513){0}{\usebox{\plotpoint}}
\put(1010.00,609.67){\usebox{\plotpoint}}
\put(1027.81,620.28){\usebox{\plotpoint}}
\multiput(1031,622)(17.270,11.513){0}{\usebox{\plotpoint}}
\put(1045.26,631.50){\usebox{\plotpoint}}
\put(1062.96,642.29){\usebox{\plotpoint}}
\multiput(1068,645)(17.270,11.513){0}{\usebox{\plotpoint}}
\put(1080.51,653.34){\usebox{\plotpoint}}
\put(1097.78,664.85){\usebox{\plotpoint}}
\put(1115.69,675.30){\usebox{\plotpoint}}
\multiput(1117,676)(17.270,11.513){0}{\usebox{\plotpoint}}
\put(1133.03,686.69){\usebox{\plotpoint}}
\put(1150.30,698.20){\usebox{\plotpoint}}
\multiput(1153,700)(18.275,9.840){0}{\usebox{\plotpoint}}
\put(1168.29,708.53){\usebox{\plotpoint}}
\put(1185.56,720.04){\usebox{\plotpoint}}
\multiput(1190,723)(18.275,9.840){0}{\usebox{\plotpoint}}
\put(1203.54,730.36){\usebox{\plotpoint}}
\put(1220.81,741.87){\usebox{\plotpoint}}
\put(1238.08,753.39){\usebox{\plotpoint}}
\multiput(1239,754)(18.275,9.840){0}{\usebox{\plotpoint}}
\put(1256.07,763.71){\usebox{\plotpoint}}
\put(1273.34,775.22){\usebox{\plotpoint}}
\multiput(1276,777)(18.275,9.840){0}{\usebox{\plotpoint}}
\put(1291.32,785.55){\usebox{\plotpoint}}
\put(1308.59,797.06){\usebox{\plotpoint}}
\multiput(1313,800)(17.270,11.513){0}{\usebox{\plotpoint}}
\put(1325.91,808.49){\usebox{\plotpoint}}
\put(1343.84,818.90){\usebox{\plotpoint}}
\put(1361.11,830.41){\usebox{\plotpoint}}
\multiput(1362,831)(18.275,9.840){0}{\usebox{\plotpoint}}
\put(1379.10,840.73){\usebox{\plotpoint}}
\put(1396.37,852.24){\usebox{\plotpoint}}
\multiput(1399,854)(17.270,11.513){0}{\usebox{\plotpoint}}
\put(1413.79,863.50){\usebox{\plotpoint}}
\put(1431.62,874.08){\usebox{\plotpoint}}
\put(1436,877){\usebox{\plotpoint}}
\end{picture}
\caption{Probabilistic Model Correlation of Accuracy for Nouns}
\label{fig:acc-noun}
\end{center}
\vskip -0.14in
\end{figure}
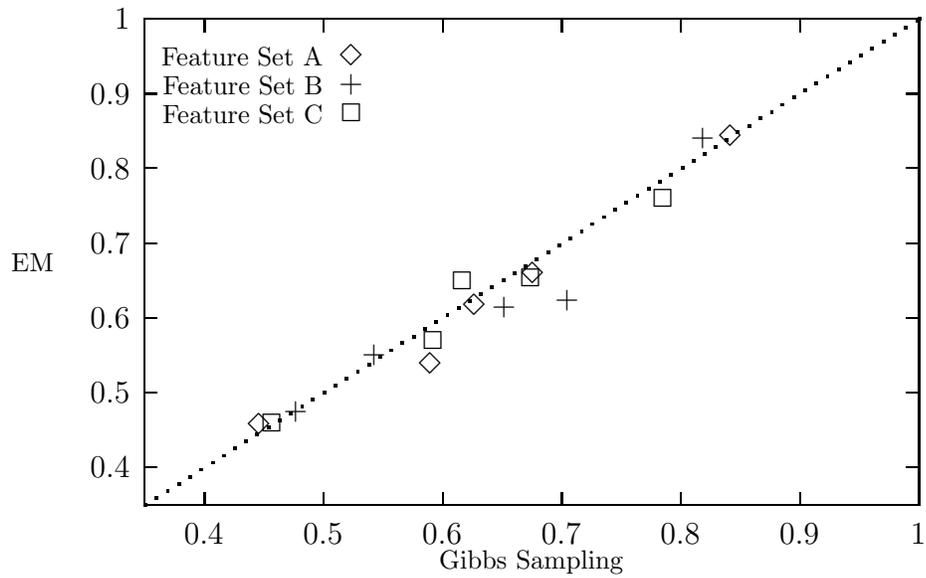

\subsection{Methodological Comparison}
\label{sec:method1}

This comparison studies the effect on disambiguation accuracy of
learning parameter estimates for a probabilistic model using Gibbs
Sampling and the EM algorithm. The motivation for the comparison
is that while the EM algorithm is easy to implement and
generally quick to converge, it is also prone to converging at local
maxima rather than a global maximum. However, while Gibbs
Sampling is guaranteed to converge at the global maximum, it does so
at greater computational expense. 

This experiment shows that there are only a few cases where a
probabilistic model with parameter estimates learned by Gibbs Sampling
results in significantly different disambiguation than the model
arrived at with the EM algorithm. Of the 39 possible pairwise
comparisons (13 words $\times$ 3 feature sets) between the EM
algorithm and Gibbs Sampling, only 7 result in significant
differences. Of those, all favor Gibbs Sampling. Those cases are shown
in bold face in Table \ref{tab:emgibbs}.   

The lack of significant differences between the EM algorithm and Gibbs
Sampling is somewhat surprising given that the EM algorithm can
converge to local maxima when the distribution of the likelihood
function is not well approximated by the normal distribution. However,
in this experiment the EM algorithm does not appear to have great
difficulty with local maxima, often converging within 20 iterations to
essentially the same estimates obtained by Gibbs Sampling.   

The use of Naive Bayes as the parametric form of the probabilistic
model provides at least a partial explanation for the comparable
results obtained with the EM algorithm and Gibbs Sampling.  Chapter
5 presents the distribution of the event counts in the
experimental data when the parametric form of the model is Naive
Bayes. These distributions prove to be relatively smooth for the three
unsupervised feature sets and are not dominated by events that are
never observed in the data; in fact, a majority of the possible events
for  each word are observed.  If the parametric form of the model were
more complex than Naive Bayes, it would certainly be the case that the
distribution of event counts would be more skewed and that the EM
algorithm would be more susceptible to becoming trapped at a local
maxima. However, the specification of Naive Bayes as the parametric
form of the model seems to avoid this difficulty.  

While the number of significant differences between the EM algorithm
and Gibbs Sampling is small, the correlation plot comparing the accuracy
of the two methods in Figure \ref{fig:accemgibb} reveals a consistent
increase in the accuracy of Gibbs Sampling relative to the EM
algorithm. Each point on this plot shows the accuracy attained by the
probabilistic model where parameter estimates are learned by the EM
algorithm and Gibbs Sampling for a given word and feature set. Figure
\ref{fig:acc-noun} again shows the correlation of accuracy, but only
for the nouns. This shows the comparable performance of both methods 
and suggests that Gibbs Sampling has a particular advantage over the
EM algorithm for the adjectives and verbs. This is not surprising
since the adjectives and verbs have the most skewed distributions of
senses and are more likely to cause difficulty for the EM algorithm
than are the nouns.   

The standard deviations associated with the two approaches  also prove
to be similar, generally falling between .03 and .10.  A standard
deviation of .00 indicates that the exact same sense group is created
by each of the 25 trials of the algorithm. The larger the deviation
the more variation there is in the sense groups created from trial to
trial.  The standard deviation observed is somewhat larger than expected,
particularly since neither method has substantial difficulties with
local maxima. This suggests that there is some degree of noise in the
data that is obscuring sense distinctions and causing the variance in
the results from trial to trial. 

There are several possible sources of noise in this data. The very
crude part--of--speech distinctions made in feature sets A and C may
not provide sufficient information to distinguish among senses. In
addition, this tagging was performed without the benefit of training
data and is likely to contain inaccuracies. 

The frequency based co--occurrences in feature sets B and C include a
value that signifies that the word at a specified position relative to
the ambiguous word is not among the 19 most frequent words that occur
at this position with all instances of the ambiguous word; this lumps
together a great many words into a single feature value and may further
blur the ability of the learning algorithm to accurately make sense
distinctions.

The difficulty in unsupervised learning is that features must be
selected from raw untagged text. The availability of accurately 
part--of--speech tagged text can not be assumed, nor can the ability
to select feature values that are indicative of minority senses. Raw
untagged text generally does not lend itself to fine grained feature
values that are able to identify particular senses; generally speaking
features must be selected simply based on frequency counts and lead to
a certain amount of noise in the data.

A noteworthy result is that only the nouns are
disambiguated with accuracy greater than the 
discussed lower bound for unsupervised learning.  The accuracy of the
probabilistic models is less than the lower bound 
when the percentage of the majority sense exceeds 68\%.
However, even in cases where the accuracy of the EM algorithm and
Gibbs Sampling is less than the lower bound, these methods are often
still providing  high accuracy disambiguation. For example, Gibbs
Sampling is able to achieve  91\% accuracy for {\it last} and 83\%
accuracy for {\it include}.   

The relative success of noun disambiguation is at least partially
explained by the fact that, as a class, the  nouns have the most
uniform distribution of senses.  However,  the distribution of senses is
not the only factor  affecting disambiguation accuracy;  the
performance of the EM algorithm and Gibbs Sampling is quite different
for {\it bill} and {\it public} despite having roughly the same sense
distributions.     

It is difficult to quantify the effect of the distribution of
senses  on a  learning algorithm,  particularly when using naturally
occurring data.  In previous unsupervised experiments with {\it
interest}, using a feature set similar to A, an increase of 36
percentage  points over the accuracy of the lower bound was
achieved when the  3 senses were evenly distributed in the training
data \cite{PedersenB97B}.  Here, the most accurate performance using
larger samples and a natural   distribution of senses  is only an
increase of 20 percentage points  over  the accuracy of the lower
bound.  

The actual distribution of senses does not closely correspond to the 
distribution of senses discovered by
either method. As examples, the distribution of senses discovered by
the EM algorithm and Gibbs Sampling relative to the known distribution
of senses is illustrated in Figures \ref{fig:concerna},
\ref{fig:interestb} and \ref{fig:helpc}.  These show the confusion
matrices associated with the disambiguation of {\it concern}, {\it
interest}, and {\it help}, using feature sets A, B, and C,
respectively. A confusion matrix shows the number of cases where the
sense  discovered by the algorithm agrees with the manually assigned
sense along the main diagonal; disagreements are shown in the rest of
the matrix.   The row totals show the actual distribution of senses
while the column totals show the discovered distributions. 

In general, these matrices show that the EM algorithm and Gibbs
Sampling result in distributions of senses that are more balanced than
those of the actual distribution. This is at  least partially due to
the assumption made prior to learning by the unsupervised methods that
each possible sense is equally likely. Adjusting this prior assumption
could result in the discovery of less balanced distributions of senses
and is an interesting direction for future research.

\begin{figure}
\begin{center}
\begin{tabular}{cr|cc|c}   
\multicolumn{2}{c}{} &
\multicolumn{2}{|c|}{Discovered} \\
\multicolumn{1}{c}{} &
\multicolumn{1}{c|}{Actual} &
\multicolumn{1}{|c}{worry} &
\multicolumn{1}{c|}{business} & \\  \hline
&worry & 384 & 63  & 447 \\
&business & 132  & 656 & 788 \\ \hline
&         & 516 & 719 & 1235  \\ \\
\multicolumn{5}{c}{EM - 1040 correct} \\ \\
\multicolumn{2}{c}{} &
\multicolumn{2}{|c|}{Discovered} \\
\multicolumn{1}{c}{} &   
\multicolumn{1}{c|}{Actual} &
\multicolumn{1}{|c}{worry} &
\multicolumn{1}{c|}{business} & \\  \hline
&worry & 384 & 63  & 447 \\  
&business & 132  & 656 & 788 \\ \hline
&         & 516 & 719 & 1235  \\ \\
\multicolumn{5}{c}{Gibbs - 1040 correct} \\ \\
\end{tabular}
\caption{concern - Feature Set A}
\label{fig:concerna}
\end{center}
\vskip -0.14in
\end{figure}

\begin{figure}
\begin{center}
\begin{tabular}{cr|ccc|c}
\multicolumn{2}{c}{} &
\multicolumn{3}{|c|}{Discovered} \\
\multicolumn{1}{c}{} &  
\multicolumn{1}{c|}{Actual} &
\multicolumn{1}{|c}{attention} &
\multicolumn{1}{c}{share} &
\multicolumn{1}{c|}{money} & \\  \hline
&attention & 127 & 230 & 4 & 361 \\
&share & 134 & 364 & 2 & 500 \\  
&money & 320 & 124 & 808 & 1252 \\ \hline
&       & 581 & 718 & 814 & 2113 \\ \\
\multicolumn{6}{c}{EM - 1299 correct} \\ \\
\multicolumn{2}{c}{} &
\multicolumn{3}{|c|}{Discovered} \\
\multicolumn{1}{c}{} &
\multicolumn{1}{c|}{Actual} &
\multicolumn{1}{|c}{attention} &
\multicolumn{1}{c}{share} &
\multicolumn{1}{c|}{money} & \\  \hline
&attention & 152 & 205 & 4 & 361 \\
&share & 134 & 364 & 2 & 500 \\
&money & 297 & 94 & 861 & 1252 \\ \hline
&       & 583 & 663 & 867 & 2113 \\ \\
\multicolumn{6}{c}{Gibbs - 1377 correct} \\ \\
\end{tabular}
\caption{interest - Feature Set B}
\label{fig:interestb}
\end{center}
\vskip -0.14in
\end{figure}

\begin{figure}
\begin{center}
\begin{tabular}{cr|cc|c}
\multicolumn{2}{c}{} &
\multicolumn{2}{|c|}{Discovered} \\
\multicolumn{1}{c}{} &
\multicolumn{1}{c|}{Actual} &
\multicolumn{1}{|c}{assist} & 
\multicolumn{1}{c|}{enhance} & \\  \hline
&assist & 119 & 160  & 279 \\
&enhance & 344  & 644 & 988 \\ \hline
&         & 463 & 804 & 1267 \\ \\
\multicolumn{5}{c}{EM - 763 correct} \\ \\
\multicolumn{2}{c}{} &
\multicolumn{2}{|c|}{Discovered} \\
\multicolumn{1}{c}{} &
\multicolumn{1}{c|}{Actual} &
\multicolumn{1}{|c}{assist} &
\multicolumn{1}{c|}{enhance} & \\  \hline
&assist & 169 & 110  & 279 \\
&enhance & 276  & 712 & 988 \\ \hline
&         & 445 & 822 & 1267 \\ \\
\multicolumn{5}{c}{Gibbs - 881 correct} \\ \\
\end{tabular}
\caption{help - Feature Set C}
\label{fig:helpc}
\end{center}
\vskip -0.14in
\end{figure}

The fact that the EM algorithm and Gibbs Sampling often arrive at
similar results suggests that a combination of the these methods might
be appropriate for this data. It is proposed in \cite{MengV97} that
the Gibbs Sampler be initialized with the parameters that the EM
algorithm converges upon rather than with randomly selected values. 
If the EM algorithm has found a local maxima then  the Gibbs Sampler
can escape it and find the global maximum. However, if the EM algorithm
has already found the global maximum  then the Gibbs Sampler will
converge quickly and confirm this result.       

\subsection{Feature Set Comparison}
\label{sec:feature1}

While there is little variation between the EM algorithm and Gibbs
Sampling given a particular word and feature set,  there are
differences in the accuracy attained for each method as they are used
with different feature sets. In general, variation in the accuracy of
a method when using different feature sets suggests that certain types
of features are more or less appropriate for particular words. 
Table \ref{tab:emgibbs} shows the maximum accuracy for each
word in parenthesis. Any accuracies that are not significantly less
than this are underlined.  

There are a number of cases where the same method attains very
different levels of accuracy when used with different feature sets. 
For example, the accuracies of the EM algorithm and Gibbs
Sampling for {\it bill} and {\it include} are much higher with feature
set B than with A or C. Less extreme examples of the same behavior are
shown by {\it agree} and {\it close}. However, the accuracies for {\it
drug} and {\it help} are much lower with feature set B than with A or
C. A less extreme example is {\it chief}. 

The separation of behavior between feature sets A and C and feature
set B is due to the nature of the features in these sets. A and C both
include part--of--speech features while feature set B does not. It
appears that the usefulness of part--of--speech features for
disambiguation varies considerably from word to word.  The fact that 3
of 4 verbs perform at higher levels of accuracy with feature set B
suggests that part--of--speech features may not be helpful when
disambiguating verbs.   

When using either the EM algorithm or Gibbs Sampling, {\it line}, {\it 
interest}, and {\it last} result in very similar disambiguation accuracy 
regardless of the feature set. In fact, for {\it interest} and
{\it line} there are no significant differences among any combination
of method and feature set.  This lack of variation shows that the
different feature sets are not able to make sufficient distinctions
among all words. It  may also point to limitations in Naive
Bayes. While it performs well in general, there may be certain words
and feature sets for which the assumptions it makes are not
appropriate.    

The performance of {\it concern} is slightly unusual; it disambiguates
most accurately with feature sets A and B. However, the only feature
in common between sets A and B is the morphological feature; it seems
unlikely that this accounts for the high disambiguation accuracy achieved using
both sets. A more likely explanation is that the part--of--speech
features from set A somewhat duplicate the information contained
in the unrestricted co--occurrences of set B. Unrestricted co--occurrences
can be largely dominated by non--content words that provide more
syntactic rather than semantic information. This same explanation may apply in
cases such as {\it line} and {\it interest} where the results for
feature sets A, B, and C are very similar. Feature sets A and C
rely heavily upon part--of--speech features which largely convey
syntactic information. If the co--occurrences in feature set B are
also essentially representing syntactic information, then similar
performance across all feature sets is possible.  

The highest average accuracy achieved for adjectives occurs when Gibbs
Sampling is used in combination with feature set C. The adjectives
have the most skewed sense distributions and set C has the largest
dimensionality of the feature sets. Given this combination of
circumstances,   it appears that the EM algorithm gets trapped at
local maxima for {\it  common} and {\it public}; Gibbs Sampling finds
a global maximum and results in significantly better accuracy in these
cases.     

Feature set B appears to be well suited for {\it bill} but fares poorly 
with {\it drug}. Otherwise there  is not a clear pattern as to which 
feature set is most accurate for the nouns.  The relatively similar
behavior across the feature sets suggests that certain features are
either essentially duplicating one another or that there are features
included in these sets that are simply not useful for the
disambiguation of nouns. While it is clear that the part--of--speech
and co--occurrence features are contributing to disambiguation
accuracy, it is less certain  that the morphological and collocation
features make significant contributions.  

\section{Analysis 2: Agglomerative Clustering}
\label{sec:anal2}

The second analysis of this experiment compares the accuracy of two
agglomerative clustering algorithms, Ward's minimum variance method and
McQuitty's similarity analysis.  Table \ref{tab:agglom} shows the
average accuracy and standard deviation of disambiguation over 25
random trials for each combination of word, method,  and feature
set. The repeated trials are necessary to determine the impact of
randomly breaking ties during clustering. As in the first analysis,
both  methodological and feature set comparisons are presented. 

\begin{table}
\begin{center}
\caption{Unsupervised Accuracy of Agglomerative Clustering}
\label{tab:agglom}
\begin{footnotesize}
\begin{tabular}{|l|c|c|c||c|c||c|c|}
\hline
\multicolumn{1}{|l}{} & 
\multicolumn{1}{|l|}{} & 
\multicolumn{2}{|c||}{Feature Set A} &
\multicolumn{2}{|c||}{Feature Set B} &
\multicolumn{2}{|c|}{Feature Set C} \\ \cline{3-8}
&Maj. & \footnotesize{McQuitty} & Ward & \footnotesize{McQuitty} & Ward &
\footnotesize{McQuitty} & Ward \\ \hline
chief  & .861 & {\bf \underline{.844 .05}} & .721 .01 &  {\bf
\underline{.831 .06}} & .611 .01 &
{\bf \underline{(.856 .00)}} & .673 .03 \\ 
common & .842 & .648 .12 & .513 .08 &  {\bf \underline{.797 .04}} & .444
.04 & {\bf \underline{(.799 .06)}} & .561 .05 \\
last   & .940 & \underline{(.791 .12)} & .598 .09 &  .541 .11 & .659 .03 &
.636 .07 &
.601 .08 \\
public & .683 & .560 .08 & .450 .05 &  {\bf .\underline{558 .07}} & .461
.03 &
{\bf \underline{(.628 .05)}} & .488 .04 \\ \hline 
\multicolumn{1}{|l|}{adjectives} &
\multicolumn{1}{|l|}{.832} &
\multicolumn{1}{|l|}{.711} &
\multicolumn{1}{|l||}{.571} &
\multicolumn{1}{|l|}{.682} &
\multicolumn{1}{|l||}{.544} &
\multicolumn{1}{|l|}{.730} &
\multicolumn{1}{|l|}{.581} \\ \hline \hline
bill   & .681 & .669 .08 & .647 .11 &  {\bf \underline{(.753 .05)}} & .600
.04 & .561
.10 & .515 .04 \\
\hline
\end{tabular} 
\end{footnotesize}
\end{center}
\vskip -0.01in
\end{table}

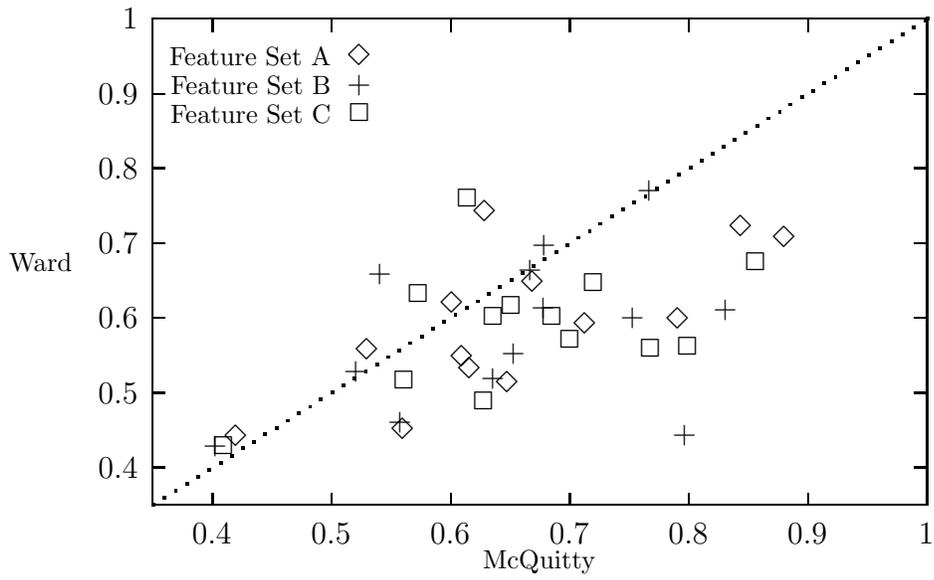
\begin{figure}
\begin{center}
\setlength{\unitlength}{0.240900pt}
\ifx\plotpoint\undefined\newsavebox{\plotpoint}\fi
\sbox{\plotpoint}{\rule[-0.200pt]{0.400pt}{0.400pt}}%
\begin{picture}(1500,900)(0,0)
\font\gnuplot=cmr10 at 10pt
\gnuplot
\sbox{\plotpoint}{\rule[-0.200pt]{0.400pt}{0.400pt}}%
\put(220.0,172.0){\rule[-0.200pt]{4.818pt}{0.400pt}}
\put(198,172){\makebox(0,0)[r]{$0.4$}}
\put(1416.0,172.0){\rule[-0.200pt]{4.818pt}{0.400pt}}
\put(220.0,289.0){\rule[-0.200pt]{4.818pt}{0.400pt}}
\put(198,289){\makebox(0,0)[r]{$0.5$}}
\put(1416.0,289.0){\rule[-0.200pt]{4.818pt}{0.400pt}}
\put(220.0,407.0){\rule[-0.200pt]{4.818pt}{0.400pt}}
\put(198,407){\makebox(0,0)[r]{$0.6$}}
\put(1416.0,407.0){\rule[-0.200pt]{4.818pt}{0.400pt}}
\put(220.0,524.0){\rule[-0.200pt]{4.818pt}{0.400pt}}
\put(198,524){\makebox(0,0)[r]{$0.7$}}
\put(1416.0,524.0){\rule[-0.200pt]{4.818pt}{0.400pt}}
\put(220.0,642.0){\rule[-0.200pt]{4.818pt}{0.400pt}}
\put(198,642){\makebox(0,0)[r]{$0.8$}}
\put(1416.0,642.0){\rule[-0.200pt]{4.818pt}{0.400pt}}
\put(220.0,759.0){\rule[-0.200pt]{4.818pt}{0.400pt}}
\put(198,759){\makebox(0,0)[r]{$0.9$}}
\put(1416.0,759.0){\rule[-0.200pt]{4.818pt}{0.400pt}}
\put(220.0,877.0){\rule[-0.200pt]{4.818pt}{0.400pt}}
\put(198,877){\makebox(0,0)[r]{$1$}}
\put(1416.0,877.0){\rule[-0.200pt]{4.818pt}{0.400pt}}
\put(314.0,113.0){\rule[-0.200pt]{0.400pt}{4.818pt}}
\put(314,68){\makebox(0,0){$0.4$}}
\put(314.0,857.0){\rule[-0.200pt]{0.400pt}{4.818pt}}
\put(501.0,113.0){\rule[-0.200pt]{0.400pt}{4.818pt}}
\put(501,68){\makebox(0,0){$0.5$}}
\put(501.0,857.0){\rule[-0.200pt]{0.400pt}{4.818pt}}
\put(688.0,113.0){\rule[-0.200pt]{0.400pt}{4.818pt}}
\put(688,68){\makebox(0,0){$0.6$}}
\put(688.0,857.0){\rule[-0.200pt]{0.400pt}{4.818pt}}
\put(875.0,113.0){\rule[-0.200pt]{0.400pt}{4.818pt}}
\put(875,68){\makebox(0,0){$0.7$}}
\put(875.0,857.0){\rule[-0.200pt]{0.400pt}{4.818pt}}
\put(1062.0,113.0){\rule[-0.200pt]{0.400pt}{4.818pt}}
\put(1062,68){\makebox(0,0){$0.8$}}
\put(1062.0,857.0){\rule[-0.200pt]{0.400pt}{4.818pt}}
\put(1249.0,113.0){\rule[-0.200pt]{0.400pt}{4.818pt}}
\put(1249,68){\makebox(0,0){$0.9$}}
\put(1249.0,857.0){\rule[-0.200pt]{0.400pt}{4.818pt}}
\put(1436.0,113.0){\rule[-0.200pt]{0.400pt}{4.818pt}}
\put(1436,68){\makebox(0,0){$1$}}
\put(1436.0,857.0){\rule[-0.200pt]{0.400pt}{4.818pt}}
\put(220.0,113.0){\rule[-0.200pt]{292.934pt}{0.400pt}}
\put(1436.0,113.0){\rule[-0.200pt]{0.400pt}{184.048pt}}
\put(220.0,877.0){\rule[-0.200pt]{292.934pt}{0.400pt}}
\put(45,495){\makebox(0,0){Ward}}
\put(828,23){\makebox(0,0){McQuitty}}
\put(220.0,113.0){\rule[-0.200pt]{0.400pt}{184.048pt}}
\put(501,818){\makebox(0,0)[r]{Feature Set A}}
\put(545,818){\raisebox{-.8pt}{\makebox(0,0){$\Diamond$}}}
\put(1144,549){\raisebox{-.8pt}{\makebox(0,0){$\Diamond$}}}
\put(777,305){\raisebox{-.8pt}{\makebox(0,0){$\Diamond$}}}
\put(1045,404){\raisebox{-.8pt}{\makebox(0,0){$\Diamond$}}}
\put(613,231){\raisebox{-.8pt}{\makebox(0,0){$\Diamond$}}}
\put(817,462){\raisebox{-.8pt}{\makebox(0,0){$\Diamond$}}}
\put(742,573){\raisebox{-.8pt}{\makebox(0,0){$\Diamond$}}}
\put(557,356){\raisebox{-.8pt}{\makebox(0,0){$\Diamond$}}}
\put(690,429){\raisebox{-.8pt}{\makebox(0,0){$\Diamond$}}}
\put(351,220){\raisebox{-.8pt}{\makebox(0,0){$\Diamond$}}}
\put(706,345){\raisebox{-.8pt}{\makebox(0,0){$\Diamond$}}}
\put(718,326){\raisebox{-.8pt}{\makebox(0,0){$\Diamond$}}}
\put(899,396){\raisebox{-.8pt}{\makebox(0,0){$\Diamond$}}}
\put(1212,533){\raisebox{-.8pt}{\makebox(0,0){$\Diamond$}}}
\put(501,773){\makebox(0,0)[r]{Feature Set B}}
\put(545,773){\makebox(0,0){$+$}}
\put(1120,420){\makebox(0,0){$+$}}
\put(1056,223){\makebox(0,0){$+$}}
\put(577,476){\makebox(0,0){$+$}}
\put(609,243){\makebox(0,0){$+$}}
\put(974,407){\makebox(0,0){$+$}}
\put(835,521){\makebox(0,0){$+$}}
\put(540,322){\makebox(0,0){$+$}}
\put(787,350){\makebox(0,0){$+$}}
\put(319,205){\makebox(0,0){$+$}}
\put(834,422){\makebox(0,0){$+$}}
\put(813,482){\makebox(0,0){$+$}}
\put(755,312){\makebox(0,0){$+$}}
\put(1000,607){\makebox(0,0){$+$}}
\put(501,728){\makebox(0,0)[r]{Feature Set C}}
\put(545,728){\raisebox{-.8pt}{\makebox(0,0){$\Box$}}}
\put(1167,493){\raisebox{-.8pt}{\makebox(0,0){$\Box$}}}
\put(1060,361){\raisebox{-.8pt}{\makebox(0,0){$\Box$}}}
\put(755,408){\raisebox{-.8pt}{\makebox(0,0){$\Box$}}}
\put(740,275){\raisebox{-.8pt}{\makebox(0,0){$\Box$}}}
\put(615,307){\raisebox{-.8pt}{\makebox(0,0){$\Box$}}}
\put(714,593){\raisebox{-.8pt}{\makebox(0,0){$\Box$}}}
\put(637,444){\raisebox{-.8pt}{\makebox(0,0){$\Box$}}}
\put(783,424){\raisebox{-.8pt}{\makebox(0,0){$\Box$}}}
\put(332,204){\raisebox{-.8pt}{\makebox(0,0){$\Box$}}}
\put(847,408){\raisebox{-.8pt}{\makebox(0,0){$\Box$}}}
\put(912,460){\raisebox{-.8pt}{\makebox(0,0){$\Box$}}}
\put(875,372){\raisebox{-.8pt}{\makebox(0,0){$\Box$}}}
\put(1002,357){\raisebox{-.8pt}{\makebox(0,0){$\Box$}}}
\sbox{\plotpoint}{\rule[-0.500pt]{1.000pt}{1.000pt}}%
\put(220,113){\usebox{\plotpoint}}
\put(220.00,113.00){\usebox{\plotpoint}}
\put(237.58,124.00){\usebox{\plotpoint}}
\put(255.25,134.84){\usebox{\plotpoint}}
\multiput(257,136)(17.270,11.513){0}{\usebox{\plotpoint}}
\put(272.52,146.35){\usebox{\plotpoint}}
\put(290.31,157.01){\usebox{\plotpoint}}
\multiput(294,159)(17.270,11.513){0}{\usebox{\plotpoint}}
\put(307.78,168.19){\usebox{\plotpoint}}
\put(325.46,179.02){\usebox{\plotpoint}}
\multiput(331,182)(17.270,11.513){0}{\usebox{\plotpoint}}
\put(343.03,190.02){\usebox{\plotpoint}}
\put(360.30,201.53){\usebox{\plotpoint}}
\put(378.19,212.02){\usebox{\plotpoint}}
\multiput(380,213)(17.270,11.513){0}{\usebox{\plotpoint}}
\put(395.56,223.37){\usebox{\plotpoint}}
\put(413.34,234.03){\usebox{\plotpoint}}
\multiput(417,236)(17.270,11.513){0}{\usebox{\plotpoint}}
\put(430.81,245.21){\usebox{\plotpoint}}
\put(448.08,256.72){\usebox{\plotpoint}}
\multiput(453,260)(18.275,9.840){0}{\usebox{\plotpoint}}
\put(466.06,267.04){\usebox{\plotpoint}}
\put(483.33,278.56){\usebox{\plotpoint}}
\put(501.22,289.04){\usebox{\plotpoint}}
\multiput(503,290)(17.270,11.513){0}{\usebox{\plotpoint}}
\put(518.59,300.39){\usebox{\plotpoint}}
\put(535.86,311.91){\usebox{\plotpoint}}
\multiput(539,314)(18.275,9.840){0}{\usebox{\plotpoint}}
\put(553.84,322.23){\usebox{\plotpoint}}
\put(571.11,333.74){\usebox{\plotpoint}}
\multiput(576,337)(17.270,11.513){0}{\usebox{\plotpoint}}
\put(588.40,345.22){\usebox{\plotpoint}}
\put(606.37,355.58){\usebox{\plotpoint}}
\put(623.64,367.09){\usebox{\plotpoint}}
\multiput(625,368)(18.275,9.840){0}{\usebox{\plotpoint}}
\put(641.62,377.41){\usebox{\plotpoint}}
\put(658.89,388.93){\usebox{\plotpoint}}
\multiput(662,391)(17.270,11.513){0}{\usebox{\plotpoint}}
\put(676.28,400.23){\usebox{\plotpoint}}
\put(694.14,410.76){\usebox{\plotpoint}}
\multiput(699,414)(17.270,11.513){0}{\usebox{\plotpoint}}
\put(711.44,422.24){\usebox{\plotpoint}}
\put(729.40,432.60){\usebox{\plotpoint}}
\put(746.67,444.11){\usebox{\plotpoint}}
\multiput(748,445)(17.270,11.513){0}{\usebox{\plotpoint}}
\put(764.17,455.24){\usebox{\plotpoint}}
\put(781.92,465.95){\usebox{\plotpoint}}
\multiput(785,468)(17.270,11.513){0}{\usebox{\plotpoint}}
\put(799.32,477.25){\usebox{\plotpoint}}
\put(817.18,487.78){\usebox{\plotpoint}}
\multiput(822,491)(17.270,11.513){0}{\usebox{\plotpoint}}
\put(834.45,499.30){\usebox{\plotpoint}}
\put(852.05,510.26){\usebox{\plotpoint}}
\put(869.70,521.13){\usebox{\plotpoint}}
\multiput(871,522)(17.270,11.513){0}{\usebox{\plotpoint}}
\put(887.20,532.26){\usebox{\plotpoint}}
\put(904.95,542.97){\usebox{\plotpoint}}
\multiput(908,545)(17.270,11.513){0}{\usebox{\plotpoint}}
\put(922.22,554.48){\usebox{\plotpoint}}
\put(939.93,565.27){\usebox{\plotpoint}}
\multiput(945,568)(17.270,11.513){0}{\usebox{\plotpoint}}
\put(957.48,576.32){\usebox{\plotpoint}}
\put(975.08,587.27){\usebox{\plotpoint}}
\put(992.73,598.15){\usebox{\plotpoint}}
\multiput(994,599)(17.270,11.513){0}{\usebox{\plotpoint}}
\put(1010.00,609.67){\usebox{\plotpoint}}
\put(1027.81,620.28){\usebox{\plotpoint}}
\multiput(1031,622)(17.270,11.513){0}{\usebox{\plotpoint}}
\put(1045.26,631.50){\usebox{\plotpoint}}
\put(1062.96,642.29){\usebox{\plotpoint}}
\multiput(1068,645)(17.270,11.513){0}{\usebox{\plotpoint}}
\put(1080.51,653.34){\usebox{\plotpoint}}
\put(1097.78,664.85){\usebox{\plotpoint}}
\put(1115.69,675.30){\usebox{\plotpoint}}
\multiput(1117,676)(17.270,11.513){0}{\usebox{\plotpoint}}
\put(1133.03,686.69){\usebox{\plotpoint}}
\put(1150.30,698.20){\usebox{\plotpoint}}
\multiput(1153,700)(18.275,9.840){0}{\usebox{\plotpoint}}
\put(1168.29,708.53){\usebox{\plotpoint}}
\put(1185.56,720.04){\usebox{\plotpoint}}
\multiput(1190,723)(18.275,9.840){0}{\usebox{\plotpoint}}
\put(1203.54,730.36){\usebox{\plotpoint}}
\put(1220.81,741.87){\usebox{\plotpoint}}
\put(1238.08,753.39){\usebox{\plotpoint}}
\multiput(1239,754)(18.275,9.840){0}{\usebox{\plotpoint}}
\put(1256.07,763.71){\usebox{\plotpoint}}
\put(1273.34,775.22){\usebox{\plotpoint}}
\multiput(1276,777)(18.275,9.840){0}{\usebox{\plotpoint}}
\put(1291.32,785.55){\usebox{\plotpoint}}
\put(1308.59,797.06){\usebox{\plotpoint}}
\multiput(1313,800)(17.270,11.513){0}{\usebox{\plotpoint}}
\put(1325.91,808.49){\usebox{\plotpoint}}
\put(1343.84,818.90){\usebox{\plotpoint}}
\put(1361.11,830.41){\usebox{\plotpoint}}
\multiput(1362,831)(18.275,9.840){0}{\usebox{\plotpoint}}
\put(1379.10,840.73){\usebox{\plotpoint}}
\put(1396.37,852.24){\usebox{\plotpoint}}
\multiput(1399,854)(17.270,11.513){0}{\usebox{\plotpoint}}
\put(1413.79,863.50){\usebox{\plotpoint}}
\put(1431.62,874.08){\usebox{\plotpoint}}
\put(1436,877){\usebox{\plotpoint}}
\end{picture}
\caption{Agglomerative Clustering Correlation of Accuracy for all words}
\label{fig:accagglom}
\end{center}
\vskip -0.14in
\end{figure}

\begin{figure}
\begin{center}
\setlength{\unitlength}{0.240900pt}
\ifx\plotpoint\undefined\newsavebox{\plotpoint}\fi
\sbox{\plotpoint}{\rule[-0.200pt]{0.400pt}{0.400pt}}%
\begin{picture}(1500,900)(0,0)
\font\gnuplot=cmr10 at 10pt
\gnuplot
\sbox{\plotpoint}{\rule[-0.200pt]{0.400pt}{0.400pt}}%
\put(220.0,172.0){\rule[-0.200pt]{4.818pt}{0.400pt}}
\put(198,172){\makebox(0,0)[r]{$0.4$}}
\put(1416.0,172.0){\rule[-0.200pt]{4.818pt}{0.400pt}}
\put(220.0,289.0){\rule[-0.200pt]{4.818pt}{0.400pt}}
\put(198,289){\makebox(0,0)[r]{$0.5$}}
\put(1416.0,289.0){\rule[-0.200pt]{4.818pt}{0.400pt}}
\put(220.0,407.0){\rule[-0.200pt]{4.818pt}{0.400pt}}
\put(198,407){\makebox(0,0)[r]{$0.6$}}
\put(1416.0,407.0){\rule[-0.200pt]{4.818pt}{0.400pt}}
\put(220.0,524.0){\rule[-0.200pt]{4.818pt}{0.400pt}}
\put(198,524){\makebox(0,0)[r]{$0.7$}}
\put(1416.0,524.0){\rule[-0.200pt]{4.818pt}{0.400pt}}
\put(220.0,642.0){\rule[-0.200pt]{4.818pt}{0.400pt}}
\put(198,642){\makebox(0,0)[r]{$0.8$}}
\put(1416.0,642.0){\rule[-0.200pt]{4.818pt}{0.400pt}}
\put(220.0,759.0){\rule[-0.200pt]{4.818pt}{0.400pt}}
\put(198,759){\makebox(0,0)[r]{$0.9$}}
\put(1416.0,759.0){\rule[-0.200pt]{4.818pt}{0.400pt}}
\put(220.0,877.0){\rule[-0.200pt]{4.818pt}{0.400pt}}
\put(198,877){\makebox(0,0)[r]{$1$}}
\put(1416.0,877.0){\rule[-0.200pt]{4.818pt}{0.400pt}}
\put(314.0,113.0){\rule[-0.200pt]{0.400pt}{4.818pt}}
\put(314,68){\makebox(0,0){$0.4$}}
\put(314.0,857.0){\rule[-0.200pt]{0.400pt}{4.818pt}}
\put(501.0,113.0){\rule[-0.200pt]{0.400pt}{4.818pt}}
\put(501,68){\makebox(0,0){$0.5$}}
\put(501.0,857.0){\rule[-0.200pt]{0.400pt}{4.818pt}}
\put(688.0,113.0){\rule[-0.200pt]{0.400pt}{4.818pt}}
\put(688,68){\makebox(0,0){$0.6$}}
\put(688.0,857.0){\rule[-0.200pt]{0.400pt}{4.818pt}}
\put(875.0,113.0){\rule[-0.200pt]{0.400pt}{4.818pt}}
\put(875,68){\makebox(0,0){$0.7$}}
\put(875.0,857.0){\rule[-0.200pt]{0.400pt}{4.818pt}}
\put(1062.0,113.0){\rule[-0.200pt]{0.400pt}{4.818pt}}
\put(1062,68){\makebox(0,0){$0.8$}}
\put(1062.0,857.0){\rule[-0.200pt]{0.400pt}{4.818pt}}
\put(1249.0,113.0){\rule[-0.200pt]{0.400pt}{4.818pt}}
\put(1249,68){\makebox(0,0){$0.9$}}
\put(1249.0,857.0){\rule[-0.200pt]{0.400pt}{4.818pt}}
\put(1436.0,113.0){\rule[-0.200pt]{0.400pt}{4.818pt}}
\put(1436,68){\makebox(0,0){$1$}}
\put(1436.0,857.0){\rule[-0.200pt]{0.400pt}{4.818pt}}
\put(220.0,113.0){\rule[-0.200pt]{292.934pt}{0.400pt}}
\put(1436.0,113.0){\rule[-0.200pt]{0.400pt}{184.048pt}}
\put(220.0,877.0){\rule[-0.200pt]{292.934pt}{0.400pt}}
\put(45,495){\makebox(0,0){Ward}}
\put(828,23){\makebox(0,0){McQuitty}}
\put(220.0,113.0){\rule[-0.200pt]{0.400pt}{184.048pt}}
\put(501,818){\makebox(0,0)[r]{Feature Set A}}
\put(545,818){\raisebox{-.8pt}{\makebox(0,0){$\Diamond$}}}
\put(817,462){\raisebox{-.8pt}{\makebox(0,0){$\Diamond$}}}
\put(742,573){\raisebox{-.8pt}{\makebox(0,0){$\Diamond$}}}
\put(557,356){\raisebox{-.8pt}{\makebox(0,0){$\Diamond$}}}
\put(690,429){\raisebox{-.8pt}{\makebox(0,0){$\Diamond$}}}
\put(351,220){\raisebox{-.8pt}{\makebox(0,0){$\Diamond$}}}
\put(501,773){\makebox(0,0)[r]{Feature Set B}}
\put(545,773){\makebox(0,0){$+$}}
\put(974,407){\makebox(0,0){$+$}}
\put(835,521){\makebox(0,0){$+$}}
\put(540,322){\makebox(0,0){$+$}}
\put(787,350){\makebox(0,0){$+$}}
\put(319,205){\makebox(0,0){$+$}}
\put(501,728){\makebox(0,0)[r]{Feature Set C}}
\put(545,728){\raisebox{-.8pt}{\makebox(0,0){$\Box$}}}
\put(615,307){\raisebox{-.8pt}{\makebox(0,0){$\Box$}}}
\put(714,593){\raisebox{-.8pt}{\makebox(0,0){$\Box$}}}
\put(637,444){\raisebox{-.8pt}{\makebox(0,0){$\Box$}}}
\put(783,424){\raisebox{-.8pt}{\makebox(0,0){$\Box$}}}
\put(332,204){\raisebox{-.8pt}{\makebox(0,0){$\Box$}}}
\sbox{\plotpoint}{\rule[-0.500pt]{1.000pt}{1.000pt}}%
\put(220,113){\usebox{\plotpoint}}
\put(220.00,113.00){\usebox{\plotpoint}}
\put(237.58,124.00){\usebox{\plotpoint}}
\put(255.25,134.84){\usebox{\plotpoint}}
\multiput(257,136)(17.270,11.513){0}{\usebox{\plotpoint}}
\put(272.52,146.35){\usebox{\plotpoint}}
\put(290.31,157.01){\usebox{\plotpoint}}
\multiput(294,159)(17.270,11.513){0}{\usebox{\plotpoint}}
\put(307.78,168.19){\usebox{\plotpoint}}
\put(325.46,179.02){\usebox{\plotpoint}}
\multiput(331,182)(17.270,11.513){0}{\usebox{\plotpoint}}
\put(343.03,190.02){\usebox{\plotpoint}}
\put(360.30,201.53){\usebox{\plotpoint}}
\put(378.19,212.02){\usebox{\plotpoint}}
\multiput(380,213)(17.270,11.513){0}{\usebox{\plotpoint}}
\put(395.56,223.37){\usebox{\plotpoint}}
\put(413.34,234.03){\usebox{\plotpoint}}
\multiput(417,236)(17.270,11.513){0}{\usebox{\plotpoint}}
\put(430.81,245.21){\usebox{\plotpoint}}
\put(448.08,256.72){\usebox{\plotpoint}}
\multiput(453,260)(18.275,9.840){0}{\usebox{\plotpoint}}
\put(466.06,267.04){\usebox{\plotpoint}}
\put(483.33,278.56){\usebox{\plotpoint}}
\put(501.22,289.04){\usebox{\plotpoint}}
\multiput(503,290)(17.270,11.513){0}{\usebox{\plotpoint}}
\put(518.59,300.39){\usebox{\plotpoint}}
\put(535.86,311.91){\usebox{\plotpoint}}
\multiput(539,314)(18.275,9.840){0}{\usebox{\plotpoint}}
\put(553.84,322.23){\usebox{\plotpoint}}
\put(571.11,333.74){\usebox{\plotpoint}}
\multiput(576,337)(17.270,11.513){0}{\usebox{\plotpoint}}
\put(588.40,345.22){\usebox{\plotpoint}}
\put(606.37,355.58){\usebox{\plotpoint}}
\put(623.64,367.09){\usebox{\plotpoint}}
\multiput(625,368)(18.275,9.840){0}{\usebox{\plotpoint}}
\put(641.62,377.41){\usebox{\plotpoint}}
\put(658.89,388.93){\usebox{\plotpoint}}
\multiput(662,391)(17.270,11.513){0}{\usebox{\plotpoint}}
\put(676.28,400.23){\usebox{\plotpoint}}
\put(694.14,410.76){\usebox{\plotpoint}}
\multiput(699,414)(17.270,11.513){0}{\usebox{\plotpoint}}
\put(711.44,422.24){\usebox{\plotpoint}}
\put(729.40,432.60){\usebox{\plotpoint}}
\put(746.67,444.11){\usebox{\plotpoint}}
\multiput(748,445)(17.270,11.513){0}{\usebox{\plotpoint}}
\put(764.17,455.24){\usebox{\plotpoint}}
\put(781.92,465.95){\usebox{\plotpoint}}
\multiput(785,468)(17.270,11.513){0}{\usebox{\plotpoint}}
\put(799.32,477.25){\usebox{\plotpoint}}
\put(817.18,487.78){\usebox{\plotpoint}}
\multiput(822,491)(17.270,11.513){0}{\usebox{\plotpoint}}
\put(834.45,499.30){\usebox{\plotpoint}}
\put(852.05,510.26){\usebox{\plotpoint}}
\put(869.70,521.13){\usebox{\plotpoint}}
\multiput(871,522)(17.270,11.513){0}{\usebox{\plotpoint}}
\put(887.20,532.26){\usebox{\plotpoint}}
\put(904.95,542.97){\usebox{\plotpoint}}
\multiput(908,545)(17.270,11.513){0}{\usebox{\plotpoint}}
\put(922.22,554.48){\usebox{\plotpoint}}
\put(939.93,565.27){\usebox{\plotpoint}}
\multiput(945,568)(17.270,11.513){0}{\usebox{\plotpoint}}
\put(957.48,576.32){\usebox{\plotpoint}}
\put(975.08,587.27){\usebox{\plotpoint}}
\put(992.73,598.15){\usebox{\plotpoint}}
\multiput(994,599)(17.270,11.513){0}{\usebox{\plotpoint}}
\put(1010.00,609.67){\usebox{\plotpoint}}
\put(1027.81,620.28){\usebox{\plotpoint}}
\multiput(1031,622)(17.270,11.513){0}{\usebox{\plotpoint}}
\put(1045.26,631.50){\usebox{\plotpoint}}
\put(1062.96,642.29){\usebox{\plotpoint}}
\multiput(1068,645)(17.270,11.513){0}{\usebox{\plotpoint}}
\put(1080.51,653.34){\usebox{\plotpoint}}
\put(1097.78,664.85){\usebox{\plotpoint}}
\put(1115.69,675.30){\usebox{\plotpoint}}
\multiput(1117,676)(17.270,11.513){0}{\usebox{\plotpoint}}
\put(1133.03,686.69){\usebox{\plotpoint}}
\put(1150.30,698.20){\usebox{\plotpoint}}
\multiput(1153,700)(18.275,9.840){0}{\usebox{\plotpoint}}
\put(1168.29,708.53){\usebox{\plotpoint}}
\put(1185.56,720.04){\usebox{\plotpoint}}
\multiput(1190,723)(18.275,9.840){0}{\usebox{\plotpoint}}
\put(1203.54,730.36){\usebox{\plotpoint}}
\put(1220.81,741.87){\usebox{\plotpoint}}
\put(1238.08,753.39){\usebox{\plotpoint}}
\multiput(1239,754)(18.275,9.840){0}{\usebox{\plotpoint}}
\put(1256.07,763.71){\usebox{\plotpoint}}
\put(1273.34,775.22){\usebox{\plotpoint}}
\multiput(1276,777)(18.275,9.840){0}{\usebox{\plotpoint}}
\put(1291.32,785.55){\usebox{\plotpoint}}
\put(1308.59,797.06){\usebox{\plotpoint}}
\multiput(1313,800)(17.270,11.513){0}{\usebox{\plotpoint}}
\put(1325.91,808.49){\usebox{\plotpoint}}
\put(1343.84,818.90){\usebox{\plotpoint}}
\put(1361.11,830.41){\usebox{\plotpoint}}
\multiput(1362,831)(18.275,9.840){0}{\usebox{\plotpoint}}
\put(1379.10,840.73){\usebox{\plotpoint}}
\put(1396.37,852.24){\usebox{\plotpoint}}
\multiput(1399,854)(17.270,11.513){0}{\usebox{\plotpoint}}
\put(1413.79,863.50){\usebox{\plotpoint}}
\put(1431.62,874.08){\usebox{\plotpoint}}
\put(1436,877){\usebox{\plotpoint}}
\end{picture}
\caption{Agglomerative Clustering Correlation of Accuracy for Nouns}
\label{fig:accagglom-noun}
\end{center}
\vskip -0.14in
\end{figure}
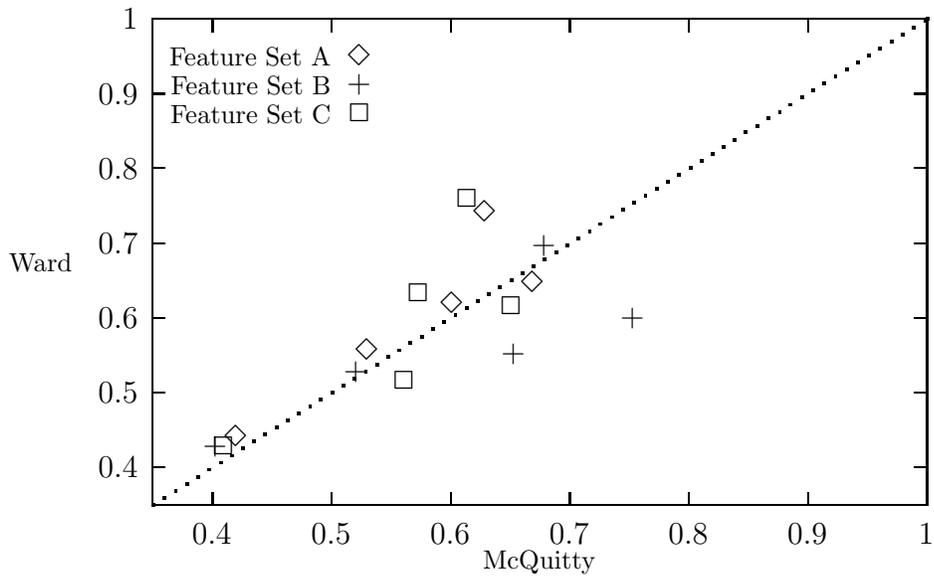

\subsection{Methodological Comparison}

Ward's and McQuitty's methods are both agglomerative clustering
algorithms and differ only in the distance measure each uses to
determine if an instance of an ambiguous word belongs in a particular
sense group. Since distance is not implicit in the features used in
this experiment, the data representing the instances of an ambiguous
word must be converted into a form where distance can be measured. In
this dissertation that representation is a dissimilarity matrix. 

Ward's method is based on  a classical measure, Euclidean distance,
while McQuitty's method employs a simple count of the number of
dissimilar features to establish group membership. The dramatic 
difference in the nature of these distance measures motivates their 
inclusion in this study. 

Unlike the probabilistic models, there are significant differences
in the accuracy of the two agglomerative clustering algorithms given a
particular feature set. Of the 39 possible pairwise comparisons, 17
result in significant differences. There are 2 cases where Ward's
method is significantly more accurate and 15 favoring McQuitty's
similarity analysis. The significant differences  are shown in bold
face in Table \ref{tab:agglom}. 

The plot of the correlation of accuracy between Ward's and
McQuitty's methods in  Figure \ref{fig:accagglom} shows 
that McQuitty's  method generally is the more accurate. However, 
in Figure \ref{fig:accagglom-noun} the correlation  plot is restricted
to the nouns and Ward's method is shown to be slightly  
more accurate. This is also illustrated in Table \ref{tab:agglom},
where there  are only three significant differences among the nouns; 
in two of those cases Ward's method is  the most accurate. Thus,
it is only for verbs and adjectives that McQuitty's  method shows a
decisive advantage.    

As is the case with the probabilistic models, only the nouns are
consistently disambiguated with accuracy greater than the majority sense. 
However, McQuitty's method achieves accuracy comparable to the majority 
sense for a few of the adjectives and verbs when the standard deviation is
taken into account. This occurs for  {\it chief} with all feature sets, 
{\it common} for sets B and C, {\it close} for set C, and {\it include} 
for set A.  

There appears to be  significant relationships among the actual sense
distributions of the ambiguous words, the inherent biases of the
agglomerative clustering algorithms, and the accuracy attained by each
method. 
Ward's method has a well--known bias towards finding balanced
distributions of sense groups \cite{Ward63}. However, McQuitty's
method has no such bias since there are no underlying parametric models
or distributional assumptions that influence the
algorithm.\footnote{In this regard McQuitty's method is unique among
the four unsupervised approaches discussed in this
dissertation. Recall that both the EM algorithm and Gibbs Sampling
also tend to find balanced distributions of senses.} 
The tendencies of both agglomerative methods are 
illustrated in Figures \ref{fig:concernaa}, \ref{fig:interestba}, and
\ref{fig:helpca}. These show the confusion matrices for the same word
and feature set combinations that are discussed in the first analysis.
These illustrate the bias of Ward's method towards the discovery of
balanced sense distributions while also showing that McQuitty's
similarity analysis tends to find more skewed distributions.  

The bias of Ward's method towards balanced distributions of senses
results in accurate disambiguation of the nouns but also leads to
rather poor performance with adjectives and verbs.  As the actual
distribution of senses grows more skewed, Ward's method becomes less
accurate. By contrast,  McQuitty's method performs fairly well with
words that have skewed distributions of senses. It has no bias
regarding the distribution of senses it discovers and is able
to learn very unbalanced distributions.  

\begin{figure}
\begin{center}
\begin{tabular}{cr|cc|c}
\multicolumn{2}{c}{} &
\multicolumn{2}{|c|}{Discovered} \\
\multicolumn{1}{c}{} &
\multicolumn{1}{c|}{Actual} &
\multicolumn{1}{|c}{worry} &
\multicolumn{1}{c|}{business} & \\  \hline
&worry & 166 & 281  & 447 \\
&business & 181  & 607 & 788 \\ \hline
&         & 347 & 888 & 1235 \\ \\
\multicolumn{5}{c}{McQuitty - 773 correct} \\ \\
\multicolumn{2}{c}{} &
\multicolumn{2}{|c|}{Discovered} \\
\multicolumn{1}{c}{} &
\multicolumn{1}{c|}{Actual} &
\multicolumn{1}{|c}{worry} &
\multicolumn{1}{c|}{business} & \\  \hline
&worry & 288 & 159  & 447 \\
&business & 155  & 633 & 788 \\ \hline
&         & 443 & 792 & 1235 \\ \\
\multicolumn{5}{c}{Ward - 921 correct} \\ \\
\end{tabular} 
\caption{concern - Feature Set A}
\label{fig:concernaa}
\end{center}
\vskip -0.14in
\end{figure}

\begin{figure}
\begin{center}
\begin{tabular}{cr|ccc|c}
\multicolumn{2}{c}{} &
\multicolumn{3}{|c|}{Discovered} \\
\multicolumn{1}{c}{} &
\multicolumn{1}{c|}{Actual} &     
\multicolumn{1}{|c}{attention} &
\multicolumn{1}{c}{share} &
\multicolumn{1}{c|}{money} & \\  \hline
&attention & 53 & 6 & 302 & 361 \\
&share & 58 & 187 & 255 & 500 \\  
&money & 108 & 4 & 1140 & 1252 \\ \hline
&       & 219 & 197 & 1697 & 2113 \\ \\
\multicolumn{6}{c}{McQuitty - 1380 correct} \\  \\
\multicolumn{2}{c}{} &
\multicolumn{3}{|c|}{Discovered} \\
\multicolumn{1}{c}{} &
\multicolumn{1}{c|}{Actual} &
\multicolumn{1}{|c}{attention} &
\multicolumn{1}{c}{share} &     
\multicolumn{1}{c|}{money} & \\  \hline
&attention & 280 & 3 & 78 & 361 \\
&share & 240 & 197 & 63 & 500 \\
&money & 559 & 0 & 693 & 1252 \\ \hline
&       & 1079 & 200 & 834 & 2113 \\ \\
\multicolumn{6}{c}{Ward - 1170 correct} \\ \\
\end{tabular}
\caption{interest - Feature Set B}
\label{fig:interestba}
\end{center}
\vskip -0.14in
\end{figure}

\begin{figure}
\begin{center}
\begin{tabular}{cr|cc|c}
\multicolumn{2}{c}{} &
\multicolumn{2}{|c|}{Discovered} \\
\multicolumn{1}{c}{} &
\multicolumn{1}{c|}{Actual} &   
\multicolumn{1}{|c}{assist} &
\multicolumn{1}{c|}{enhance} & \\  \hline
&assist & 45 & 234  & 279 \\
&enhance & 146  & 842 & 988 \\ \hline
&         & 191 & 1076 & 1267 \\ \\
\multicolumn{5}{c}{McQuitty - 887 correct} \\ \\
\multicolumn{2}{c}{} &
\multicolumn{2}{|c|}{Discovered} \\
\multicolumn{1}{c}{} &
\multicolumn{1}{c|}{Actual} &
\multicolumn{1}{|c}{assist} &
\multicolumn{1}{c|}{enhance} & \\  \hline
&assist & 88 & 191  & 279 \\
&enhance & 354  & 634 & 988 \\ \hline
&         & 442 & 825 & 1267 \\ \\ 
\multicolumn{5}{c}{Ward - 722 correct} \\ \\
\end{tabular}
\caption{help - Feature Set C}
\label{fig:helpca}
\end{center}
\vskip -0.14in
\end{figure}

The standard deviations in Table \ref{tab:agglom} measure the impact
of randomly breaking ties during the clustering process. 
A standard deviation of .00 indicates that no ties occurred 
during clustering; in this case the agglomerative algorithm is
deterministic and the sense groups discovered from trial to trial are
identical. As clustering becomes more influenced by random
breaking of ties, the standard deviation will increase since the 
sense groups created will vary from trial to trial. 

Overall, the standard deviation for McQuitty's similarity analysis
is greater than that of Ward's method. This is not surprising given
the simplicity of McQuitty's approach; distances are based on a
count of the dissimilar features between two instances
of an ambiguous word.  Ties are common given these sets since they
contain relatively small numbers of features; A has 8, B has
5, and C has 7.    

However, Ward's method computes Euclidean distances in $n$--space. 
While this more detailed measure results in fewer ties,
there are still enough to cause relatively large amounts of deviation
for some words. This suggests that the conversion of raw text into a
dissimilarity matrix representation results in a reduction in
the discriminating power of the feature set to the point where ties
are still common.

In general, these standard deviations suggest that the feature sets
need to be expanded to provide more distinctions between instances of
an ambiguous word when using the agglomerative clustering algorithms.     

\subsection{Feature Set Comparison}
\label{sec:feature2}

When using probabilistic models, feature sets A and C result in
similar performance and often have an inverse relationship to the
accuracy attained with feature set B. For example, if set B results in
high accuracy then A and C may not. When feature set A and C 
result in high accuracy then set B often does not. These patterns hold
for 7 of 13 words when disambiguating with  probabilistic models.
This suggests that the features common to sets A and C, the
part--of--speech of surrounding words, are a main contributor to
disambiguation accuracy when using probabilistic models.  

However, when an agglomerative approach is employed these patterns are
much less pronounced. The only cases where results from set A and set
C are more accurate than those from set B are when McQuitty's 
method is used to disambiguate {\it help} and when Ward's method is
used for {\it concern}. The only case where feature set B is most
accurate is when  McQuitty's method is used to disambiguate {\it
bill}. 

This change in behavior suggests that the data representation employed
by the agglomerative methods blurs some distinctions that are present
in the frequency count data used by the probabilistic models. 
While both the probabilistic models and agglomerative methods use the
same feature sets, the agglomerative methods convert the data into a
dissimilarity matrix representation that shows how many
features differ between instances of an ambiguous word; however, it
makes no distinctions as to which features differ. Thus the
part--of--speech distinctions that appear to have significant impact
on disambiguation accuracy when using probabilistic models are not
distinguishable from other features in the dissimilarity matrix
representation.   

Overall the combination of McQuitty's similarity analysis with
feature set C results in consistently high accuracy for the adjectives
and verbs. Feature set C has the highest dimensionality of the
three feature sets. Thus, the number of dissimilar features between
two instances of an ambiguous word will likely be fairly high most of
the time since there are a large number of possible values for the
features. This results in the creation of a large sense group
where all members have fairly high dissimilarity counts.  The
distribution of discovered senses in this case will be skewed and
likely correspond fairly well with the actual distributions.  

The only cases where McQuitty's method and feature set C fares poorly
for the verbs and adjectives are for {\it last} and {\it include}.
These are interesting exceptions in that these two words
have the largest majority senses, .94\% and .91\% respectively. They
are both most accurately disambiguated with feature set A; no other
combination of feature set and method results in comparable
performance. Agglomerative clustering based on dissimilarity counts of
the features in set A is particularly effective with these words. This
suggests that the combination of a low dimensional feature set with a
word that has an extremely skewed distribution of senses may be
appropriate.  

There is not a clear pattern as to which feature sets lead to accurate
disambiguation of nouns. For {\it concern}, Ward's method in
conjunction with feature sets A and C achieves the highest accuracy.
Ward's method is also most accurate for {\it drug} when used with
feature set C. The success of Ward's method for these two nouns is
related to the general tendency of Ward's method to find balanced
distributions of senses. As is the case in the probabilistic models,
{\it interest} and {\it line} do not show great variation from one
feature set to the next. This again suggests that the dissimilarity
matrix may be reducing the granularity of the information available to
the clustering algorithm by reducing the distinctions that the
feature sets are able to represent.  

\section{Analysis 3: Gibbs Sampling and McQuitty's Similarity Analysis }

The first analysis in Section \ref{sec:anal1} shows that Gibbs
Sampling offers some improvement over the EM algorithm, particularly
for adjectives and verbs. The second analysis in Section
\ref{sec:anal2}  shows that McQuitty's similarity analysis is often
more accurate than Ward's method; primarily when disambiguating
adjectives and verbs.  This final analysis compares McQuitty's method
and Gibbs Sampling. This section only contains a methodological
comparison since  the feature set comparisons for McQuitty's method
and Gibbs Sampling are included in the first two analyses.  

Table \ref{tab:gibbsmcq} reformats the accuracies reported in the
previous two analyses for easy comparison. As before, significant
differences between the two methods for a particular word and feature
set are shown in bold face. The maximum accuracy for a word is in
parenthesis and any accuracies that are not significantly less than
this maximum are underlined. 

There are a relatively large number of significant differences between
Gibbs Sampling and McQuitty's similarity analysis given a particular
feature set. Of the 39 pairwise comparisons, 19 show significant 
differences. Gibbs Sampling is more accurate in 10 of those cases and 
McQuitty's method is more accurate in 9. These significant differences 
are shown in  bold face in Table \ref{tab:gibbsmcq}.  

\begin{table}
\begin{center}
\caption{Unsupervised Accuracy of Gibbs and McQuitty's}
\label{tab:gibbsmcq}
\begin{footnotesize}
\begin{tabular}{|l|c|c|c||c|c||c|c|} \hline
\multicolumn{1}{|l}{} & 
\multicolumn{1}{|l|}{} & 
\multicolumn{2}{|c||}{Feature Set A} & 
\multicolumn{2}{|c||}{Feature Set B} &
\multicolumn{2}{|c|}{Feature Set C} \\ \cline{3-8} 
& Maj.  &  Gibbs  & \footnotesize{McQuitty}  & Gibbs   &
\footnotesize{McQuitty} & Gibbs & \footnotesize{McQuitty} \\ \hline
chief    &.861   & .719 .01 & {\bf \underline{.844 .05}} & .648 .00 & 
{\bf \underline{.831 .06}} &
.728 .04 & {\bf \underline{(.856 .00)}} \\ 
common   &.842   & .522 .00 & {\bf .648 .12} & .507 .07 & 
{\bf \underline{.797 .04}} & .670 .11 & {\bf \underline{(.799 .06)}} \\
last     &.940   & {\bf.900 .00} & .791 .12 & 
{\bf \underline{(.912 .00)}} 
& .541 .11 & {\bf .908 .00} & .636 .07 \\
public   &.683   & .514 .00 & .560 .08 & .478 .04 & .558 .07 & .578 .00 &
{\bf \underline{(.628 .05)}} \\ \hline 
\multicolumn{1}{|l|}{adjectives} &
\multicolumn{1}{|l|}{.832} &
\multicolumn{1}{|l|}{.663} &
\multicolumn{1}{|l||}{.711} &
\multicolumn{1}{|l|}{.636} &
\multicolumn{1}{|l||}{.682} &
\multicolumn{1}{|l|}{.721} &
\multicolumn{1}{|l|}{.730} \\ \hline \hline
bill     & .681  & .590 .04 & .669 .08 & \underline{.705 .10} &
\underline{(.753 .05)} & .592 .04 & .561 .10 \\
\hline
\end{tabular} 
\end{footnotesize}
\end{center}
\vskip -0.01in
\end{table}

The correlation of accuracy between Gibbs Sampling and McQuitty's method  
is shown in Figure \ref{fig:accboth}. Since there is not a clear
pattern associated with the performance of the methods,  this data is
broken down into separate correlation plots for adjectives, nouns, and
verbs in Figures \ref{fig:accboth-adj}, \ref{fig:accboth-noun},  and
\ref{fig:accboth-verb}.   

Figure \ref{fig:accboth-adj} shows that McQuitty's method is generally
more accurate for adjectives, the exception being {\it last} for all three
feature sets. Figure \ref{fig:accboth-noun} suggests that Gibbs Sampling 
is more accurate for the nouns. Figure \ref{fig:accboth-verb} shows that
McQuitty's method is generally more accurate for the verbs, although
not so dramatically as is the case with the adjectives.  

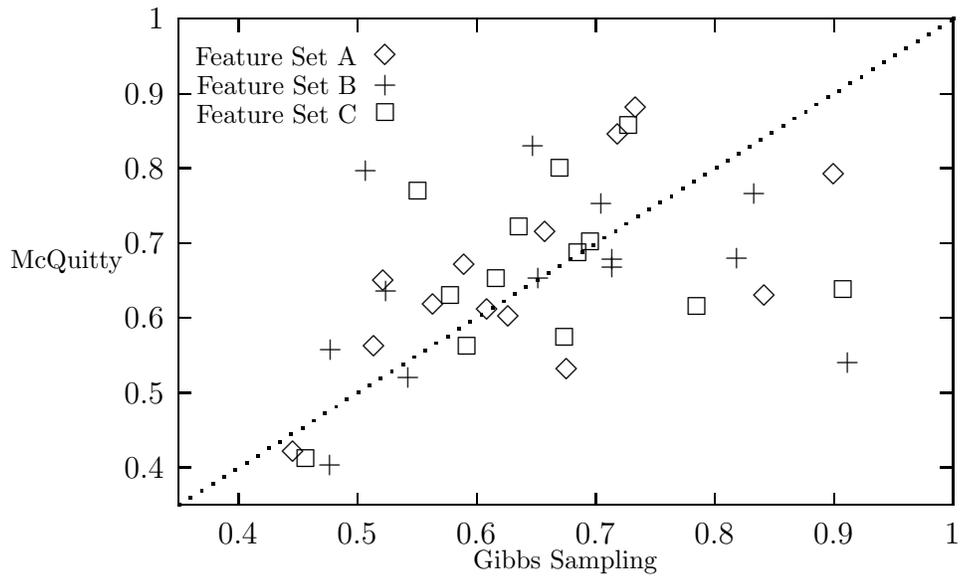
\begin{figure}
\begin{center}
\setlength{\unitlength}{0.240900pt}
\ifx\plotpoint\undefined\newsavebox{\plotpoint}\fi
\sbox{\plotpoint}{\rule[-0.200pt]{0.400pt}{0.400pt}}%
\begin{picture}(1500,900)(0,0)
\font\gnuplot=cmr10 at 10pt
\gnuplot
\sbox{\plotpoint}{\rule[-0.200pt]{0.400pt}{0.400pt}}%
\put(220.0,172.0){\rule[-0.200pt]{4.818pt}{0.400pt}}
\put(198,172){\makebox(0,0)[r]{$0.4$}}
\put(1416.0,172.0){\rule[-0.200pt]{4.818pt}{0.400pt}}
\put(220.0,289.0){\rule[-0.200pt]{4.818pt}{0.400pt}}
\put(198,289){\makebox(0,0)[r]{$0.5$}}
\put(1416.0,289.0){\rule[-0.200pt]{4.818pt}{0.400pt}}
\put(220.0,407.0){\rule[-0.200pt]{4.818pt}{0.400pt}}
\put(198,407){\makebox(0,0)[r]{$0.6$}}
\put(1416.0,407.0){\rule[-0.200pt]{4.818pt}{0.400pt}}
\put(220.0,524.0){\rule[-0.200pt]{4.818pt}{0.400pt}}
\put(198,524){\makebox(0,0)[r]{$0.7$}}
\put(1416.0,524.0){\rule[-0.200pt]{4.818pt}{0.400pt}}
\put(220.0,642.0){\rule[-0.200pt]{4.818pt}{0.400pt}}
\put(198,642){\makebox(0,0)[r]{$0.8$}}
\put(1416.0,642.0){\rule[-0.200pt]{4.818pt}{0.400pt}}
\put(220.0,759.0){\rule[-0.200pt]{4.818pt}{0.400pt}}
\put(198,759){\makebox(0,0)[r]{$0.9$}}
\put(1416.0,759.0){\rule[-0.200pt]{4.818pt}{0.400pt}}
\put(220.0,877.0){\rule[-0.200pt]{4.818pt}{0.400pt}}
\put(198,877){\makebox(0,0)[r]{$1$}}
\put(1416.0,877.0){\rule[-0.200pt]{4.818pt}{0.400pt}}
\put(314.0,113.0){\rule[-0.200pt]{0.400pt}{4.818pt}}
\put(314,68){\makebox(0,0){$0.4$}}
\put(314.0,857.0){\rule[-0.200pt]{0.400pt}{4.818pt}}
\put(501.0,113.0){\rule[-0.200pt]{0.400pt}{4.818pt}}
\put(501,68){\makebox(0,0){$0.5$}}
\put(501.0,857.0){\rule[-0.200pt]{0.400pt}{4.818pt}}
\put(688.0,113.0){\rule[-0.200pt]{0.400pt}{4.818pt}}
\put(688,68){\makebox(0,0){$0.6$}}
\put(688.0,857.0){\rule[-0.200pt]{0.400pt}{4.818pt}}
\put(875.0,113.0){\rule[-0.200pt]{0.400pt}{4.818pt}}
\put(875,68){\makebox(0,0){$0.7$}}
\put(875.0,857.0){\rule[-0.200pt]{0.400pt}{4.818pt}}
\put(1062.0,113.0){\rule[-0.200pt]{0.400pt}{4.818pt}}
\put(1062,68){\makebox(0,0){$0.8$}}
\put(1062.0,857.0){\rule[-0.200pt]{0.400pt}{4.818pt}}
\put(1249.0,113.0){\rule[-0.200pt]{0.400pt}{4.818pt}}
\put(1249,68){\makebox(0,0){$0.9$}}
\put(1249.0,857.0){\rule[-0.200pt]{0.400pt}{4.818pt}}
\put(1436.0,113.0){\rule[-0.200pt]{0.400pt}{4.818pt}}
\put(1436,68){\makebox(0,0){$1$}}
\put(1436.0,857.0){\rule[-0.200pt]{0.400pt}{4.818pt}}
\put(220.0,113.0){\rule[-0.200pt]{292.934pt}{0.400pt}}
\put(1436.0,113.0){\rule[-0.200pt]{0.400pt}{184.048pt}}
\put(220.0,877.0){\rule[-0.200pt]{292.934pt}{0.400pt}}
\put(45,495){\makebox(0,0){McQuitty}}
\put(828,23){\makebox(0,0){Gibbs Sampling}}
\put(220.0,113.0){\rule[-0.200pt]{0.400pt}{184.048pt}}
\put(501,818){\makebox(0,0)[r]{Feature Set A}}
\put(545,818){\raisebox{-.8pt}{\makebox(0,0){$\Diamond$}}}
\put(910,694){\raisebox{-.8pt}{\makebox(0,0){$\Diamond$}}}
\put(542,463){\raisebox{-.8pt}{\makebox(0,0){$\Diamond$}}}
\put(1249,631){\raisebox{-.8pt}{\makebox(0,0){$\Diamond$}}}
\put(527,360){\raisebox{-.8pt}{\makebox(0,0){$\Diamond$}}}
\put(669,488){\raisebox{-.8pt}{\makebox(0,0){$\Diamond$}}}
\put(1140,441){\raisebox{-.8pt}{\makebox(0,0){$\Diamond$}}}
\put(830,325){\raisebox{-.8pt}{\makebox(0,0){$\Diamond$}}}
\put(738,408){\raisebox{-.8pt}{\makebox(0,0){$\Diamond$}}}
\put(400,195){\raisebox{-.8pt}{\makebox(0,0){$\Diamond$}}}
\put(705,419){\raisebox{-.8pt}{\makebox(0,0){$\Diamond$}}}
\put(620,426){\raisebox{-.8pt}{\makebox(0,0){$\Diamond$}}}
\put(796,540){\raisebox{-.8pt}{\makebox(0,0){$\Diamond$}}}
\put(938,736){\raisebox{-.8pt}{\makebox(0,0){$\Diamond$}}}
\put(501,773){\makebox(0,0)[r]{Feature Set B}}
\put(545,773){\makebox(0,0){$+$}}
\put(777,678){\makebox(0,0){$+$}}
\put(514,638){\makebox(0,0){$+$}}
\put(1271,337){\makebox(0,0){$+$}}
\put(459,357){\makebox(0,0){$+$}}
\put(884,587){\makebox(0,0){$+$}}
\put(1097,500){\makebox(0,0){$+$}}
\put(581,314){\makebox(0,0){$+$}}
\put(785,469){\makebox(0,0){$+$}}
\put(458,175){\makebox(0,0){$+$}}
\put(901,499){\makebox(0,0){$+$}}
\put(901,486){\makebox(0,0){$+$}}
\put(546,449){\makebox(0,0){$+$}}
\put(1124,603){\makebox(0,0){$+$}}
\put(501,728){\makebox(0,0)[r]{Feature Set C}}
\put(545,728){\raisebox{-.8pt}{\makebox(0,0){$\Box$}}}
\put(927,708){\raisebox{-.8pt}{\makebox(0,0){$\Box$}}}
\put(819,641){\raisebox{-.8pt}{\makebox(0,0){$\Box$}}}
\put(1264,449){\raisebox{-.8pt}{\makebox(0,0){$\Box$}}}
\put(647,440){\raisebox{-.8pt}{\makebox(0,0){$\Box$}}}
\put(673,361){\raisebox{-.8pt}{\makebox(0,0){$\Box$}}}
\put(1034,423){\raisebox{-.8pt}{\makebox(0,0){$\Box$}}}
\put(826,375){\raisebox{-.8pt}{\makebox(0,0){$\Box$}}}
\put(719,467){\raisebox{-.8pt}{\makebox(0,0){$\Box$}}}
\put(420,184){\raisebox{-.8pt}{\makebox(0,0){$\Box$}}}
\put(847,507){\raisebox{-.8pt}{\makebox(0,0){$\Box$}}}
\put(755,548){\raisebox{-.8pt}{\makebox(0,0){$\Box$}}}
\put(867,524){\raisebox{-.8pt}{\makebox(0,0){$\Box$}}}
\put(596,604){\raisebox{-.8pt}{\makebox(0,0){$\Box$}}}
\sbox{\plotpoint}{\rule[-0.500pt]{1.000pt}{1.000pt}}%
\put(220,113){\usebox{\plotpoint}}
\put(220.00,113.00){\usebox{\plotpoint}}
\put(237.58,124.00){\usebox{\plotpoint}}
\put(255.25,134.84){\usebox{\plotpoint}}
\multiput(257,136)(17.270,11.513){0}{\usebox{\plotpoint}}
\put(272.52,146.35){\usebox{\plotpoint}}
\put(290.31,157.01){\usebox{\plotpoint}}
\multiput(294,159)(17.270,11.513){0}{\usebox{\plotpoint}}
\put(307.78,168.19){\usebox{\plotpoint}}
\put(325.46,179.02){\usebox{\plotpoint}}
\multiput(331,182)(17.270,11.513){0}{\usebox{\plotpoint}}
\put(343.03,190.02){\usebox{\plotpoint}}
\put(360.30,201.53){\usebox{\plotpoint}}
\put(378.19,212.02){\usebox{\plotpoint}}
\multiput(380,213)(17.270,11.513){0}{\usebox{\plotpoint}}
\put(395.56,223.37){\usebox{\plotpoint}}
\put(413.34,234.03){\usebox{\plotpoint}}
\multiput(417,236)(17.270,11.513){0}{\usebox{\plotpoint}}
\put(430.81,245.21){\usebox{\plotpoint}}
\put(448.08,256.72){\usebox{\plotpoint}}
\multiput(453,260)(18.275,9.840){0}{\usebox{\plotpoint}}
\put(466.06,267.04){\usebox{\plotpoint}}
\put(483.33,278.56){\usebox{\plotpoint}}
\put(501.22,289.04){\usebox{\plotpoint}}
\multiput(503,290)(17.270,11.513){0}{\usebox{\plotpoint}}
\put(518.59,300.39){\usebox{\plotpoint}}
\put(535.86,311.91){\usebox{\plotpoint}}
\multiput(539,314)(18.275,9.840){0}{\usebox{\plotpoint}}
\put(553.84,322.23){\usebox{\plotpoint}}
\put(571.11,333.74){\usebox{\plotpoint}}
\multiput(576,337)(17.270,11.513){0}{\usebox{\plotpoint}}
\put(588.40,345.22){\usebox{\plotpoint}}
\put(606.37,355.58){\usebox{\plotpoint}}
\put(623.64,367.09){\usebox{\plotpoint}}
\multiput(625,368)(18.275,9.840){0}{\usebox{\plotpoint}}
\put(641.62,377.41){\usebox{\plotpoint}}
\put(658.89,388.93){\usebox{\plotpoint}}
\multiput(662,391)(17.270,11.513){0}{\usebox{\plotpoint}}
\put(676.28,400.23){\usebox{\plotpoint}}
\put(694.14,410.76){\usebox{\plotpoint}}
\multiput(699,414)(17.270,11.513){0}{\usebox{\plotpoint}}
\put(711.44,422.24){\usebox{\plotpoint}}
\put(729.40,432.60){\usebox{\plotpoint}}
\put(746.67,444.11){\usebox{\plotpoint}}
\multiput(748,445)(17.270,11.513){0}{\usebox{\plotpoint}}
\put(764.17,455.24){\usebox{\plotpoint}}
\put(781.92,465.95){\usebox{\plotpoint}}
\multiput(785,468)(17.270,11.513){0}{\usebox{\plotpoint}}
\put(799.32,477.25){\usebox{\plotpoint}}
\put(817.18,487.78){\usebox{\plotpoint}}
\multiput(822,491)(17.270,11.513){0}{\usebox{\plotpoint}}
\put(834.45,499.30){\usebox{\plotpoint}}
\put(852.05,510.26){\usebox{\plotpoint}}
\put(869.70,521.13){\usebox{\plotpoint}}
\multiput(871,522)(17.270,11.513){0}{\usebox{\plotpoint}}
\put(887.20,532.26){\usebox{\plotpoint}}
\put(904.95,542.97){\usebox{\plotpoint}}
\multiput(908,545)(17.270,11.513){0}{\usebox{\plotpoint}}
\put(922.22,554.48){\usebox{\plotpoint}}
\put(939.93,565.27){\usebox{\plotpoint}}
\multiput(945,568)(17.270,11.513){0}{\usebox{\plotpoint}}
\put(957.48,576.32){\usebox{\plotpoint}}
\put(975.08,587.27){\usebox{\plotpoint}}
\put(992.73,598.15){\usebox{\plotpoint}}
\multiput(994,599)(17.270,11.513){0}{\usebox{\plotpoint}}
\put(1010.00,609.67){\usebox{\plotpoint}}
\put(1027.81,620.28){\usebox{\plotpoint}}
\multiput(1031,622)(17.270,11.513){0}{\usebox{\plotpoint}}
\put(1045.26,631.50){\usebox{\plotpoint}}
\put(1062.96,642.29){\usebox{\plotpoint}}
\multiput(1068,645)(17.270,11.513){0}{\usebox{\plotpoint}}
\put(1080.51,653.34){\usebox{\plotpoint}}
\put(1097.78,664.85){\usebox{\plotpoint}}
\put(1115.69,675.30){\usebox{\plotpoint}}
\multiput(1117,676)(17.270,11.513){0}{\usebox{\plotpoint}}
\put(1133.03,686.69){\usebox{\plotpoint}}
\put(1150.30,698.20){\usebox{\plotpoint}}
\multiput(1153,700)(18.275,9.840){0}{\usebox{\plotpoint}}
\put(1168.29,708.53){\usebox{\plotpoint}}
\put(1185.56,720.04){\usebox{\plotpoint}}
\multiput(1190,723)(18.275,9.840){0}{\usebox{\plotpoint}}
\put(1203.54,730.36){\usebox{\plotpoint}}
\put(1220.81,741.87){\usebox{\plotpoint}}
\put(1238.08,753.39){\usebox{\plotpoint}}
\multiput(1239,754)(18.275,9.840){0}{\usebox{\plotpoint}}
\put(1256.07,763.71){\usebox{\plotpoint}}
\put(1273.34,775.22){\usebox{\plotpoint}}
\multiput(1276,777)(18.275,9.840){0}{\usebox{\plotpoint}}
\put(1291.32,785.55){\usebox{\plotpoint}}
\put(1308.59,797.06){\usebox{\plotpoint}}
\multiput(1313,800)(17.270,11.513){0}{\usebox{\plotpoint}}
\put(1325.91,808.49){\usebox{\plotpoint}}
\put(1343.84,818.90){\usebox{\plotpoint}}
\put(1361.11,830.41){\usebox{\plotpoint}}
\multiput(1362,831)(18.275,9.840){0}{\usebox{\plotpoint}}
\put(1379.10,840.73){\usebox{\plotpoint}}
\put(1396.37,852.24){\usebox{\plotpoint}}
\multiput(1399,854)(17.270,11.513){0}{\usebox{\plotpoint}}
\put(1413.79,863.50){\usebox{\plotpoint}}
\put(1431.62,874.08){\usebox{\plotpoint}}
\put(1436,877){\usebox{\plotpoint}}
\end{picture}
\caption{Gibbs and McQuitty's Correlation of Accuracy for all words}
\label{fig:accboth}
\end{center}
\vskip -0.14in
\end{figure}

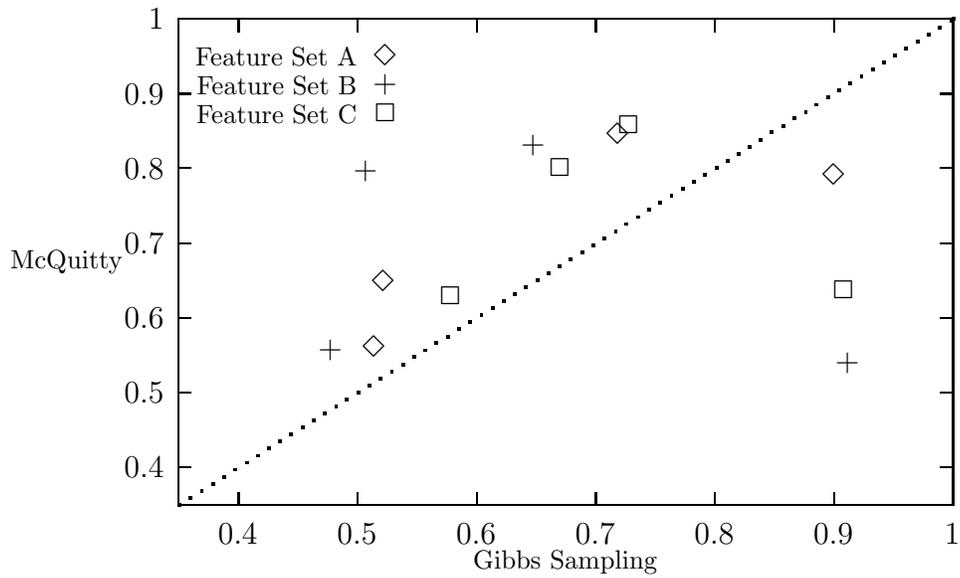
\begin{figure}
\begin{center}
\setlength{\unitlength}{0.240900pt}
\ifx\plotpoint\undefined\newsavebox{\plotpoint}\fi
\sbox{\plotpoint}{\rule[-0.200pt]{0.400pt}{0.400pt}}%
\begin{picture}(1500,900)(0,0)
\font\gnuplot=cmr10 at 10pt
\gnuplot
\sbox{\plotpoint}{\rule[-0.200pt]{0.400pt}{0.400pt}}%
\put(220.0,172.0){\rule[-0.200pt]{4.818pt}{0.400pt}}
\put(198,172){\makebox(0,0)[r]{$0.4$}}
\put(1416.0,172.0){\rule[-0.200pt]{4.818pt}{0.400pt}}
\put(220.0,289.0){\rule[-0.200pt]{4.818pt}{0.400pt}}
\put(198,289){\makebox(0,0)[r]{$0.5$}}
\put(1416.0,289.0){\rule[-0.200pt]{4.818pt}{0.400pt}}
\put(220.0,407.0){\rule[-0.200pt]{4.818pt}{0.400pt}}
\put(198,407){\makebox(0,0)[r]{$0.6$}}
\put(1416.0,407.0){\rule[-0.200pt]{4.818pt}{0.400pt}}
\put(220.0,524.0){\rule[-0.200pt]{4.818pt}{0.400pt}}
\put(198,524){\makebox(0,0)[r]{$0.7$}}
\put(1416.0,524.0){\rule[-0.200pt]{4.818pt}{0.400pt}}
\put(220.0,642.0){\rule[-0.200pt]{4.818pt}{0.400pt}}
\put(198,642){\makebox(0,0)[r]{$0.8$}}
\put(1416.0,642.0){\rule[-0.200pt]{4.818pt}{0.400pt}}
\put(220.0,759.0){\rule[-0.200pt]{4.818pt}{0.400pt}}
\put(198,759){\makebox(0,0)[r]{$0.9$}}
\put(1416.0,759.0){\rule[-0.200pt]{4.818pt}{0.400pt}}
\put(220.0,877.0){\rule[-0.200pt]{4.818pt}{0.400pt}}
\put(198,877){\makebox(0,0)[r]{$1$}}
\put(1416.0,877.0){\rule[-0.200pt]{4.818pt}{0.400pt}}
\put(314.0,113.0){\rule[-0.200pt]{0.400pt}{4.818pt}}
\put(314,68){\makebox(0,0){$0.4$}}
\put(314.0,857.0){\rule[-0.200pt]{0.400pt}{4.818pt}}
\put(501.0,113.0){\rule[-0.200pt]{0.400pt}{4.818pt}}
\put(501,68){\makebox(0,0){$0.5$}}
\put(501.0,857.0){\rule[-0.200pt]{0.400pt}{4.818pt}}
\put(688.0,113.0){\rule[-0.200pt]{0.400pt}{4.818pt}}
\put(688,68){\makebox(0,0){$0.6$}}
\put(688.0,857.0){\rule[-0.200pt]{0.400pt}{4.818pt}}
\put(875.0,113.0){\rule[-0.200pt]{0.400pt}{4.818pt}}
\put(875,68){\makebox(0,0){$0.7$}}
\put(875.0,857.0){\rule[-0.200pt]{0.400pt}{4.818pt}}
\put(1062.0,113.0){\rule[-0.200pt]{0.400pt}{4.818pt}}
\put(1062,68){\makebox(0,0){$0.8$}}
\put(1062.0,857.0){\rule[-0.200pt]{0.400pt}{4.818pt}}
\put(1249.0,113.0){\rule[-0.200pt]{0.400pt}{4.818pt}}
\put(1249,68){\makebox(0,0){$0.9$}}
\put(1249.0,857.0){\rule[-0.200pt]{0.400pt}{4.818pt}}
\put(1436.0,113.0){\rule[-0.200pt]{0.400pt}{4.818pt}}
\put(1436,68){\makebox(0,0){$1$}}
\put(1436.0,857.0){\rule[-0.200pt]{0.400pt}{4.818pt}}
\put(220.0,113.0){\rule[-0.200pt]{292.934pt}{0.400pt}}
\put(1436.0,113.0){\rule[-0.200pt]{0.400pt}{184.048pt}}
\put(220.0,877.0){\rule[-0.200pt]{292.934pt}{0.400pt}}
\put(45,495){\makebox(0,0){McQuitty}}
\put(828,23){\makebox(0,0){Gibbs Sampling}}
\put(220.0,113.0){\rule[-0.200pt]{0.400pt}{184.048pt}}
\put(501,818){\makebox(0,0)[r]{Feature Set A}}
\put(545,818){\raisebox{-.8pt}{\makebox(0,0){$\Diamond$}}}
\put(910,694){\raisebox{-.8pt}{\makebox(0,0){$\Diamond$}}}
\put(542,463){\raisebox{-.8pt}{\makebox(0,0){$\Diamond$}}}
\put(1249,631){\raisebox{-.8pt}{\makebox(0,0){$\Diamond$}}}
\put(527,360){\raisebox{-.8pt}{\makebox(0,0){$\Diamond$}}}
\put(501,773){\makebox(0,0)[r]{Feature Set B}}
\put(545,773){\makebox(0,0){$+$}}
\put(777,678){\makebox(0,0){$+$}}
\put(514,638){\makebox(0,0){$+$}}
\put(1271,337){\makebox(0,0){$+$}}
\put(459,357){\makebox(0,0){$+$}}
\put(501,728){\makebox(0,0)[r]{Feature Set C}}
\put(545,728){\raisebox{-.8pt}{\makebox(0,0){$\Box$}}}
\put(927,708){\raisebox{-.8pt}{\makebox(0,0){$\Box$}}}
\put(819,641){\raisebox{-.8pt}{\makebox(0,0){$\Box$}}}
\put(1264,449){\raisebox{-.8pt}{\makebox(0,0){$\Box$}}}
\put(647,440){\raisebox{-.8pt}{\makebox(0,0){$\Box$}}}
\sbox{\plotpoint}{\rule[-0.500pt]{1.000pt}{1.000pt}}%
\put(220,113){\usebox{\plotpoint}}
\put(220.00,113.00){\usebox{\plotpoint}}
\put(237.58,124.00){\usebox{\plotpoint}}
\put(255.25,134.84){\usebox{\plotpoint}}
\multiput(257,136)(17.270,11.513){0}{\usebox{\plotpoint}}
\put(272.52,146.35){\usebox{\plotpoint}}
\put(290.31,157.01){\usebox{\plotpoint}}
\multiput(294,159)(17.270,11.513){0}{\usebox{\plotpoint}}
\put(307.78,168.19){\usebox{\plotpoint}}
\put(325.46,179.02){\usebox{\plotpoint}}
\multiput(331,182)(17.270,11.513){0}{\usebox{\plotpoint}}
\put(343.03,190.02){\usebox{\plotpoint}}
\put(360.30,201.53){\usebox{\plotpoint}}
\put(378.19,212.02){\usebox{\plotpoint}}
\multiput(380,213)(17.270,11.513){0}{\usebox{\plotpoint}}
\put(395.56,223.37){\usebox{\plotpoint}}
\put(413.34,234.03){\usebox{\plotpoint}}
\multiput(417,236)(17.270,11.513){0}{\usebox{\plotpoint}}
\put(430.81,245.21){\usebox{\plotpoint}}
\put(448.08,256.72){\usebox{\plotpoint}}
\multiput(453,260)(18.275,9.840){0}{\usebox{\plotpoint}}
\put(466.06,267.04){\usebox{\plotpoint}}
\put(483.33,278.56){\usebox{\plotpoint}}
\put(501.22,289.04){\usebox{\plotpoint}}
\multiput(503,290)(17.270,11.513){0}{\usebox{\plotpoint}}
\put(518.59,300.39){\usebox{\plotpoint}}
\put(535.86,311.91){\usebox{\plotpoint}}
\multiput(539,314)(18.275,9.840){0}{\usebox{\plotpoint}}
\put(553.84,322.23){\usebox{\plotpoint}}
\put(571.11,333.74){\usebox{\plotpoint}}
\multiput(576,337)(17.270,11.513){0}{\usebox{\plotpoint}}
\put(588.40,345.22){\usebox{\plotpoint}}
\put(606.37,355.58){\usebox{\plotpoint}}
\put(623.64,367.09){\usebox{\plotpoint}}
\multiput(625,368)(18.275,9.840){0}{\usebox{\plotpoint}}
\put(641.62,377.41){\usebox{\plotpoint}}
\put(658.89,388.93){\usebox{\plotpoint}}
\multiput(662,391)(17.270,11.513){0}{\usebox{\plotpoint}}
\put(676.28,400.23){\usebox{\plotpoint}}
\put(694.14,410.76){\usebox{\plotpoint}}
\multiput(699,414)(17.270,11.513){0}{\usebox{\plotpoint}}
\put(711.44,422.24){\usebox{\plotpoint}}
\put(729.40,432.60){\usebox{\plotpoint}}
\put(746.67,444.11){\usebox{\plotpoint}}
\multiput(748,445)(17.270,11.513){0}{\usebox{\plotpoint}}
\put(764.17,455.24){\usebox{\plotpoint}}
\put(781.92,465.95){\usebox{\plotpoint}}
\multiput(785,468)(17.270,11.513){0}{\usebox{\plotpoint}}
\put(799.32,477.25){\usebox{\plotpoint}}
\put(817.18,487.78){\usebox{\plotpoint}}
\multiput(822,491)(17.270,11.513){0}{\usebox{\plotpoint}}
\put(834.45,499.30){\usebox{\plotpoint}}
\put(852.05,510.26){\usebox{\plotpoint}}
\put(869.70,521.13){\usebox{\plotpoint}}
\multiput(871,522)(17.270,11.513){0}{\usebox{\plotpoint}}
\put(887.20,532.26){\usebox{\plotpoint}}
\put(904.95,542.97){\usebox{\plotpoint}}
\multiput(908,545)(17.270,11.513){0}{\usebox{\plotpoint}}
\put(922.22,554.48){\usebox{\plotpoint}}
\put(939.93,565.27){\usebox{\plotpoint}}
\multiput(945,568)(17.270,11.513){0}{\usebox{\plotpoint}}
\put(957.48,576.32){\usebox{\plotpoint}}
\put(975.08,587.27){\usebox{\plotpoint}}
\put(992.73,598.15){\usebox{\plotpoint}}
\multiput(994,599)(17.270,11.513){0}{\usebox{\plotpoint}}
\put(1010.00,609.67){\usebox{\plotpoint}}
\put(1027.81,620.28){\usebox{\plotpoint}}
\multiput(1031,622)(17.270,11.513){0}{\usebox{\plotpoint}}
\put(1045.26,631.50){\usebox{\plotpoint}}
\put(1062.96,642.29){\usebox{\plotpoint}}
\multiput(1068,645)(17.270,11.513){0}{\usebox{\plotpoint}}
\put(1080.51,653.34){\usebox{\plotpoint}}
\put(1097.78,664.85){\usebox{\plotpoint}}
\put(1115.69,675.30){\usebox{\plotpoint}}
\multiput(1117,676)(17.270,11.513){0}{\usebox{\plotpoint}}
\put(1133.03,686.69){\usebox{\plotpoint}}
\put(1150.30,698.20){\usebox{\plotpoint}}
\multiput(1153,700)(18.275,9.840){0}{\usebox{\plotpoint}}
\put(1168.29,708.53){\usebox{\plotpoint}}
\put(1185.56,720.04){\usebox{\plotpoint}}
\multiput(1190,723)(18.275,9.840){0}{\usebox{\plotpoint}}
\put(1203.54,730.36){\usebox{\plotpoint}}
\put(1220.81,741.87){\usebox{\plotpoint}}
\put(1238.08,753.39){\usebox{\plotpoint}}
\multiput(1239,754)(18.275,9.840){0}{\usebox{\plotpoint}}
\put(1256.07,763.71){\usebox{\plotpoint}}
\put(1273.34,775.22){\usebox{\plotpoint}}
\multiput(1276,777)(18.275,9.840){0}{\usebox{\plotpoint}}
\put(1291.32,785.55){\usebox{\plotpoint}}
\put(1308.59,797.06){\usebox{\plotpoint}}
\multiput(1313,800)(17.270,11.513){0}{\usebox{\plotpoint}}
\put(1325.91,808.49){\usebox{\plotpoint}}
\put(1343.84,818.90){\usebox{\plotpoint}}
\put(1361.11,830.41){\usebox{\plotpoint}}
\multiput(1362,831)(18.275,9.840){0}{\usebox{\plotpoint}}
\put(1379.10,840.73){\usebox{\plotpoint}}
\put(1396.37,852.24){\usebox{\plotpoint}}
\multiput(1399,854)(17.270,11.513){0}{\usebox{\plotpoint}}
\put(1413.79,863.50){\usebox{\plotpoint}}
\put(1431.62,874.08){\usebox{\plotpoint}}
\put(1436,877){\usebox{\plotpoint}}
\end{picture}
\caption{Gibbs and McQuitty's Correlation of Accuracy for Adjectives}
\label{fig:accboth-adj}
\end{center}
\vskip -0.14in
\end{figure}

\begin{figure}
\begin{center}
\setlength{\unitlength}{0.240900pt}
\ifx\plotpoint\undefined\newsavebox{\plotpoint}\fi
\sbox{\plotpoint}{\rule[-0.200pt]{0.400pt}{0.400pt}}%
\begin{picture}(1500,900)(0,0)
\font\gnuplot=cmr10 at 10pt
\gnuplot
\sbox{\plotpoint}{\rule[-0.200pt]{0.400pt}{0.400pt}}%
\put(220.0,172.0){\rule[-0.200pt]{4.818pt}{0.400pt}}
\put(198,172){\makebox(0,0)[r]{$0.4$}}
\put(1416.0,172.0){\rule[-0.200pt]{4.818pt}{0.400pt}}
\put(220.0,289.0){\rule[-0.200pt]{4.818pt}{0.400pt}}
\put(198,289){\makebox(0,0)[r]{$0.5$}}
\put(1416.0,289.0){\rule[-0.200pt]{4.818pt}{0.400pt}}
\put(220.0,407.0){\rule[-0.200pt]{4.818pt}{0.400pt}}
\put(198,407){\makebox(0,0)[r]{$0.6$}}
\put(1416.0,407.0){\rule[-0.200pt]{4.818pt}{0.400pt}}
\put(220.0,524.0){\rule[-0.200pt]{4.818pt}{0.400pt}}
\put(198,524){\makebox(0,0)[r]{$0.7$}}
\put(1416.0,524.0){\rule[-0.200pt]{4.818pt}{0.400pt}}
\put(220.0,642.0){\rule[-0.200pt]{4.818pt}{0.400pt}}
\put(198,642){\makebox(0,0)[r]{$0.8$}}
\put(1416.0,642.0){\rule[-0.200pt]{4.818pt}{0.400pt}}
\put(220.0,759.0){\rule[-0.200pt]{4.818pt}{0.400pt}}
\put(198,759){\makebox(0,0)[r]{$0.9$}}
\put(1416.0,759.0){\rule[-0.200pt]{4.818pt}{0.400pt}}
\put(220.0,877.0){\rule[-0.200pt]{4.818pt}{0.400pt}}
\put(198,877){\makebox(0,0)[r]{$1$}}
\put(1416.0,877.0){\rule[-0.200pt]{4.818pt}{0.400pt}}
\put(314.0,113.0){\rule[-0.200pt]{0.400pt}{4.818pt}}
\put(314,68){\makebox(0,0){$0.4$}}
\put(314.0,857.0){\rule[-0.200pt]{0.400pt}{4.818pt}}
\put(501.0,113.0){\rule[-0.200pt]{0.400pt}{4.818pt}}
\put(501,68){\makebox(0,0){$0.5$}}
\put(501.0,857.0){\rule[-0.200pt]{0.400pt}{4.818pt}}
\put(688.0,113.0){\rule[-0.200pt]{0.400pt}{4.818pt}}
\put(688,68){\makebox(0,0){$0.6$}}
\put(688.0,857.0){\rule[-0.200pt]{0.400pt}{4.818pt}}
\put(875.0,113.0){\rule[-0.200pt]{0.400pt}{4.818pt}}
\put(875,68){\makebox(0,0){$0.7$}}
\put(875.0,857.0){\rule[-0.200pt]{0.400pt}{4.818pt}}
\put(1062.0,113.0){\rule[-0.200pt]{0.400pt}{4.818pt}}
\put(1062,68){\makebox(0,0){$0.8$}}
\put(1062.0,857.0){\rule[-0.200pt]{0.400pt}{4.818pt}}
\put(1249.0,113.0){\rule[-0.200pt]{0.400pt}{4.818pt}}
\put(1249,68){\makebox(0,0){$0.9$}}
\put(1249.0,857.0){\rule[-0.200pt]{0.400pt}{4.818pt}}
\put(1436.0,113.0){\rule[-0.200pt]{0.400pt}{4.818pt}}
\put(1436,68){\makebox(0,0){$1$}}
\put(1436.0,857.0){\rule[-0.200pt]{0.400pt}{4.818pt}}
\put(220.0,113.0){\rule[-0.200pt]{292.934pt}{0.400pt}}
\put(1436.0,113.0){\rule[-0.200pt]{0.400pt}{184.048pt}}
\put(220.0,877.0){\rule[-0.200pt]{292.934pt}{0.400pt}}
\put(45,495){\makebox(0,0){McQuitty}}
\put(828,23){\makebox(0,0){Gibbs Sampling}}
\put(220.0,113.0){\rule[-0.200pt]{0.400pt}{184.048pt}}
\put(501,818){\makebox(0,0)[r]{Feature Set A}}
\put(545,818){\raisebox{-.8pt}{\makebox(0,0){$\Diamond$}}}
\put(669,488){\raisebox{-.8pt}{\makebox(0,0){$\Diamond$}}}
\put(1140,441){\raisebox{-.8pt}{\makebox(0,0){$\Diamond$}}}
\put(830,325){\raisebox{-.8pt}{\makebox(0,0){$\Diamond$}}}
\put(738,408){\raisebox{-.8pt}{\makebox(0,0){$\Diamond$}}}
\put(400,195){\raisebox{-.8pt}{\makebox(0,0){$\Diamond$}}}
\put(501,773){\makebox(0,0)[r]{Feature Set B}}
\put(545,773){\makebox(0,0){$+$}}
\put(884,587){\makebox(0,0){$+$}}
\put(1097,500){\makebox(0,0){$+$}}
\put(581,314){\makebox(0,0){$+$}}
\put(785,469){\makebox(0,0){$+$}}
\put(458,175){\makebox(0,0){$+$}}
\put(501,728){\makebox(0,0)[r]{Feature Set C}}
\put(545,728){\raisebox{-.8pt}{\makebox(0,0){$\Box$}}}
\put(673,361){\raisebox{-.8pt}{\makebox(0,0){$\Box$}}}
\put(1034,423){\raisebox{-.8pt}{\makebox(0,0){$\Box$}}}
\put(826,375){\raisebox{-.8pt}{\makebox(0,0){$\Box$}}}
\put(719,467){\raisebox{-.8pt}{\makebox(0,0){$\Box$}}}
\put(420,184){\raisebox{-.8pt}{\makebox(0,0){$\Box$}}}
\sbox{\plotpoint}{\rule[-0.500pt]{1.000pt}{1.000pt}}%
\put(220,113){\usebox{\plotpoint}}
\put(220.00,113.00){\usebox{\plotpoint}}
\put(237.58,124.00){\usebox{\plotpoint}}
\put(255.25,134.84){\usebox{\plotpoint}}
\multiput(257,136)(17.270,11.513){0}{\usebox{\plotpoint}}
\put(272.52,146.35){\usebox{\plotpoint}}
\put(290.31,157.01){\usebox{\plotpoint}}
\multiput(294,159)(17.270,11.513){0}{\usebox{\plotpoint}}
\put(307.78,168.19){\usebox{\plotpoint}}
\put(325.46,179.02){\usebox{\plotpoint}}
\multiput(331,182)(17.270,11.513){0}{\usebox{\plotpoint}}
\put(343.03,190.02){\usebox{\plotpoint}}
\put(360.30,201.53){\usebox{\plotpoint}}
\put(378.19,212.02){\usebox{\plotpoint}}
\multiput(380,213)(17.270,11.513){0}{\usebox{\plotpoint}}
\put(395.56,223.37){\usebox{\plotpoint}}
\put(413.34,234.03){\usebox{\plotpoint}}
\multiput(417,236)(17.270,11.513){0}{\usebox{\plotpoint}}
\put(430.81,245.21){\usebox{\plotpoint}}
\put(448.08,256.72){\usebox{\plotpoint}}
\multiput(453,260)(18.275,9.840){0}{\usebox{\plotpoint}}
\put(466.06,267.04){\usebox{\plotpoint}}
\put(483.33,278.56){\usebox{\plotpoint}}
\put(501.22,289.04){\usebox{\plotpoint}}
\multiput(503,290)(17.270,11.513){0}{\usebox{\plotpoint}}
\put(518.59,300.39){\usebox{\plotpoint}}
\put(535.86,311.91){\usebox{\plotpoint}}
\multiput(539,314)(18.275,9.840){0}{\usebox{\plotpoint}}
\put(553.84,322.23){\usebox{\plotpoint}}
\put(571.11,333.74){\usebox{\plotpoint}}
\multiput(576,337)(17.270,11.513){0}{\usebox{\plotpoint}}
\put(588.40,345.22){\usebox{\plotpoint}}
\put(606.37,355.58){\usebox{\plotpoint}}
\put(623.64,367.09){\usebox{\plotpoint}}
\multiput(625,368)(18.275,9.840){0}{\usebox{\plotpoint}}
\put(641.62,377.41){\usebox{\plotpoint}}
\put(658.89,388.93){\usebox{\plotpoint}}
\multiput(662,391)(17.270,11.513){0}{\usebox{\plotpoint}}
\put(676.28,400.23){\usebox{\plotpoint}}
\put(694.14,410.76){\usebox{\plotpoint}}
\multiput(699,414)(17.270,11.513){0}{\usebox{\plotpoint}}
\put(711.44,422.24){\usebox{\plotpoint}}
\put(729.40,432.60){\usebox{\plotpoint}}
\put(746.67,444.11){\usebox{\plotpoint}}
\multiput(748,445)(17.270,11.513){0}{\usebox{\plotpoint}}
\put(764.17,455.24){\usebox{\plotpoint}}
\put(781.92,465.95){\usebox{\plotpoint}}
\multiput(785,468)(17.270,11.513){0}{\usebox{\plotpoint}}
\put(799.32,477.25){\usebox{\plotpoint}}
\put(817.18,487.78){\usebox{\plotpoint}}
\multiput(822,491)(17.270,11.513){0}{\usebox{\plotpoint}}
\put(834.45,499.30){\usebox{\plotpoint}}
\put(852.05,510.26){\usebox{\plotpoint}}
\put(869.70,521.13){\usebox{\plotpoint}}
\multiput(871,522)(17.270,11.513){0}{\usebox{\plotpoint}}
\put(887.20,532.26){\usebox{\plotpoint}}
\put(904.95,542.97){\usebox{\plotpoint}}
\multiput(908,545)(17.270,11.513){0}{\usebox{\plotpoint}}
\put(922.22,554.48){\usebox{\plotpoint}}
\put(939.93,565.27){\usebox{\plotpoint}}
\multiput(945,568)(17.270,11.513){0}{\usebox{\plotpoint}}
\put(957.48,576.32){\usebox{\plotpoint}}
\put(975.08,587.27){\usebox{\plotpoint}}
\put(992.73,598.15){\usebox{\plotpoint}}
\multiput(994,599)(17.270,11.513){0}{\usebox{\plotpoint}}
\put(1010.00,609.67){\usebox{\plotpoint}}
\put(1027.81,620.28){\usebox{\plotpoint}}
\multiput(1031,622)(17.270,11.513){0}{\usebox{\plotpoint}}
\put(1045.26,631.50){\usebox{\plotpoint}}
\put(1062.96,642.29){\usebox{\plotpoint}}
\multiput(1068,645)(17.270,11.513){0}{\usebox{\plotpoint}}
\put(1080.51,653.34){\usebox{\plotpoint}}
\put(1097.78,664.85){\usebox{\plotpoint}}
\put(1115.69,675.30){\usebox{\plotpoint}}
\multiput(1117,676)(17.270,11.513){0}{\usebox{\plotpoint}}
\put(1133.03,686.69){\usebox{\plotpoint}}
\put(1150.30,698.20){\usebox{\plotpoint}}
\multiput(1153,700)(18.275,9.840){0}{\usebox{\plotpoint}}
\put(1168.29,708.53){\usebox{\plotpoint}}
\put(1185.56,720.04){\usebox{\plotpoint}}
\multiput(1190,723)(18.275,9.840){0}{\usebox{\plotpoint}}
\put(1203.54,730.36){\usebox{\plotpoint}}
\put(1220.81,741.87){\usebox{\plotpoint}}
\put(1238.08,753.39){\usebox{\plotpoint}}
\multiput(1239,754)(18.275,9.840){0}{\usebox{\plotpoint}}
\put(1256.07,763.71){\usebox{\plotpoint}}
\put(1273.34,775.22){\usebox{\plotpoint}}
\multiput(1276,777)(18.275,9.840){0}{\usebox{\plotpoint}}
\put(1291.32,785.55){\usebox{\plotpoint}}
\put(1308.59,797.06){\usebox{\plotpoint}}
\multiput(1313,800)(17.270,11.513){0}{\usebox{\plotpoint}}
\put(1325.91,808.49){\usebox{\plotpoint}}
\put(1343.84,818.90){\usebox{\plotpoint}}
\put(1361.11,830.41){\usebox{\plotpoint}}
\multiput(1362,831)(18.275,9.840){0}{\usebox{\plotpoint}}
\put(1379.10,840.73){\usebox{\plotpoint}}
\put(1396.37,852.24){\usebox{\plotpoint}}
\multiput(1399,854)(17.270,11.513){0}{\usebox{\plotpoint}}
\put(1413.79,863.50){\usebox{\plotpoint}}
\put(1431.62,874.08){\usebox{\plotpoint}}
\put(1436,877){\usebox{\plotpoint}}
\end{picture}
\caption{Gibbs and McQuitty's Correlation of Accuracy for Nouns}
\label{fig:accboth-noun}
\end{center}
\vskip -0.14in
\end{figure}
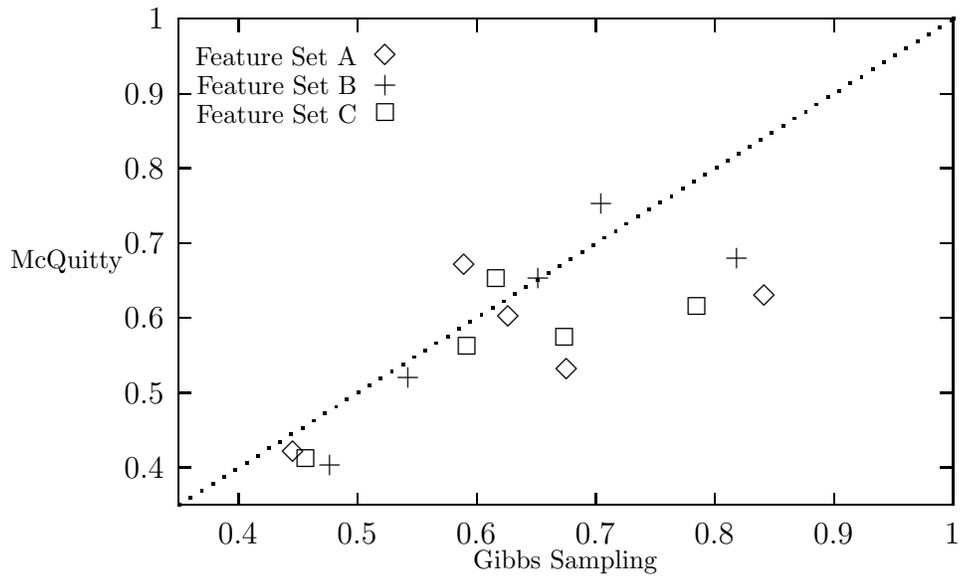

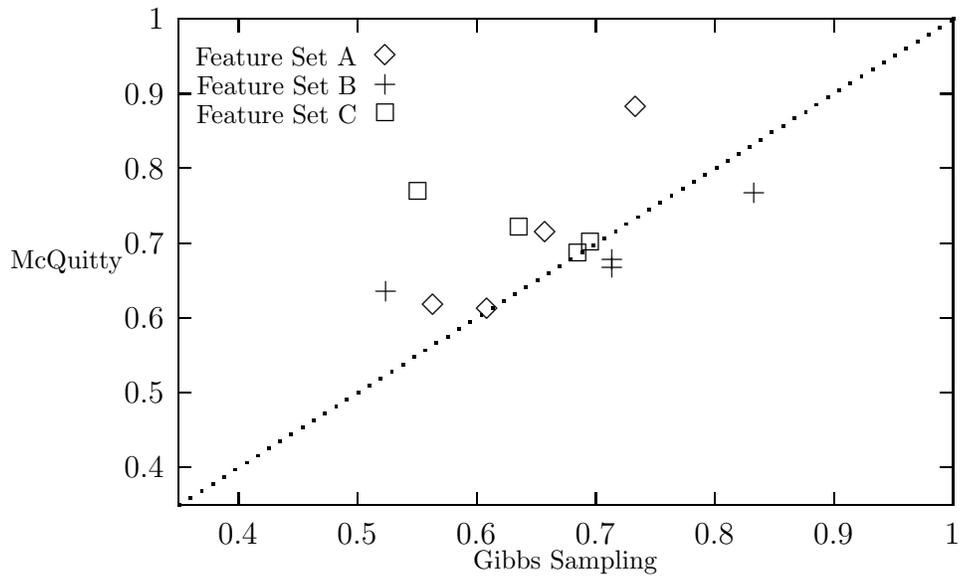
\begin{figure}
\begin{center}
\setlength{\unitlength}{0.240900pt}
\ifx\plotpoint\undefined\newsavebox{\plotpoint}\fi
\sbox{\plotpoint}{\rule[-0.200pt]{0.400pt}{0.400pt}}%
\begin{picture}(1500,900)(0,0)
\font\gnuplot=cmr10 at 10pt
\gnuplot
\sbox{\plotpoint}{\rule[-0.200pt]{0.400pt}{0.400pt}}%
\put(220.0,172.0){\rule[-0.200pt]{4.818pt}{0.400pt}}
\put(198,172){\makebox(0,0)[r]{$0.4$}}
\put(1416.0,172.0){\rule[-0.200pt]{4.818pt}{0.400pt}}
\put(220.0,289.0){\rule[-0.200pt]{4.818pt}{0.400pt}}
\put(198,289){\makebox(0,0)[r]{$0.5$}}
\put(1416.0,289.0){\rule[-0.200pt]{4.818pt}{0.400pt}}
\put(220.0,407.0){\rule[-0.200pt]{4.818pt}{0.400pt}}
\put(198,407){\makebox(0,0)[r]{$0.6$}}
\put(1416.0,407.0){\rule[-0.200pt]{4.818pt}{0.400pt}}
\put(220.0,524.0){\rule[-0.200pt]{4.818pt}{0.400pt}}
\put(198,524){\makebox(0,0)[r]{$0.7$}}
\put(1416.0,524.0){\rule[-0.200pt]{4.818pt}{0.400pt}}
\put(220.0,642.0){\rule[-0.200pt]{4.818pt}{0.400pt}}
\put(198,642){\makebox(0,0)[r]{$0.8$}}
\put(1416.0,642.0){\rule[-0.200pt]{4.818pt}{0.400pt}}
\put(220.0,759.0){\rule[-0.200pt]{4.818pt}{0.400pt}}
\put(198,759){\makebox(0,0)[r]{$0.9$}}
\put(1416.0,759.0){\rule[-0.200pt]{4.818pt}{0.400pt}}
\put(220.0,877.0){\rule[-0.200pt]{4.818pt}{0.400pt}}
\put(198,877){\makebox(0,0)[r]{$1$}}
\put(1416.0,877.0){\rule[-0.200pt]{4.818pt}{0.400pt}}
\put(314.0,113.0){\rule[-0.200pt]{0.400pt}{4.818pt}}
\put(314,68){\makebox(0,0){$0.4$}}
\put(314.0,857.0){\rule[-0.200pt]{0.400pt}{4.818pt}}
\put(501.0,113.0){\rule[-0.200pt]{0.400pt}{4.818pt}}
\put(501,68){\makebox(0,0){$0.5$}}
\put(501.0,857.0){\rule[-0.200pt]{0.400pt}{4.818pt}}
\put(688.0,113.0){\rule[-0.200pt]{0.400pt}{4.818pt}}
\put(688,68){\makebox(0,0){$0.6$}}
\put(688.0,857.0){\rule[-0.200pt]{0.400pt}{4.818pt}}
\put(875.0,113.0){\rule[-0.200pt]{0.400pt}{4.818pt}}
\put(875,68){\makebox(0,0){$0.7$}}
\put(875.0,857.0){\rule[-0.200pt]{0.400pt}{4.818pt}}
\put(1062.0,113.0){\rule[-0.200pt]{0.400pt}{4.818pt}}
\put(1062,68){\makebox(0,0){$0.8$}}
\put(1062.0,857.0){\rule[-0.200pt]{0.400pt}{4.818pt}}
\put(1249.0,113.0){\rule[-0.200pt]{0.400pt}{4.818pt}}
\put(1249,68){\makebox(0,0){$0.9$}}
\put(1249.0,857.0){\rule[-0.200pt]{0.400pt}{4.818pt}}
\put(1436.0,113.0){\rule[-0.200pt]{0.400pt}{4.818pt}}
\put(1436,68){\makebox(0,0){$1$}}
\put(1436.0,857.0){\rule[-0.200pt]{0.400pt}{4.818pt}}
\put(220.0,113.0){\rule[-0.200pt]{292.934pt}{0.400pt}}
\put(1436.0,113.0){\rule[-0.200pt]{0.400pt}{184.048pt}}
\put(220.0,877.0){\rule[-0.200pt]{292.934pt}{0.400pt}}
\put(45,495){\makebox(0,0){McQuitty}}
\put(828,23){\makebox(0,0){Gibbs Sampling}}
\put(220.0,113.0){\rule[-0.200pt]{0.400pt}{184.048pt}}
\put(501,818){\makebox(0,0)[r]{Feature Set A}}
\put(545,818){\raisebox{-.8pt}{\makebox(0,0){$\Diamond$}}}
\put(705,419){\raisebox{-.8pt}{\makebox(0,0){$\Diamond$}}}
\put(620,426){\raisebox{-.8pt}{\makebox(0,0){$\Diamond$}}}
\put(796,540){\raisebox{-.8pt}{\makebox(0,0){$\Diamond$}}}
\put(938,736){\raisebox{-.8pt}{\makebox(0,0){$\Diamond$}}}
\put(501,773){\makebox(0,0)[r]{Feature Set B}}
\put(545,773){\makebox(0,0){$+$}}
\put(901,499){\makebox(0,0){$+$}}
\put(901,486){\makebox(0,0){$+$}}
\put(546,449){\makebox(0,0){$+$}}
\put(1124,603){\makebox(0,0){$+$}}
\put(501,728){\makebox(0,0)[r]{Feature Set C}}
\put(545,728){\raisebox{-.8pt}{\makebox(0,0){$\Box$}}}
\put(847,507){\raisebox{-.8pt}{\makebox(0,0){$\Box$}}}
\put(755,548){\raisebox{-.8pt}{\makebox(0,0){$\Box$}}}
\put(867,524){\raisebox{-.8pt}{\makebox(0,0){$\Box$}}}
\put(596,604){\raisebox{-.8pt}{\makebox(0,0){$\Box$}}}
\sbox{\plotpoint}{\rule[-0.500pt]{1.000pt}{1.000pt}}%
\put(220,113){\usebox{\plotpoint}}
\put(220.00,113.00){\usebox{\plotpoint}}
\put(237.58,124.00){\usebox{\plotpoint}}
\put(255.25,134.84){\usebox{\plotpoint}}
\multiput(257,136)(17.270,11.513){0}{\usebox{\plotpoint}}
\put(272.52,146.35){\usebox{\plotpoint}}
\put(290.31,157.01){\usebox{\plotpoint}}
\multiput(294,159)(17.270,11.513){0}{\usebox{\plotpoint}}
\put(307.78,168.19){\usebox{\plotpoint}}
\put(325.46,179.02){\usebox{\plotpoint}}
\multiput(331,182)(17.270,11.513){0}{\usebox{\plotpoint}}
\put(343.03,190.02){\usebox{\plotpoint}}
\put(360.30,201.53){\usebox{\plotpoint}}
\put(378.19,212.02){\usebox{\plotpoint}}
\multiput(380,213)(17.270,11.513){0}{\usebox{\plotpoint}}
\put(395.56,223.37){\usebox{\plotpoint}}
\put(413.34,234.03){\usebox{\plotpoint}}
\multiput(417,236)(17.270,11.513){0}{\usebox{\plotpoint}}
\put(430.81,245.21){\usebox{\plotpoint}}
\put(448.08,256.72){\usebox{\plotpoint}}
\multiput(453,260)(18.275,9.840){0}{\usebox{\plotpoint}}
\put(466.06,267.04){\usebox{\plotpoint}}
\put(483.33,278.56){\usebox{\plotpoint}}
\put(501.22,289.04){\usebox{\plotpoint}}
\multiput(503,290)(17.270,11.513){0}{\usebox{\plotpoint}}
\put(518.59,300.39){\usebox{\plotpoint}}
\put(535.86,311.91){\usebox{\plotpoint}}
\multiput(539,314)(18.275,9.840){0}{\usebox{\plotpoint}}
\put(553.84,322.23){\usebox{\plotpoint}}
\put(571.11,333.74){\usebox{\plotpoint}}
\multiput(576,337)(17.270,11.513){0}{\usebox{\plotpoint}}
\put(588.40,345.22){\usebox{\plotpoint}}
\put(606.37,355.58){\usebox{\plotpoint}}
\put(623.64,367.09){\usebox{\plotpoint}}
\multiput(625,368)(18.275,9.840){0}{\usebox{\plotpoint}}
\put(641.62,377.41){\usebox{\plotpoint}}
\put(658.89,388.93){\usebox{\plotpoint}}
\multiput(662,391)(17.270,11.513){0}{\usebox{\plotpoint}}
\put(676.28,400.23){\usebox{\plotpoint}}
\put(694.14,410.76){\usebox{\plotpoint}}
\multiput(699,414)(17.270,11.513){0}{\usebox{\plotpoint}}
\put(711.44,422.24){\usebox{\plotpoint}}
\put(729.40,432.60){\usebox{\plotpoint}}
\put(746.67,444.11){\usebox{\plotpoint}}
\multiput(748,445)(17.270,11.513){0}{\usebox{\plotpoint}}
\put(764.17,455.24){\usebox{\plotpoint}}
\put(781.92,465.95){\usebox{\plotpoint}}
\multiput(785,468)(17.270,11.513){0}{\usebox{\plotpoint}}
\put(799.32,477.25){\usebox{\plotpoint}}
\put(817.18,487.78){\usebox{\plotpoint}}
\multiput(822,491)(17.270,11.513){0}{\usebox{\plotpoint}}
\put(834.45,499.30){\usebox{\plotpoint}}
\put(852.05,510.26){\usebox{\plotpoint}}
\put(869.70,521.13){\usebox{\plotpoint}}
\multiput(871,522)(17.270,11.513){0}{\usebox{\plotpoint}}
\put(887.20,532.26){\usebox{\plotpoint}}
\put(904.95,542.97){\usebox{\plotpoint}}
\multiput(908,545)(17.270,11.513){0}{\usebox{\plotpoint}}
\put(922.22,554.48){\usebox{\plotpoint}}
\put(939.93,565.27){\usebox{\plotpoint}}
\multiput(945,568)(17.270,11.513){0}{\usebox{\plotpoint}}
\put(957.48,576.32){\usebox{\plotpoint}}
\put(975.08,587.27){\usebox{\plotpoint}}
\put(992.73,598.15){\usebox{\plotpoint}}
\multiput(994,599)(17.270,11.513){0}{\usebox{\plotpoint}}
\put(1010.00,609.67){\usebox{\plotpoint}}
\put(1027.81,620.28){\usebox{\plotpoint}}
\multiput(1031,622)(17.270,11.513){0}{\usebox{\plotpoint}}
\put(1045.26,631.50){\usebox{\plotpoint}}
\put(1062.96,642.29){\usebox{\plotpoint}}
\multiput(1068,645)(17.270,11.513){0}{\usebox{\plotpoint}}
\put(1080.51,653.34){\usebox{\plotpoint}}
\put(1097.78,664.85){\usebox{\plotpoint}}
\put(1115.69,675.30){\usebox{\plotpoint}}
\multiput(1117,676)(17.270,11.513){0}{\usebox{\plotpoint}}
\put(1133.03,686.69){\usebox{\plotpoint}}
\put(1150.30,698.20){\usebox{\plotpoint}}
\multiput(1153,700)(18.275,9.840){0}{\usebox{\plotpoint}}
\put(1168.29,708.53){\usebox{\plotpoint}}
\put(1185.56,720.04){\usebox{\plotpoint}}
\multiput(1190,723)(18.275,9.840){0}{\usebox{\plotpoint}}
\put(1203.54,730.36){\usebox{\plotpoint}}
\put(1220.81,741.87){\usebox{\plotpoint}}
\put(1238.08,753.39){\usebox{\plotpoint}}
\multiput(1239,754)(18.275,9.840){0}{\usebox{\plotpoint}}
\put(1256.07,763.71){\usebox{\plotpoint}}
\put(1273.34,775.22){\usebox{\plotpoint}}
\multiput(1276,777)(18.275,9.840){0}{\usebox{\plotpoint}}
\put(1291.32,785.55){\usebox{\plotpoint}}
\put(1308.59,797.06){\usebox{\plotpoint}}
\multiput(1313,800)(17.270,11.513){0}{\usebox{\plotpoint}}
\put(1325.91,808.49){\usebox{\plotpoint}}
\put(1343.84,818.90){\usebox{\plotpoint}}
\put(1361.11,830.41){\usebox{\plotpoint}}
\multiput(1362,831)(18.275,9.840){0}{\usebox{\plotpoint}}
\put(1379.10,840.73){\usebox{\plotpoint}}
\put(1396.37,852.24){\usebox{\plotpoint}}
\multiput(1399,854)(17.270,11.513){0}{\usebox{\plotpoint}}
\put(1413.79,863.50){\usebox{\plotpoint}}
\put(1431.62,874.08){\usebox{\plotpoint}}
\put(1436,877){\usebox{\plotpoint}}
\end{picture}
\caption{Gibbs and McQuitty's Correlation of Accuracy for Verbs}
\label{fig:accboth-verb}
\end{center}
\vskip -0.14in
\end{figure}

Of the 19 significant differences, 10 occur among the adjectives, 7
occur among the nouns, and 2 occur among the verbs.  The distribution
of senses plays a role in these results, particularly for the
adjectives. Of the 10 significant differences among the adjectives, 7
favor McQuitty's method. Given the tendency of McQuitty's method to
discover skewed sense distributions this is not surprising. Of the 7
significant differences observed among the nouns, all favor Gibbs
Sampling. Again, this is somewhat expected given the bias of Gibbs
Sampling towards discovering balanced distributions of
senses. Finally, the two significant differences in the verbs favor
McQuitty's method. Despite having rather skewed sense distributions, 
McQuitty's method and Gibbs Sampling perform at comparable levels of
accuracy for the verbs. This indicates that the greater granularity of
the data representation used by Gibbs Sampling is sometimes sufficient
to offset the bias of McQuitty's method towards discovering skewed
sense distributions.  

Direct comparison of McQuitty's method and
Gibbs Sampling must take into account the differences in the data
representations employed.  McQuitty's similarity analysis is based
upon counts of the number of dissimilar features between multiple
instances of the ambiguous word. A probabilistic model is based upon
frequency counts of the marginal events as defined by its parametric
form. Given these rather different representations, it is not
surprising that the accuracies for the two methods for a given word
and feature set are often quite different.  However, this makes the
few cases where the two methods achieve nearly identical results all the more
intriguing. For example, {\it agree} with feature sets A and C, {\it
interest} with feature set B, and {\it help} with feature set C, all
achieve very similar levels of accuracy despite the differences in
representation.  Understanding the conditions that lead to these
results is an interesting area for future work. 

In general, these results suggest that the characteristics of
a feature set must be compatible with the learning algorithm in order
to achieve good results.  Gibbs Sampling benefits from feature sets
that contain a small number of features that each have a limited
number of possible values. This allows for accurate parameter
estimation, particularly when the parametric form is a simple model
such as Naive Bayes. For example, the ability learn reliable parameter
estimates contributes to the high accuracy that Gibbs Sampling
achieves with feature set A for the nouns. By contrast, McQuitty's
method generally  benefits from larger numbers of features and higher
dimensional  spaces. Given such data, a dissimilarity matrix becomes a
richer source of information that can be used to make more fine
grained  distinctions than is the case with a small number of
features.  This data representation contributes to the overall high
accuracy attained with feature set C for adjectives and verbs.


\chapter{RELATED WORK} 

Much of the early work in word sense disambiguation relied on the use
of rich, manually--crafted knowledge sources such as semantic networks
and concept hierarchies (e.g., \cite{Hirst87}, \cite{SmallR82},
\cite{Wilks75}). While these systems were very successful in limited
domains, they tended to be difficult to scale up or port to new domains. 

As the difficulty in creating knowledge--rich resources for larger
domains became apparent, research shifted to exploiting
online lexical resources that were already constructed such as
dictionaries, thesaruses, and encyclopedias (e.g., \cite{Lesk86},
\cite{Yarowsky92}). While these are rich sources of knowledge that offer 
relatively  broad  coverage of both language and topic, they are 
not designed for use with a mechanical inferencing algorithm. 
Rather, these resources are intended
for a human user who will apply their own inferencing
methods to find and understand the information in the lexical
resource. 

Recent work in disambiguation has been geared towards corpus--based,
statistical methods (e.g., \cite{BruceW94A}, \cite{BruceW94B}, 
\cite{NgL96}, \cite{Ng97}, \cite{PedersenB97A},
\cite{PedersenBW97}, \cite{WilksS97}). These approaches often employ
supervised learning algorithms and require the availability of
manually created training examples from which to learn. However,
sense--tagged generally does not exist in large quantities
and it proves expensive to create. 

The difficulties in building semantic networks, the lack of
automatic inferencing algorithms appropriate for lexical resources 
designed for human use, and the time consuming nature of manually
sense--tagging text; all these factors lead to the  realization 
that the only truly broad--coverage knowledge resource currently available
for word sense disambiguation  
is raw untagged text.  However, the lack of any systematic structure and 
the absence of points of reference to external knowledge sources makes 
untagged text a very challenging resource from which to learn. 

It is difficult to precisely quantify the degree of structure and
richness in a knowledge source for word sense disambiguation.  
The following is an approximate and subjective ranking, beginning 
with the richest and most structured sources of knowledge and ending 
with raw untagged text, the most impoverished and unstructured source
considered here. 

\begin{enumerate}
\item semantic networks, concept hierarchies
\item machine readable dictionaries, thesaruses
\item parallel translations 
\item sense--tagged corpora
\item raw untagged corpora 
\end{enumerate}

This chapter discusses representative approaches to word sense
disambiguation that employ each of these  different kinds of knowledge 
resources.
    
\section{Semantic Networks}

A semantic network is a highly structured knowledge source where nodes
represent concepts and related concepts are connected by links 
of various types. Common examples of links include {\it is--a}, {\it
has--part}, and {\it is--made--of}. 

Semantic networks are often used to model and enforce selectional
restrictions,
a concept that finds its roots in Case Grammar \cite{Fillmore68}. 
This is a lexically based linguistic formalism where verbs are defined
based on the roles, i.e, case frames,  of the words that they may be
validly used with. As a simple example, suppose that the verb {\it
hit} is defined as follows:    
\begin{quote}
{\bf hit} :: [AGENT:human] [OBJECT:projectile] [INSTRUMENT:club]
\end{quote}

AGENT, OBJECT and INSTRUMENT are just a few examples of possible case
frames.  The selectional restrictions on these frames are specified
in lower case letters. This definition tells us that the verb {\it
hit} expects that the AGENT who performs the hitting is a human,
that the OBJECT that AGENT hits is a projectile, and that the INSTRUMENT the
AGENT uses to hit the OBJECT with is a club. 

In these approaches nouns are often defined in terms of subsuming
relations as shown in a IS-A hierarchy. A {\it bat} is a club, a
{\it boy} is a human, and a {\it ball} is a projectile. A Case
Grammar parser will accept the sentence {\it The boy hit the ball
with a bat} since all of the selectional restrictions imposed by the
verb are honored.  

Case Grammar and selectional restrictions are conveniently mapped onto
a semantic network. The nodes of the network represent concepts and
the links between nodes enforce the selectional restrictions. A
semantic network representation of the Case Grammar for {\it hit} 
appears in  Figure \ref{fig:semnet}. Here the nouns and verbs in the
sentence are shown in boxes, the concepts are in ovals, and the links
are labeled appropriately. 

\begin{figure}
\centerline{\epsfbox{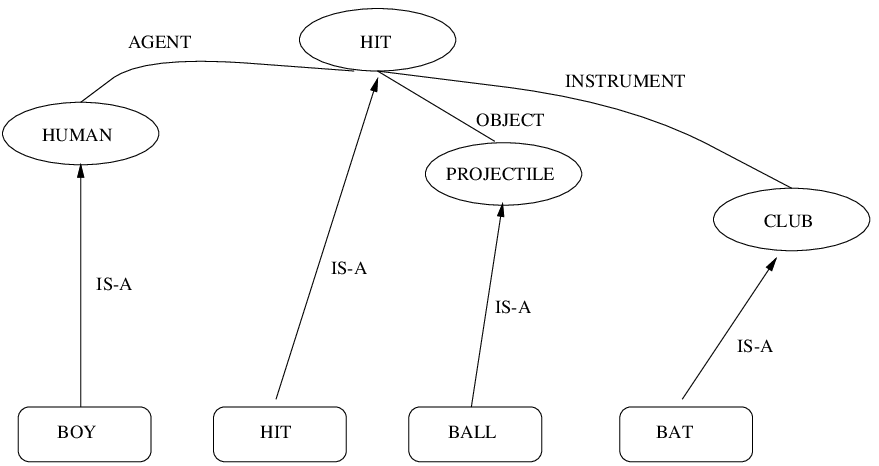}}
\caption{Simple Semantic Network}
\label{fig:semnet}
\myendfig


Once a semantic network is constructed, word sense disambiguation can
be performed using marker passing as an inference mechanism. Marker
passing was introduced in \cite{Quillian69} as a means of spreading
activation on semantic memory. Marker passing was extended to serve as
an inferencing mechanism by \cite{Charniak83}. 

Markers are able to travel through the semantic network, visiting the
nodes and moving along the links. Markers are restricted as to what
types of links they may travel along.  Inferencing is achieved by
propagating markers from the concepts of interest and determining at
what concepts they intersect in the network. These points of
intersection will reveal some kind of relationship between the two
concepts that may not have been previously realized.  

Marker passing has been widely used in language processing for various
inferencing problems, including word sense disambiguation
(e.g. \cite{ChungM93}, \cite{Hirst87}, \cite{MoldovanLLC92},
\cite{Norvig89}, and \cite{YuS90}).  Generally the words in a sentence
activate the concepts that they are linked to by passing a marker. The
activated concepts continue to propagate markers to other concepts
until the network eventually stabilizes. This stabilized network
represents the disambiguated sentence.

Marker passing offers tremendous opportunities to exploit parallel
computer architectures. It is also an intuitively appealing approach
that may ultimately allow for the development of reasonable cognitive
models of disambiguation. However, the question of how to construct
the underlying representations remains problematic. One approach
that has proven successful is to learn selectional constraints via
interactive training with a user (e.g. \cite{Cardie93},
\cite{KimM92}).  Another option is to automatically  construct these
representations from existing resources such  as a machine readable
dictionary (e.g. \cite{BruceWGSD92}, \cite{ChodorowBH85},
\cite{VeronisI90}).  

\section{Machine Readable Dictionaries}

Machine readable dictionaries were first applied to word sense
disambiguation in \cite{Lesk86}. There, {\it pine cone} was
disambiguated based on the dictionary definitions of {\it pine} and
{\it cone}. It was  noted that the definitions of  {\it pine} and 
{\it cone} both contained references to the concept of a tree:
\begin{quote}
{\bf pine:} any of a genus of coniferous evergreen {\it trees} which have 
slender elongated needles and some of which are valuable timber {\it
trees} or ornamentals \\
{\bf cone:} a mass of ovule-bearing or pollen-bearing scales or bracts in
{\it trees}  of the pine family or in cycads that are arranged usually on 
a  somewhat elongated axis
\end{quote}
By unifying these references to {\it tree} a computer program 
inferred that {\it pine cone} is the fruit of a tree rather than an
edible receptacle for a pine tree. Experimental results for this
approach are reported at 50\%--70\% accuracy for short passages from
{\it Pride and Prejudice} and an Associated Press news story.  

\cite{Yarowsky92} presents an approach where ambiguous words that occur in 
encyclopedia entries are disambiguated with respect to categories defined 
in Roget's Thesaurus. 
A Naive Bayes model is developed that contains 100 feature
variables and a single variable representing the sense of the
ambiguous word.  The feature variables are the 50 words to the left
and right of the ambiguous word. The parameter estimates for this
model are made using category information from Roget's
Thesaurus. There are 1042 categories in Roget's. Typical examples of
categories  include
{\it tools--machinery} or {\it animal--insect},  and each category is
described by a broad set of relations (similar to those represented
by links in a semantic network) 
that typically consist of over 3,000 words. After the 
parameter estimates are made from the entries describing Roget's 
categories, this probabilistic model is used to disambiguate instances of 
twelve ambiguous words found in the June 1991 version of 
Grolier's Encyclopedia. Accuracy is reported at above 90\%  for  11 of 12 
words with between 2 and 6 possible senses.   

While machine readable dictionaries are a promising resource for
disambiguation, it can sometimes be the case that dictionary
entries are too brief to provide all of the salient collocations or other
clues that might identify the sense of an ambiguous word. However, as
online dictionaries grow more extensive their usefulness as a knowledge
source in corpus--based language processing will likewise increase. 

\section{Parallel Translations}

Given the expense of manually tagging ambiguous words with senses, it
is natural to ask if there are clever means of obtaining sense--tagged
text that avoid the need for manual intervention.  In fact, the use of
parallel translations is such an approach. This methodology relies
upon the premise that while a word may be ambiguous in one language,
the various senses may have distinct word forms in another language.  

Consider the word {\it bill} in English. It has many possible senses,
among them pending legislation and statement requesting payment. In
Spanish these two senses have distinct word forms, {\it proyecto de
ley} and {\it cuenta}. Suppose the following usages of {\it bill} are
found in parallel English and Spanish text:
\begin{quote}
1E) The bill is too much. \\
1S) La cuenta es demasiado.  \\
2E) The bill to save the banks is good.  \\
2S) El proyecto de ley para salvar los bancos es bueno.
\end{quote}

From the Spanish text it is clear that usage of {\it bill} in sentence
1E) refers to a statement requesting payment while the usage in
sentence 2E) refers to pending legislation. Thus, the sense distinction
made in Spanish is utilized to assign the appropriate sense--tags to {\it
bill} in English.  

This approach to creating sense--tagged text has been pursued mainly
in French and English due to the availability of parallel translations
of the Canadian Parliamentary Proceedings, i.e., the Hansards,
(e.g. \cite{BrownPPM91}, \cite{GaleCY92}). Once the sense--tags
are obtained from a parallel translation, supervised
learning methods can be employed as if the tagging had been performed
manually. Naive Bayes with a  large window of context is employed in
\cite{GaleCY92} while  \cite{BrownPPM91} identify a single binary
feature that makes the sense  distinction. 

However, given the nature of the Hansards, it is in fact rather
difficult to locate many words that are truly ambiguous within
that domain.  For example,\cite{GaleCY92} point out that while 
{\it bank} is highly ambiguous in general text, in the Hansards it
nearly always is used to refer to a financial institution. Indeed, the
location of more diverse parallel bilingual texts remains the main
obstacle to wider use of this approach.  

A related method described in \cite{DaganI94} finds translations
between Hebrew and English using co--occurrence statistics from
independent Hebrew and English corpora. This approach is somewhat more
flexible in that it does not require the availability of diverse
parallel bilingual corpora.  

An unanswered question is key to determining the viability of
parallel corpora approaches; how large is the set of words that are
ambiguous in both languages? If it is small then this approach
is certainly viable. If not, then it may suffer from scaling problems
much like other resources. 

\section{Sense--Tagged Corpora}

The earliest use of sense--tagged text to create models of word sense
disambiguation may have been that of \cite{KellyS75}. They built 1,815
models of disambiguation manually, focusing on words that occur at
least 20 times in a corpus of 510,976 words. Their models
consist of sets of rules and use features that are
found within four positions of the ambiguous word. These features
include the part--of--speech of surrounding words, the morphology of
the ambiguous word, and membership of surrounding words into one of
sixteen possible semantic categories: Animate, Human,
Collective, Abstract Noun, Social Place, Body Part, Political, Economic,
Color, Communications, Emotions, Frequency, Evaluative Adjective,
Dimensionality Adjective, Position Adjective, and Degree Adverb. 

An early automatic approach where models are learned from 
sense--tagged 
text is presented in \cite{Black88}. Two--thousand  sense--tagged
instances for each of five  words were created, where each word had
three  or four possible senses. A decision tree learner was provided
with 1,500 training examples for each word, where each example was
characterized by 81 binary features representing the presence or
absence of certain ``contextual categories''.  There are three
varieties of contextual category; subject categories from Longman's
Dictionary of Contemporary English, the 41 words that occur most
frequently within two positions of the ambiguous word, and the 40
content words that occur most frequently in the sentence with an
ambiguous word. It was found that the dictionary categories resulted
in 47\% accuracy, the 41 most frequent words resulted in 72\%
accuracy, and the 40 most frequent content words resulted in 75\%
accuracy.  

Early probabilistic approaches typically attempted to identify and
exploit a single very powerful contextual feature to perform
disambiguation. For example \cite{BrownPPM91}, \cite{DaganIS91}, and
\cite{Yarowsky93} all present methods for identifying a single feature
that is sufficient to make highly accurate disambiguation decisions. In
\cite{Yarowsky93} for example, it is reported that a single collocation
feature, {\it content--word--to--the--right}, results in accuracy well
over 90\% for binary sense distinctions.  

In order to utilize probabilistic models with more complicated
interactions among feature variables, \cite{BruceW94B} introduced the
use of sequential model selection and decomposable models for word
sense disambiguation. Prior to this,  statistical analysis of natural
language data was often limited to the application of standard
models, such as n-grams and Naive Bayes.  They developed  a
sequential model selection procedure using backward search and the
exact conditional test in combination  with a test for model
predictive power. In their procedure, the  exact conditional test is used 
to guide the generation of new models and  a test of model predictive 
power was used to select the final model from  among those generated 
during the search. 

The supervised learning portion of this dissertation largely consists
of extensions to the work of Bruce and Wiebe. As such, their methods
are discussed rather extensively in Chapter 3 and their feature set
and sense--tagged text is described in Chapter 5.  

What emerges throughout the literature of corpus--based approaches to word 
sense disambiguation is considerable variation in the methodologies, 
a wide range of feature sets, and a great variety in the types of text
that have been disambiguated. Unfortunately, comparative studies  of these 
approaches have been relatively rare. 

As mentioned in Chapter 6, \cite{LeacockTV93} compare  a neural 
network, a Naive Bayes classifier,  and a content vector when 
disambiguating six senses of {\it line}. It is reported that all three 
methods are equally accurate. This same data is utilized by
\cite{Mooney96} and  applied to an even  wider range of approaches; a
Naive Bayes  classifier, a perceptron, a  decision--tree, a
nearest--neighbor  classifier, a logic based Disjunctive  Normal Form
learner, a logic based  Conjunctive Normal Form learner, and a  decision
list learner are all compared. It is found  that the Naive Bayes
classifier and the perceptron  prove to be the most accurate of these
approaches.  

Both  studies employ the same feature set for the {\it line} data. It
consists of  binary features that represent  the occurrence of all words 
within approximately a 50 word window of the ambiguous word, resulting in 
nearly 3,000 binary features. Given the vast size of the event space, 
representations of training data created by simple approaches such as 
Naive Bayes and the perceptron capture the same information as those
created by more sophisticated methods. 

A comparative study of the nearest neighbor classifier PEBLS and the
backward sequential model selection method of
\cite{BruceW94B} is presented by  \cite{NgL96}. They 
compare the performance of the two methods at disambiguating 6 senses of 
{\it interest}. They report that PEBLS achieves
accuracy of 87\% while  \cite{BruceW94B} report accuracy of
78\%.\footnote{The same data is employed in this dissertation and  the 
highest accuracy attained is 76$\pm2$\%.} 
They expand upon the feature set used in \cite{BruceW94B} (feature
set BW) by including collocation features and verb--object 
relationships. Their feature set consists of the following:
\begin{enumerate}
\item collocations that occur within one word of the ambiguous word
\item part--of--speech of words $\pm$ 3 positions of ambiguous word
\item morphology of the ambiguous word
\item unordered set of surrounding key--words, i.e., co--occurrences
\item verb--object syntactic relations
\end{enumerate}

\cite{NgL96} evaluate the relative contribution of
each type of feature to the overall disambiguation accuracy. They
report that the collocations provide nearly all of the
disambiguation accuracy, while the part--of--speech and morphological
information also prove useful. The unordered sets of surrounding
key--words and verb--object syntactic relations tended to contribute
very little to disambiguation accuracy. Thus, the improvement in
accuracy that they report may be  due to their use of
collocations features.  

The fundamental limitation of supervised learning approaches to word
sense disambiguation is the availability of sense--tagged text. The
largest available source of sense--tagged text  is
the Defense System Organization  192,800 sense--tagged word corpus
\cite{NgL96}.  There are 191 different nouns and verbs that are
sense--tagged. The  average number of senses per noun is 7.8 and 12.0
senses per verb.  The only other large source of sense--tagged text
that is widely  available is a 100,000 word subset of the Brown Corpus
\cite{FrancisK82}.  Both of these corpora are tagged with WordNet
senses. By way of speculation, if all of the ``privately held''
sense--tagged text  was added to the 300,000 words provided by the two
corpora above, it seems unlikely that the total number of
sense--tagged instances would exceed one--million words.  

\section{Raw Untagged Corpora} 

There are in fact relatively few ``pure'' unsupervised methodologies for 
word sense disambiguation that rely strictly on raw untagged text (e.g.,
\cite{PedersenB97C}, \cite{PedersenB98}, \cite{Schutze92},
\cite{Schutze98}).  More
typically, {\it bootstrapping} approaches have been employed. The
first such example is  described in \cite{Hearst91}.  There a
supervised learning algorithm is trained with a small amount of
manually sense--tagged text and applied to a held out test set. Those
examples in the test set that are most confidently disambiguated are
added to the training sample and the supervised learning algorithm is
re--trained with this larger collection of examples. 

\cite{Yarowsky95} describes a more recent bootstrapping approach. This
method takes advantage of the {\it one sense per collocation}
hypothesis put forth in \cite{Yarowsky93}, where it is observed that 
words have a strong tendency to be used only in one sense in a given
collocation.  This is an extension of the observation made in
 \cite{GaleCY92}  that words tend to be used only in one sense in a
given discourse or document, i.e., the {\it one sense per discourse}
hypothesis.  

This algorithm requires a small number of training examples to serve
as a seed. There are a variety of options discussed for automatically
selecting seeds; one is to identify  collocations that uniquely
distinguish between senses. For {\it plant},  the collocations {\it
manufacturing plant} and {\it living plant} make  such a
distinction. Based on 106 examples of {\it manufacturing plant}  and
82 examples of {\it living plant} this algorithm is able to
distinguish between two senses of {\it plant} for 7,350 examples with
97 percent accuracy. Experiments with 11 other words using collocation
seeds result in an average accuracy of 96 percent where each word had
two possible senses. 

There are relatively few approaches that attempt to perform
disambiguation only using information found in raw untagged text. 
One of the first such efforts is described  \cite{Schutze92}. There
words are represented in terms of the co-occurrence statistics of four
letter sequences.  This representation uses 97 features to
characterize a word, where each feature is a linear combination of
letter four-grams formulated by a singular value decomposition of a
5000 by 5000 matrix of letter four-gram co-occurrence frequencies.
The weight associated with each feature reflects all usages of the
word in the sample.  A context vector is formed for each occurrence of
an ambiguous word by summing the vectors of the contextual words. The
set of context vectors for the word to be disambiguated are then
clustered, and the clusters are manually sense-tagged.  

A related method is described in \cite{Schutze98}. However, here
ambiguous words are clustered into sense  groups based on 
second--order co--occurrences; two instances of an ambiguous word are
assigned to the same sense if the words that they co--occur with
likewise co--occur with similar words in the training data. In  the
previous approach the assignment to sense groups was based on
first--order co--occurrences where an ambiguous word was represented
by the four--grams it directly occurs with. It is reported that
second--order co--occurrences reduce sparsity and allow for the use of
smaller  matrices of co--occurrence frequencies. In this approach the
evaluation is performed relative to information retrieval tasks that
utilize the  sense group and do not require sense--tags. This results
in a fully automatic approach where no manual intervention is
required. 

While similar in spirit, the unsupervised work in this dissertation and 
that of Sch\"utze are somewhat distinct. The features employed in this
dissertation occupy a much smaller event space and rely mainly on 
collocations, part--of--speech and morphological 
information. While he also employs agglomerative clustering to form sense 
groups, the data is represented in terms of context vectors while the data  
here is represented in terms of dissimilarity matrices.


\chapter{CONCLUSIONS} 

This dissertation presents methods of learning probabilistic
models of word sense disambiguation that use both supervised and
unsupervised techniques. This chapter summarizes the contributions of
this research and outlines directions for future work. 

\section{Supervised Learning}

Supervised learning approaches to word sense disambiguation depend upon
the availability of sense--tagged text. While the amount of such text
is still limited, there has been a definite increase in quantity
in recent years. The largest contribution to this has been the release
of the DSO corpus, discussed in the previous chapter. Given the
likelihood that even larger amounts of sense--tagged text will become
available, continuing to develop and improve supervised learning
approaches for word sense disambiguation is an important issue. 

Indeed, while the cost of manually tagging text with senses 
is high, it is still a less expensive enterprise than 
creating the resources utilized by knowledge--intensive
approaches to disambiguation. These more elaborate representations of
knowledge bring with them an additional problem; suitable inferencing
mechanisms must also be developed to reason from this data.  When viewed
against these alternatives, the cost of manually annotating text
is actually quite modest. 

\subsection{Contributions}

This dissertation advances the state of supervised learning as applied
to word sense disambiguation in the following ways:  

\newpage

{\em The information criteria are introduced as evaluation criteria
for sequential model selection as applied to word sense disambiguation.}
These are alternatives to significance tests that result in a fully 
automatic selection process. The information criteria do not require 
manually tuned values to stop the model selection process; such a 
mechanism is inherent in their formulation. 

In particular, Akaike's Information Criteria is shown to result in a
model selection process that automatically selects accurate models of
disambiguation using either backward or forward search.  

{\em Forward sequential search is introduced as a search strategy for
sequential model selection as applied to word sense disambiguation.} This 
is an alternative to backward search that is especially well suited for 
the sparse data typical in language processing. Forward sequential search 
has the advantage that the search process starts  with models of very low 
complexity. This results in candidate models that have a small number of 
parameters whose estimates are well supported even in relatively small 
quantities of training data.   This ensures that the selection process 
makes decisions based upon the best available information at the time.

{\em This dissertation also introduces the Naive Mix, a new supervised
learning algorithm for word sense disambiguation.}  
The Naive Mix averages an entire sequence of decomposable models
generated during a sequential selection process to create a
probabilistic model.  It is more typical that model selection methods
only find a single best model. However, this dissertation shows that
there are usually several different models that result in similar
levels of accuracy; this suggests a degree of uncertainly in model
selection that is accommodated by the Naive Mix.   

Empirically, the Naive Mix is shown to result in improved accuracy
over single best selected models and also proves to be competitive
with leading machine learning algorithms. It is also observed that the
learning rate of the Naive Mix is very fast. It often learns models of
high accuracy using small amounts of training data, sometimes with 
as few as 10 or 50 sense--tagged examples. 

Despite making rather broad assumptions about the dependencies among
features in models of disambiguation, Naive Bayes consistently results
in accuracy that is  competitive with a host of other methods. 
{\em This dissertation presents an analysis of Naive Bayes that includes a
study of the learning rate as well as a bias--variance decomposition
of classification error. }   

The learning rate reveals that Naive Bayes has poor accuracy when the
training sample sizes are small. Given its fixed parametric form it is
easily mislead by spurious patterns in very small amounts of
sense--tagged text. However, as the amount of training data is
increased, it quickly achieves levels of accuracy comparable to
methods that build more representative models of the training
data. 

This behavior  is analyzed via a bias variance decomposition and
reveals that the nature of the errors made by the Naive Bayes model
are substantially different than those made by a more representative
model of the training data, here represented as a decision tree.  The
bulk of classification errors made by Naive Bayes are due to the
assumptions conveyed in the parametric form of the model. However, it
also tends to be very robust to differences between the test and
training data. By contrast, the errors made by a decision tree learner
are largely due to a failure to generalize well enough to 
accommodate differences between test and  training instances. However, 
despite these different sources of error, the total level of 
classification accuracy achieved by both methods is comparable.     

\subsection{Future Work}

The continued viability of supervised learning for word sense
disambiguation  is largely dependent on the availability of 
sense--tagged text. Thus, the creation of such text at relatively low
cost must be a high priority; future improvments in supervised
learning methodologies will be of little interest if sufficient
quantities of training data are not readily available. 

In general, supervised learning is a well developed area of
research. However, natural language poses peculiar problems that have
not necessarily been accounted for in previous work. Continued
refinement of supervised learning methodologies as applied to natural
language processing problems is an important area of future work. 

{\em Creation of Sense--Tagged Text:}
The manual annotation of text with sense tags is the clearest route to
expanding the current pool of sense--tagged text. Results from this
dissertation suggest that even relatively small amounts of
sense--tagged text can result in high levels of disambiguation accuracy. 
This is encouraging news, suggesting that even small additions to the 
available quantity of sense--tagged text will prove to be a valuable
resource for word sense disambiguation. 

Traditional manual annotation efforts will benefit greatly from the
development of tools that provide some degree automated assistance. As
an example, the Alembic workbench \cite{alembic}, provides support for
discourse process tagging tasks. A similar tool devoted to word
sense disambiguation would considerably ease the burden of manual 
annotation. If a human tagger noticed a particularly salient
co--occurrence or collocation, such a tool could allow for the rapid
tagging of a large number of similar instances. For example, suppose
that a human tagger notes that any time {\it interest rate} occurs, it
is nearly certain that {\it interest} refers to the
cost of borrowing money. After the first such instance is manually
sense--tagged, an annotation tool locates all the sentences in a
corpus where {\it interest rate} occurs and applies that same sense
tag automatically. 

When a commitment is made to manual annotation, there are related
questions that arise. Which sense inventory should be used for a
particular domain?  Are the sense distinctions in any dictionary clear
enough so that only one sense can be assigned to a particular word in
a particular context? How can tagger uncertainty be incorporated into
sense tagging? How large a factor is human error in manual annotation
efforts?  All of these questions open up new areas of future research.

As an alternative to manual sense--tagging, 
the large amount of text linked together via the World Wide Web can be
viewed as a source of alternative sources of knowledge to apply to
language processing problems.  A hyperlink
connecting a word or a phrase to a related web page is not nearly as
precise a source of information as is the link from a word to a sense
inventory, i.e., a sense tag. However, this diversity brings richness;
hyperlinks from {\it mallard} could lead to photos, stories from duck
watching expeditions, or maps showing migratory patterns. Short
summaries generated from these various resources (or provided by the
web page creator by way of a title or introductory comment) can then
serve as definitions or descriptions of the word or phrase in the
referring web page. This process ultimately results in an abstracted
and simplified version of the relevant portion of the Web that can
then be treated as a knowledge representation structure from which
inferences about other bodies of text can be made. 

{\em Varying Search Strategies:}
To date only backward and forward search strategies have been utilized
with sequential model selection for word sense
disambiguation. However, these are greedy approaches that conduct very
focused searches that can bypass models that are worthy of
consideration.  Developing approaches that combine backward and
forward search is a potential solution to this problem.    

Given the success of Naive Bayes, an alternative strategy is to begin
forward searches  at Naive Bayes rather than the model of
independence. However, this strategy presumes that all the features
are relevant to disambiguation and disables the ability to perform
feature selection. In order to allow model selection to disregard
irrelevant features, the process could begin at Naive Bayes and perform
a backward search to determine if any dependencies can safely be
removed. The model that results from this backward search then serves
as the starting point for a forward search. At various intervals the
strategy could be reversed from forward to backward, backward to
forward, and so on,  before arriving at a selected model. 

An alternative to starting the forward searches at Naive Bayes is to
generate a model of moderate complexity randomly and then search
backward for some number of steps, then forward, and so on until until a
model is selected. This entire process is repeated some number
of times so that a variety of random starting models are employed. The
models that are ultimately  selected presumably differ somewhat and
could be averaged together in a randomized variant of the Naive Mix.

If a reversible search strategy is adopted then the information 
criteria present certain advantages over the significance tests as
evaluation criteria. The information criteria perform at roughly the
same levels of accuracy during backward and forward search and do not
require any adjustment when changing search direction. However, the
significance tests are somewhat sensitive and  require that the
pre--determined cutoff value, $\alpha$, be reset as the direction of
the search changes.   

{\em Extending Feature Sets:}
The feature set employed for supervised learning in this dissertation
relies upon part--of--speech of the surrounding words, morphology of
the ambiguous word, and collocations that occur anywhere in the
sentence. A potential extension to this feature set is to incorporate
co--occurrence features. Preliminary experimental results with
feature sets made up entirely of co--occurrences that occur within 1
or 2 positions of an ambiguous word result in disambiguation accuracy
that is at least comparable to that of the supervised learning feature
set. This result, mentioned briefly in \cite{PedersenB97B}, largely
inspired the use of co--occurrence features in the unsupervised
learning experiments.  

The feature set could also be extended beyond the sentence boundary to
include features that occur in the same paragraph or even the same
document as the ambiguous word. This would allow for the inclusion of
features that provide information about earlier occurrences of a word
and the sense it was determined to have in that previous context. For
example, if an instance of {\it bill} is being disambiguated and it is
known that two sentences earlier {\it bill} refers to a bird jaw then
it seems unlikely that the current occurrence is being used in the
sense of pending legislation.  

\section{Unsupervised Learning}

The development and improvement of unsupervised learning techniques is
an important issue in natural language processing given the difficulty
in obtaining training data for supervised learning. The lack of
sense--tagged text poses a considerable bottleneck when porting
supervised learning methods to new domains and unsupervised methods
offer a way to eliminate this need for sense--tagged text.     

\subsection{Contributions}

The contributions of this dissertation to unsupervised learning of
word senses are as follows:       

{\em Several feature sets appropriate for unsupervised learning of word
senses from raw text are developed.} Feature sets designed for use
with supervised approaches are not directly applicable in an
unsupervised setting since they often contain features whose values
are based on information only available in sense--tagged text.    

The local context features developed for unsupervised learning
are co--occurrences that occur within a few positions of the ambiguous
word.  It is more common for  unsupervised approaches learning from
raw text to rely upon a much wider window of context. However, such
approaches result in high dimensional event spaces that can press the
limits of computing resources. The use of local context features in
this dissertation has led to acceptable levels of disambiguation
accuracy while still maintaining a relatively modest event space. 

{\em This dissertation develops probabilistic models for word sense
disambiguation without utilizing sense--tagged text.} 
The EM algorithm and Gibbs Sampling are used to estimate the
parameters of probabilistic models of disambiguation based strictly
upon information available in raw untagged text. 

Empirical comparison shows that Gibbs Sampling results in limited
improvement over the  accuracy of models learned via the EM
algorithm. The similar results are somewhat surprising given the
tendency of the EM algorithm to find local maxima. However, the
combination of local context features and the parametric form of Naive
Bayes results in relatively small event spaces where parameter
estimation is still fairly reliable. 

Despite its widespread popularly in a wide range of other
applications, Gibbs Sampling has not previously been applied to word
sense disambiguation. The introduction of this technique is an
important contribution since it is a general purpose methodology that
can be used in a variety of language processing problems. 
\newpage

Finally, the comparable accuracy of the EM algorithm and Gibbs
Sampling suggests that rather than competing methodologies these
should be treated as complimentary. The EM algorithm appears to
provide a reasonably good and very efficient first pass through
untagged data. It may be reasonable to approach unsupervised learning
using the EM algorithm first and then allowing Gibbs
Sampling to continue from there. A similar suggestion is made 
in \cite{MengV97}.

{\em McQuitty's similarity analysis has not been applied to word sense
disambiguation previously.} It is a simple agglomerative clustering
algorithm that makes no assumptions about the nature of the data it is
processing and yet results in accurate disambiguation in an unsupervised
setting. This approach requires that the data to be disambiguated be
converted into a dissimilarity matrix representation that shows the
number of mismatched features between observations. 

Despite the simplicity of both the algorithm and the data
representation, McQuitty's method  is shown to consistently result in
more accurate disambiguation than a well--known agglomerative
clustering algorithm, Ward's minimum--variance method. It also
outperforms the EM algorithm and Gibbs Sampling when disambiguating
words with very skewed sense distributions such as the adjectives and
verbs in these experiments. 

\subsection{Future Work} 

Unsupervised approaches to word sense disambiguation are of interest
because they eliminate the need for sense--tagged text. Disambiguation
can be performed based solely on information found in raw untagged
text. However, the lack of sense--tagged text impacts much more than
the learning algorithm itself. Both the feature sets and the
evaluation methodology must be formulated somewhat differently than in
the case of supervised learning. 

{\em Meaningful Labeling of Sense Groups:}
In the process of eliminating the need for  sense--tagged text to
learn probabilistic models of disambiguation, unsupervised approaches
also remove the link between the text and a sense inventory
established by a dictionary or some other lexical resource. Thus, the
sense groups that are created by an unsupervised learner do not have
meaningful sense labels  or definitions automatically attached to
them; the sense groups are tagged with meaningless names. This poses a
problem if the evaluation of the unsupervised learner is relative to 
human  sense judgments which are in turn based on knowledge of an
established sense inventory. 

This dissertation addresses this problem 
by developing an evaluation methodology where a post--processing step
is performed that maps sense groups to entries in a sense inventory
via sense--tagged text. While this allows for very exact measurements
of the agreement between the unsupervised learner and a human judge,
it also imposes a requirement for sense--tagged text on the evaluation
methodology. 

A more automatic alternative is to generate some form of sense
descriptions from the sense groups themselves. It is unlikely that
definitions as precise as those found in a dictionary could be
created. However, some meaningful labeling of the sense groups based
on the content of the sentences assigned to the sense group is
possible and might provide enough additional information to make a
link to a known entry in a sense inventory. 

This is perhaps best viewed as another  manifestation of a 
text summarization
problem. Given the sentences that make up a sense group, generate a 
statement that summarizes those sentences. Suppose the following
usages of the ambiguous word {\it bank} are found in a sense group:  
\begin{quote}
I went to the {\it bank} to deposit the money. \\
The {\it bank} extended a loan to the Martinez family. \\
Chase Manhattan bought my {\it bank}. \\
The Federal Reserve {\it Bank} controls the money supply.
\end{quote}

While it seems improbable that a formal dictionary definition could be
generated from these examples, it is possible to imagine the creation
of a generalized description such as {\it An entity concerned with
financial matters}. This description can then be used to choose from
the entries in a sense inventory for {\it bank}, resulting in
the selection of an entry that includes some mention of finances, for
example. 

There are several potential problems in this approach. First, creating
these  generalized descriptions implies that some external knowledge
source is available. It is possible that this sort of external
knowledge will be just as difficult to acquire as  sense--tagged text. 
Second, incorrectly grouped instances could cause a description to become 
overly general, i.e., {\it An entity concerned with objects.}  

{\em Feature Selection:}
The frequency based features developed for unsupervised learning of
word senses result in reasonably accurate performance, however they do
not tend to provide much information about minority senses.

When using raw text features, values are usually  selected based on 
frequency of occurrence. This results in features that are often skewed
towards the majority sense, particularly if the majority sense is a
large one. The development of feature selection methods that pick out
values indicative of minority senses is a key issue for improving the
performance of unsupervised approaches. 

{\em Part--of--Speech Ambiguity:}
Since the early work of Kelly and Stone, sense disambiguation has
been detached from the problem of part--of--speech ambiguity. It has
generally been assumed that part--of--speech ambiguity is resolved
before sense disambiguation is performed. In supervised learning this
is a reasonable assumption since a human tagger must make a
part--of--speech judgment before assigning a sense--tag. However, 
in unsupervised learning where no such examples are employed, the
decoupling of sense and part--of--speech ambiguity may in fact
gloss over the fact that reliable part--of--speech information may not
be available in a truly unsupervised setting. 

There are two alternatives. In this dissertation part--of--speech
ambiguity is resolved by a rule based part--of--speech tagger that is
applied to the text before unsupervised learning begins. However, the
quality of this tagging is dubious, even though the part--of--speech
distinctions are rather coarse. The second option is to simply assume
that part--of--speech information will not be available for
unsupervised learning problems. This requires disambiguation using a
wider range of possible senses that will cross over multiple
parts--of--speech. This assumption would also suggest that
unsupervised disambiguation be based only on collocations,
co--occurrences, and any other immediately apparent lexical feature.
In fact, one of the feature sets in this dissertation (B) takes this
approach and performs as well as those features sets that rely more
heavily on syntactic information.





\end{thesis}
 
\end{document}